\documentclass{article}

\usepackage{arxiv}

\usepackage{color}
\usepackage[x11names]{xcolor}
\usepackage[utf8]{inputenc}
\usepackage[T1]{fontenc}

\usepackage{natbib}
\usepackage{amsmath, amsfonts, amssymb, amsthm, mathtools, empheq, dsfont}
\usepackage{multirow, booktabs}
\usepackage{siunitx}

\usepackage[ruled, lined, linesnumbered, noend]{algorithm2e}

\usepackage{float, wrapfig, subcaption}
\usepackage{tikz}

\usepackage{hyperref}
\RequirePackage[all]{hypcap}
\usepackage{url}

\usetikzlibrary{arrows, arrows.meta, patterns, automata, calc, positioning}
\allowdisplaybreaks
\newenvironment{algo}[1][1.0\textwidth]{%
    \begin{center}
        \begin{minipage}{#1}
            \begin{algorithm}[H]}{%
            \end{algorithm}
        \end{minipage}
    \end{center}
}

\def\texpdf#1#2{\texorpdfstring{#1}{#2}}

\def\ie{i.e.\@}

\def\epsilon{\varepsilon}

\def\A{\mathcal A}
\def\B{\mathcal B}
\def\C{\mathcal C}
\def\D{\mathcal D}

\def\F{\mathcal F}

\def\H{\mathcal H}

\def\L{\mathcal L}
\def\M{\mathcal M}
\def\N{\mathcal N}
\def\O{\mathcal O}
\def\P{\mathcal P}
\def\Q{\mathcal Q}

\def\S{\mathcal S}

\def\U{\mathcal U}

\def\X{\mathcal X}

\def\a{\mathbf a}
\def\b{\mathbf b}
\def\c{\mathbf c}

\def\f{\mathbf f}

\def\h{\mathbf h}
\def\i{\mathbf i}

\def\m{\mathbf m}

\def\o{\mathbf o}

\def\r{\mathbf r}
\def\s{\mathbf s}

\def\u{\mathbf u}
\def\v{\mathbf v}

\def\x{\mathbf x}
\def\y{\mathbf y}
\def\z{\mathbf z}

\def\NN{\mathbb N}

\def\RR{\mathbb R}

\def\0{\mathbf 0}
\def\1{\mathbf 1}

\def\->{\rightarrow}
\def\=>{\Rightarrow}
\def\<->{\leftrightarrow}
\def\<=>{\Leftrightarrow}

\DeclareMathOperator*{\expect}{\mathlarger{\mathbb{E}}}
\DeclareMathOperator*{\argmax}{arg\,max}

\makeatletter
\newcommand{\subalign}[1]{%
  \vcenter{%
    \Let@ \restore@math@cr \default@tag
    \baselineskip\fontdimen10 \scriptfont\tw@
    \advance\baselineskip\fontdimen12 \scriptfont\tw@
    \lineskip\thr@@\fontdimen8 \scriptfont\thr@@
    \lineskiplimit\lineskip
    \ialign{\hfil$\m@th\scriptstyle##$&$\m@th\scriptstyle{}##$\hfil\crcr
      #1\crcr
    }%
  }%
}
\makeatother

\definecolor{DarkBlue}{HTML}{36587E}
\definecolor{BrightOrange}{HTML}{EB811B}

\definecolor{C1}{HTML}{1f77b4}
\definecolor{C2}{HTML}{ff7f0e}
\definecolor{C3}{HTML}{2ca02c}
\definecolor{C4}{HTML}{d62728}
\definecolor{C5}{HTML}{9467bd}
\definecolor{C6}{HTML}{8c564b}
\definecolor{C7}{HTML}{e377c2}
\definecolor{C8}{HTML}{7f7f7f}
\definecolor{C9}{HTML}{bcbd22}
\definecolor{C10}{HTML}{17becf}

\def\vaa{\text{VAA}}
\def\vaastar{\text{VAA}\textsuperscript{*}}
\def\cqf{\texpdf{$\mathcal{Q}$}{Q}-function}
\def\abs#1{\left\lvert#1\right\rvert}

\begin{document}

\title{Warming up recurrent neural networks to maximise reachable multistability greatly improves learning}

\author{%
    Gaspard Lambrechts  \\
        \texttt{gaspard.lambrechts@uliege.be} \And
    Florent De Geeter \\
        \texttt{florent.degeeter@uliege.be} \And
    Nicolas Vecoven \\
        \texttt{nicolas.vecoven@uliege.be} \And
    Damien Ernst \\
        \texttt{dernst@uliege.be} \And 
    Guillaume Drion \\
        \texttt{gdrion@uliege.be}}

\date{}

\hypersetup{
pdftitle={Warming up recurrent neural networks to maximise reachable multistability greatly improves learning},
pdfauthor={Gaspard Lambrechts, Florent De Geeter, Nicolas Vecoven, Damien Ernst, Guillaume Drion},
pdfkeywords={recurrent neural network, multistability, initialisation procedure, long-term memory, warmup},
}

\maketitle

\begin{abstract}
    Training recurrent neural networks is known to be difficult when time dependencies become long.
    In this work, we show that most standard cells only have one stable equilibrium at initialisation, and that learning on tasks with long time dependencies generally occurs once the number of network stable equilibria increases; a property known as multistability.
    Multistability is often not easily attained by initially monostable networks, making learning of long time dependencies between inputs and outputs difficult.
    This insight leads to the design of a novel way to initialise any recurrent cell connectivity through a procedure called “warmup” to improve its capability to learn arbitrarily long time dependencies.
    This initialisation procedure is designed to maximise network reachable multistability, \ie{}, the number of equilibria within the network that can be reached through relevant input trajectories, in few gradient steps.
    We show on several information restitution, sequence classification, and reinforcement learning benchmarks that warming up greatly improves learning speed and performance, for multiple recurrent cells, but sometimes impedes precision.
    We therefore introduce a double-layer architecture initialised with a partial warmup that is shown to greatly improve learning of long time dependencies while maintaining high levels of precision.
    This approach provides a general framework for improving learning abilities of any recurrent cell when long time dependencies are present.
    We also show empirically that other initialisation and pretraining procedures from the literature implicitly foster reachable multistability of recurrent cells.
\end{abstract}


\section{Introduction}
\label{sec:introduction}

Despite their performances and widespread use, recurrent neural networks (RNNs) are known to be blackbox models with extremely complex internal dynamics.
A growing body of work has focused on understanding the internal dynamics of trained RNNs \citep{sussillo2013opening,ceni2020interpreting, maheswaranathan2019reverse}, providing invaluable intuition into the RNN prediction process.
This viewpoint has already been used to understand the difficulties for RNNs to capture longer time dependencies \citep{bengio1993problem, doya1993bifurcations}.
In particular, recent work has highlighted the important role played by fixed points in RNN state spaces, that are defined as hidden states that updates to themself for a given input \citep{sussillo2013opening, katz2017using}.
This line of work has argued that locating such fixed points efficiently could provide insights into RNN dynamics and input-output properties.
Here, we build upon this line of work by studying the impact of the number of reachable fixed points in an RNN on the ability to learn long time dependencies.
Moreover, we highlight how maximizing the number of reachable fixed points at initialisation can improve RNN learning, in particular in the presence of arbitrarily long dependencies.

More precisely, we introduce a fast-to-compute measure of the multistability of a network called variability amongst attractors (VAA). This measure gives the number of reachable attractors for a set of initial states.
We show that loss decrease during learning in tasks with long time dependencies is highly correlated with an increase in VAA, highlighting both the relevance of the measure and the importance of multistability for efficient learning.
Second, we use stochastic gradient ascent on a differentiable proxy of the VAA, called \vaastar{}, as a way of maximizing the number of reachable attractors within the network at initialisation.
We show that this technique strongly improves performance on long time dependencies benchmarks, at the cost of precision, the latter relying on the richness of network transient dynamics.
Third, we propose a parallel recurrent network structure with a partial warmup that enables one to combine long-term memory through multistability with precision through rich transient dynamics.
Finally, we show empirically that other methods from the literature such as the chrono initialisation and the bistable recurrent cells implicitly achieve the same goal of maximising the number of reachable attractors.
Another pretraining procedure, the auxiliary losses proposed by \citet{trinh_learning_2018}, are also shown to foster multistability and to achieve good results on benchmarks with long time dependencies, using a much heavier procedure.
For the sake of clarity, those results are only reported in \autoref{app:aux_losses}.

In \autoref{sec:background}, RNNs are introduced as dynamical systems and the concept of multistability is introduced for those systems.
In \autoref{sec:benchmarks}, the supervised learning and reinforcement learning benchmarks are introduced.
In \autoref{sec:correlating_multistability_and_learning}, the VAA is introduced along with the estimation procedure of the multistability of an RNN for a set of initial states.
The correlation between multistability and learning is shown empirically on the benchmarks with long time dependencies.
In \autoref{sec:fostering_multistability_at_initialisation}, the \vaastar{} is introduced along with the warmup procedure that fosters multistability at initialisation.
The benefits of warmup are shown empirically on benchmarks with long time dependencies.
In addition, the double-layer architecture is introduced and shown to achieve a better performance on all benchmarks.
Finally, \autoref{sec:conclusion} concludes and proposes several future works.

\section{Related works}
\label{sec:related_works}


Training RNNs is known to be difficult when time dependencies become too long \citep{pascanu2013difficulty}.
Indeed, the most used algorithm to train RNNs is the backpropagation through time (BPTT) algorithm \citep{werbos1990backpropagation}, which unrolls the RNN to see it as a feedforward neural network with shared weights before applying the backpropagation.
However, the longer the sequence, the deeper the corresponding feedforward neural network is.
Backpropagating through such deep networks often leads to vanishing or exploding gradients, and different methods have been proposed to tackle this issue.
These methods usually act on one of three different levels: the training, the initialisation/pretraining and the network architecture.

\paragraph{Training} These methods modify the training of RNNs.
For instance, clipping the gradients \citep{pascanu2013difficulty} prevents the gradients from exploding.
Another example is the truncated variant of BPTT \citep{williams_gradient-based_1995}, which does not propagate gradients through the whole sequences, but rather through parts of these sequences, leading to gradients that vanish or explode less often.
It is likely that truncating the BPTT prevents from learning long time dependencies efficiently.
Finally, \citet{trinh_learning_2018} propose adding auxiliary losses at some timesteps, to avoid having only one loss computed at the end of the sequences.
These losses are computed in an unsupervised fashion: either a decoder has to reconstruct a part of the sequence (\textit{reconstruction} loss), or a network has to predict the next input (\textit{prediction} loss).
This method can also be used as a pretraining to first train the RNN to encode correctly the sequences.
This work achieved good results on very long sequences, which motivated the aforementioned comparison with our work in \autoref{app:aux_losses}.

\paragraph{Initialisation/pretraining} The goal of these methods is to bring the network weights to a better place in the parameters space where the learning will be better and faster.
Notably, the chrono-initialisation \citep{tallec2018can, van2018unreasonable} changes the initial biases parameters to improve the learning of long time dependencies.
Some pretraining methods rely on autoencoders: \citet{pasa_pre-training_2014} use the parameters of a linear encoder as initial weights for the RNN, \citet{sagheer_unsupervised_2019} train a LSTM-based stacked autoencoder layer-wise before adding a output layer and fine-tuning on the dataset and \citet{ong_dynamic_2014} introduce a dynamic pretraining of AE specifically made for time-series.
\citet{pasa_neural_2015} pretrain the RNN on a smoothed version of the dataset produced by a first-order hidden Markov model and then fine-tunes on the original dataset.
\citet{tang_recurrent_2016} first train a DNN before using it as a teacher to train the RNN.
\citet{ienco_supervised_2019} focus on multi-class sequences classification.
A trained RNN is used to rank the classes by decreasing order of complexity, then a new RNN is pretrained to predict the most complex class, then the second one, etc.
All these pretraining methods have improved the performance of RNNs either on classification or on time-series prediction tasks. While making the final training of the network easier and better, none of them seems to directly promote the learning of long time dependencies.

\paragraph{Network architectures} The most notable improvement made in the RNN architectures is the introduction of the gates, which are used to control the flow of information in the network and to help the gradients to propagate through the time.
These gates have led to the development of the long-short term memory (LSTM) \citep{hochreiter1997long} and the gated recurrent unit (GRU) \citep{cho2014properties}, which are now the most used RNNs in practice.
In the experiments, we also consider the minimal gated unit (MGU) \citep{zhou2016minimal}, a minimal design among gated recurrent units that only has one gate.
Other approaches include the introduction of different time-scales inside the RNN.
The segmented-memory RNN \citep{jinmiao_chen_segmented-memory_2009} splits the sequences into segments and uses a two-layers RNN, where the first layer is reset at the end of each segment, while the second one is updated when a new segment begins.
The hierarchical RNN \citep{hihi_hierarchical_1995}, the hierarchical multiscale RNN \citep{chung_hierarchical_2017} and the clockwork RNN (CW-RNN) \citep{koutnik_clockwork_2014} stack recurrent layers that are updated at different frequencies.
The structurally constrained recurrent network (SCRN) \citep{mikolov_learning_2015} imposes some constraints on a subset of the recurrent weights, forcing some neurons hidden states to be slowly updated.
The nonlinear autoregressive with exogenous inputs (NARX) RNN \citep{lin_learning_1996, menezes_long-term_2008} uses the $n$ previous hidden states as inputs, making it a $n^{th}$-order RNN.
Likewise, novel recurrent cell dynamics, such as the bistable recurrent cell (BRC) and the neuromodulated BRC (NBRC) \citep{vecoven2021bio}, have been introduced to help tackle long time dependencies benchmarks.
NBRCs were specifically designed to maximise reachability of cellular bistability, providing a way to create never-fading memory at the cellular level.
These results highlighted how dynamics of untrained RNNs, \ie{}, at initialisation, can strongly impact learning performance of RNNs.
In this work, we extend this approach at the network level by maximizing multistability of any recurrent cell type prior to learning.
To this end, we propose a novel RNN pretraining procedure called “warmup” that is designed to maximise the number of RNN attractors that can be reached from hidden states resulting from input sequences.
Compared to pretraining methods, this method is very efficient since it only requires a few gradient steps before reaching a multistable regime for the RNN.

\section{Background} \label{sec:background}

In this section, RNNs are introduced as dynamical systems.
The fixed points of these systems are defined, and the notions of attractors, reachable attractors and multistability are introduced.

\subsection{Recurrent neural networks} \label{sec:recurrent_neural_networks}

RNNs are parametric function approximators that are often used to tackle problems with temporal structure.
Indeed, RNNs process the inputs sequentially, exhibiting memory through hidden states that are outputted after each timestep, and processed at the next timestep along with the following input.
These connections allow RNNs to memorise relevant information that should be captured over multiple timesteps.
More formally, an RNN architecture is defined by its update function $f$, its output function $g$ and its initialisation function $h$ that are parameterised by a parameter vector $\theta \in \RR^d$.
Let $\u_{1:T} = [\u_1, \dots, \u_T]$, with $T \in \NN$ and $\u_t \in \RR^n$, an input sequence.
RNNs maintain an internal memory state $\x_t$ through an update rule $\x_{t} = f(\x_{t-1}, \u_t; \theta)$ and output a value $\o_t = g(\x_t; \theta)$, where the initial hidden state $\x_0 = h(\theta)$ is often chosen to be zero.
We note that often, the output of the RNN is simply its hidden state $\x_t$, \ie{} $g$ is the identity function.
RNNs can be composed of only one recurrent layer, or they can be built with $L$ layers that are linked sequentially through $\u^i_t = \o^{i-1}_t$ with $\u^1_t = \u_t$ and $\o_t = \o^L_t$, where $\o^i_t$ denotes the output of layer $i$ and $\u^i_t$ its input.
In this case, each layer $i$ has its own update function $f^i$, output function $g^i$ and initialisation function $h^i$.
Backpropagation through time is used to train these networks where gradients are computed through the complete sequence via the hidden states \citep{werbos1990backpropagation}.
The following recurrent architectures are considered in the experiments: LSTM, GRU, BRC, NBRC, MGU. The specific update functions of those RNNs can be found in \autoref{app:recurrent_neural_network_architectures}. In addition, we consider the chrono-initialised LSTM.

\subsection{Fixed points in recurrent neural networks} \label{sec:fixed_points}

\paragraph{Fixed points in $\u$.}

In dynamical systems, fixed points are defined as points in the state space that map to themselves through the update function, for a given input $\u$.
For a system $f$, we say that a state $\x^*$ is a fixed point in $\u$ if and only if
\begin{equation}
    \x^* = f(\x^*,\u).
\end{equation}

\paragraph{Attractors in $\u$.}

Fixed points can either be fully attractive (attractors), fully repulsive (repellors), or combine attractive and repulsive manifolds (saddle points).
For a constant input $\u$, the set of starting states for which the system converges to the fixed point $\x^*$ is called basin of attraction of $\x^*$ in $\u$ and is written as
\begin{align}
    \mathcal{B}^\u_{\x^*}     & = \left\{ \x \;\middle\vert\; \lim_{n\to\infty} f^n(\x, \u) = \x^*\right\} \\
    \text{ with } f^n(\x, \u) & =
    \begin{cases}
        f\left(f^{n-1}(\x, \u), \u\right) \text{ if $n > 1$}, \\
        f(\x, \u) \text{ if $n = 1$}.
    \end{cases}
\end{align}
If the limit is not defined for some point $\x$, then this point does not belong to any basin of attraction in $\u$.
Mathematically, $\x^*$ is an attractor in $\u$ if its basin of attraction in $\u$, $\mathcal{B}^\u_{\x^*}$, has a positive measure.

\paragraph{Reachable attractors in $\u$.}

In particular, we say that an attractor $\x^*$ in $\u$ is reachable from some state $\x$ if, and only if $\x \in \mathcal{B}^\u_{\x^*}$.

\paragraph{Monostability and multistability in $\u$.}

Given a set of states $\X = \{ \x_1, \dots, \x_n \}$, a system that has a unique reachable attractor in $\u$ for all states is said to be monostable in $\u$ for this set, whereas a system that has multiple reachable attractors in $\u$ is said to be multistable in $\u$ for this set. More formally, $f$ is said to be monostable in $\u$ for $\X$ if, and only if, there exists a unique attractor $\x^*
$, such that $\forall \x \in \X, \; \x \in \mathcal{B}^\u_{\x^*}$.
On the contrary, $f$ is said to be multistable in $\u$ is, and only if, there exists at least two attractors $\x^*_1$ and $\x^*_2$ such that $\x^*_1 \neq \x^*_2$ and $\exists \x_1, \x_2 \in \X, \; \x_1 \in \B^\u_{\x^*_1}, \; \x_2 \in \B^\u_{\x^*_2}$.

\paragraph{Recurrent neural networks.}

Due to their temporal nature and update rules, RNNs can be seen as discrete-time non-linear dynamical systems.
Formally, given a parameter vector $\theta$, the system $f$ is given by the update function of the RNN, such that $f(\x, \u) = f(\x, \u; \, \theta)$.
Since attractors correspond to network steady states, they are thought to be the allowing factor for RNNs to retain information over a long period of time \citep{pascanu2013difficulty, sussillo2013opening, maheswaranathan2019reverse}.

\section{Benchmarks} \label{sec:benchmarks}

In this section, the different benchmarks are introduced.
First, the supervised learning tasks are introduced, including long-term information restitution benchmarks in \autoref{sec:information_restitution_benchmark} and sequence classification benchmarks in \autoref{sec:sequence_classification_benchmarks}.
In \autoref{sec:rl_benchmarks}, a reinforcement learning benchmark with partially observable environment is introduced.
This environment contains long time dependencies.

\subsection{Long-term information restitution benchmarks}
\label{sec:information_restitution_benchmark}

The benchmarks introduced in this subsection contain long time dependencies, and therefore require networks able to remember relevant information for a long period.
For those benchmarks, RNNs are trained on a dataset of \num{40000} sample sequences and evaluated on a dataset of \num{40000} sample sequences.
During training, \num{20}\% of the training set is used as a validation set.

\paragraph{Copy first input benchmark.}

In this benchmark, the network is presented with a one-dimensional sequence of $T$ timesteps
$\u_{1:T}$, where $\u_t \sim \mathcal{N}(0, 1), \; t = 1, \dots, T$, and is tasked at approximating the target $\y_T = \u_1$.
This benchmark thus consists of memorizing the initial input for $T$ timesteps.
It allows one to measure the ability of recurrent architectures to bridge long time dependencies when the length $T$ is large.
Given the output $\o_T$ of the network, we seek to minimise the squared error $\L(\o_T, \y_T) = (\o_T - \y_T)^2$.

\paragraph{Denoising benchmark.}

In this benchmark, the network is presented with a two-dimensional sequence of $T$ timesteps.
The first dimension is a noised input stream $\u_{1:T}^1$, where $\u_t^1 \sim \mathcal{N}(0, 1), \; t = 1, \dots, T$.
Five timesteps of this stream should be remembered and outputted one by one by the network at timesteps $\{T - 4, \dots, T\}$.
These five timesteps $\S = \{t_1, t_2, t_3, t_4, t_5\}$, with $t_1 < t_2 < t_3 < t_4 < t_5$, are sampled without replacement in $\{1, \dots, T - N\}$ with $N \geq 5$.
$N$ is a hyperparameter that allows one to tune how long the network should be able to retain the information at a minimum.
The five timesteps are communicated to the network through the second dimension of the input $\u_{1:T}^2$, where $\u_t^2 = 1$ if $t \in \S$, and $\u_t^2 = 0$ otherwise, for $t = 1, \dots, T$.
The target is thus given by $\y_{T-4:T} = [\u_{t_1}, \u_{t_2}, \u_{t_3}, \u_{t_4}, \u_{t_5}]$.
Given the output sequence $\o_{T-4:T}$ of the network, we seek to minimise the mean squared error $\L(\o_{T-4:T}, \y_{T-4:T}) = \sum_{t = T - 4}^{T} (\o_t - \y_t)^2$.

\subsection{Sequence classification benchmarks}
\label{sec:sequence_classification_benchmarks}

The benchmarks introduced in this subsection are sequence classification problems and therefore require networks able to use the information received in the sequence in order to infer the class.
For those benchmarks, RNNs are trained on datasets derived from the usual train and test sets of the original MNIST dataset.
During training, \num{20}\% of the training set is used as a validation set.

\paragraph{Permuted sequential MNIST.}

In this benchmark, the network is presented with the MNIST images, where pixels are presented to the network one by one as a sequence of length $T = 28 \times 28 = 784$.
It differs from the regular sequential MNIST in that pixels are shuffled in a random order.
Note that all images are shuffled according to the same random order.\footnote{\label{fn:permutation}The permutation is given by: \texttt{np.random.seed(42); np.random.permutation(28*28);} (NumPy 1.23.2).}
The network is tasked at outputting a probability for each possible digit that could be represented in the initial image.
This benchmark is known to be a more complex challenge than the regular one.
Given the output $\o_T \in \RR^{10}$ of the network and the true digit index $\y_T \in \{1, \dots, 10\}$, we seek to minimise the negative log likelihood loss $\L(\o_T, \y_T) = - \log(\o_T^{\y_T})$.

\paragraph{Permuted line-sequential MNIST.}

This benchmark is the same as the permuted sequential MNIST benchmark, except that the pixels are fed \num{28} by \num{28}, which corresponds to one line of the permuted image.\textsuperscript{\ref{fn:permutation}}
The input dimension is thus \num{28} instead of one.
$N$ black lines are added at the end of the sequence such that the total length of the sequence is $T = 28+N$.
This has the effect of a forgetting period, such that any relevant information for predicting the class probabilities will be farther from the prediction timestep $T$.

\subsection{Reinforcement learning benchmark}
\label{sec:rl_benchmarks}

In reinforcement learning, the function approximators also process sequences as input when considering partially observable Markov decision processes (POMDPs).
Indeed, in such environments, the optimal policies, as well as the value functions, are functions of the complete sequence of observations and past actions, called the history.
In this work, we focus on the approximation of the history-action value function, or \cqf{}, in order to derive a near-optimal policy in the considered POMDP.
The deep recurrent Q-network (DRQN) algorithm is used to approximate this \cqf{} with an RNN.
From this approximation, we derive the fully greedy policy by taking the action that maximises the \cqf{} for any given history.
See \autoref{app:pomdp} for the formal definition of POMDPs and their \cqf{}s, and see \autoref{app:drqn} for the detailed DRQN algorithm.

\begin{wrapfigure}{r}{0.4\textwidth}
    \centering
    \vspace{-2em}
    \includegraphics[width=0.38\textwidth]{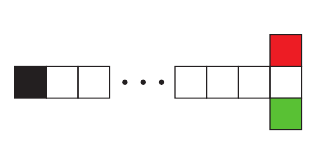}
    \vspace{-1em}
    \caption{%
        T-Maze layout example, with the initial position of the agent in black, the treasure in green and the cell to avoid in red.}
    \vspace{-1em}
    \label{fig:small_tmaze}
\end{wrapfigure}

The partially observable environment that is considered is the T-Maze environment \citep{bakker2001reinforcement}.
The T-Maze is a POMDP where the agent is tasked with finding the treasure in a T-shaped maze (see \autoref{fig:small_tmaze}).
The state is given by the position of the agent in the maze and the maze layout that indicates whether the goal lies up or down after the crossroads.
The initial state determines the maze layout, and it never changes afterwards.
The initial observation made by the agent indicates the layout.
Navigating in the maze provides zero reward, except when bouncing onto a wall, in which case a reward of $-0.1$ is received.
While traveling along the maze, the agent only receives the information that it has not yet reached the junction.
Once the junction reached, the agent is notified: it must now choose a direction depending on the past information it remembers.
Finding the treasure provides a reward of \num{4}.
Passed the crossroads, the states are always terminal.
The optimal policy thus consists of going through the maze, while remembering the initial observation in order to take the correct direction at the crossroads.
This POMDP is parameterised by the corridor length $L \in \NN$ that determines the number of timesteps for which the agent should remember the initial observation.
The discount factor is $\gamma = 0.98$.
This POMDP is formally defined in \autoref{app:tmaze}.

\section{Correlating multistability and learning}
\label{sec:correlating_multistability_and_learning}

This section aims at showing the correlation that exists between multistability properties of RNNs and their ability to learn long time dependencies.
To this end, in \autoref{sec:variability_amongst_attractors} we first introduce the \vaa{}, a measure of the number of basins of attraction that are spanned by a set of states.
In \autoref{sec:estimating_vaa}, we show how to estimate the multistability of an RNN using \vaa{} by estimating the number of reachable attractors for a set of states resulting from the input sequences.
We then carry out a number of experiments in \autoref{sec:vaa_experiments} to show the correlation between multistability and learning with different types of RNN on the benchmarks previously introduced.

\subsection{Variability amongst attractors}
\label{sec:variability_amongst_attractors}

One way to quantify the multistability in $\u$ of a system for a set of states $\X$ is to count the number of different attractors that can be reached starting from those states.
Our measure is named Variability Amongst Attractors (\vaa).
The \vaa{} of a system $f$ for a set of initial states $\X$ and an input $\u$ is defined as
\begin{equation}
    \vaa(f, \X, \u) = \frac{1}{\abs{\X}} \sum^{\abs{\X}}_{i = 1}
    \frac{
    1
    }{
    \sum\limits_{j = 1}^{\abs{\X}}
    \delta \left(
    \limsup\limits_{n\to\infty}
    \left\lVert
    f^n(\x_i, \u) - f^n(\x_j, \u)
    \right\lVert = 0
    \right)
    } \label{eq:theoretical_vaa}
\end{equation}
where $\delta(x)$ is the Kronecker delta function that returns \num{1} when condition $x$ is met, and \num{0} otherwise.
It can be noted that this definition does not exclude limit cycles and considers states that are on the same limit cycle but far from each others as different attractors.
This is a limitation that we discuss in our conclusion.
In the following, we make the hypothesis that such limit cycles are not encountered in practice.

The denominator of \autoref{eq:theoretical_vaa} gives the number of states in $\X$ that converge towards the same attractor as $\x_i$.
The sum of this fraction over all states that converge towards a given attractor is thus equal to one, such that the sum of this fraction over all states gives the number of different attractors.
$\vaa(f, \X, \u)$ is thus equal to the number of different attractors in $\u$ reached from the initial states contained in $\X$ divided by the number of initial states $\abs{\X}$.
Its maximal value is thus \num{1}, when all reached attractors are different, and its minimal value is $\frac{1}{\abs{\X}}$, when all the states have converged towards the same attractor (\ie, the system is monostable).

In practice, since it is impossible to evaluate the limits to infinity in the \vaa{}, we fix a finite number of timesteps $M$ for state convergence, called the stabilization period.
As a consequence, the system may not have completely converged towards the attractor after this period.
We thus define a tolerance $\epsilon$ below which two final states are considered to correspond to the same attractor.
This truncated $\vaa$ is written as
\begin{equation}
    \vaa_{M, \epsilon}(f, \X, \u) = \frac{1}{\abs{\X}}\sum^{\abs{\X}}_{i = 1} \dfrac{1}{\overset{\abs{\X}}{\underset{j = 1}{\sum}} \, \delta\left(\left\lVert f^M(\x_i, \u) - f^M(\x_j, \u) \right\lVert \leq \epsilon \right)} .
    \label{eq:practical_vaa}
\end{equation}

\subsection{Estimating the multistability of an RNN for a set of input sequences} \label{sec:estimating_vaa}

RNNs can exhibit a long-lasting memory through multistability in their hidden states \citep{vecoven2021bio}.
Indeed, having multiple attractors that are reachable from different input sequences probably allows one to encode information about these sequences over the long term.
We propose estimating the multistability of an RNN for a set of input sequences by computing the number of different reachable attractors for hidden states resulting from different input sequences.
More precisely, we propose to compute $\vaa(f, \X, \u)$ for hidden states $\X$ sampled from different input sequences.
In practice, it is not feasible to estimate the \vaa{} for all hidden states resulting from the set of input sequences.
Indeed, computing the \vaa{} is quadratic in the number of hidden states because of the pairwise distances.
We thus propose to estimate the \vaa{} by averaging its value over several small batches of hidden states sampled at random time steps in different sequences sampled from the set of input sequences.
Moreover, we still have to choose the stable input $\u$ according to which we want to measure the multistability in $\u$.
In order to measure the multistability of the network for a wide range of stable inputs, we propose to measure the multistability on average for several inputs sampled according to a standard normal distribution.
Note that for each batch of hidden states, a unique $\u \sim \N(\0, \1)$ is sampled and kept constant during the convergence period of $M$ timesteps.
The resulting procedure for estimating the multistability of an RNN for a set of input sequences is given in \autoref{algo:vaa}.

\begin{algo}
    \capstart \DontPrintSemicolon \footnotesize
    \caption{
        Estimating the proportion of reachable attractors of an RNN for a set of input sequences}
    \label{algo:vaa}

    \SetKwInOut{Parameters}{Parameters}
    \SetKwInOut{Output}{Output}
    \SetKwFunction{RandomHiddenStates}{RandomHiddenStates}

    \Parameters{%
        $I \in \NN$ the number of iterations to compute the mean of the \vaa{}. \\
        $M \in \NN$ the stabilization period. \\
        $\varepsilon \in \RR^+$ tolerance when considering state similarity. \\
        $\theta \in \RR^{d_\theta}$ the parameters of the network.
    }

    \KwData{%
        $\D = \{\u_{1:T_1}^1,\dots,\u_{1:T_n}^N\}$
        a set of $N$ input sequences.
    }

    Let $f \leftarrow f(\cdot, \cdot; \; \theta)$ the dynamical system. \;
    Initialise mean value $\overline{\vaa{}} \leftarrow 0$. \;
    \For{$i = 1, \dots, I$}{
        Sample a batch of input sequences $\B \sim \D$. \;

        Sample a random hidden state in each input sequence $\X \leftarrow \RandomHiddenStates(\B, \theta)$. \;

        Sample $\u \sim \mathcal{N}(\mathbf{0}, \mathbf{1})$ \;

        $\overline{\vaa{}} \leftarrow \overline{\vaa{}} + \frac{1}{I} \, \vaa{}_{M, \varepsilon}(f, \X, \u)$ \;
    }

    \Return{$\overline{\vaa{}}$}

\end{algo}

\begin{algo}
    \capstart \DontPrintSemicolon \footnotesize
    \caption{Random Hidden States}
    \label{algo:rhs}

    \SetKwInOut{Parameters}{Parameters}
    \SetKwInOut{Output}{Output}

    \Parameters{%
        $\theta \in \RR^{d_\theta}$ the parameters of the network.

    }

    \KwData{%
        $\B = \{\u_{1:T_1}^1,\dots,\u_{1:T_n}^n\}$
        a batch of $n$ input sequence sampled in the training set.
    }

    $\X \leftarrow \{\}$ \;

    \ForEach {$\u_{1:T_i}^i \in \B$ } {

    Sample a timestep $t \sim \U(\left\{1, \dots, T_i\right\})$ \;

    Set $\x^i_0 = h(\theta)$ where $h$ is the RNN's initialisation function \;

    \For{$k = 1, \dots, t$}{
    Set $\x^i_k = f(\x^i_{k-1},\u_k^i;\,\theta)$
    where $f$ is the RNN's update function. \;
    }

    Update $\X \leftarrow \X \bigcup \left\{\x^i_t\right\}$ \;
    }

    \Return{$\X$}

\end{algo}

\subsection{Experiments}
\label{sec:vaa_experiments}

In this subsection, we observe how the multistability of RNNs evolves when they are trained on the long-term information restitution and reinforcement learning benchmarks introduced in \autoref{sec:benchmarks}.
The multistability of these networks is estimated throughout the training procedure, using \autoref{algo:vaa}.
For the copy first input benchmark, networks are made up of one \num{128} neurons recurrent layer.
For the other benchmarks, networks are made up of two recurrent layers, each of \num{256} neurons.
All averages and standard deviations reported were computed over five different training sessions.
Training was done using the Adam optimizer \citep{kingma2014adam} with a learning rate of \num{1e-3} and a batch size of \num{32}.
All hyperparameters have been chosen a priori to standard values and are kept fixed.
The goal here is not to measure the best performance of each architecture but rather to study, for a given architecture and optimization procedure, whether there is a link between learning and multistability for different benchmarks.
In \autoref{app:generalisation_vaa}, we show that those results also hold with other hyperparameters for the copy first input benchmark.
In all experiments, the multistability is estimated with $M = \num{10000}$, $\varepsilon = \num{1e-4}$, and $I = 10$.

\paragraph{Copy first input benchmark}

\begin{figure}[ht]
    \centering
    \begin{subfigure}{.28\textwidth}
        \centering
        \includegraphics[width=\textwidth]{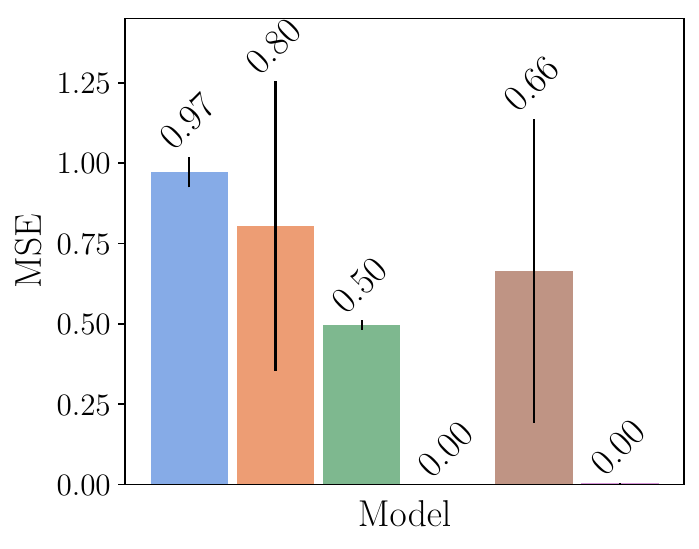}
        \caption{$T = 50$}
    \end{subfigure}
    \begin{subfigure}{.28\textwidth}
        \centering
        \includegraphics[width=\textwidth]{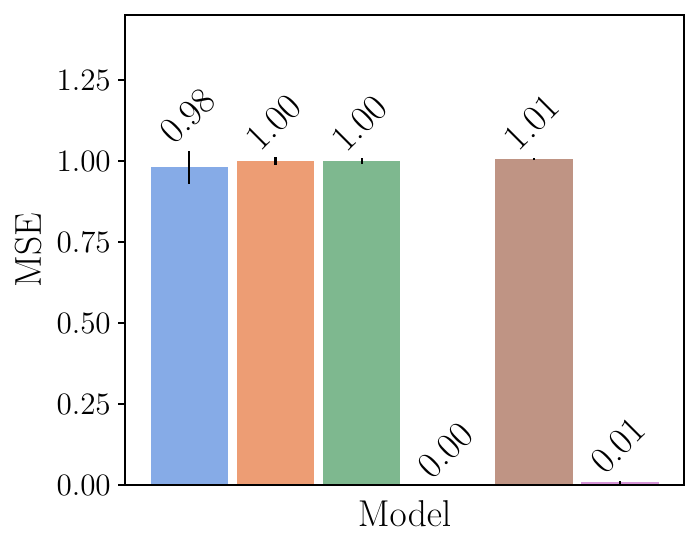}
        \caption{$T = 300$}
    \end{subfigure}
    \begin{subfigure}{.375\textwidth}
        \centering
        \includegraphics[width=\textwidth]{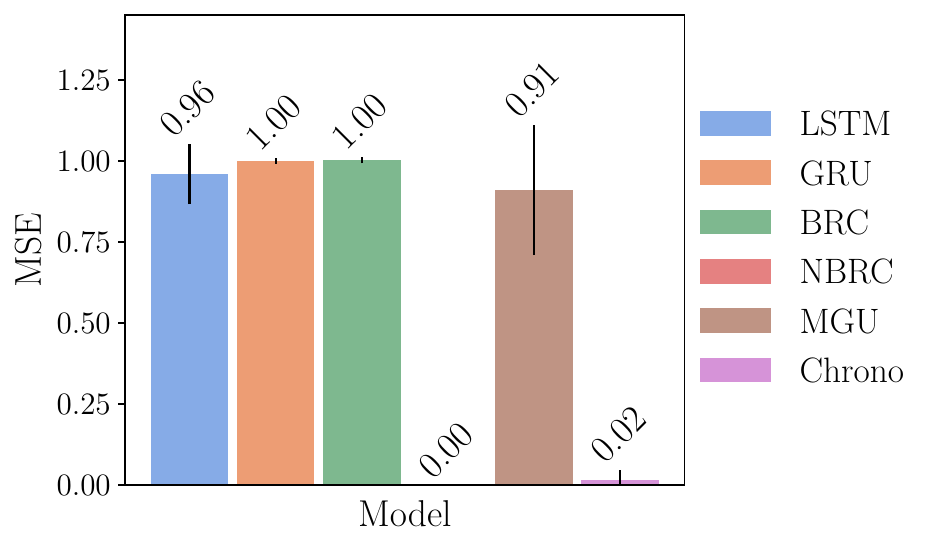}
        \caption{$T = 600$}
    \end{subfigure}
    \caption{%
        Test MSE loss for the copy first input benchmark with different sequence lengths $T$.
        Mean and standard deviation are reported after 50 epochs.}
    \label{table:vaa_copybench}
\end{figure}

\begin{figure}[ht]
    \centering
    \begin{subfigure}{.35\textwidth}
        \centering
        \includegraphics[width=\textwidth]{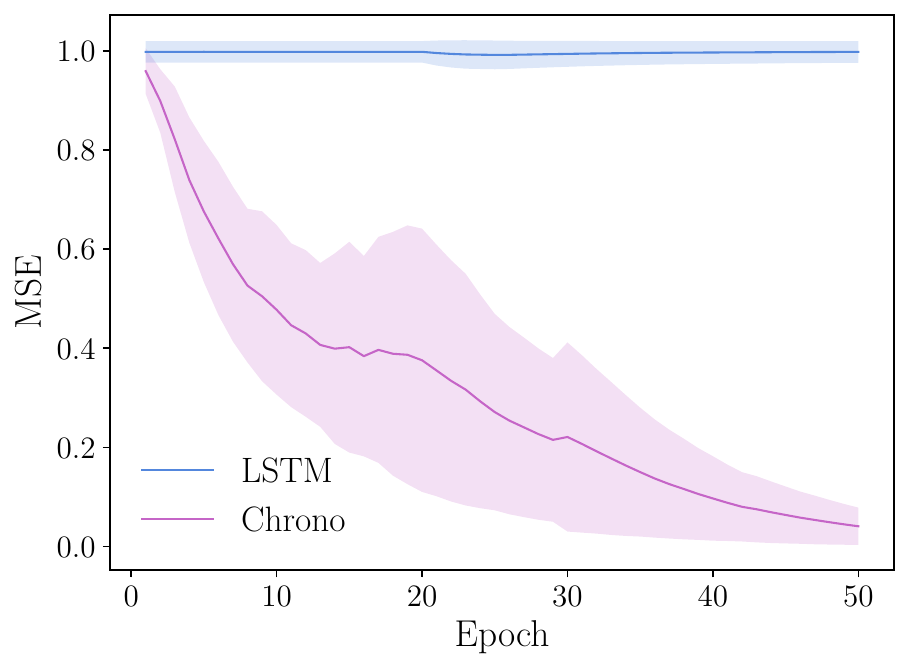}
        \label{fig:copyfirstinput_lstm_chrono_mse}
    \end{subfigure}
    \begin{subfigure}{.35\textwidth}
        \centering
        \includegraphics[width=\textwidth]{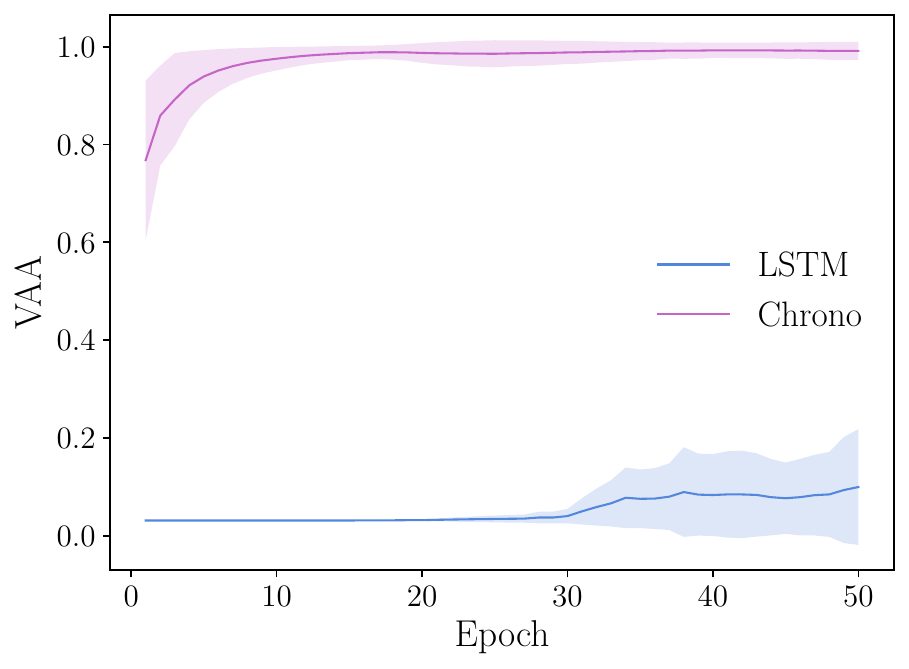}
        \label{fig:copyfirstinput_lstm_chrono_vaa}
    \end{subfigure}
    \caption{%
        Evolution of the validation loss (left) and of the VAA (right) of LSTM networks, with and without chrono initialisation, for the copy first input benchmark with $T = 50$.
        Mean and standard deviation are reported after 50 epochs.}
    \label{fig:copyfirstinput_lstm_chrono}
\end{figure}

\begin{figure}[ht]
    \centering
    \begin{subfigure}{.35\textwidth}
        \centering
        \includegraphics[width=\textwidth]{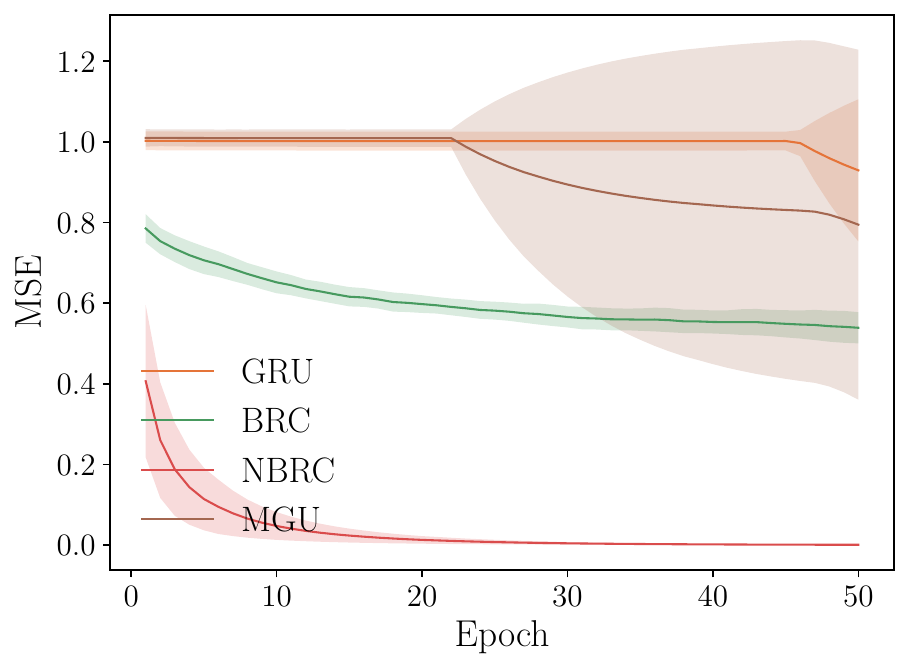}
        \label{fig:copyfirstinput_others_mse}
    \end{subfigure}
    \begin{subfigure}{.35\textwidth}
        \centering
        \includegraphics[width=\textwidth]{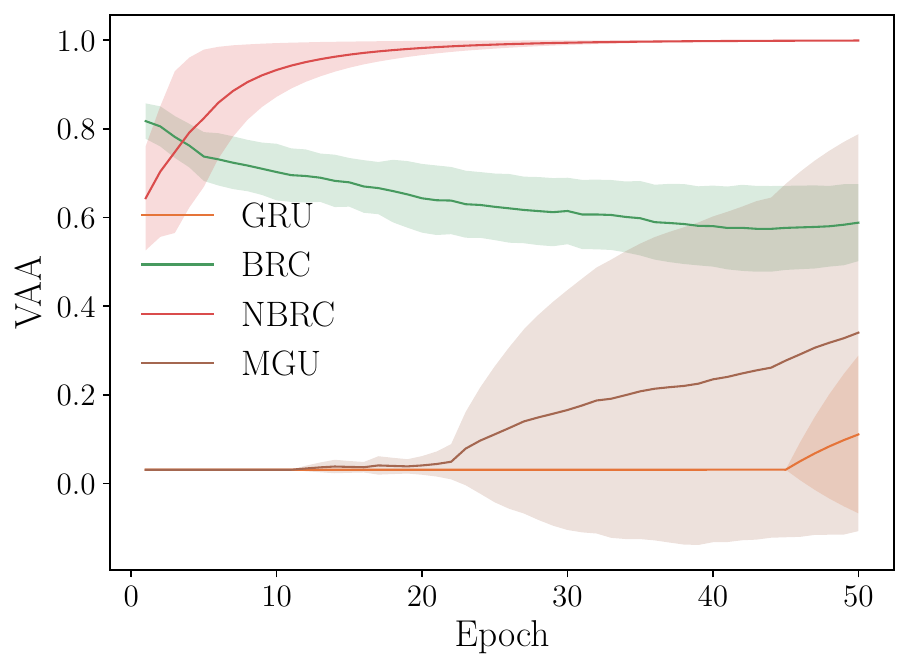}
        \label{fig:copyfirstinput_others_vaa}
    \end{subfigure}
    \caption{%
        Evolution of the validation loss (left) and of the VAA (right) of GRU, MGU, BRC and NBRC networks, for the copy first input benchmark with $T = 50$.
        Mean and standard deviation are reported after 50 epochs.}
    \label{fig:copyfirstinput_others}
\end{figure}

\autoref{table:vaa_copybench} shows the performance of the different cells on this benchmark for different sequence lengths $T \in \{50, 300, 600\}$.
The best-performing cell is the NBRC, whose performance is not affected by the length of the sequences.
In comparison, the classical cells, MGU, LSTM and GRU, struggle to decrease their losses.
Surprisingly, the BRC, a bistable cell, does not succeed in decreasing its loss.
Generally speaking, the longer the sequences are, the worse their performances are.
The last cell, the chrono-initialised LSTM, competes with the NBRC with its hyperparameter $T_{max}$ chosen to $600$.
\autoref{fig:copyfirstinput_lstm_chrono} illustrates the correlation between the VAA and the validation loss for the LSTM and CHRONO cells.
The LSTM cell, whose VAA increases late and little, fails to learn.
On the other hand, the chrono initialised LSTM cell sees its loss decreasing while its VAA increases.
This figure also shows that the chrono initialisation promotes the learning of long time dependencies through multistability.
\autoref{fig:copyfirstinput_others} illustrates the correlation between the VAA and the validation loss for the other cells.
It is clear from this figure that the bistability mechanism introduced in the BRC and NBRC cells also promote multistability.
Moreover, as for the LSTMs and chrono-initialised LSTMs, the loss only starts decreasing when the VAA increases.

\paragraph{Denoising benchmark}

\begin{figure}[ht]
    \centering
    \begin{subfigure}{.28\textwidth}
        \centering
        \includegraphics[width=\textwidth]{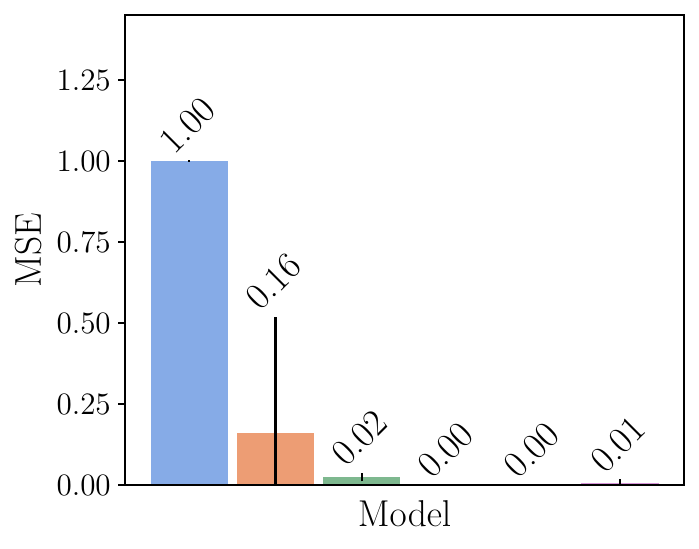}
        \caption{$N = 5$}
    \end{subfigure}
    \begin{subfigure}{.375\textwidth}
        \centering
        \includegraphics[width=\textwidth]{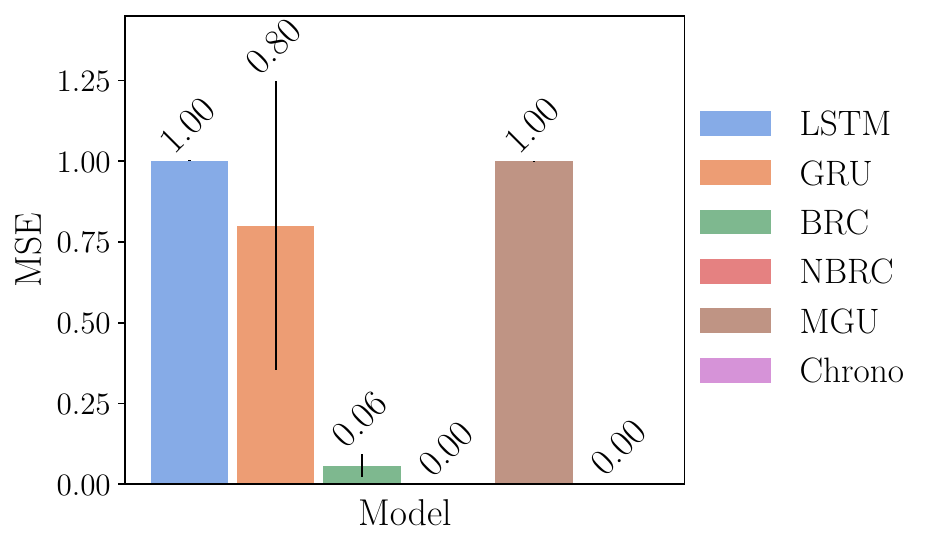}
        \caption{$N = 100$}
    \end{subfigure}
    \caption{%
        Test MSE loss for the denoising benchmark with different forgetting periods $N$ and $T = 200$.
        Mean and standard deviation are reported after 50 epochs.}
    \label{table:vaa_denoising}
\end{figure}

\begin{figure}[ht]
    \centering
    \begin{subfigure}{.35\textwidth}
        \centering
        \includegraphics[width=\textwidth]{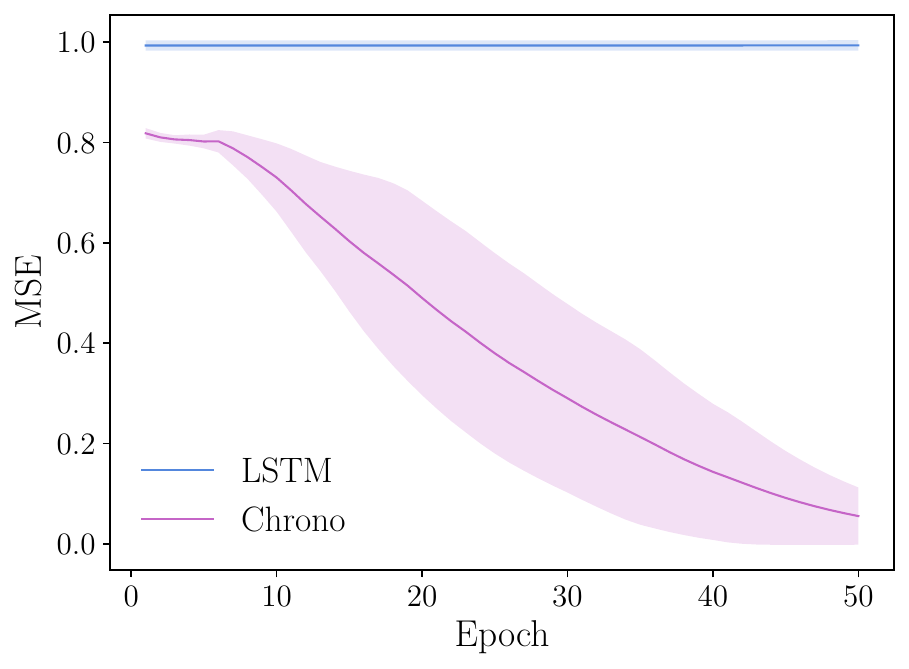}
        \label{fig:denoising_lstm_chrono_mse}
    \end{subfigure}
    \begin{subfigure}{.35\textwidth}
        \centering
        \includegraphics[width=\textwidth]{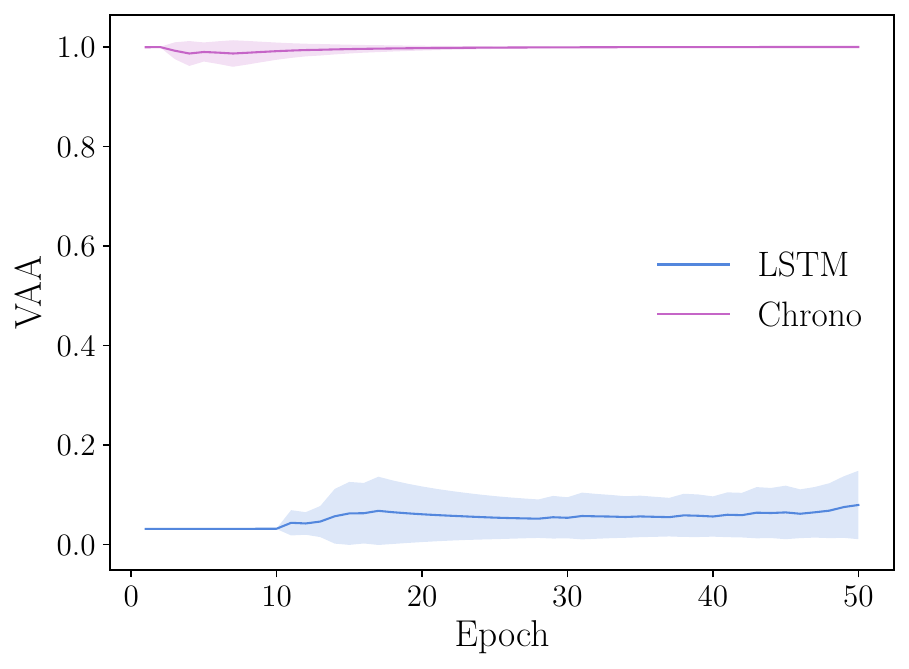}
        \label{fig:denoising_lstm_chrono_vaa}
    \end{subfigure}
    \caption{%
        Evolution of the validation loss (left) and of the VAA (right) of LSTM networks, with and without chrono initialisation, for the denoising benchmark with $N = 100$ and $T = 200$.
        Mean and standard deviation are reported after 50 epochs.}
    \label{fig:denoising_lstm_chrono}
\end{figure}

\begin{figure}[ht]
    \centering
    \begin{subfigure}{.35\textwidth}
        \centering
        \includegraphics[width=\textwidth]{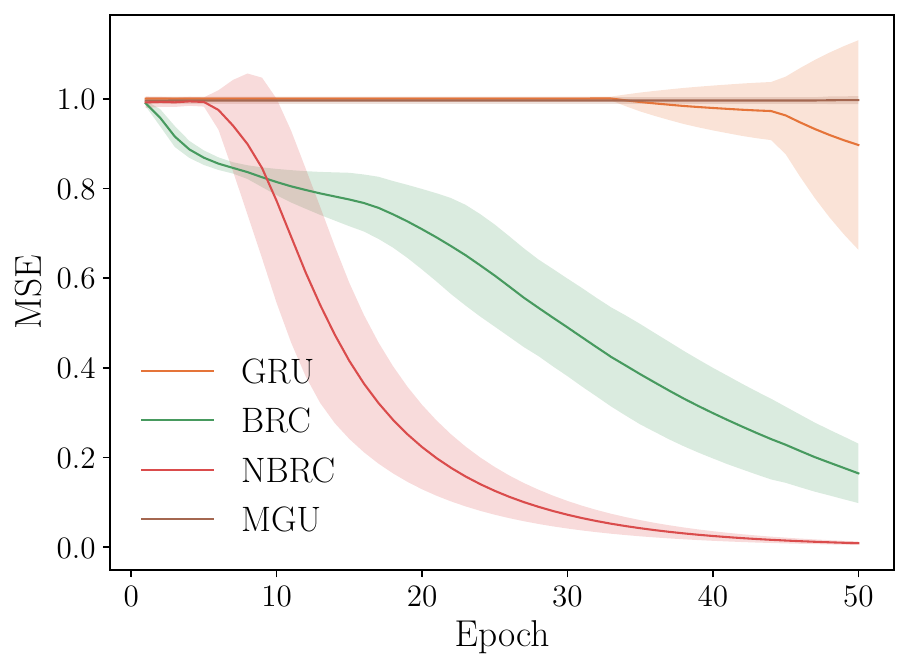}
        \label{fig:denoising_others_mse}
    \end{subfigure}
    \begin{subfigure}{.35\textwidth}
        \centering
        \includegraphics[width=\textwidth]{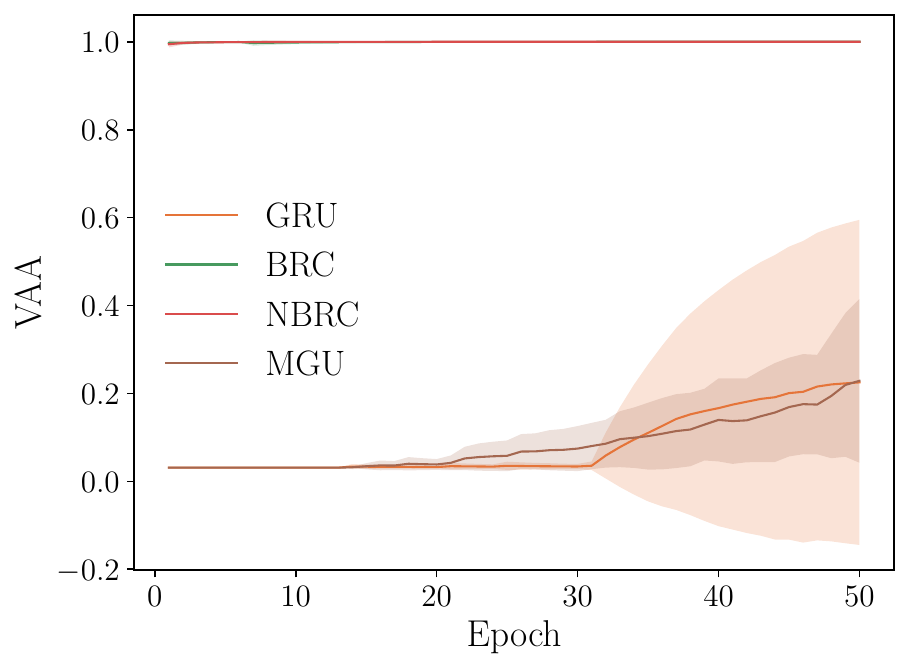}
        \label{fig:denoising_others_vaa}
    \end{subfigure}
    \caption{%
        Evolution of the validation loss (left) and of the VAA (right) of GRU, MGU, BRC and NBRC networks, for the denoising benchmark with $N = 100$ and $T = 200$.
        Mean and standard deviation are reported after 50 epochs.}
    \label{fig:denoising_others}
\end{figure}

\begin{figure}[ht]
    \centering
    \begin{subfigure}{.35\textwidth}
        \centering
        \includegraphics[width=\textwidth]{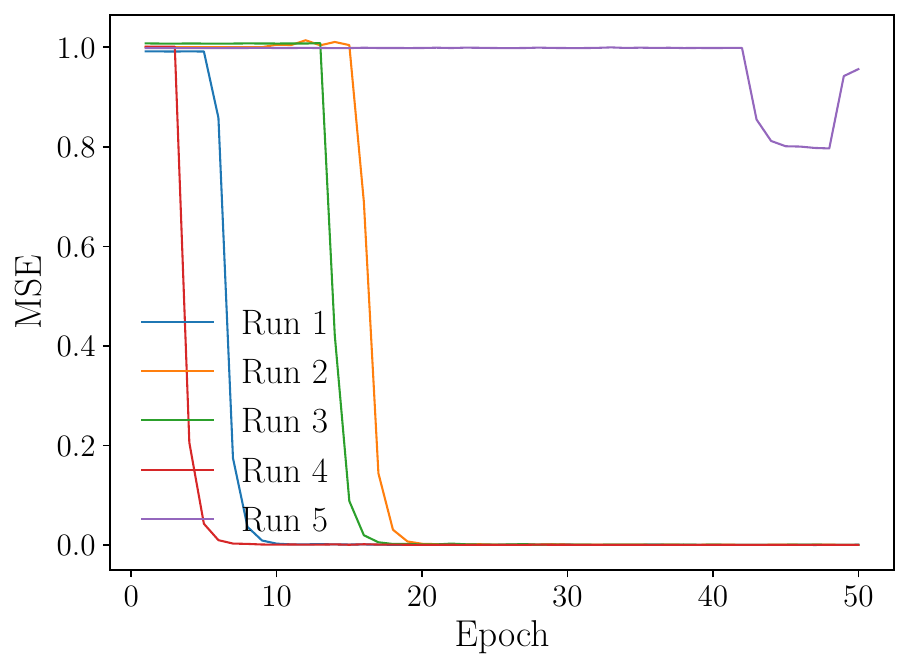}
        \label{fig:denoising_3gru_mse}
    \end{subfigure}
    \begin{subfigure}{.35\textwidth}
        \centering
        \includegraphics[width=\textwidth]{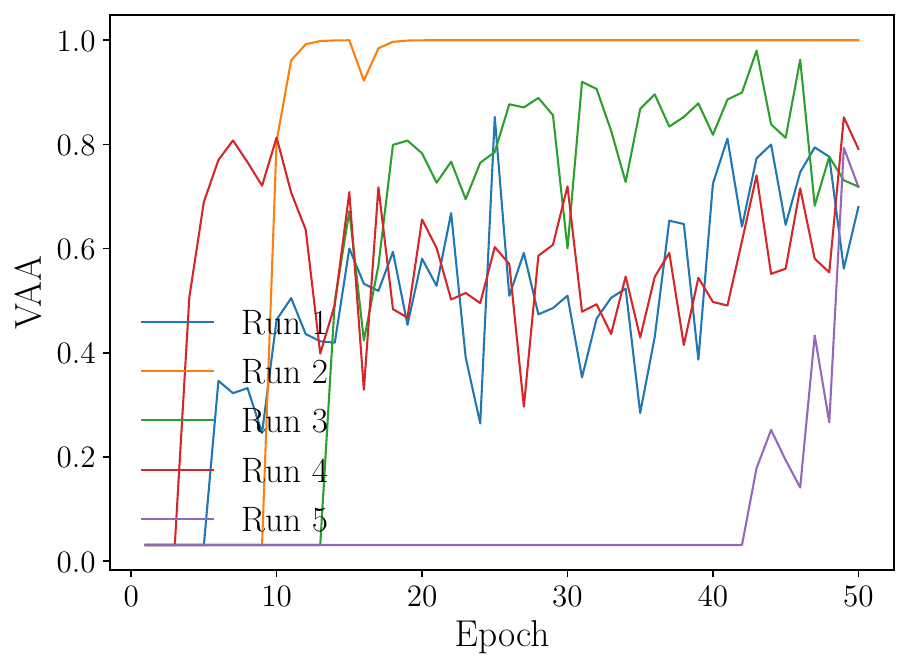}
        \label{fig:denoising_3gru_vaa}
    \end{subfigure}
    \caption{%
        Evolution of the validation loss (left) and of the VAA (right) of multiple GRU networks, for the denoising benchmark with $N = 5$ and $T = 200$.
        Mean and standard deviation are reported after 50 epochs.
        Loss decrease only start when the network becomes multistable (VAA greater than $\frac{1}{\abs{\X}}$).}
    \label{fig:denoising_3gru}
\end{figure}

\autoref{table:vaa_denoising} shows the performance of the different cells on this benchmark for different forgetting periods $N \in \{5, 100\}$.
Once again, the NBRC has the best performance, closely followed by the chrono-initialised LSTM.
On this benchmark, the BRC also reaches a very low loss.
Once again, we can see that all classical cells (LSTM, GRU, and MGU) generally fail in learning when longer time dependencies are present ($N = 100$).
\autoref{fig:denoising_lstm_chrono} shows the evolution of the VAA and the validation loss of multiple LSTM cells, with and without chrono initialisation, during the training on this benchmark.
As for the previous benchmark, only the chrono-initialised LSTMs have a high VAA and efficiently decrease their loss.
It can be noted that classically initialised LSTMs have a VAA close to zero throughout the learning on this harder benchmark.
\autoref{fig:denoising_others} shows these results for the GRU, BRC, NBRC and MGU cells.
It is observed that the GRU network has a very low VAA, and learning does not start before its VAA increases.
The MGU network does not manage to learn on this benchmark while its VAA only slowly increases at the end of the training procedure.
As far as the bistable networks (BRC and NBRC) are concerned, their VAA is directly maximised and learning starts directly, indicating that those indeed promote the learning of long time dependencies through multistability.
Finally, \autoref{fig:denoising_3gru} shows the validation loss and the VAA of five different trainings of the GRU cell on the denoising benchmark with $N = 5$.
It is clear that the GRU cell only starts decreasing its loss when its VAA has started increasing.
This proves once more the correlation between the VAA and the learning on long-term information restitution benchmarks.

\paragraph{T-Maze benchmark}

\begin{figure}[ht]
    \centering
    \begin{subfigure}{.35\textwidth}
        \centering
        \includegraphics[width=\textwidth]{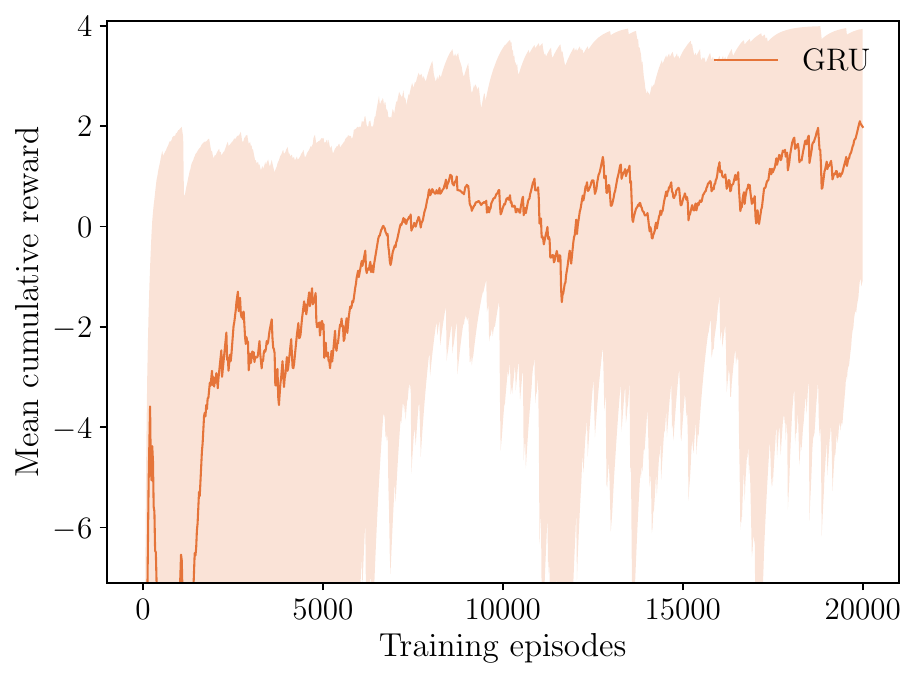}
        \label{fig:tmaze200_gru_reward}
    \end{subfigure}%
    \begin{subfigure}{.35\textwidth}
        \centering
        \includegraphics[width=\textwidth]{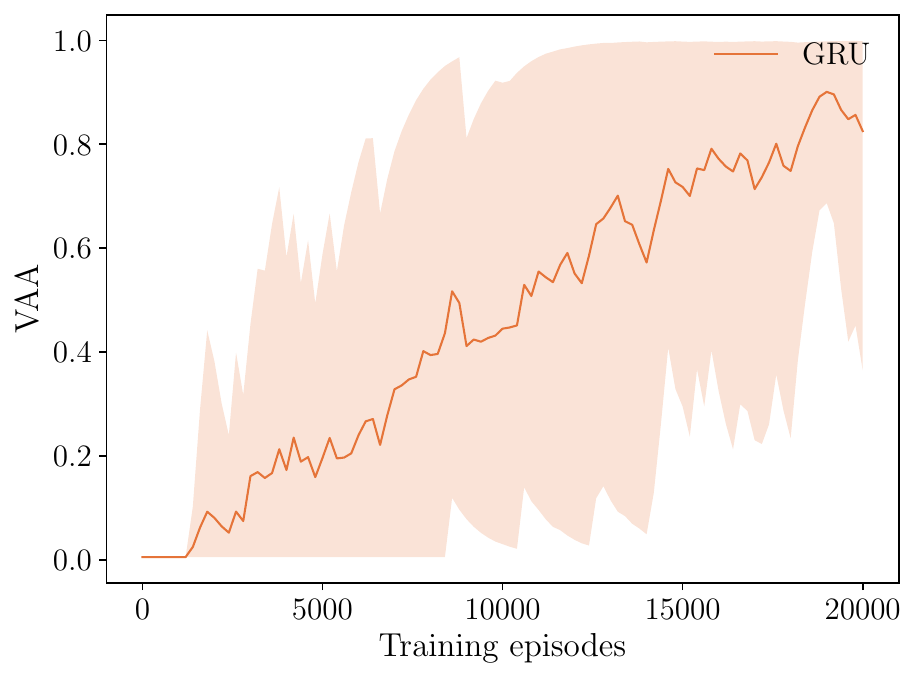}
        \label{fig:tmaze200_gru_vaa}
    \end{subfigure}
    \caption{%
        Evolution of the mean cumulative reward (left) and their VAA (right) obtained by GRU agents during DRQN training on a T-Maze of length 200.
        Mean and standard deviation are estimated over 3 training sessions.}
    \label{fig:tmaze200_gru}
\end{figure}

In this reinforcement learning setting, a policy is derived from the approximation of the \cqf{}.
The hyperparameters of the DRQN algorithm used for approximating the \cqf{} are given in \autoref{app:drqn}.
On the left in \autoref{fig:tmaze200_gru}, we can see the mean non-discounted cumulative reward obtained by the policies derived from GRU cells approximating the \cqf{}.
On the right in \autoref{fig:tmaze200_gru}, we can see the VAA of these cells estimated with \autoref{algo:vaa} using the histories of the replay buffer as input sequences.
Those value are clearly correlated.
Indeed, the better the agent plays, the higher its VAA is.

\section{Fostering multistability at initialisation}
\label{sec:fostering_multistability_at_initialisation}

In \autoref{sec:warming_up_rnns}, we describe the warmup initialisation procedure that allows one to maximise the estimated multistability of a network for a dataset of input sequences.
Then, in \autoref{sec:warmup_experiments}, we compare classic cells to warmed-up cells on information restitution, sequence classification, and RL benchmarks and show the benefits of the warmup in tasks with long time dependencies, when considering the same standard hyperparameters as in the previous section.
However, we also show that the warmup procedure does not improve the results in the sequence classification tasks.
In \autoref{sec:recurrent_double_layers}, we introduce the double-layer architecture, that has both multistable and transient dynamics.
We show that this architecture reaches a better performance both on information restitution and sequence classification benchmarks.
Finally, in \autoref{subsec:model_selection}, we show that the advantage of the warmup and the double-layer architecture, shown for standard hyperparameters in \autoref{sec:warmup_experiments} and \autoref{sec:recurrent_double_layers}, also holds when optimizing the hyperparameters for each cell version (number of recurrent layers $L$, number of hidden units $H$, batch size $B$, learning rate $\alpha$).

\subsection{Warming up RNNs}
\label{sec:warming_up_rnns}

The previous observations, that show a correlation between the multistability of a network and its ability to learn long time dependencies, suggest that fostering multistability could ease learning in this case.
In order to promote the multistability of a network, we propose maximizing the number of reachable attractors for hidden states resulting from the set of input sequences.
As for the estimation of the multistability, computing the VAA for all hidden states is not feasible because of its quadratic complexity.
In practice, we propose using stochastic gradient descend (SGD) to maximise the number of reachable attractors for batches of hidden states from different input sequences.
As for the estimation of the multistability, we sample a different stable input $\u \sim \N(\0, \1)$ for each batch of hidden states.
We note however that SGD cannot be used directly on the estimation of the proportion of reachable attractors detailed in \autoref{algo:vaa}, for two different reasons.
First, the VAA and the $\vaa_{M, \varepsilon}$ are not differentiable because of the Kronecker delta, which prevents from computing the gradient.
Second, it is likely that hidden states convergence is slow when several RNNs are stacked.
Indeed, the first layers must have reached stability for the following one to receive a stable input.

In order to solve the first problem, we introduce a differentiable proxy $\vaa^*_{M, \varepsilon}$ of the $\vaa_{M, \varepsilon}$.
Instead of the denominator
\begin{equation}
    C_{i, j} = \delta\left(\left\lVert f^M(\x_i, \u) - f^M(\x_j, \u) \right\lVert \leq \epsilon \right),
\end{equation}
that is equal to \num{1} when the final states after truncated convergence are close enough, we use
\begin{equation}
    C_{i, j}^* = 1 - \frac{\max(0, \left\lVert \tanh f^M(\x_i, \u) - \tanh f^M(\x_j, \u) \right\lVert - \varepsilon)}{\left\lVert \tanh f^M(\x_i, \u) - \tanh f^M(\x_j, \u) \right\lVert}.
\end{equation}
We note that the value of $C^*_{i, j}$ is strictly equal to \num{1} if $f^M(\x_i, u)$ is close enough in Euclidian distance to $f^M(\x_j, u)$.
On the other hand, $C^*_{i, j}$ will be close to \num{0} when they are far away.
We also note that $C^*_{i, j}$ will never be strictly equal to \num{0}, but will get closer as the distance increases, since the fraction tends towards \num{1}.
It can be noted that we are not interested in states being far apart from each other, but just in them being different.
However, we noticed in the experiments that this small bias provides a good direction for the gradient in order to reach multistability.
For this same reason, we need to apply a saturating function (hyperbolic tangent in this case) to the states in order to avoid extreme states when maximizing \vaastar.
The resulting differentiable proxy of the \vaa{} is given by
\begin{equation}
    \vaa^*_{M, \varepsilon}(f, \X, \u) =
    \frac{1}{\abs{\X}}\sum^{\abs{\X}}_{i = 1} \dfrac{1}{\overset{\abs{\X}}{\underset{j = 1}{\sum}} \,
        1 - \frac{\max(0, \left\lVert \tanh f^M(\x_i, \u) - \tanh f^M(\x_j, \u)
            \right\lVert - \varepsilon)}{\left\lVert \tanh f^M(\x_i, \u) - \tanh
            f^M(\x_j, \u) \right\lVert}} .
    \label{eq:vaa_star}
\end{equation}

\begin{figure}[ht]
    \begin{subfigure}{.32\textwidth}
        \centering
        \includegraphics[width=\textwidth]{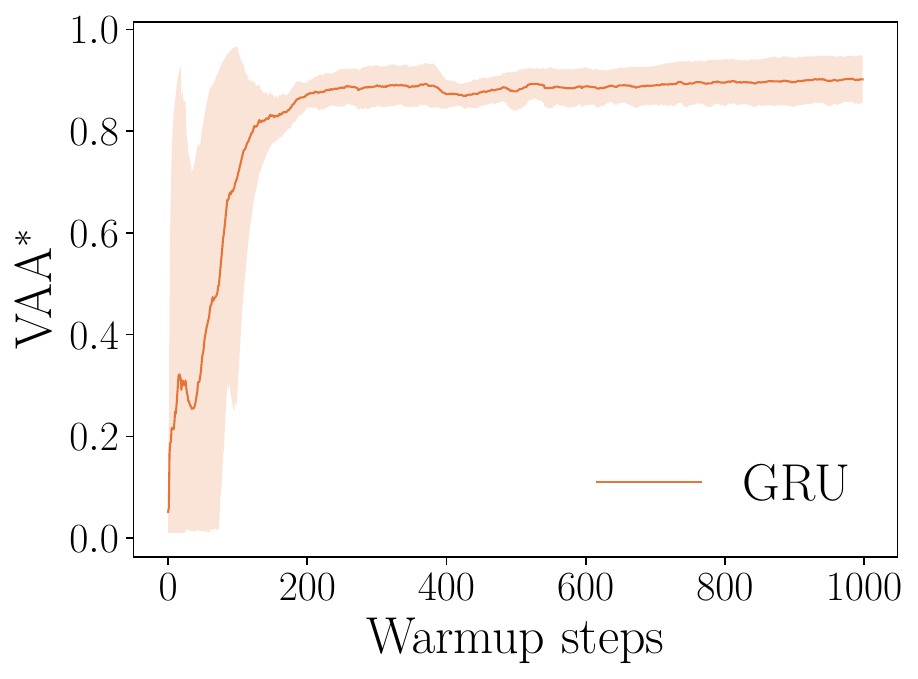}
        \label{fig:warmup_denoising_gru_layer0}
    \end{subfigure}
    \begin{subfigure}{.32\textwidth}
        \centering
        \includegraphics[width=\textwidth]{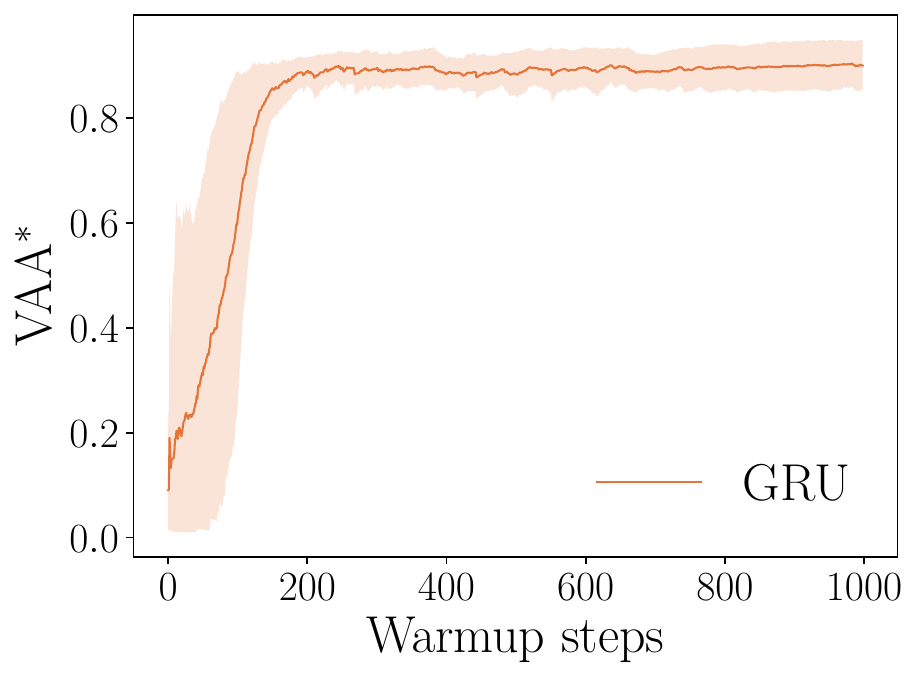}
        \label{fig:warmup_denoising_gru_layer1}
    \end{subfigure}
    \begin{subfigure}{.32\textwidth}
        \centering
        \includegraphics[width=\textwidth]{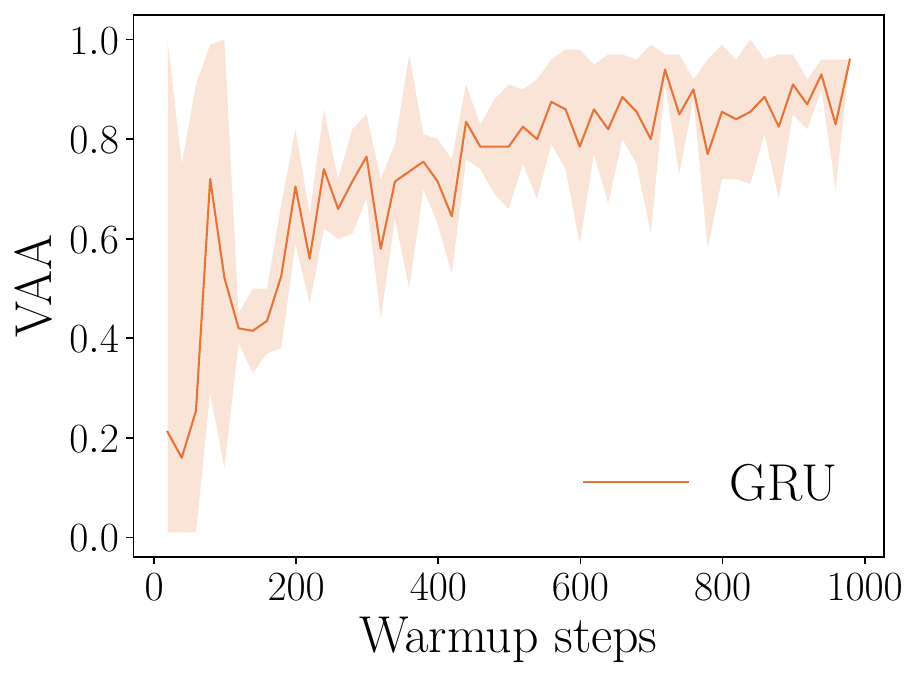}
        \label{fig:warmup_denoising_gru_network}
    \end{subfigure}
    \caption{Evolution of the \vaastar{} for a two-layer GRU (left and middle) and of the \vaa{} of the network (right) during warmup. This network is warmed up on the denoising dataset and results were averaged over three runs.}
    \label{fig:warmup_denoising_gru}
\end{figure}

For maximizing the multistability of an RNN for a given dataset of input sequences, we thus propose to maximise by SGD the \vaastar{} of batches of hidden states resulting from different input sequences, at random time steps.
For each batch of hidden states, a constant input perturbation is randomly sampled from $\u \sim \N(\0, \1)$ in order to stabilise the RNN hidden states over $M$ time steps.
However, as can be seen from equation~\eqref{eq:vaa_star}, maximising the \vaastar{} only occurs when all hidden states are infinitely distant, which is not desirable for learning efficiently.
In practice, we thus use SGD to get the \vaastar{} of each layer as close as possible to $k = 0.95$, as this proved empirically to maximise the number of attractors (see \autoref{fig:warmup_denoising_gru}) while avoiding too extreme states that could arise from the approximation of the $\vaa$ with $C^*$.
In \autoref{app:impact_k}, we show on the copy first input benchmark with $T \in \{50, 300, 600\}$ that the warmup procedure improves learning for a wide range of $k$.
It shows the robustness of our findings with respect to some hyperparameter variation.
The loss used is thus given by
\begin{equation}
    \L(\v, k) = \frac{1}{L} \sum_{i=1}^{L}(v_i-k)^2 \quad
\end{equation}
where $v_i = \vaa^*_{M, \varepsilon}(f^i, \X, \u)$ is the estimated multistability of layer $i$ and $L$ is the number of layers in the RNN.
Maximizing the \vaastar{} of each layer separately allows one to tackle the problem of layer convergence as identified above.
To avoid over-fitting problems, $M$ is sampled uniformly in $\{1, \dots, M_\text{max}(s)\}$ at gradient step $s$, where $M_\text{max}(s) = \min(M^*, 1 + c \cdot s)$ with $M^*$ the maximum stabilization period and $c$ the stabilization period increment.
This progressive increase is required for reaching multistability smoothly, avoiding gradients problems.
For the supervised learning tasks, the batches of input sequences are sampled in the training set.
For the reinforcement learning tasks, batches of input sequences are sampled from the exploration policy.
\autoref{algo:warmup} details the whole warmup procedure for a dataset $\D$ of input sequences.

\begin{algo}
    \capstart \DontPrintSemicolon \footnotesize
    \caption{Warming up an RNN}
    \label{algo:warmup}

    \SetKwInOut{Parameters}{Parameters}
    \SetKwInOut{Output}{Output}

    \Parameters{%
        $S \in \NN$ the number of gradient steps. \\
        $n \in \NN$ the batch size. \\
        $\alpha \in \RR^+$ the learning rate. \\
        $k \in [0, 1]$ the target average $\vaa^*_{M, \varepsilon}$. \\
        $M^* \in \NN$ the maximum stabilization period. \\
        $c \in \NN$ the stabilization period increment. \\
        $\varepsilon \in \RR^+$ tolerance when considering state similarity. \\
        $\theta \in \RR^{d_\theta}$ the parameters of the network. \\
        $L$ the number of layers in the RNN.
    }

    \KwData{%
        $\D = \{\u_{1:T_1}^1,\dots,\u_{1:T_N}^N\}$ a training set of $N$ input sequences.
    }

    \For{$s = 1, \dots, S$}{
        Sample a batch $\B$ of $n$ sequences in $\D$ without replacement $\B \sim \U^n(\D)$. \;

        Sample a random hidden state in each input sequence $\X \leftarrow \RandomHiddenStates(\B, \theta)$. \;

        Sample $M \sim \U(\{1, \dots, \min(M, 1 + s \cdot c)\})$. \;

        \For{$i = 1, \dots, L$}{
            Sample $\u \sim \mathcal{N}(\mathbf{0}, \mathbf{1})$. \;
            Set $v_i = \vaa^*_{M, \varepsilon}(f^i, \X, \u)$ where $f^i$ is the update function of the $i$\textsuperscript{th} RNN layer. \;
        }

        Compute loss $L \leftarrow \L(\v, k)$ where $\v = \begin{pmatrix}v_1 & \cdots & v_L\end{pmatrix}$. \;

        Compute gradient $g \leftarrow \nabla_\theta L$ with BPTT (over stabilization period and input sequence). \;

        Update parameters $\theta \leftarrow \theta - \alpha g$. \;

        Update maximum stabilization period $M^* \leftarrow M^* + c$. \;
    }
\end{algo}

We show in \autoref{fig:warmup_denoising_gru} that the warmup procedure effectively increases the \vaastar{} of each layer in an RNN.
Furthermore, we can also see on the right in \autoref{fig:warmup_denoising_gru} that as the warmup procedure is carried out, the true VAA measure of the RNN is increasing as well, even reaching \num{1} as the warmup procedure ends.

\subsection{Experiments}
\label{sec:warmup_experiments}

To demonstrate the impact of warming up RNNs on information restitution tasks, sequence classification tasks, and in partially observable RL environment, we tackle all benchmarks introduced in \autoref{sec:benchmarks}.
We train the LSTM, GRU and MGU cells with and without warmup and show that their performance is greatly improved with warmup.
As chrono-initialised LSTMs are known to work well, we also compare our results to such cells, with and without warmup.
The hyperparameters have been chosen to the same values as in previous section.
The goal here is not to measure the best performance of each architecture with or without warmup but rather to measure, for a given architecture and optimization procedure with fixed hyperparameters, whether the warmup initialisation procedure provides a better learning for different benchmarks.
In \autoref{app:generalisation_warmup}, we show that those results also hold for other hyperparameters for the permuted row sequential MNIST benchmark.
In addition, in the \autoref{subsec:model_selection}, we compare the performance of all cells with and without warmup with optimized hyperparameters.
All averages and standard deviations reported were computed over three different training sessions.
The optimal parameters for warming up can vary depending on architectures and needs, but we found $\alpha = 1e^{-2}$, $c=10$, $S = 100$, $n = 200$ and $M^*=200$ to be a good choice.

\paragraph{Copy first input benchmark}

\begin{figure}[ht]
    \begin{subfigure}{.32\textwidth}
        \centering
        \includegraphics[width=\textwidth]{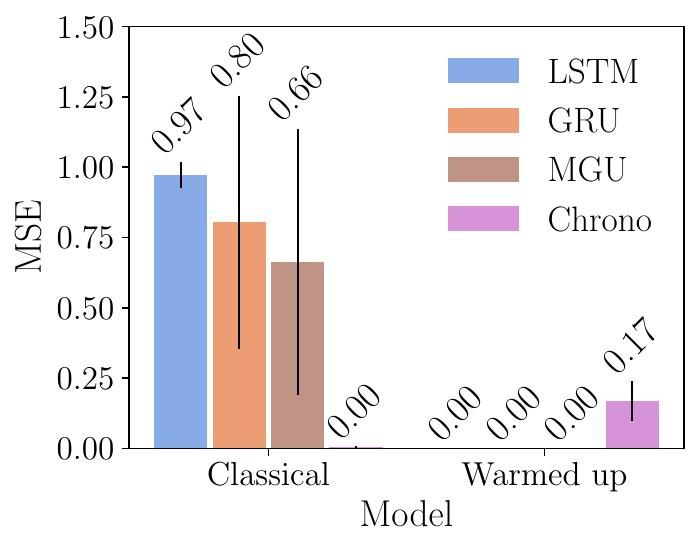}
        \caption{$T = 50$}
        \label{fig:bar_copy_50_nodouble}
    \end{subfigure}
    \begin{subfigure}{.32\textwidth}
        \centering
        \includegraphics[width=\textwidth]{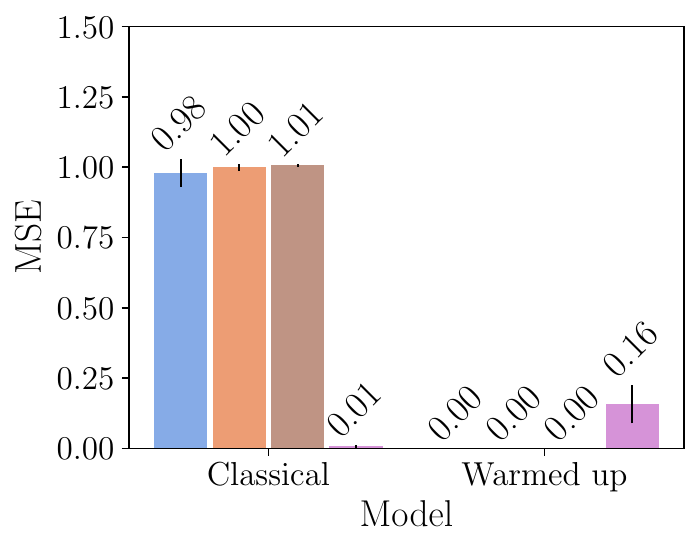}
        \caption{$T = 300$}
        \label{fig:bar_copy_300_nodouble}
    \end{subfigure}
    \begin{subfigure}{.32\textwidth}
        \centering
        \includegraphics[width=\textwidth]{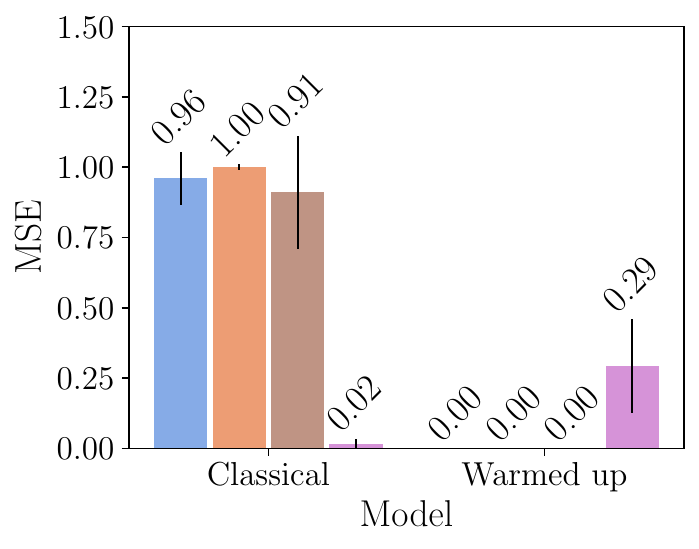}
        \caption{$T = 600$}
        \label{fig:bar_copy_600_nodouble}
    \end{subfigure}
    \caption{%
        Test MSE loss for the copy first input benchmark with different sequence lengths $T$.
        Mean and standard deviation are reported after 50 epochs.}
    \label{tab:warmup_bench1_nodouble}
\end{figure}

As can be seen from \autoref{tab:warmup_bench1_nodouble}, warming up RNNs greatly improves performances in the copy first input benchmark, for any sequence length $T \in \{50, 300, 600\}$.
Indeed, classically initialised RNNs have an average loss above \num{0.500} after \num{50} epochs, while all warmed-up RNNs have an average loss below \num{0.001} after \num{50} epochs.
On the other hand, the chrono-initialised LSTM performs better when it is not warmed up. Even if the chrono-initialisation and the warmup both promote the learning of long-term dependencies, combining them seems to have the opposite effect, leading to less performant model.

\paragraph{Denoising benchmark}

\begin{figure}[ht]
    \centering
    \begin{subfigure}{.32\textwidth}
        \centering
        \includegraphics[width=\textwidth]{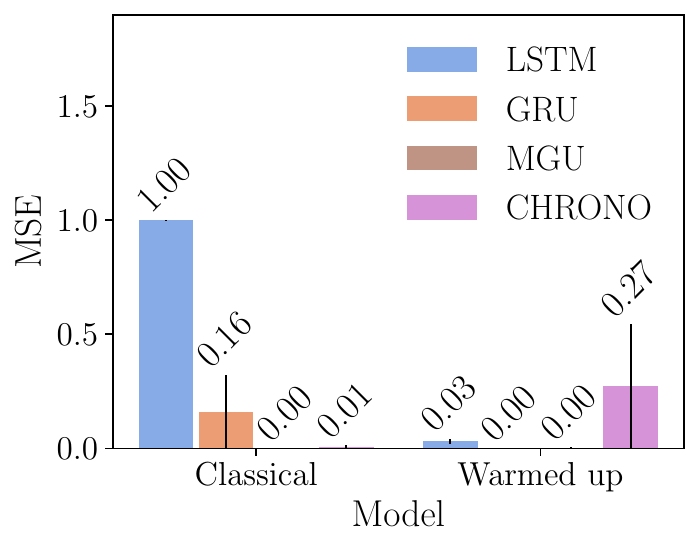}
        \caption{$N = 5$}
        \label{fig:bar_denoising_5_nodouble}
    \end{subfigure}
    \begin{subfigure}{.32\textwidth}
        \centering
        \includegraphics[width=\textwidth]{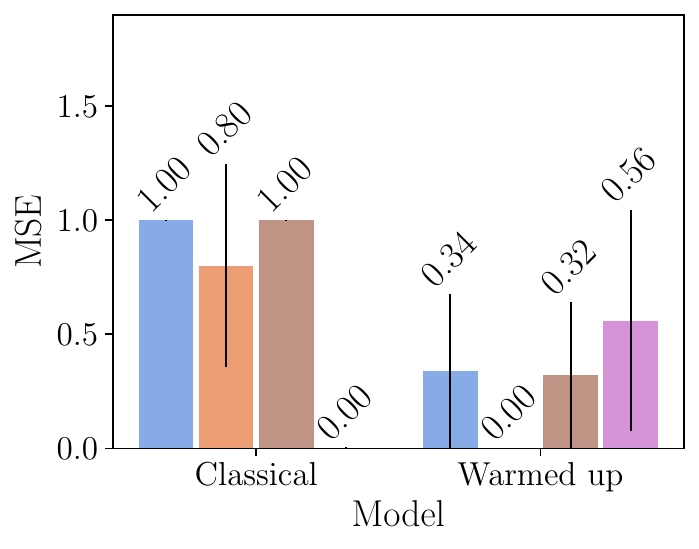}
        \caption{$N = 100$}
        \label{fig:bar_denoising_100_nodouble}
    \end{subfigure}
    \caption{%
        Test MSE loss for the denoising benchmark with different forgetting periods $N$ and $T = 200$.
        Mean and standard deviation are reported after 50 epochs.}
    \label{tab:warmup_bench2_nodouble}
\end{figure}

As far as the denoising benchmark is concerned, \autoref{tab:warmup_bench2_nodouble} shows that warmed-up cells always perform better than classically initialised ones, on sequences of length $T = 200$.
However, it can be noted that the average loss is still quite significant after \num{50} epochs for the LSTM and MGU cells, in the case of a forgetting period of $N = 100$.
As for the copy first input benchmark, the chrono-initialised cells perform worse when warmed-up which suggests once again that the chrono initialisation interacts disadvantageously with the warmup procedure.

\paragraph{T-Maze benchmark}

\begin{figure}[ht]
    \begin{subfigure}{.32\textwidth}
        \centering
        \includegraphics[width=\textwidth]{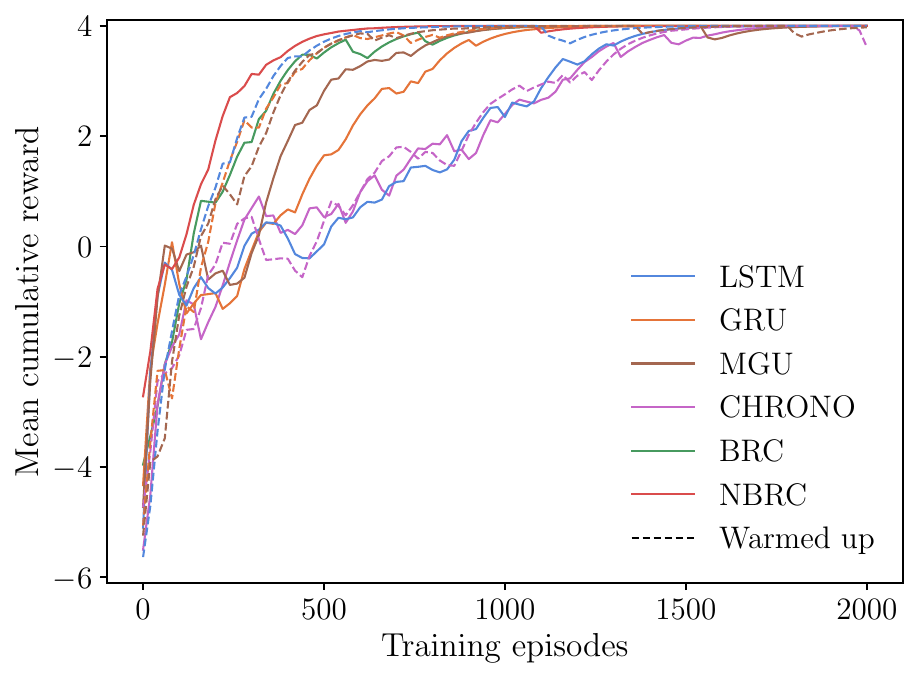}
        \label{fig:tmaze20_warmup_reward}
    \end{subfigure}
    \begin{subfigure}{.32\textwidth}
        \centering
        \includegraphics[width=\textwidth]{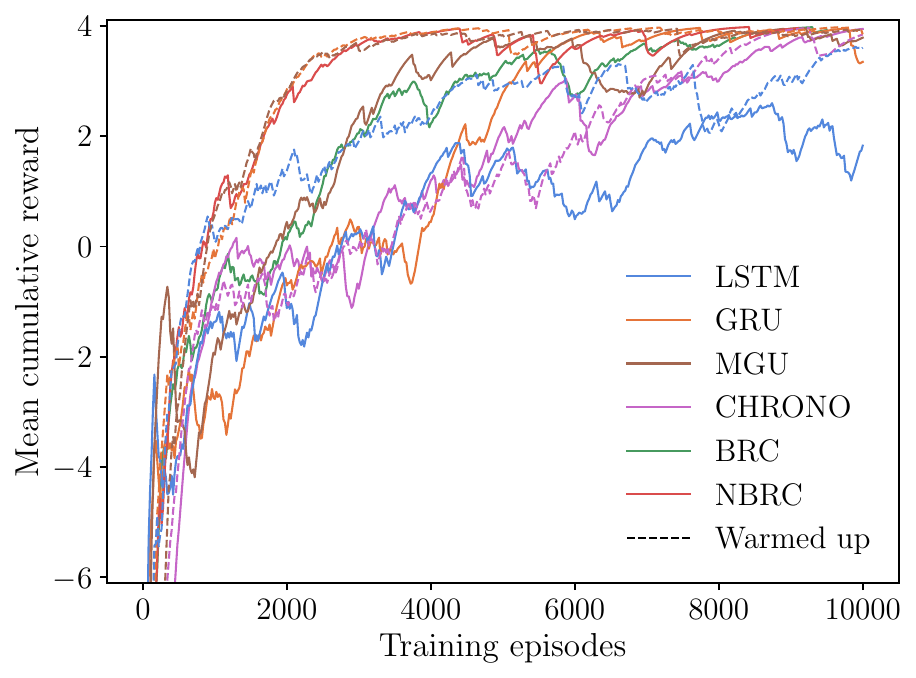}
        \label{fig:tmaze100_warmup_reward}
    \end{subfigure}
    \begin{subfigure}{.32\textwidth}
        \centering
        \includegraphics[width=\textwidth]{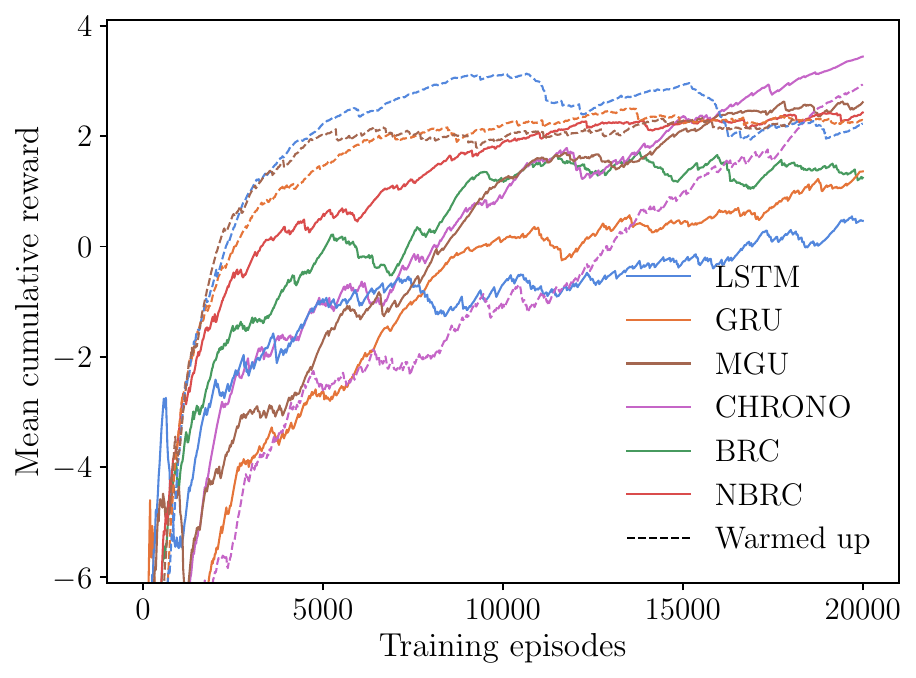}
        \label{fig:tmaze200_warmup_reward}
    \end{subfigure}
    \begin{subfigure}{.32\textwidth}
        \centering
        \includegraphics[width=\textwidth]{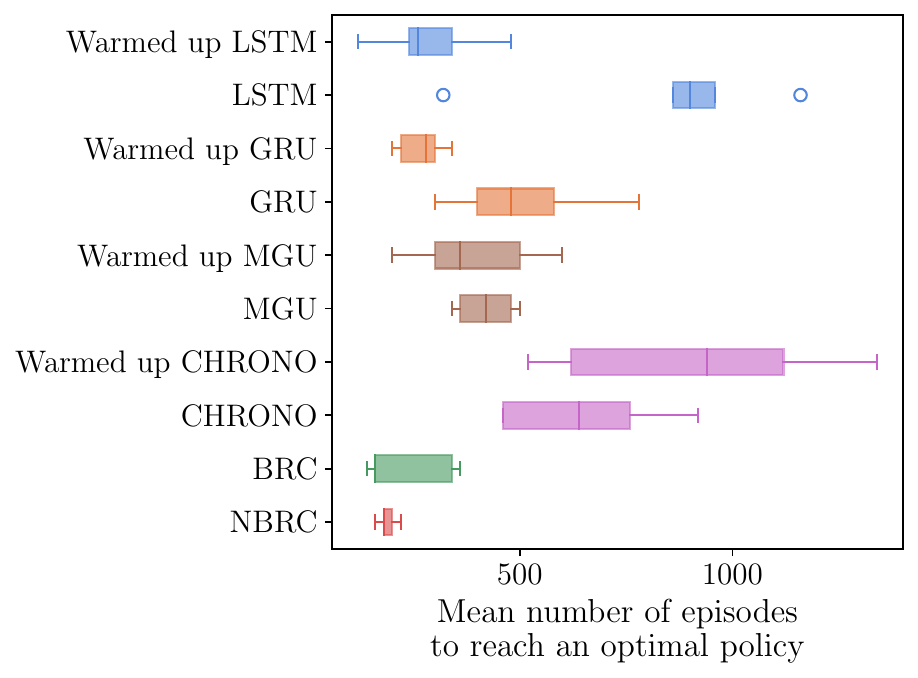}
        \label{fig:tmaze20_warmup_optimal}
    \end{subfigure}
    \begin{subfigure}{.32\textwidth}
        \centering
        \includegraphics[width=\textwidth]{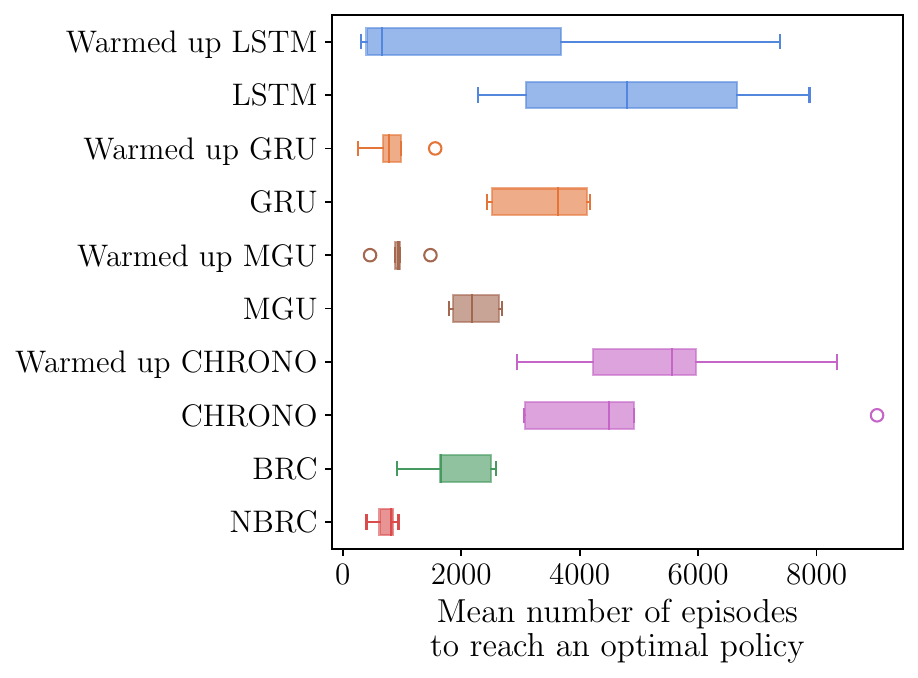}
        \label{fig:tmaze100_warmup_optimal}
    \end{subfigure}
    \begin{subfigure}{.32\textwidth}
        \centering
        \includegraphics[width=\textwidth]{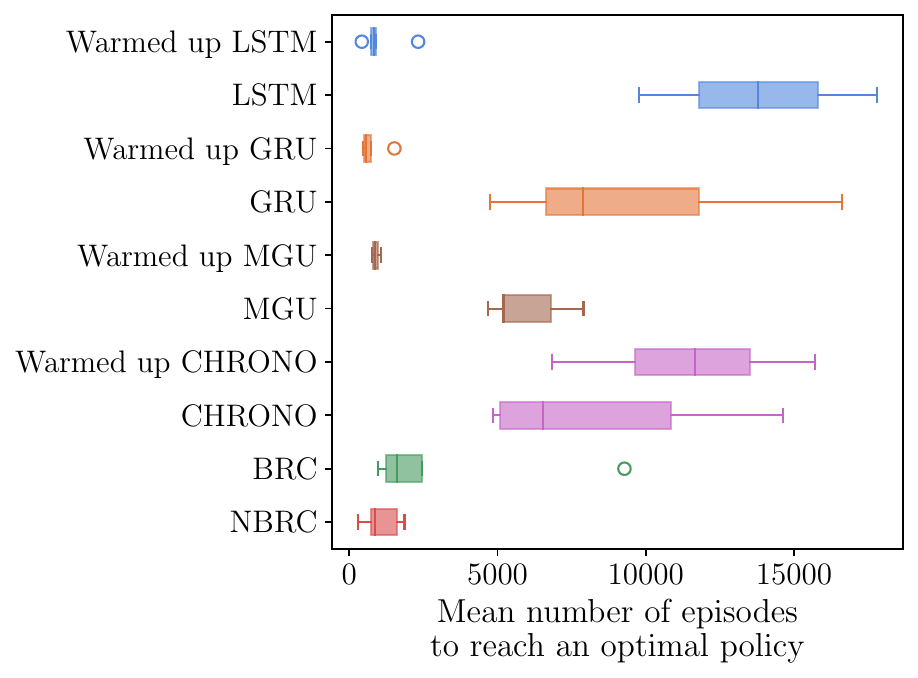}
        \label{fig:tmaze200_warmup_optimal}
    \end{subfigure}
    \caption{%
        Evolution of the mean cumulative reward obtained by warmed-up and classic agents during their training (up) and mean number of episodes required to reach the optimal policy (down) on T-Mazes of length 20 (left), 100 (center) and 200 (right).}
    \label{fig:tmaze_warmup}
\end{figure}

On the left in \autoref{fig:tmaze_warmup}, we can see the evolution of the expected cumulative reward of the DRQN policy for the T-Maze environment as a function of the number of episodes of interaction.
It is more than clear that all warmed-up cells and bistable cells (\ie, BRC and NBRC), are better than the classically initialised ones on this RL benchmark.
As for the other benchmarks, the chrono-initialised LSTMs seem to interact disadvantageously with the warmup procedure.
In any case, it can be noted that the chrono-initialised LSTMs are always among the worse cells for this benchmark, with and without warmup.
Furthermore, we can see that warming up cells improves their performance even more as the length of the T-Maze increases, suggesting that the warmup procedure and the multistability of an RNN indeed help to tackle tasks with long time dependencies.
On the right in \autoref{fig:tmaze_warmup}, we can see the number of episodes required to reach the optimal policy for each cell.
It is clear that warming up a cell speeds up the convergence towards the optimal policy when time dependencies become large.
Indeed, for $L = 200$, all warmed-up cells reach the optimal policy before any classically initialised cell, except for the chrono-initialised LSTM.

\paragraph{Permuted sequential MNIST}

\begin{figure}[ht]
    \centering
    \includegraphics[width=.45\linewidth]{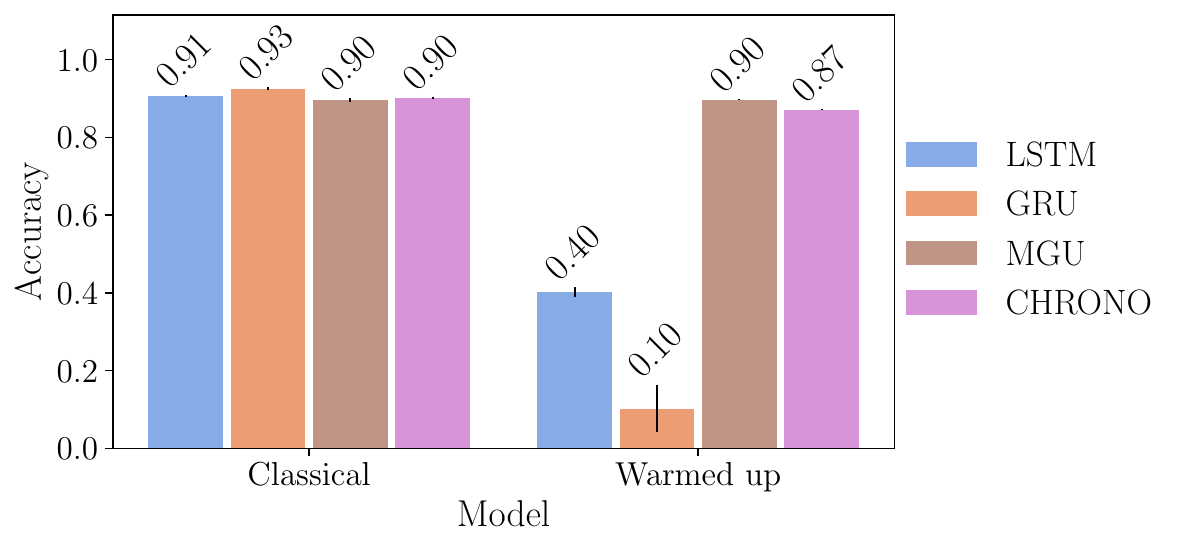}
    \caption{%
        Test accuracy for the permuted sequential MNIST benchmark.
        Mean and standard deviation are reported after 70 epochs.}
    \label{tab:warmup_bench3_nodouble}
\end{figure}

In \autoref{tab:warmup_bench3_nodouble}, we can see the test accuracies after \num{70} epochs on the permuted sequential MNIST benchmark.
It is clear that the warmup initialisation does not help in this task.
For the LSTM and GRU, the warmed-up cells are even worse than the classic cells.
This confirms that some tasks such as this sequence classification benchmark needs more transient dynamics instead of multistable ones.

\paragraph{Permuted line-sequential MNIST}

\begin{figure}[ht]
    \centering
    \begin{subfigure}{.42\linewidth}
        \centering
        \includegraphics[width=\textwidth]{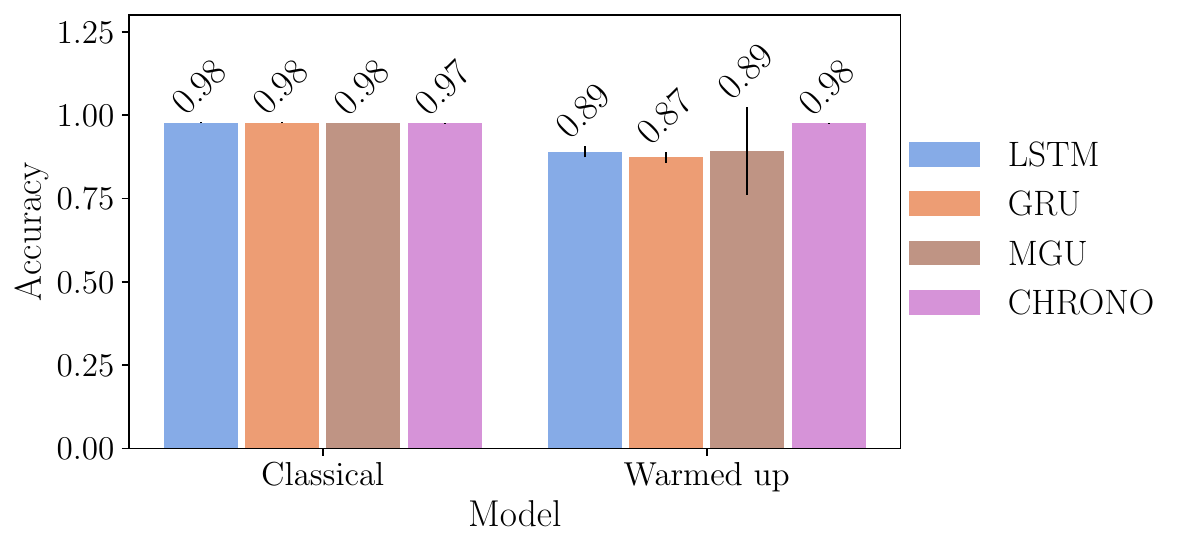}
        \caption{$N = 72$}
        \label{fig:bar_row_mnist_72_nodouble}
    \end{subfigure}
    \begin{subfigure}{.33\linewidth}
        \centering
        \includegraphics[width=\textwidth]{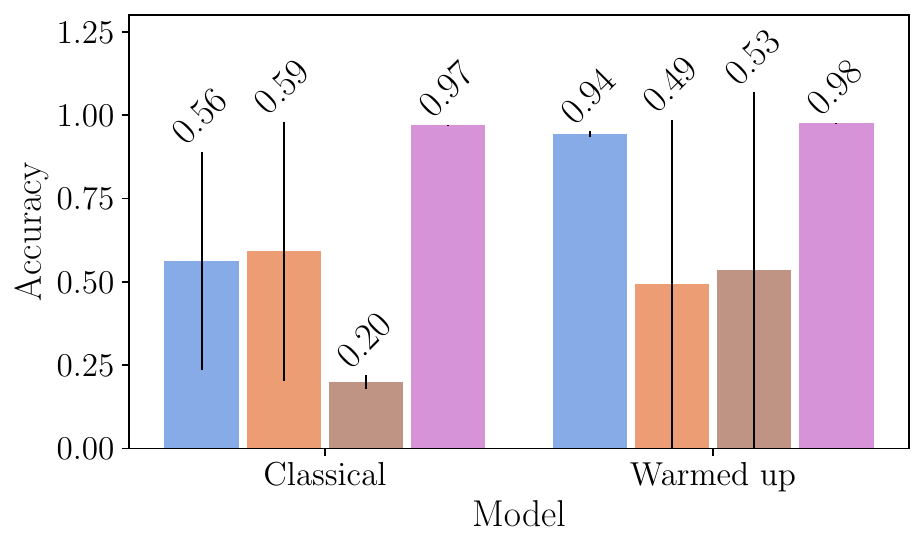}
        \caption{$N = 472$}
        \label{fig:bar_row_mnist_472_nodouble}
    \end{subfigure}
    \caption{%
        Test accuracy for the permuted line-sequential MNIST benchmark for different forgetting periods $N$.
        Mean and standard deviation are reported after 70 epochs.
        We note that when $N$ equals \num{72} (\num{472}) the resulting image has \num{100} (\num{500}) lines.}
    \label{tab:warmup_bench4_nodouble}
\end{figure}

In \autoref{tab:warmup_bench4_nodouble}, we can see the accuracies of each cell after \num{70} epochs on the test set of the permuted line-sequential MNIST benchmark.
For a sequence length of \num{100} (\ie, $N = 72$), it is clear that the classically initialised cells are better at this task.
As for the permuted sequential MNIST, this shows that transient dynamics are important for those sequence classification tasks, as opposed to information restitution tasks.

\subsection{Recurrent double-layers}
\label{sec:recurrent_double_layers}

As shown in the previous section and mentioned in the literature \citep{sussillo2013opening}, the importance of the transient dynamics of RNNs should not be neglected for prediction.
Indeed, it is easy to see why transient dynamics can be of importance when trying to tackle a regression task.
If information is only stored in the form of attractors, then there can only be a limited number of states the network can take, making it very hard to get precise predictions.
We observe that when warming up neural networks they tend to lose predictive accuracy, at the benefit of easier training on longer sequences.
This leads one to think that RNNs should be built to have both rich transient and multistable dynamics.
We thus propose using a double-layer architecture that allows one to get precise predictions while maintaining the benefits of warmup.
We simply split each recurrent layer in two equal parts and only warmup one of them.
In this double architecture, the hidden states sizes are divided by two compared to the simple architecture, for a fairer comparison.
This allows to endow some part of each layer with multistability, while the other remains monostable with richer transient dynamics.
A double-layer structure is depicted in \autoref{fig:double_arch}.

\begin{figure}[ht]
    \centering
    \resizebox{0.75\textwidth}{!}{\begin{tikzpicture}

    \node[draw, rectangle, minimum width=2cm, minimum height=0.6cm, rounded corners] (input) at (0, 0) {Input};
    \node[draw, rectangle, minimum width=3cm, minimum height=1cm, rounded corners] (rnn1) at (4, 1) {RNN};
    \node[draw, rectangle, minimum width=3cm, minimum height=1cm, rounded corners] (wrnn1) at (4, -1) {Warmed-up RNN};
    \node[draw, rectangle, minimum width=2cm, minimum height=0.6cm, rounded corners] (concat) at (8, 0) {$\oplus$};
    \node[draw, rectangle, minimum width=3cm, minimum height=1cm, rounded corners] (rnn2) at (12, 1) {RNN};
    \node[draw, rectangle, minimum width=3cm, minimum height=1cm, rounded corners] (wrnn2) at (12, -1) {Warmed-up RNN};
    \node[draw, rectangle, minimum width=2cm, minimum height=0.6cm, rounded corners] (dots) at (16, 0) {$\dots$};

    \draw[-{Stealth}] (input.east) -- (rnn1.west) ;
    \draw[-{Stealth}] (input.east) -- (wrnn1.west) ;

    \draw[-{Stealth}] (rnn1.east) -- (concat.west);
    \draw[-{Stealth}] (wrnn1.east) -- (concat.west);

    \draw[-{Stealth}] (concat.east) -- (rnn2.west) ;
    \draw[-{Stealth}] (concat.east) -- (wrnn2.west) ;

    \draw[-{Stealth}] (rnn2.east) -- (dots.west);
    \draw[-{Stealth}] (wrnn2.east) -- (dots.west);

    \node[right = 0.30cm of rnn1.north] (loop1) {}; 
    \node[right = 0.30cm of wrnn1.north] (wloop1) {};
    \node[right = 0.30cm of rnn2.north] (loop2) {};
    \node[right = 0.30cm of wrnn2.north] (wloop2) {};

    \draw[-{Stealth}] (loop1) arc (-45:225:0.5);
    \draw[-{Stealth}] (wloop1) arc (-45:225:0.5);
    \draw[-{Stealth}] (loop2) arc (-45:225:0.5);
    \draw[-{Stealth}] (wloop2) arc (-45:225:0.5);

    \coordinate (A) at (1.5, -1.8);
    \coordinate (B) at (6.5, 2.6);
    \coordinate (C) at (9.5, -1.8);
    \coordinate (D) at (14.5, 2.6);

    \draw[dashed, fill=DarkBlue, opacity=0.2, rounded corners] (A) rectangle (B);
    \draw[dashed, fill=DarkBlue, opacity=0.2, rounded corners] (C) rectangle (D);

    \node[below left = 0cm of B, align=right] {Layer 1};
    \node[below left = 0cm of D, align=right] {Layer 2};

\end{tikzpicture}}
    \caption{Double layer architecture}
    \label{fig:double_arch}
\end{figure}

\begin{figure}[ht]
    \centering
    \begin{subfigure}{.49\textwidth}
        \centering
        \includegraphics[width=\textwidth]{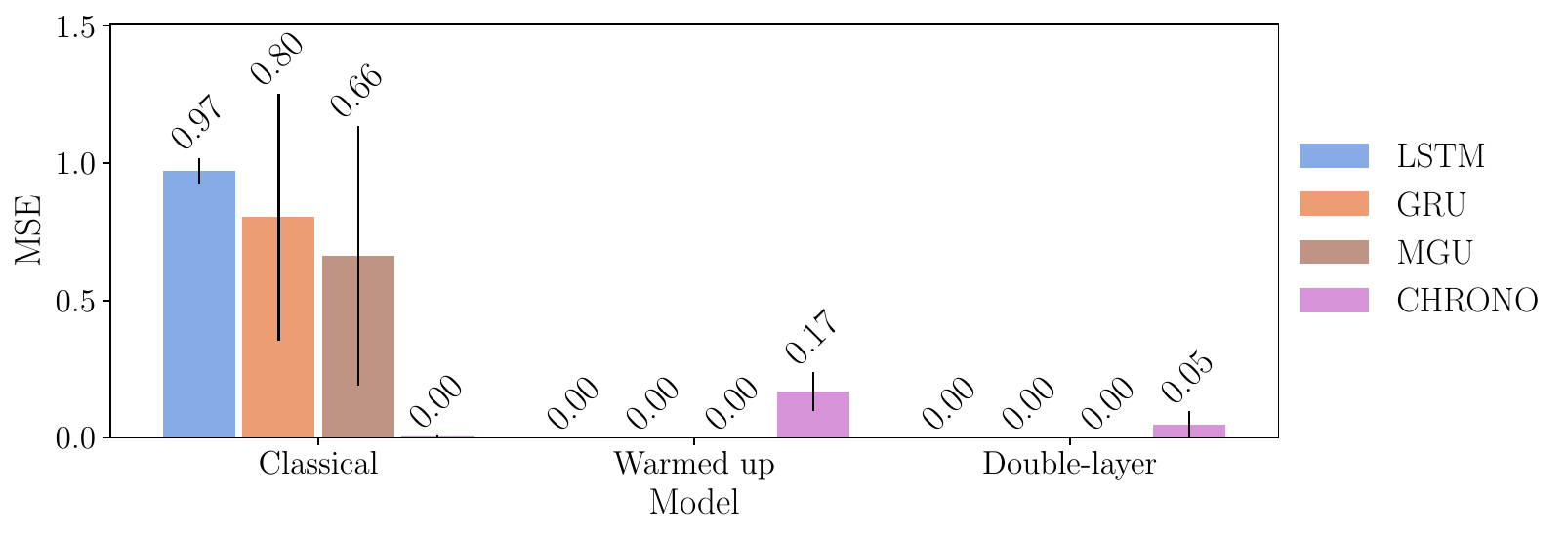}
        \caption{$T = 50$}
        \label{fig:bar_copy_50_warmup}
    \end{subfigure}
    \begin{subfigure}{.49\textwidth}
        \centering
        \includegraphics[width=\textwidth]{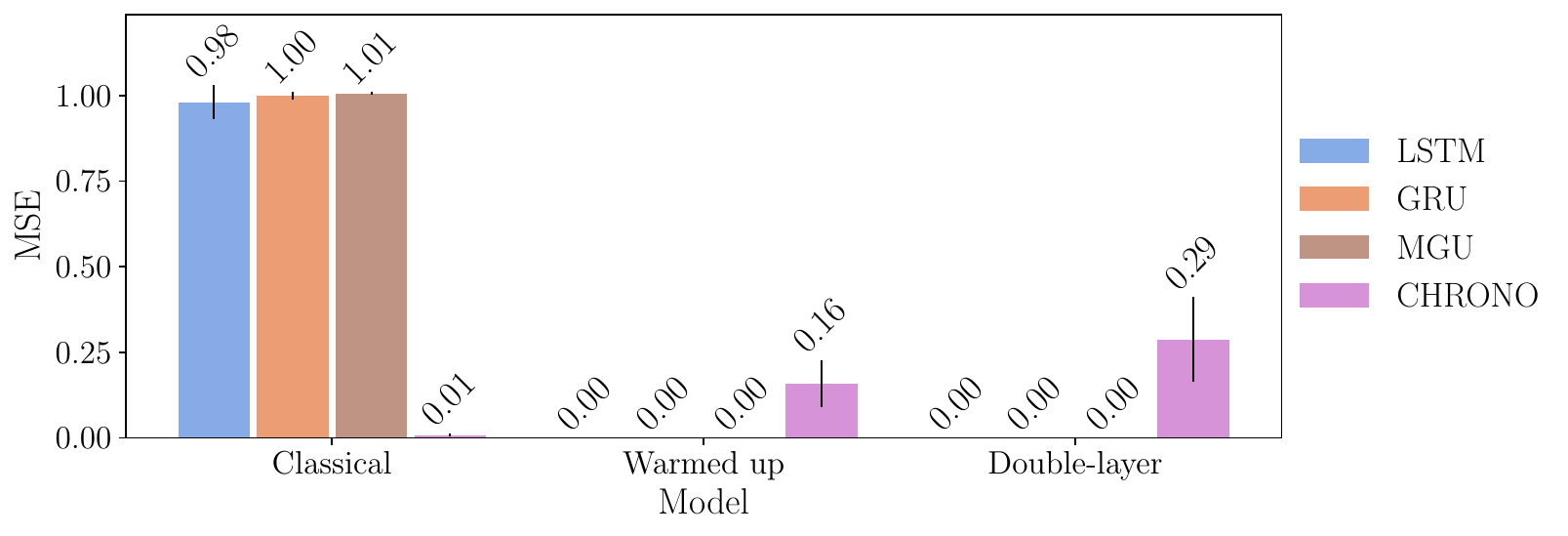}
        \caption{$T = 300$}
        \label{fig:bar_copy_300_warmup}
    \end{subfigure}
    \begin{subfigure}{.49\textwidth}
        \centering
        \includegraphics[width=\textwidth]{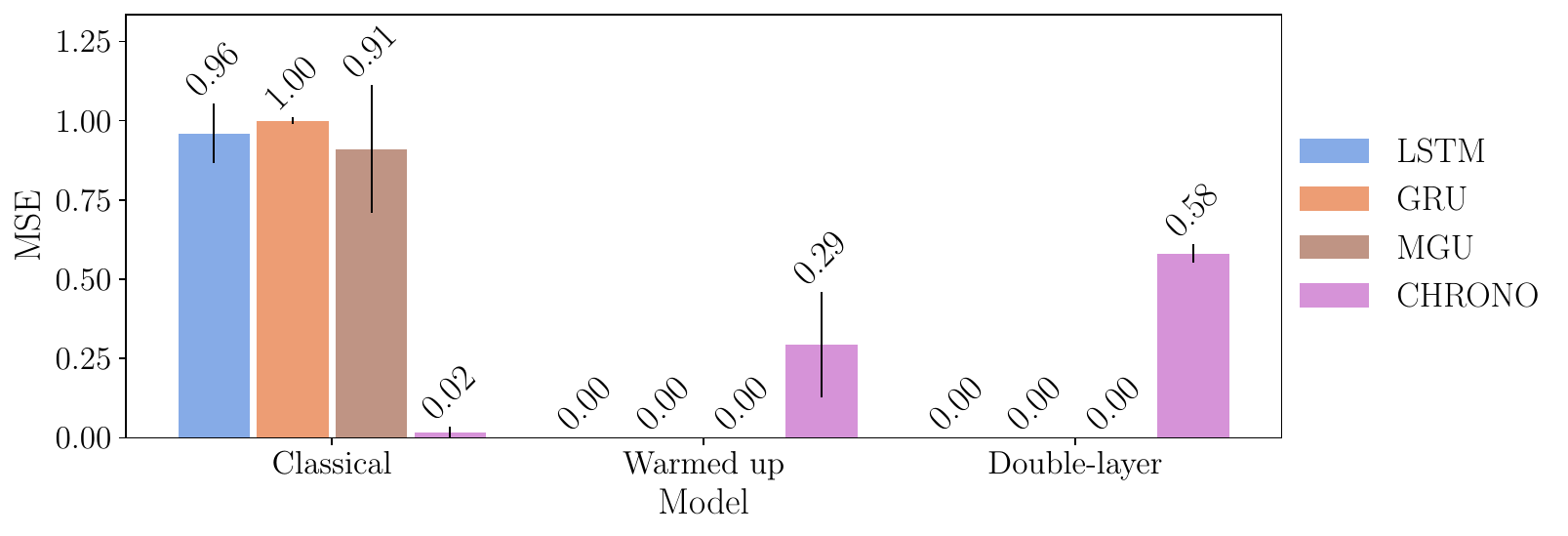}
        \caption{$T = 600$}
        \label{fig:bar_copy_600_warmup}
    \end{subfigure}
    \caption{%
        Test MSE loss for the copy first input benchmark with different sequence lengths $T$.
        Mean and standard deviation are reported after 50 epochs.}
    \label{tab:warmup_bench1}
\end{figure}

\begin{figure}[ht]
    \centering
    \begin{subfigure}{.49\textwidth}
        \centering
        \includegraphics[width=\textwidth]{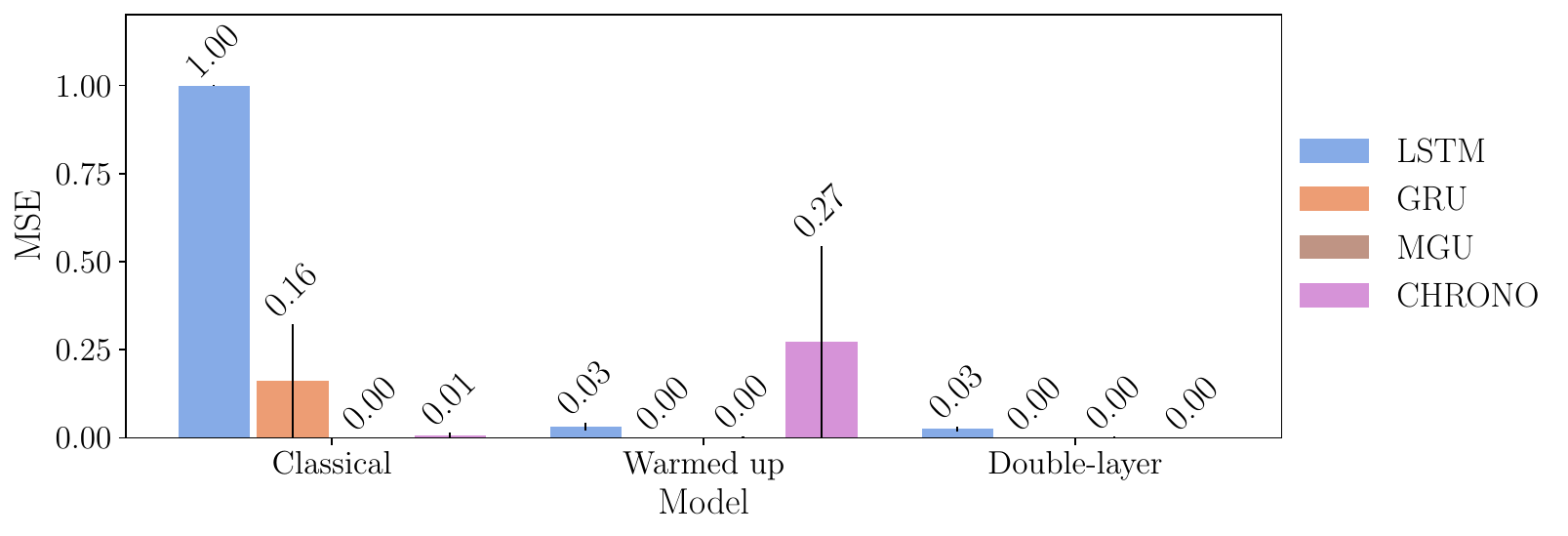}
        \caption{$N = 5$}
        \label{fig:bar_denoising_5_warmup}
    \end{subfigure}
    \begin{subfigure}{.49\textwidth}
        \centering
        \includegraphics[width=\textwidth]{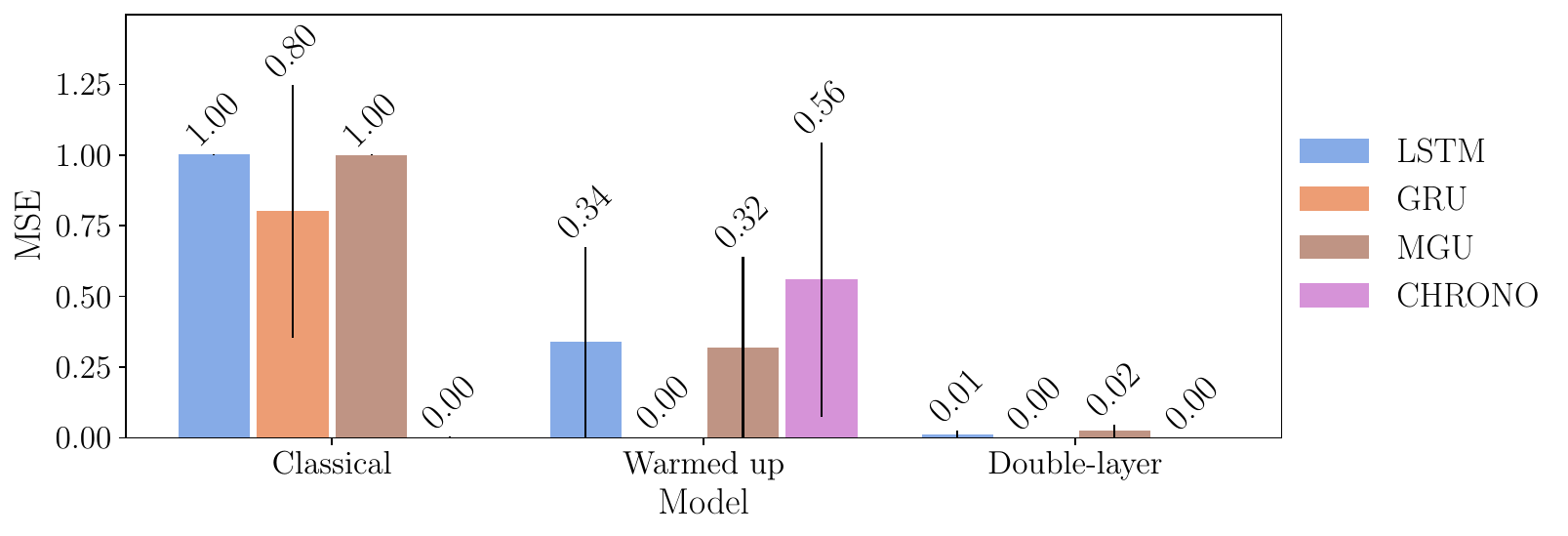}
        \caption{$N = 100$}
        \label{fig:bar_denoising_100_warmup}
    \end{subfigure}
    \caption{%
        Test MSE loss for the denoising benchmark with different forgetting periods $N$ and $T = 200$.
        Mean and standard deviation are reported after 50 epochs.}
    \label{tab:warmup_bench2}
\end{figure}

\begin{figure}[ht]
    \centering
    \includegraphics[width=.49\textwidth]{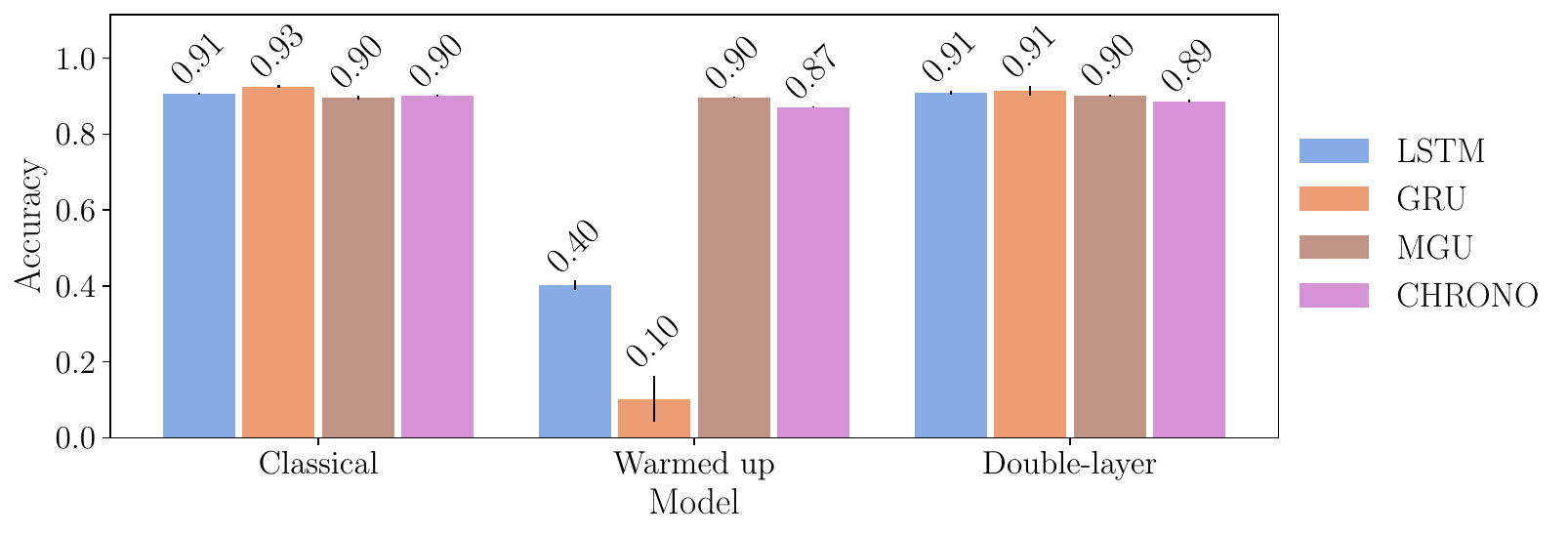}
    \caption{%
        Test accuracy for the permuted sequential MNIST benchmark.
        Mean and standard deviation are reported after 70 epochs.}
    \label{tab:warmup_bench3}
\end{figure}

\begin{figure}[p]
    \centering
    \begin{subfigure}{.49\textwidth}
        \centering
        \includegraphics[width=\textwidth]{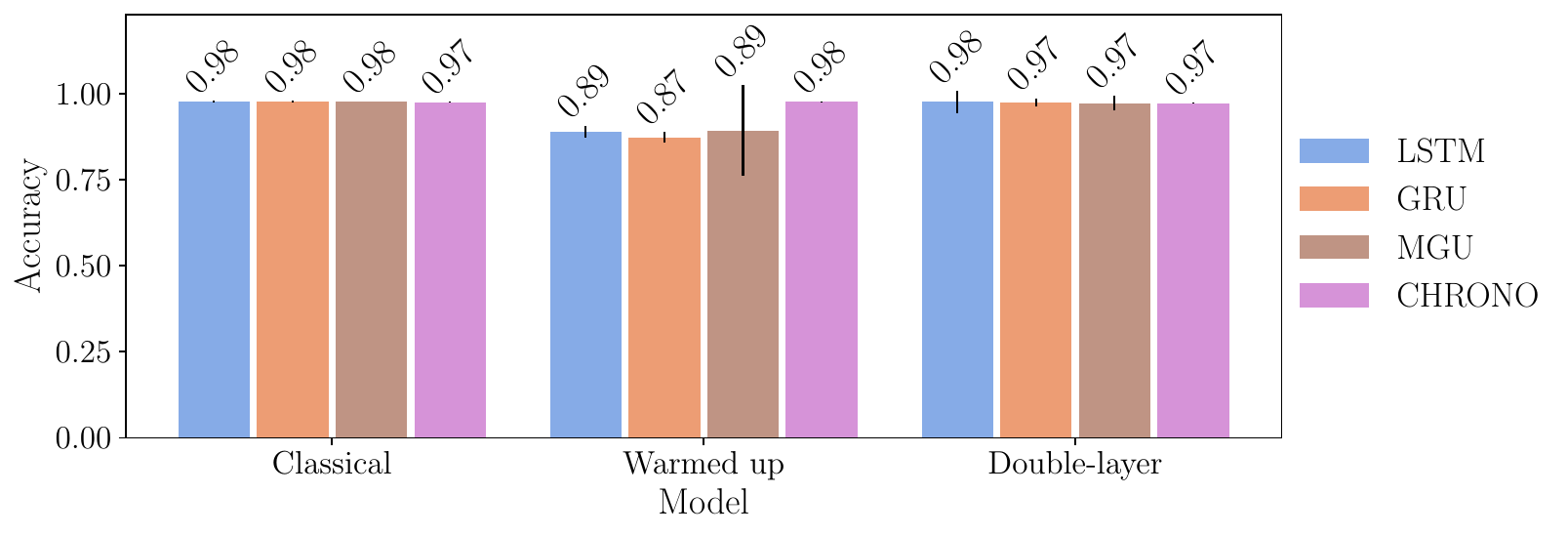}
        \caption{$N = 72$}
        \label{fig:bar_row_mnist_72_warmup}
    \end{subfigure}
    \begin{subfigure}{.49\textwidth}
        \centering
        \includegraphics[width=\textwidth]{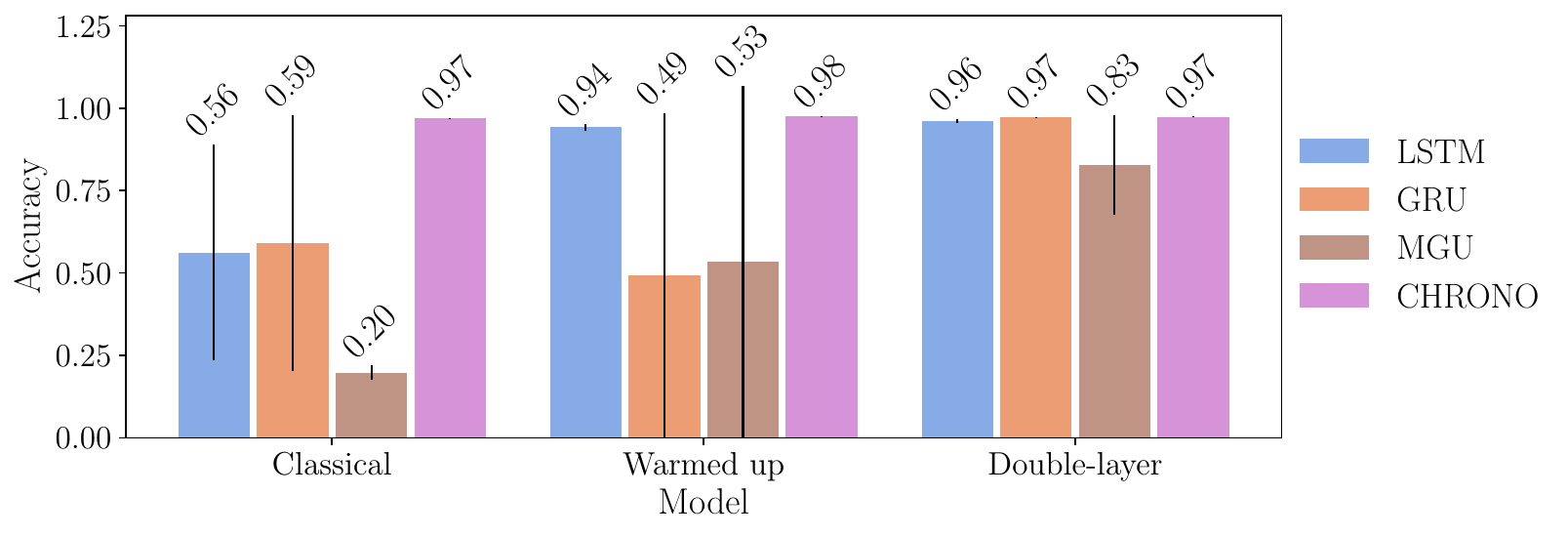}
        \caption{$N = 472$}
        \label{fig:bar_row_mnist_472_warmup}
    \end{subfigure}
    \caption{%
        Test accuracy for the permuted line-sequential MNIST benchmark for different forgetting periods $N$.
        Mean and standard deviation are reported after 70 epochs.
        We note that when $N$ equals \num{72} (\num{472}) the resulting image has \num{100} (\num{500}) lines.}
    \label{tab:warmup_bench4}
\end{figure}


\begin{figure}[p]
    \centering
    \begin{subfigure}{.35\textwidth}
        \centering
        \includegraphics[width=\textwidth]{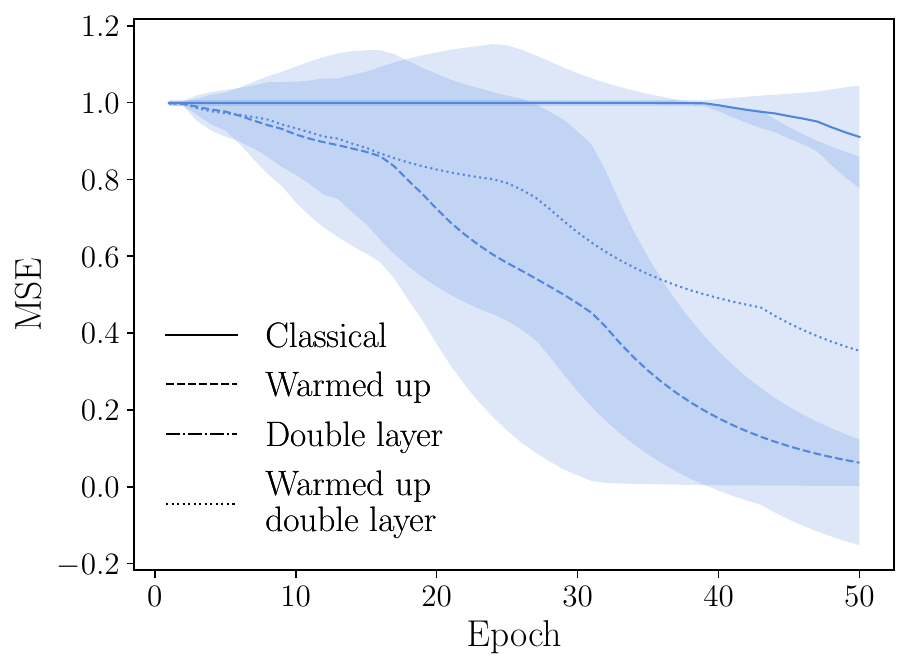}
        \label{fig:denoising_lstm2_double_warmup}
    \end{subfigure}
    \begin{subfigure}{.35\textwidth}
        \centering
        \includegraphics[width=\textwidth]{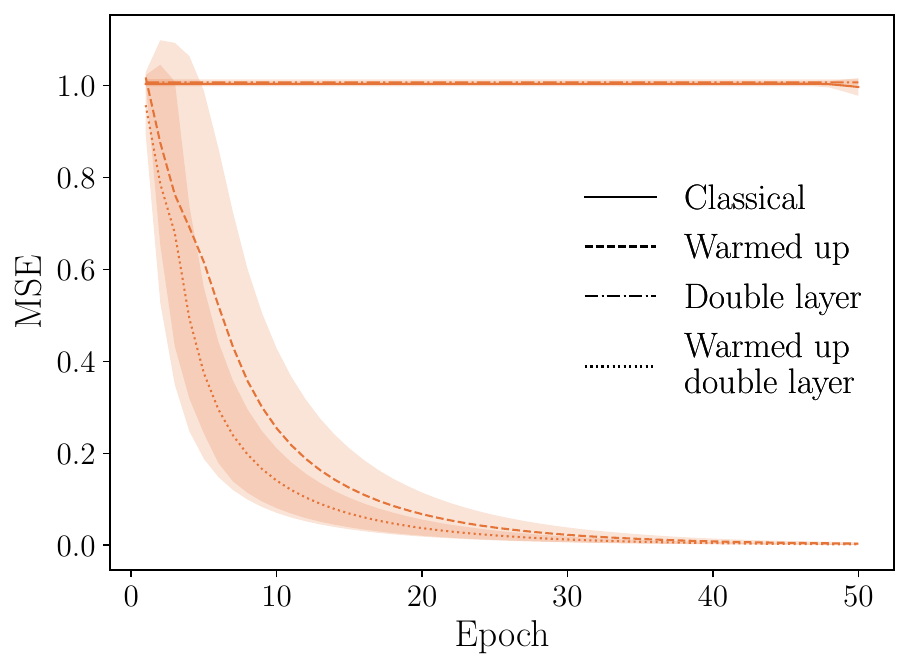}
        \label{fig:denoising_gru_double_warmup}
    \end{subfigure}
    \begin{subfigure}{.35\textwidth}
        \centering
        \includegraphics[width=\textwidth]{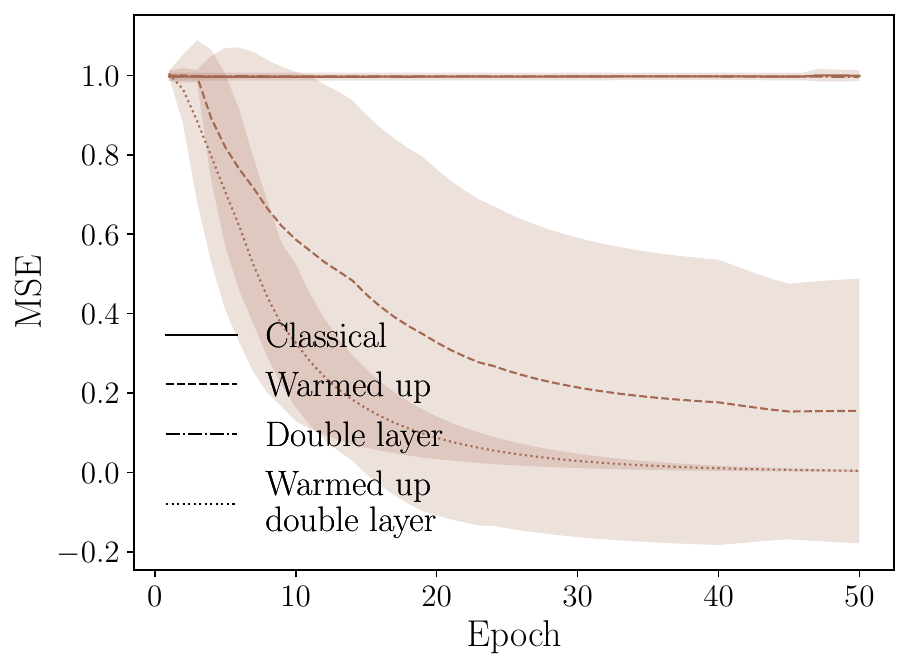}
        \label{fig:denoising_mgu_double_warmup}
    \end{subfigure}
    \caption{%
        Evolution of the validation loss on the denoising benchmark for LSTM, GRU and MGU networks, with $N = 100$ and $T = 200$.
        For each cell, four versions are considered: the classical one, the warmed-up one and the double-layer one, with and without partial warmup.}
    \label{fig:denoising_double_warmup}
\end{figure}

As can be seen from \autoref{tab:warmup_bench1}, \autoref{tab:warmup_bench2}, \autoref{tab:warmup_bench3} and \autoref{tab:warmup_bench4}, the double-layer architecture is always among the best performing architecture, for all four supervised learning benchmarks and for the LSTM, GRU and MGU cells, when using the same standard hyperparameters of the previous sections.
Even the chrono-initialised LSTMs perform well with the double-layer architecture except on the copy first input benchmark.
It shows that the double-layer architecture combines both the transient and multistable features of an RNN.
In addition, we can see in \autoref{tab:warmup_bench4} that the double-layer architecture is significantly better than the other architecture, for all types of cell, on the permuted line-sequential MNIST benchmark with a forgetting length of $N = 472$, a problem that requires both transient and multistable dynamics.
In addition, we show in \autoref{app:dlu_control} that the double-layer architectures without partial warmup generally perform worse that the classic architectures.
This ensures that the partial warmup is the most important factor for the performance of the double-layer architecture.
In \autoref{fig:denoising_double_warmup}, we can visualise the evolution of the validation loss averaged over 5 training sessions on the denoising benchmark for the LSTM, GRU and MGU cells, with the three architectures (\ie, classic, warmed up and double).
It is clear that the warmed-up and double-layer architectures are better.
Additionally, we can see that the double-layer architecture is significantly faster at learning this task for the GRU and MGU cells.

\subsection{Hyperparameter optimization} \label{subsec:model_selection}

In this section, we study the performance of the different cells in their different versions (\ie, classic, warmed up and double), when the hyperparameters are optimized.
In \autoref{sec:fostering_multistability_at_initialisation} and \autoref{app:generalisation}, we have shown that the warmup procedure and double-layer architecture provides a nice improvement in performance for a wide range of hyperparameters.
Here, we consider a more practical setting in which the hyperparameters of a considered cell version can be optimised according to the learning set.
We consider a standard hyperparameter selection procedure where the hyperparameters are selected according to the loss on a selection set, averaged over $5$ training sessions (see \autoref{app:hyperparameters} for details).
Those hyperparameters are then selected for $5$ training sessions according to the standard procedure, and the average loss on the test set is reported.
Due to the computational cost of such an optimization procedure, we only consider the most challenging benchmarks of each category, that is the denoising benchmark with $N = 100$ and the permuted line-sequential MNIST benchmark with $N = 472$.

\begin{figure}[p]
    \centering
    \begin{subfigure}{.49\textwidth}
        \centering
        \includegraphics[width=\textwidth]{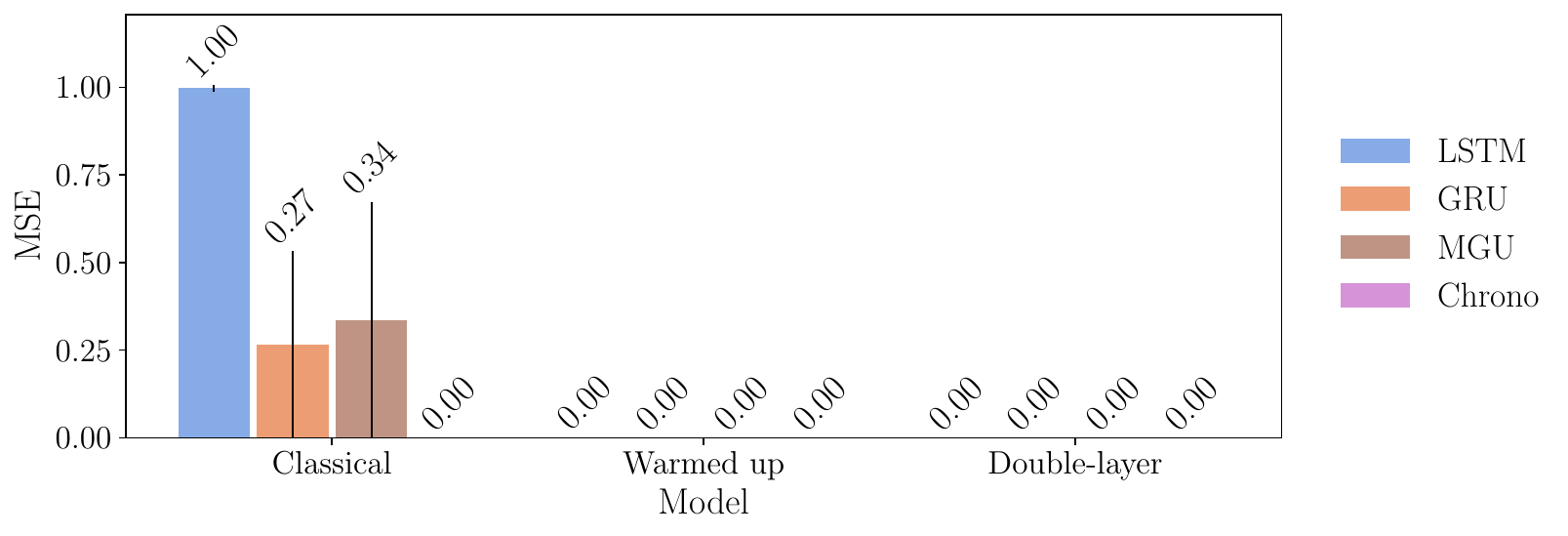}
        \caption{Denoising benchmark with $T = 200$ and $N = 100$.}
        \label{fig:ms_bar_denoising}
    \end{subfigure}
    \begin{subfigure}{.49\textwidth}
        \centering
        \includegraphics[width=\textwidth]{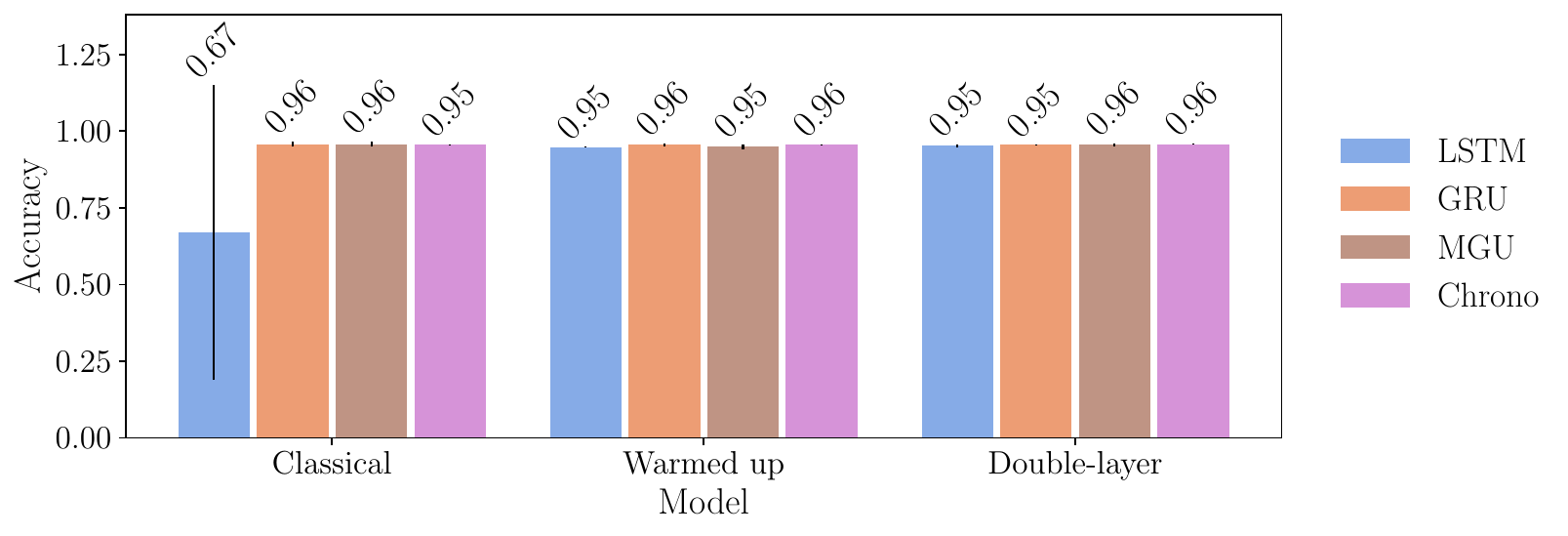}
        \caption{Permuted line-sequential MNIST with $N = 472$.}
        \label{fig:ms_bar_mnist}
    \end{subfigure}
    \caption{%
        Test accuracy for the denoising benchmark and the permuted line-sequential MNIST benchmark with hyperparameter selection on the learning set.
        Mean and standard deviation are reported after 50 epochs.}
    \label{tab:ms}
\end{figure}

\begin{figure}[p]
    \centering
    \begin{subfigure}{.35\textwidth}
        \centering
        \includegraphics[width=\textwidth]{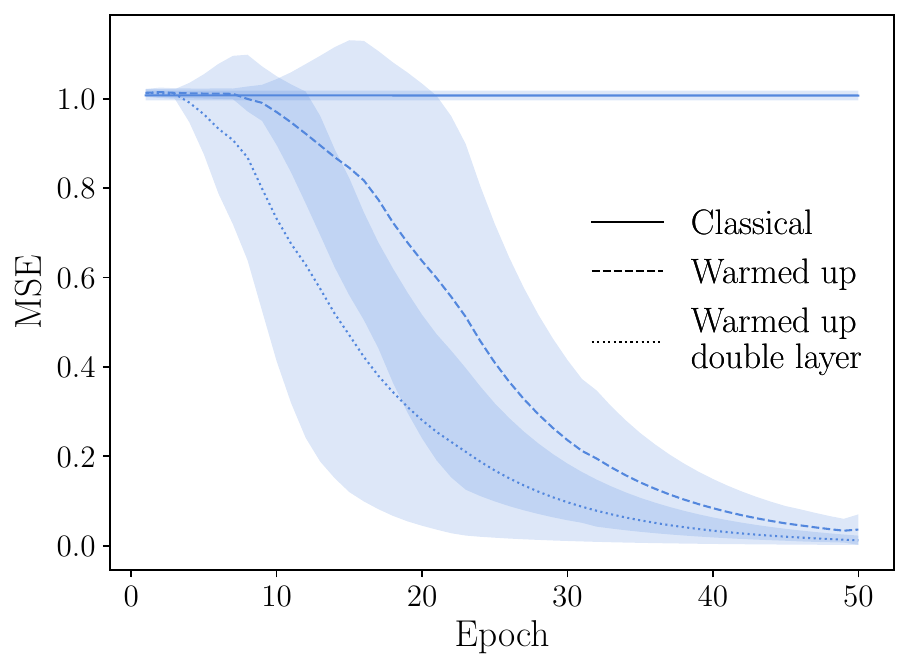}
    \end{subfigure}
    \begin{subfigure}{.35\textwidth}
        \centering
        \includegraphics[width=\textwidth]{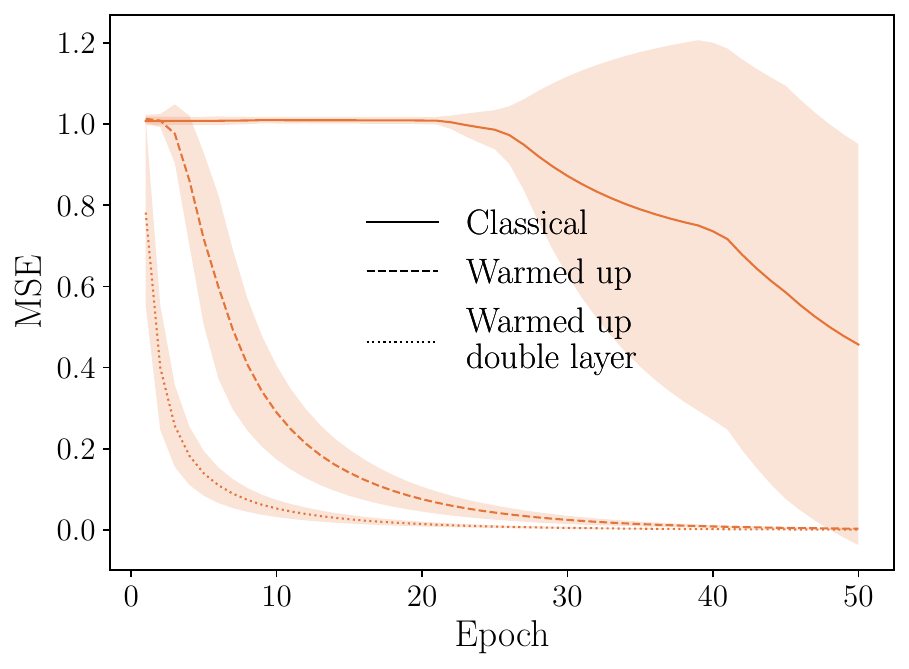}
    \end{subfigure}
    \begin{subfigure}{.35\textwidth}
        \centering
        \includegraphics[width=\textwidth]{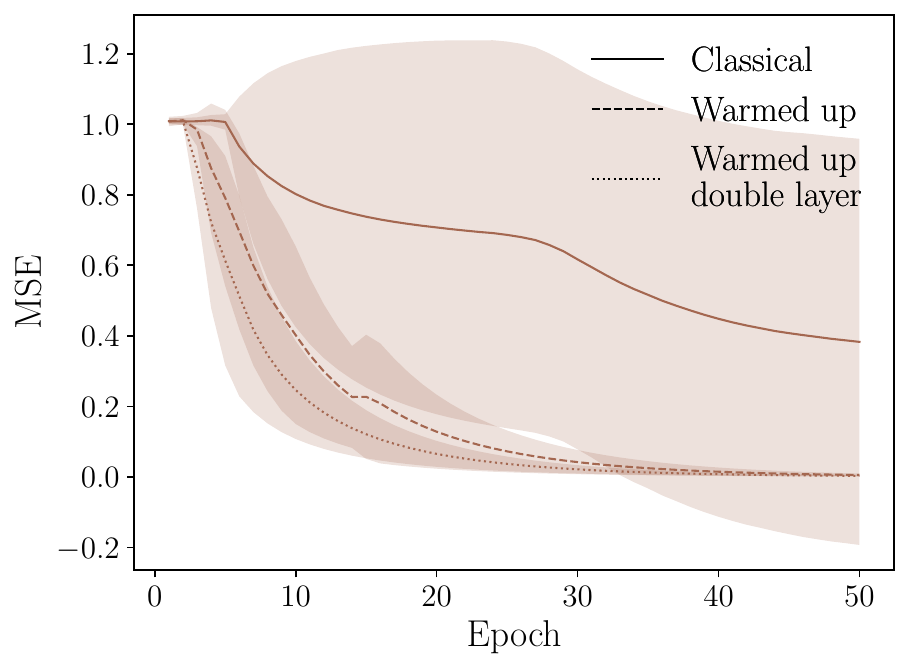}
    \end{subfigure}
    \caption{%
        Evolution of the validation loss on the denoising benchmark for LSTM, GRU and MGU networks, with $N = 100$ and $T = 200$.
        For each cell, three versions are considered: the classical one, the warmed-up one and the double-layer one with partial warmup.
        The hyperparameters of each cell version were optimized on the learning set.
        }
    \label{fig:hp_denoising}
\end{figure}

\begin{figure}[p]
    \centering
    \begin{subfigure}{.35\textwidth}
        \centering
        \includegraphics[width=\textwidth]{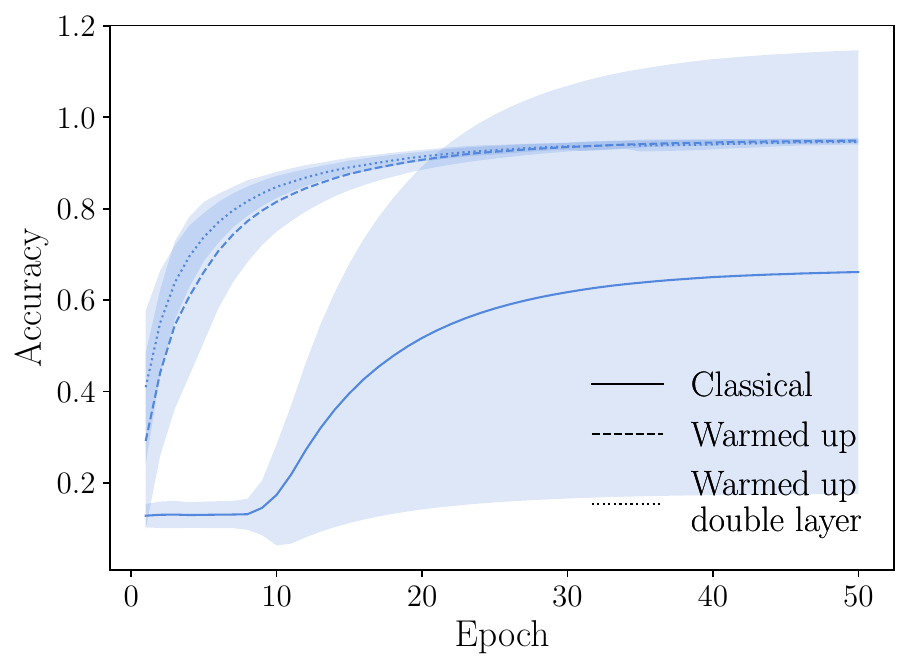}
    \end{subfigure}
    \begin{subfigure}{.35\textwidth}
        \centering
        \includegraphics[width=\textwidth]{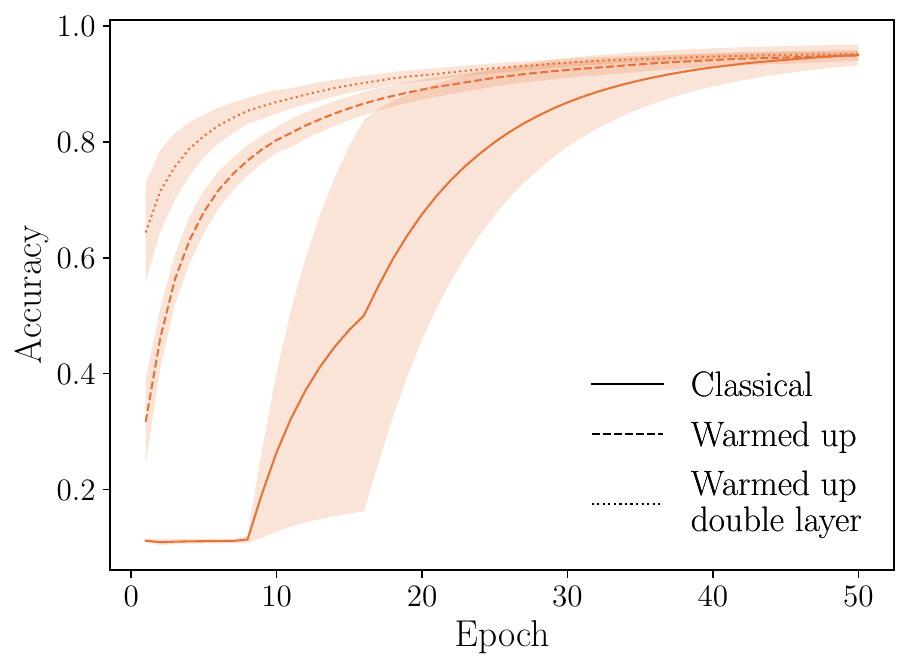}
    \end{subfigure}
    \begin{subfigure}{.35\textwidth}
        \centering
        \includegraphics[width=\textwidth]{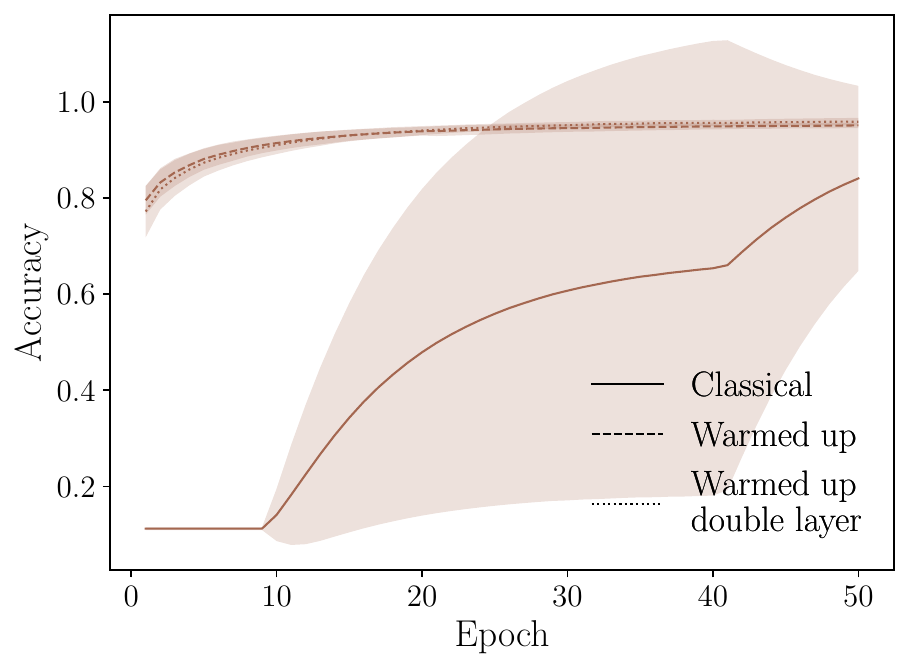}
    \end{subfigure}
    \caption{%
        Evolution of the validation loss on the line-sequential MNIST benchmark for LSTM, GRU and MGU networks, with $N = 100$ and $T = 200$.
        For each cell, three versions are considered: the classical one, the warmed-up one and the double-layer one with partial warmup.
        The hyperparameters of each cell version were optimized on the learning set.
        }
    \label{fig:hp_mnist}
\end{figure}

The best hyperparameters are reported in \autoref{app:hyperparameters} for both benchmarks.
The test losses obtained using those hyperparameters are given in \autoref{tab:ms}.
As can be seen by putting \autoref{fig:ms_bar_denoising} in perspective with \autoref{fig:bar_denoising_100_warmup}, the hyperparameter selection allows all cell versions to reach a lower test MSE for the denoising benchmark.
Similarly, by putting \autoref{fig:ms_bar_mnist} in perspective with \autoref{fig:bar_row_mnist_472_warmup}, it can be seen that all cell versions reach a higher test accuracy for the MNIST benchmark, when the hyperparameters have been optimized.
\autoref{fig:hp_denoising} shows the evolution of the validation losses throughout the training procedure for the best hyperparameters of each cell version, averaged over the 5 training sessions, for the denoising benchmark with $N = 100$.
It can be seen that the warmup procedure and the double cell architecture still provide a significant advantage in term of convergence speed and final performance.
\autoref{fig:hp_mnist} shows the evolution of the validation losses throughout the training procedure using the best hyperparameters, for the line-sequential MNIST benchmark with $N = 472$.
As for the denoising benchmark, the warmup and the double layer architecture still provide a very significant improvement in term of convergence speed.

\section{Conclusion}
\label{sec:conclusion}

In this work, we introduced a new initialisation procedure, called warmup, that improve the ability of recurrent neural networks to learn long time dependencies.
This procedure is motivated by recent work that showed the importance of fixed points and attractors for the prediction process of trained RNNs.
More precisely, we introduced a lightweight measure called VAA, that can be optimized at initialisation in few gradient steps to endow RNNs with multistable dynamics.
Warmup can be used with any type of recurrent cell and we show that it vastly improves their performance on problems with long time dependencies.
In addition, we introduced a new architecture that combines transient and multistable dynamics through partial warmup.
This architecture was shown to reach a better performance than both classic and warmed-up cells on several tasks, including information restitution and sequence classifications tasks.

This work also motivates several future works.
First, it can be noted that the double-layer architecture might be worth exploring with different types of cell.
We showed here that there are benefits of using different types of initialisation for the same type of cell.
This might hint at the possibility of having similar benefits when combining different types of cell that have different dynamical properties in a single recurrent neural network.
Furthermore, in this paper we have aimed at maximizing the number of attractors through warmup before training.
We noticed however that in some rare cases, networks loose multistability properties when training.
Using VAA as a regularization loss to avoid this could be interesting.
For online reinforcement learning too, a regularization loss throughout the learning procedure might make more sense than warming up a priori on random trajectories.
Moreover, we note that not all benchmarks would benefit from warming up.
In fact, it is likely that for several benchmarks, having only a few attractors could be better.
In this regard, it would be interesting to try to warm up in order to reach a specific number of attractors, rather than for maximizing them.
Finally, the warmup procedure maximises reachable multistability for a particular dataset of input sequences.
Warming up on totally random input sequences would result in a simpler procedure that might still provide a good initialisation for reaching multistability.

This work also present some limitations.
First, the VAA is not discriminating limit cycles from fixed point attractors.
In addition, states that are on the same limit cycles but far from each others are not considered in the basin.
Moreover, the warmup maximises the number of attractors present in all hidden states, while we might want the hidden states from a same input sequence to belong to a single basin of attraction.
Finally, the stability of the RNN is measured for a stable input, an assumption that is unrealistic in our experiments and in general.
It might be worth exploring those problems in future works.

\section*{Acknowledgements}

Gaspard Lambrechts and Nicolas Vecoven gratefully acknowledge the financial support of the \emph{Wallonia-Brussels Federation} for their FRIA grant.
Florent De Geeter and Gaspard Lambrechts gratefully acknowledge the financial support of the \emph{Walloon Region} for Grant No.\@ 2010235 – ARIAC by DW4AI.
Computational resources have been provided by the \emph{Consortium des Équipements de Calcul Intensif} (CÉCI), funded by the \emph{Fonds de la Recherche Scientifique de Belgique} (F.R.S.-FNRS) under Grant No.\@ 2502011 and by the Walloon Region.

\bibliography{references.bib}

\appendix
\newpage

\section{Recurrent neural network architectures}
\label{app:recurrent_neural_network_architectures}

Formally, an RNN architecture is defined by its update function $f$, its output function $g$ and its initialisation function $h$ that are parameterised by a vector $\theta \in \RR^d$.
Given a sequence of inputs $\u_{1:T} = [\u_1, \dots, \u_T]$, with $T \in \NN$ and $\u_t \in \RR^n$, the RNN maintains a hidden state $\x_t$ and output a prediction $\o_t$ according to
\begin{align}
    \x_{t} & = f(\x_{t-1}, \u_t; \theta), \; t = 1, \dots, T, \label{eq:rnn_update} \\
    \o_t   & = g(\x_t; \theta), \; t = 1, \dots, T, \label{eq:rnn_output}           \\
    \x_0   & = h(\theta). \label{eq:rnn_initialisation}
\end{align}
RNNs can be composed of $L$ layers that are linked sequentially through $\u^i_t = \o^{i-1}_t$ with $\u^1_t = \u_t$ and $\o_t = \o^L_t$, where $\o^i_t$ denotes the output of layer $i$ and $\u^i_t$ its input.
In this case, each layer $i$ has its own update function $f^i$, output function $g^i$ and initialisation function $h^i$.

In the following, we give the update function $f$ and output function $g$ of a single layer for each architecture considered in this work.
As far as the initial hidden state is concerned, it is always chosen to zero, \ie{}, $h(\theta) = \0$.
Note that $\sigma(x) = \frac{1}{1+e^{-x}}$ denote the sigmoid activation function, and $\odot$ to denote the Hadamard product.

\paragraph{Long short-term memory.}

The LSTM update and output functions are defined from the following intermediate values.
\begin{align}
    \f_t         & = \sigma(\mathbf{W}_{fu} \u_t + \mathbf{W}_{fh} \h_{t-1} + \b_t) \\
    \i_t         & = \sigma(\mathbf{W}_{iu} \u_t + \mathbf{W}_{ih} \h_{t-1} + \b_i) \\
    \r_t         & = \sigma(\mathbf{W}_{ou} \u_t + \mathbf{W}_{oh} \h_{t-1} + \b_r) \\
    \tilde{\c}_t & = \tanh(\mathbf{W}_{cu} \u_t + \mathbf{W}_{ch} \h_{t-1} + \b_c)  \\
    \c_t         & = \f_t \odot \c_{t-1} + \i_t \odot \tilde{\c}_t                  \\
    \h_t         & = \r_t \odot \tanh(\c_t)
\end{align}
The hidden state is given by $\x_t = f(\x_{t-1}, \u_t; \theta) = [\h_t, \c_t]$, and the output is given by $\o_t = g(\x_t ; \theta) = \h_t$.
The parameters of the LSTM network are $\theta = (
    \mathbf{W}_{fu}, \allowbreak
    \mathbf{W}_{fh}, \allowbreak
    \mathbf{W}_{iu}, \allowbreak
    \mathbf{W}_{ih}, \allowbreak
    \mathbf{W}_{ou}, \allowbreak
    \mathbf{W}_{oh}, \allowbreak
    \mathbf{W}_{cu}, \allowbreak
    \mathbf{W}_{ch}, \allowbreak
    \b_t, \b_i, \b_r, \b_c
    )$.

\paragraph{Gated recurrent unit.}

The GRU update and output functions are defined from the following intermediate values.
\begin{align}
    \z_t & = \sigma(\mathbf{W}_{zu} \u_t + \mathbf{W}_{zh} \h_{t-1} + \b_z)                                                  \\
    \r_t & = \sigma(\mathbf{W}_{ru} \u_t + \mathbf{W}_{rh} \h_{t-1} + \b_r)                                                  \\
    \h_t & = \z_t \odot \h_{t-1} + (1 - \z_t) \odot \tanh(\mathbf{W}_{hu} \u_t + \r_t \odot \mathbf{W}_{hh} \h_{t-1} + \b_h)
\end{align}
The hidden state is given by $\x_t = f(\x_{t-1}, \u_t; \theta) = \h_t$, and the output is given by $\o_t = g(\x_t ; \theta) = \h_t$.
The parameters of the GRU network are $\theta = (
    \mathbf{W}_{zu}, \allowbreak
    \mathbf{W}_{zh}, \allowbreak
    \mathbf{W}_{ru}, \allowbreak
    \mathbf{W}_{rh}, \allowbreak
    \mathbf{W}_{hu}, \allowbreak
    \mathbf{W}_{hh}, \allowbreak
    \b_z, \b_r, \b_h
    )$.

\paragraph{Bistable recurrent cell.}

The BRC update and output functions are defined from the following intermediate values.
\begin{align}
    \c_t & = \sigma(\mathbf{W}_{cu} \u_t + w_c \odot \h_{t-1} + \b_c)                                   \\
    \a_t & = 1 + \tanh(\mathbf{W}_{au} \u_t + w_a \odot \h_{t-1} + \b_a)                                \\
    \h_t & = \c_t \odot \h_{t-1} + (1 - \c_t) \odot \tanh(\mathbf{W}_{hu} + \a_t \odot \h_{t-1} + \b_h)
\end{align}
The hidden state is given by $\x_t = f(\x_{t-1}, \u_t; \theta) = \h_t$, and the output is given by $\o_t = g(\x_t ; \theta) = \h_t$.
The parameters of the BRC network are $\theta = (
    \mathbf{W}_{cu}, \allowbreak
    \mathbf{w}_{c}, \allowbreak
    \mathbf{W}_{au}, \allowbreak
    \mathbf{w}_{a}, \allowbreak
    \mathbf{W}_{hu}, \allowbreak
    \b_c, \b_a, \b_h
    )$.

\paragraph{Neuromodulated bistable recurrent cell.}

The NBRC update and output functions are defined from the following intermediate values.
\begin{align}
    \c_t & = \sigma(\mathbf{W}_{cu} \u_t + \mathbf{W}_{ch} \h_{t-1} + \b_c)                             \\
    \a_t & = 1 + \tanh(\mathbf{W}_{au} \u_t + \mathbf{W}_{ah} \h_{t-1} + \b_a)                          \\
    \h_t & = \c_t \odot \h_{t-1} + (1 - \c_t) \odot \tanh(\mathbf{W}_{hu} + \a_t \odot \h_{t-1} + \b_h)
\end{align}
The hidden state is given by $\x_t = f(\x_{t-1}, \u_t; \theta) = \h_t$, and the output is given by $\o_t = g(\x_t ; \theta) = \h_t$.
The parameters of the NBRC network are $\theta = (
    \mathbf{W}_{cu}, \allowbreak
    \mathbf{W}_{ch}, \allowbreak
    \mathbf{W}_{au}, \allowbreak
    \mathbf{W}_{ah}, \allowbreak
    \mathbf{W}_{hu}, \allowbreak
    \b_c, \b_a, \b_h
    )$.

\paragraph{Minimal gated unit.}

The MGU update and output functions are defined from the following intermediate values.
\begin{align}
    \f_t         & = \sigma(\mathbf{W}_{fu} \u_t + \mathbf{W}_{fh} \h_{t-1} + \b_f)             \\
    \tilde{\h}_t & = \tanh(\mathbf{W}_{hu} \u_t + \mathbf{W}_{hh} (\f_t \odot \h_{t-1}) + \b_h) \\
    \h_t         & = \f_t \odot \tilde{\h} + (1 - \f_t) \odot \h_{t-1}
\end{align}
The hidden state is given by $\x_t = f(\x_{t-1}, \u_t; \theta) = \h_t$, and the output is given by $\o_t = g(\x_t ; \theta) = \h_t$.
The parameters of the MGU network are $\theta = (
    \mathbf{W}_{fu}, \allowbreak
    \mathbf{W}_{fh}, \allowbreak
    \mathbf{W}_{hu}, \allowbreak
    \mathbf{W}_{hh}, \allowbreak
    \b_f, \b_h
    )$.

\section{Partially observable Markov decision process} \label{app:pomdp}

Formally, a POMDP $P$ is an $8$-tuple $P = (\S, \A, \O, p_0, T, R, O, \gamma)$ where $\S$ is the state space, $\A$ is the action space, and $\O$ is the observation space.
The initial state distribution $p_0$ gives the probability $p_0(\s_0)$ of $\s_0 \in \S$ being the initial state of the decision process.
The dynamics are described by the transition distribution $T$ that gives the probability $T(\s_{t+1} \mid \s_t, \a_t)$ of $\s_{t+1} \in \S$ being the state resulting from action $\a_t \in \A$ in state $\s_t \in \S$.
The reward function $R$ gives the immediate reward $r_t = R(\s_t, \a_t, \s_{t+1})$ obtained after each transition.
The observation distribution $O$ gives the probability $O(\o_t \mid \s_t)$ to get observation $\o_t \in \O$ in state $\s_t \in \S$.
Finally, the discount factor $\gamma \in [0, 1[$ gives the relative importance of future rewards.

Taking a sequence of $t$ actions ($\a_{0:t-1}$) in the POMDP conditions its execution and provides a sequence of $t+1$ observations ($\o_{0:t}$).
Together, they compose the history $\eta_{0:t} = (\o_{0:t}, \a_{0:t-1}) \in \H_{0:t}$ until timestep $t$, where $\H_{0:t}$ is the set of such histories.
Let $\eta \in \H$ denote a history of arbitrary length sampled in the POMDP, and let $\H = \bigcup_{t = 0}^\infty \H_{0:t}$ denote the set of histories of arbitrary length.

A policy $\pi \in \Pi$ in a POMDP is a mapping from histories to actions, where $\Pi = \H \-> \A$ is the set of such mappings.
A policy $\pi^* \in \Pi$ is said to be optimal when it maximises the expected discounted sum of future rewards starting from any history $\eta_{0:t} \in \H_{0:t}$ at time $t \in \NN_0$
\begin{equation}
    \pi^* \in \argmax_{\pi \in \Pi} \expect_{\pi, P} \left[
        \sum_{t' = t}^{\infty} \gamma^{t' - t} r_{t'}
        \;\middle\vert\; \eta_{0:t}
        \right], \; \forall \eta_{0:t} \in \H_{0:t}, \; \forall t \in \NN_0 .
    \label{eq:optimal_policy}
\end{equation}
The history-action value function, or \cqf{}, is defined as the maximal expected discounted reward that can be gathered, starting from a history $\eta_{0:t} \in \H_{0:t}$ at time $t \in \NN_0$ and an action $\a_t \in \A$
\begin{equation}
    \Q(\eta_{0:t}, \a_t) = \max_{\pi \in \Pi} \expect_{\pi, P} \left[
        \sum_{t' = t}^{\infty} \gamma^{t' - t} r_{t'}
        \;\middle\vert\; \eta_{0:t}, \a_t
        \right], \;
    \forall \eta_{0:t} \in \H_{0:t}, \;
    \forall \a_t \in \A, \;
    \forall t \in \NN_0 .
    \label{eq:cqf}
\end{equation}
The \cqf{} is also the unique solution of the Bellman equation
\citep{smallwood1973optimal, kaelbling1998planning, porta2004value}
\begin{equation}
    \Q(\eta, \a) = \expect_{P} \left[
        r + \gamma \max_{\a' \in \A} \Q(\eta', \a')
        \;\middle\vert\; \eta, \a
        \right], \;
    \forall \eta \in \H, \;
    \forall \a \in \A
    \label{eq:cqf_bellman}
\end{equation}
where $\eta' = \eta \cup (\a, \o')$ and $r$ is the immediate reward obtained when taking action $\a$ in history $\eta$.
From \eqref{eq:optimal_policy} and \eqref{eq:cqf}, it can be noticed that any optimal policy satisfies
\begin{equation}
    \pi^*(\eta) \in \argmax_{\a \in \A} \Q(\eta, \a),
    \; \forall \eta \in \H .
    \label{eq:q_policy}
\end{equation}

\section{Deep Recurrent Q-learning} \label{app:drqn}

The DRQN \citep{hausknecht2015deep} algorithm aims at learning a parametric approximation $\Q_\theta$ of the \cqf{}, where $\theta \in \RR^{d_\theta}$ is the parameter vector of a recurrent neural network.
This algorithm is motivated by equation \eqref{eq:q_policy} that shows that an optimal policy can be derived from the \cqf{}.
The strategy consists of minimising with respect to $\theta$, for all $(\eta, \a)$, the distance between the estimation $\Q_\theta(\eta, \a)$ of the LHS of equation \eqref{eq:cqf_bellman}, and the estimation of the expectation $\mathbb{E}_P[r + \gamma \max_{\a' \in \A} \Q_\theta(\eta', \a')]$ of the RHS of equation \eqref{eq:cqf_bellman}.
This is done by using transitions $(\eta, \a, r, \o', \eta')$ sampled in the POMDP, with $\eta' = \eta \cup (\a, \o')$.

In practice, this algorithm interleaves the generation of episodes and the update of the estimation $\Q_\theta$.
Indeed, in the DRQN algorithm, the episodes are generated with the $\varepsilon$-greedy policy derived from the current estimation $\Q_\theta$.
This stochastic policy selects actions according to $\argmax_{\a \in \A} \Q_\theta(\cdot, \a)$ with probability $1 - \varepsilon$, and according to an exploration policy with probability $\varepsilon$.
This exploration policy is defined by a probability distribution $\mathcal{E}(\A) \in \P(\A)$ over the actions, where $\P(\A)$ is the set of probability measures over the action space $\A$.
The DRQN algorithm also introduces a truncation horizon $H$ such that the histories generated in the POMDP have a maximum length of $H$.
Moreover, a replay buffer of histories is used and the gradient is evaluated on a batch of histories sampled from this buffer.
Furthermore, the parameters $\theta$ are updated with the Adam algorithm \citep{kingma2014adam}.
Finally, the target $r_t + \gamma \max_{\a \in \A} \Q_{\theta'}(\eta_{0:t+1}, \a)$ is computed using a past version $\Q_{\theta'}$ of the estimation $\Q_\theta$ with parameters $\theta'$ that are updated to $\theta$ less frequently, which eases the convergence towards the target, and ultimately towards the \cqf{}.

\begin{algo}
    \DontPrintSemicolon \footnotesize
    \caption{DRQN - \cqf{} approximation}
    \label{algo:drqn}

    \SetKwInOut{Parameters}{Parameters}
    \SetKwInOut{Output}{Output}
    \SetKwFunction{Warmup}{Warmup}

    \Parameters{%
        $N \in \NN$ the buffer capacity. \\
        $C \in \NN$ the target update period in term of episodes. \\
        $E \in \NN$ the number of episodes. \\
        $H \in \NN$ the truncation horizon. \\
        $I \in \NN$ the number of gradient steps after each episode. \\
        $\varepsilon \in \RR$ the exploration rate. \\
        $\mathcal{E}(\A) \in \P(\A)$ the exploration policy probability distribution. \\
        $\alpha \in \RR$ the learning rate. \\
        $B \in \NN$ the batch size. \\
        $\theta \in \RR^{d_\theta}$ the initial parameters of the network. \\
        $\theta' \in \RR^{d_\theta}$ the initial parameters of the target
        network.}

    Initialise weights $\theta$ randomly \;

    Fill replay buffer $\B$ with random transitions from the exploration policy $\mathcal{E}(\A)$. \;
    \If{warmup}{
        Let $\D$ be the set of histories $\eta$ (input sequences) in replay buffer $\B$. \;
        \Warmup($\D, \theta$) using default parameters of the \Warmup algorithm.
    }

    \For{$e = 0, \dots, E - 1$}{

        \If{$e \bmod C = 0$}{%
            Update target network with $\theta' \leftarrow \theta$
        }

        \tcp{Generate new episode, store history and rewards}

        Draw an initial state $\s_0$ according to $p_0$ and observe $\o_0$ \;

        Let $\eta_{0:0} = (\o_0)$  \;

        \For{$t = 0, \dots, H - 1$}{
            Select $\a_t \sim \mathcal{E}(\A)$ with
            probability $\varepsilon$, otherwise select
            $\a_t = \argmax_{\a \in \A} \left\{
                \Q_\theta(\eta_{0:t}, \a )
                \right\}$ \;

            Take action $\a_t$ and observe $r_t$ and $\o_{t+1}$ \;
            Let $\eta_{0:t+1} = (\o_0, \a_0, \o_1, \dots, \o_{t+1})$ \;

            \lIf{$|\B| < N$}{
                add $(\eta_{0:t}, \a_t, r_t, \o_{t+1}, \eta_{0:t+1})$ in replay
                buffer $\B$
            }
            \lElse{%
                replace oldest transition in replay buffer $\B$ by
                $(\eta_{0:t}, \a_t, r_t, \o_{t+1}, \eta_{0:t+1})$
            }

            \If{$\o_{t+1}$ is terminal}{%
                \textbf{break}
            }
        }

        \tcp{Optimize recurrent Q-network}

        \For{$i = 0, \dots, I - 1$}{

            Sample $B$ transitions $(\eta_{0:t}^b, \a_t^b, r_t^b,
                \o_{t + 1}^b, \eta_{0:t+1}^b)$ uniformly from the
            replay buffer $\B$ \;

            Compute targets $y^b = \begin{cases}
                    r_t^b + \gamma \max_{\a \in \A}
                    \left\{
                    \Q_{\theta'} (\eta_{0:t+1}^b, \a)
                    \right\}
                          & \text{if } \o_{t + 1}^b \text{ is not terminal} \\
                    r_t^b & \text{otherwise}
                \end{cases}$ \;

            Compute loss $L = \sum_{b = 0}^{B - 1} \left( y^b -
                \Q_\theta(\eta_{0:t}^b, \a_t^b) \right)^2$ \;

            Compute direction $g$ using Adam optimizer, perform gradient step
            $\theta \leftarrow \theta + \alpha g$ \;
        }
    }
\end{algo}

The DRQN training procedure is detailed in \autoref{algo:drqn}.
In this algorithm, the output of the RNN is $\y_t = g(\h_t; \theta) \in \RR^{|\A|}$, and it gives $\Q_\theta(\eta_{0:t}, \a), \; \forall \a \in \A$.
The hidden states are given by $\h_k = f(\h_{k-1}, \x_k; \theta), \; \forall k \in \NN_0$, with the inputs given by $\x_k = (\a_{k-1}, \o_k), \; \forall t \in \NN$ and $\x_0 = (\mathbf{0}, \o_0)$.
From the approximation $\Q_\theta$, the policy $\pi_\theta$ is given by $\pi_\theta(\eta) = \argmax_{\a \in \A} \Q_\theta(\eta, \a)$.

In the experiments, the following hyperparameters have been chosen: $N = \num{8192}$, $C = 20$, $I = 10$, $\varepsilon = 0.2$, $\alpha = \num{1e-3}$, $B = 32$. The exploration policy and truncation horizon depend on the environment and are thus detailed in the following appendix.

\section{T-Maze environment} \label{app:tmaze}

The T-Maze environment is a POMDP $(\S, \A, \O, p_0, T, R, O, \gamma)$ parameterised by the maze length $L \in \NN$.
The formal definition of this environment is given below.

\begin{figure}[ht]
    \centering
    \resizebox{0.6\textwidth}{!}{\begin{tikzpicture}[scale=2]

    \pgfmathsetmacro{\L}{8}

    \node (mup) at (0.5, 5.5) {$\mathbf{m} = \text{Up}$} ;
    \node (mup) at (0.5, 1.5) {$\mathbf{m} = \text{Down}$} ;

    \foreach \s in {0, 4} {
        \draw[lightgray] (0, \s) grid (\L, \s + 1) ;
        \draw[lightgray] (\L, \s - 1) grid (\L + 1, \s + 2) ;

        \fill[blue, opacity=0.2] (0, \s) rectangle (1, \s + 1) ;
        \fill[pattern=north west lines, pattern color=blue!20]
            (\L, 6 * \s / 4 - 1) rectangle (\L + 1, 6 * \s / 4) ;
        \fill[gray, opacity=0.2] (\L, \s + 1) rectangle (\L + 1, \s + 2) ;
        \fill[gray, opacity=0.2] (\L, \s - 1) rectangle (\L + 1, \s + 0) ;

        \pgfmathsetmacro{\Lminustwo}{\L-2}
        \foreach \i in {0,...,\Lminustwo} {
            \node (c\i) at (\i + 0.5, \s + 0.5) {$\mathbf{c} = (\i, 0)$} ;
        }

        \node (cdot) at (\L - 0.5, \s + 0.5) {$\dots$} ;
        \node (cL) at (\L + 0.5, \s + 0.5) {$\mathbf{c} = (L, 0)$} ;
        \node (cL+1) at (\L + 0.5, \s + 1.5) {$\mathbf{c} = (L, 1)$} ;
        \node (cL+2) at (\L + 0.5, \s - 0.5) {$\mathbf{c} = (L, -1)$} ;
    }

\end{tikzpicture}}
    \caption{%
        T-Maze state space. Initial states in blue, terminal states in grey, and treasure states hatched.}
    \label{fig:tmaze}
\end{figure}

\paragraph{State space}

The discrete state space $\S$ is composed of the set of positions $\C$ for the agent in each of the two maze layouts $\M$.
The maze layout determines the position of the treasure.
Formally, we have
\begin{empheq}[left = \empheqlbrace]{align}
    \S &= \M \times \C \\
    \M &= \{\text{Up}, \text{Down}\} \\
    \C &= \left\{ (0 ,  0), \dots, (L ,  0) \right\} \cup \{(L, 1), (L, -1)\}
\end{empheq}

A state $\s_t \in \S$ is thus defined by $\s_t = (\m_t, \c_t)$ with $\m_t \in \M$ and $\c_t \in \C$.
Let us also define $\F = \left\{ \s_t = (\m_t, \c_t) \in \S \mid \c_t \in \{(L, 1), (L, -1)\} \right\}$ the set of terminal states, four in number.

\paragraph{Action space}

The discrete action space $\A$ is composed of the four possible moves that the agent can take
\begin{equation}
    \A = \left\{
    (1,  0), (0,  1), (-1,  0), (0, -1)
    \right\}
\end{equation}
that correspond to Right, Up, Left and Down, respectively.

\paragraph{Observation space}

The discrete observation space $\O$ is composed of the four partial observations of the state that the agent can perceive
\begin{equation}
    \O = \left\{
    \text{Up}, \text{Down}, \text{Corridor}, \text{Junction}
    \right\} .
\end{equation}

\paragraph{Initial state distribution}

The two possible initial states are $\s_0^\text{Up} = (\text{Up}, (0, 0))$ and $\s_0^\text{Down} = (\text{Down}, (0 ,  0))$, depending on the maze in which the agent lies.
The initial state distribution $p_0: \S \-> [0, 1]$ is thus given by
\begin{equation}
    p_0(\s_0) =
    \begin{cases}
        0.5 & \text{ if } \s_0 = \s_0^\text{Up}   \\
        0.5 & \text{ if } \s_0 = \s_0^\text{Down} \\
        0   & \text{ otherwise }
    \end{cases}
\end{equation}

\paragraph{Transition distribution}

The transition distribution function $T\colon \S \times \A \times \S \-> [0, 1]$ is given by
\begin{equation}
    T(\s_{t+1} \mid \s_t, \a_t) = \delta_{f(\s_t, \a_t)}(\s_{t+1})
\end{equation}
where $\s_t \in \S, \a_t \in \A$ and $\s_{t+1} \in \S$, and $f$ is given by
\begin{equation}
    f(\s_t, \a_t) =
    \begin{cases}
        \s_{t+1} = (\m_t, \c_t + \a_t)
         & \text{ if } \s_t \not \in \F, \c_t + \a_t \in \C \\
        \s_{t+1} = (\m_t, \c_t)
         & \text{ otherwise }
    \end{cases}
\end{equation}
where $\s_t = (\m_t, \c_t) \in \S$ and $\a_t \in \A$.

\paragraph{Reward function}

The reward function $R: \S \times \A \times \S \-> \RR$ is given by
\begin{equation}
    R(\s_t, \a_t, \s_{t+1}) =
    \begin{cases}
        0     & \text{ if } \s_t \in \F                                   \\
        0     & \text{ if } \s_t \not \in \F, \s_{t+1} \not \in \F,
        \s_t \neq \s_{t+1}                                                \\
        - 0.1 & \text { if } \s_t \not \in \F, \s_{t+1} \not \in \F,
        \s_t = \s_{t+1}                                                   \\
        4     & \text{ if } \s_t \not \in \F, \s_{t+1} \in \F, \c_{t+1} =
        \begin{cases}
            (L, 1)  & \text{ if } \m_{t+1} = \text{Up}   \\
            (L, -1) & \text{ if } \m_{t+1} = \text{Down}
        \end{cases}                      \\
        -0.1  & \text{ if } \s_t \not \in \F, \s_{t+1} \in \F, \c_{t+1}
        = \begin{cases}
              (L, -1) & \text{ if } \m_{t+1} = \text{Up}   \\
              (L, +1) & \text{ if } \m_{t+1} = \text{Down}
          \end{cases}
    \end{cases}
\end{equation}
where $\s_t = (\m_t, \c_t) \in \S, \a_t \in \A$ and $\s_{t+1} = (\m_{t+1},
    \c_{t+1}) \in \S$.

\paragraph{Observation distribution}

In the T-Maze, the observations are deterministic. The observation distribution $O\colon \S \times \O \-> [0, 1]$ is given by
\begin{equation}
    O(\o_t \mid \s_t) =
    \begin{cases}
        1 & \text{ if } \o_t = \text{Up}, \c_t = (0, 0), \m_t = \text{Up} \\
        1 & \text{ if } \o_t = \text{Down}, \c_t = (0, 0),
        \m_t = \text{Down}                                                \\
        1 & \text{ if } \o_t = \text{Corridor}, \c_t \in \left\{
        (1 ,  0), \dots, (L-1 ,  0) \right\}                              \\
        1 & \text{ if } \o_t = \text{Junction}, \c_t \in \left\{
        (L ,  0), (L, 1), (L, -1) \right\}                                \\
        0 & \text{ otherwise }
    \end{cases}
\end{equation}
where $\s_t = (\m_t, \c_t) \in \S$ and $\o_t \in \O$.

\paragraph{Exploration policy}

The exploration policy $\mathcal{E}: \A \-> [0, 1]$ is a stochastic policy that is given by $\mathcal{E}(\text{Right}) = 1/2$ and $\mathcal{E}(\text{Other}) = 1/6$ where $\text{Other} \in \left\{ \text{Up}, \text{Left}, \text{Down} \right\}$.
It enforces to explore the right of the maze layouts.
This exploration policy, tailored to the T-Maze environment, allows one to speed up the training procedure, without interfering with the study of this work.

\paragraph{Truncation horizon}

The truncation horizon $H$ of the DRQN algorithm is chosen such that the expected displacement of an agent moving according to the exploration policy in a T-Maze with an infinite corridor on both sides is greater than $L$.
Let $r = \mathcal{E}(\text{Right})$ and $l = \mathcal{E}(\text{Left})$.
In this infinite T-Maze, starting at \num{0}, the expected displacement after one timestep is $\bar{x}_1 = r - l$.
By independence, $\bar{x}_H = H \bar{x}_1$ such that, for $\bar{x}_H \geq L$, the time horizon is given by
\begin{equation}
    H = \left\lceil \frac{L}{r - l} \right\rceil .
\end{equation}

\section{Generalisation to other hyperparameters} \label{app:generalisation}

In this section, we study the generalisation of the results of this work to other hyperparameters.
More precisely, we vary the number of recurrent layers, the number of neurons in each layer, the batch size, and the learning rate.
In \autoref{app:generalisation_vaa}, we study if the VAA increases when learning occurs for the copy first input benchmark with $T = 50$.
In \autoref{app:generalisation_warmup}, we study if the warmup procedure and the double layer architecture improve learning for the permuted row sequential MNIST benchmark with $N = 472$.
Finally, in \autoref{app:impact_k}, we study the impact of the warmup procedure on the copy first input benchmark with $T = 300$ for different values of $k$.
All averages and standard deviations reported were computed over three different training sessions.

\subsection{Generalisation of the correlation between multistability and learning} \label{app:generalisation_vaa}

\begin{figure}[p]
    \centering
    \begin{subfigure}{.49\textwidth}
        \centering
        \includegraphics[width=.49\textwidth]{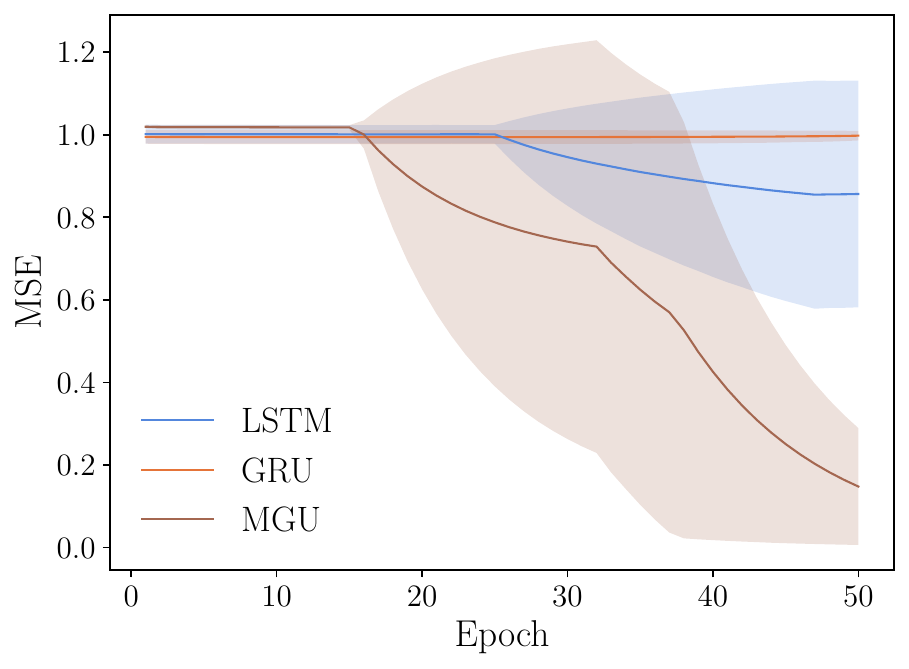}
        \includegraphics[width=.49\textwidth]{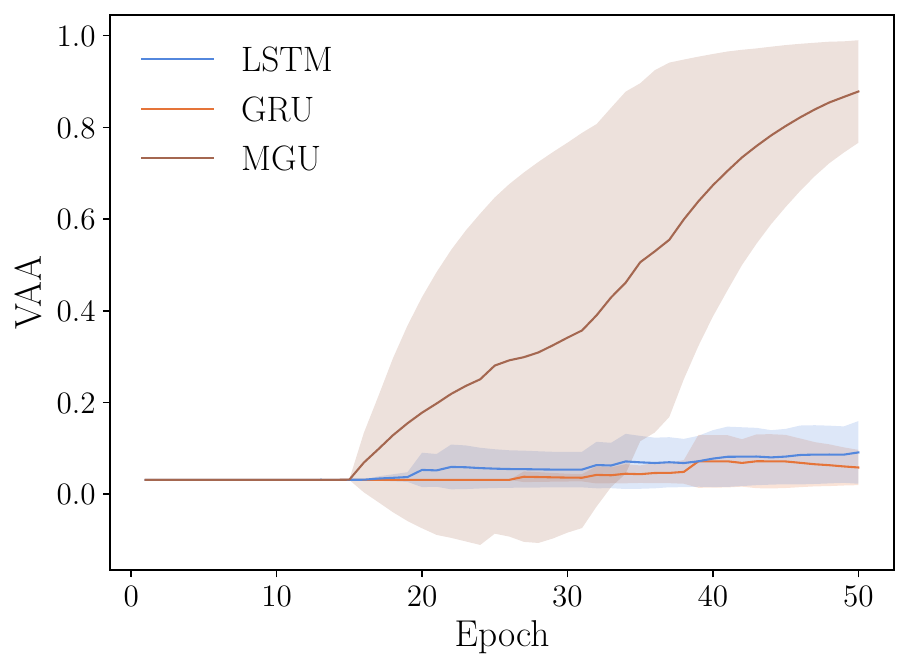}
        \caption{$S = 1, H = 64, \alpha = 1e-3, B = 32$}
    \end{subfigure}
    \begin{subfigure}{.49\textwidth}
        \centering
        \includegraphics[width=.49\textwidth]{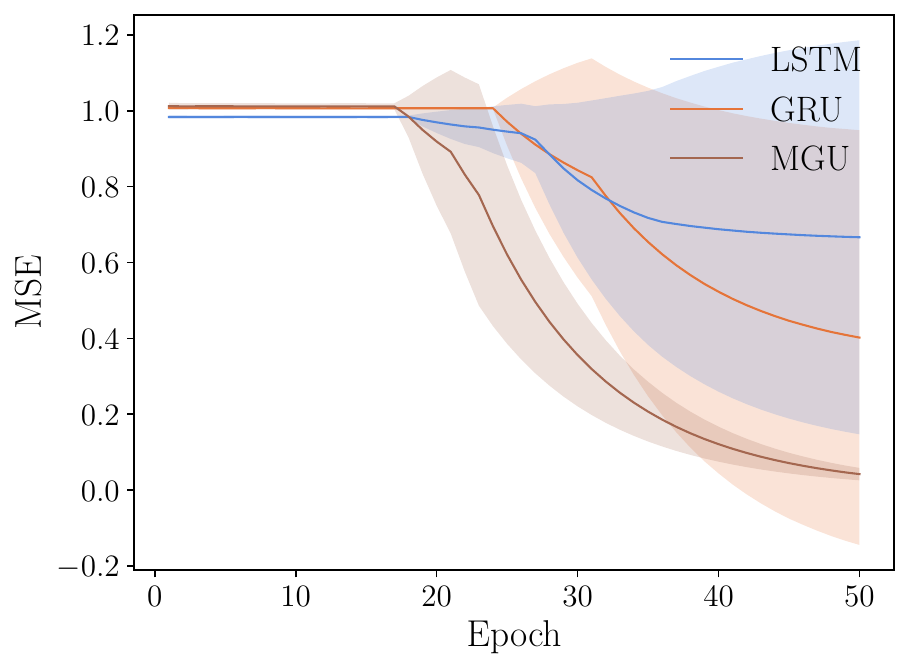}
        \includegraphics[width=.49\textwidth]{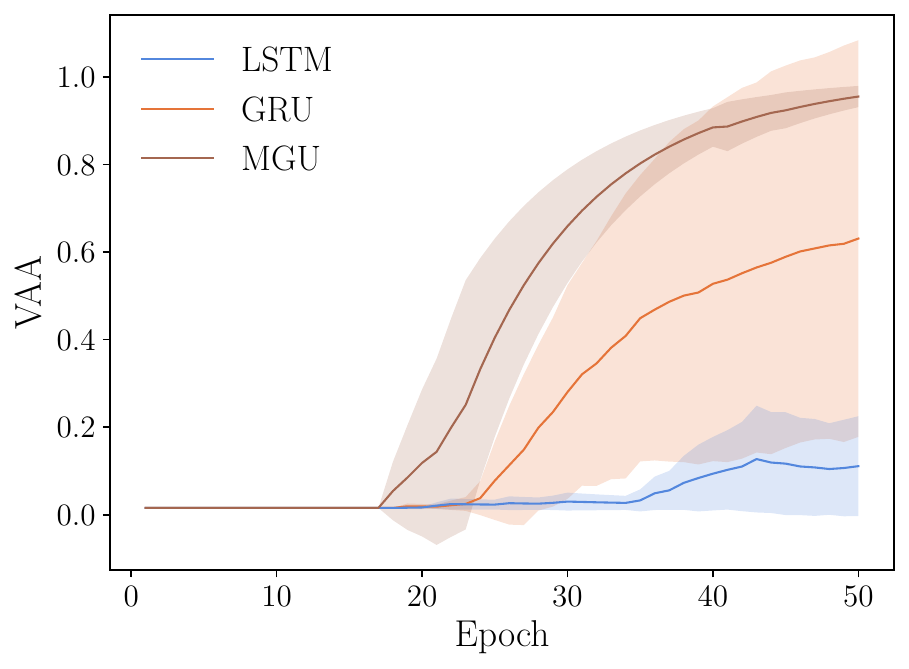}
        \caption{$S = 1, H = 64, \alpha = 1e-3, B = 64$}
    \end{subfigure}
    \begin{subfigure}{.49\textwidth}
        \centering
        \includegraphics[width=.49\textwidth]{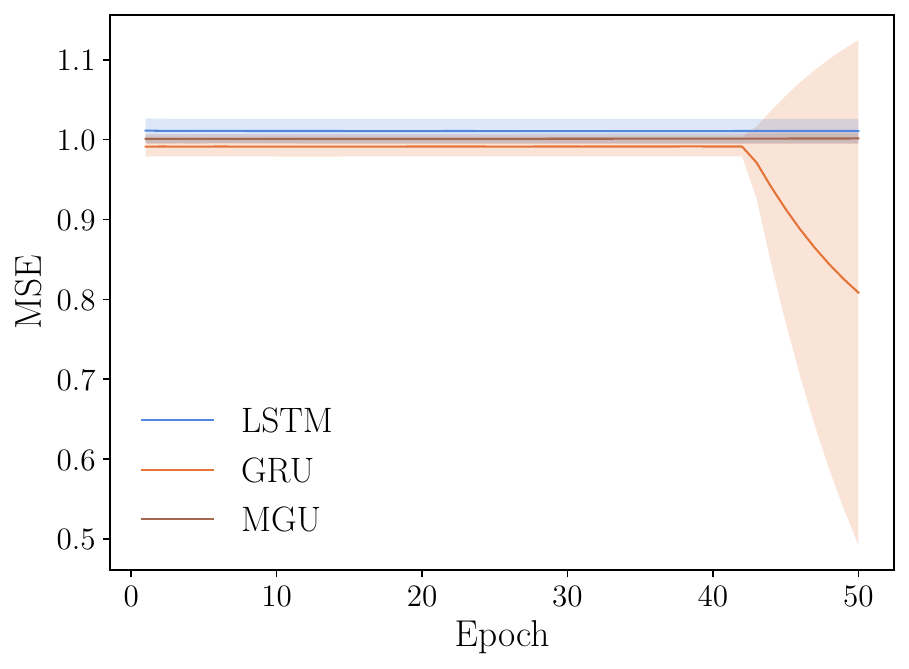}
        \includegraphics[width=.49\textwidth]{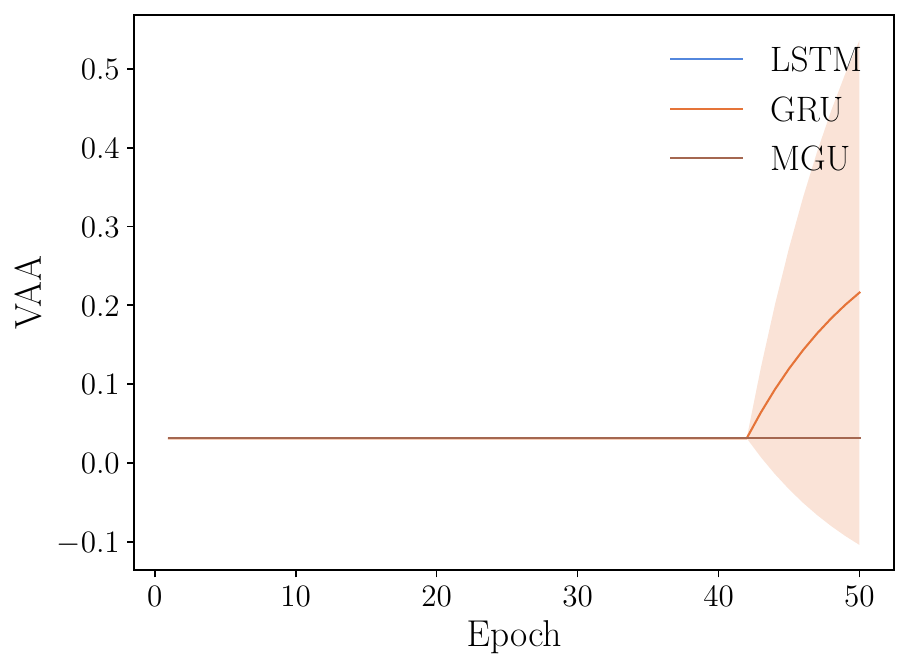}
        \caption{$S = 1, H = 64, \alpha = 1e-4, B = 32$}
    \end{subfigure}
    \begin{subfigure}{.49\textwidth}
        \centering
        \includegraphics[width=.49\textwidth]{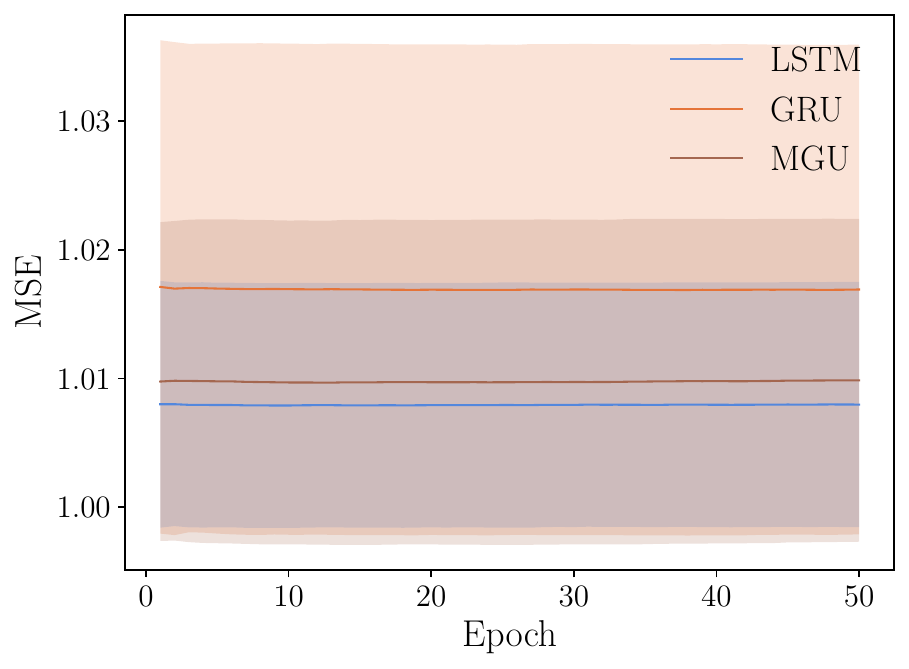}
        \includegraphics[width=.49\textwidth]{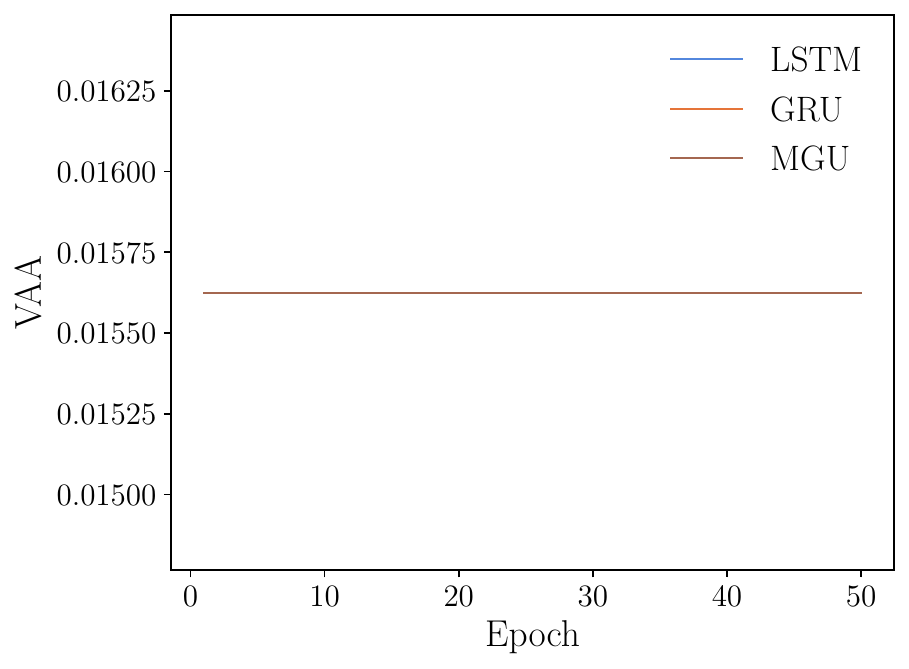}
        \caption{$S = 1, H = 64, \alpha = 1e-4, B = 64$}
    \end{subfigure}
    \begin{subfigure}{.49\textwidth}
        \centering
        \includegraphics[width=.49\textwidth]{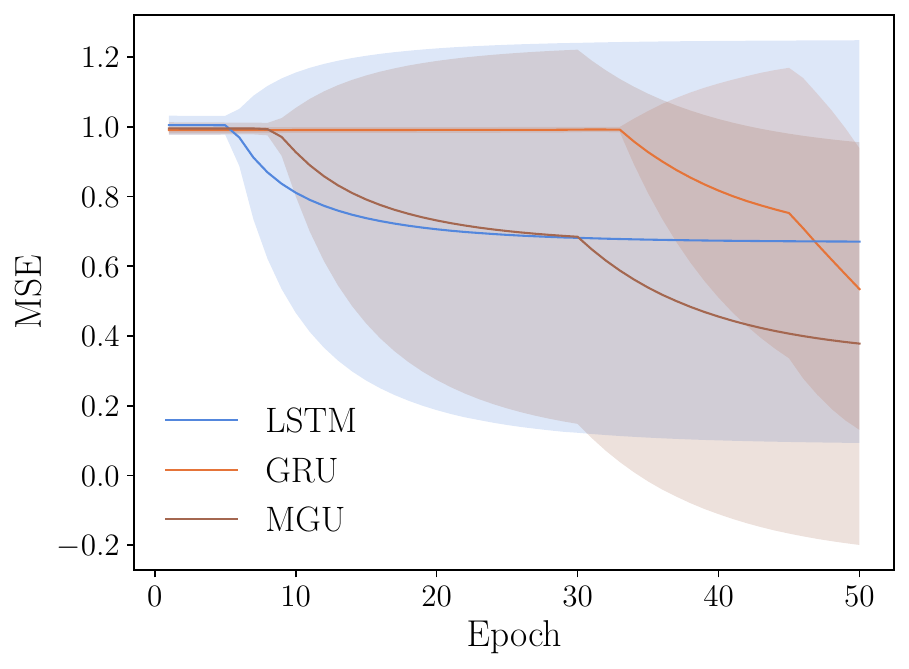}
        \includegraphics[width=.49\textwidth]{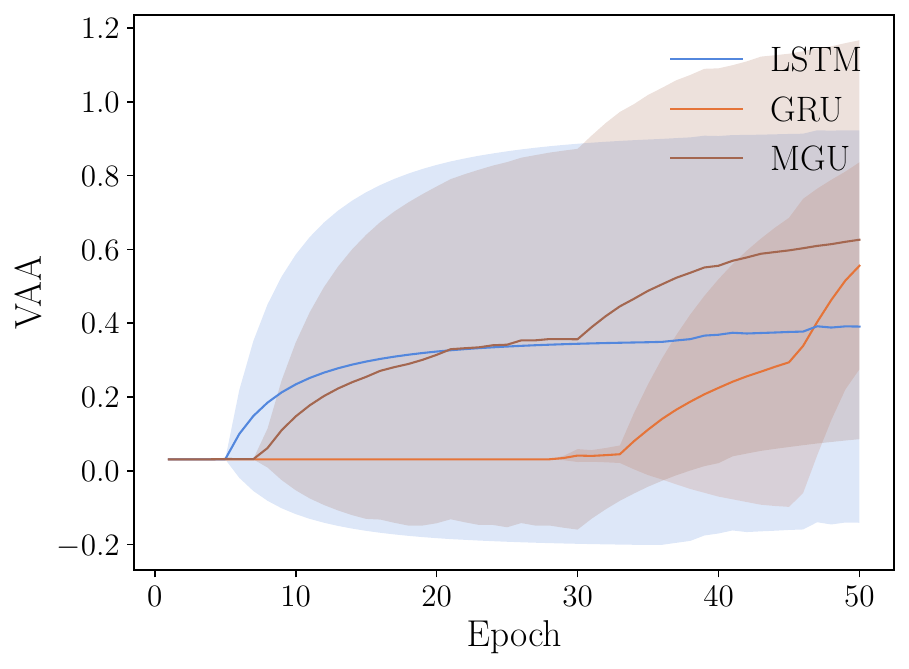}
        \caption{$S = 1, H = 256, \alpha = 1e-3, B = 32$}
    \end{subfigure}
    \begin{subfigure}{.49\textwidth}
        \centering
        \includegraphics[width=.49\textwidth]{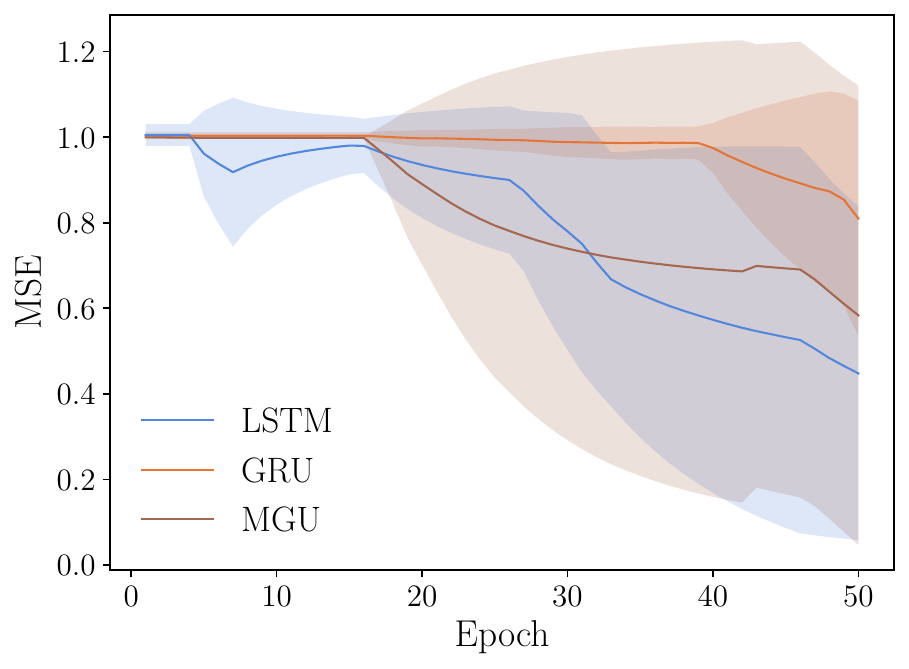}
        \includegraphics[width=.49\textwidth]{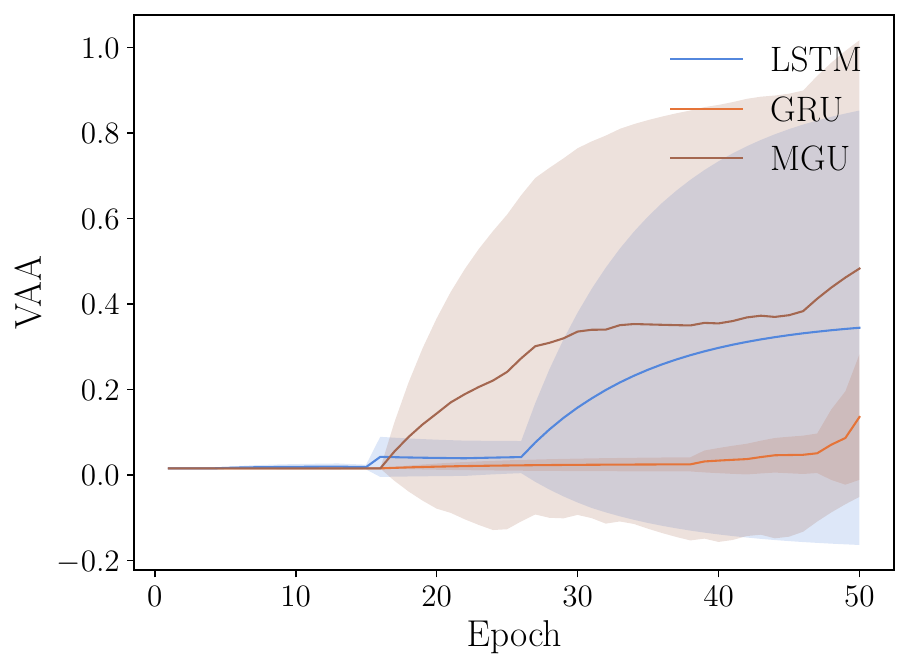}
        \caption{$S = 1, H = 256, \alpha = 1e-3, B = 64$}
    \end{subfigure}
    \begin{subfigure}{.49\textwidth}
        \centering
        \includegraphics[width=.49\textwidth]{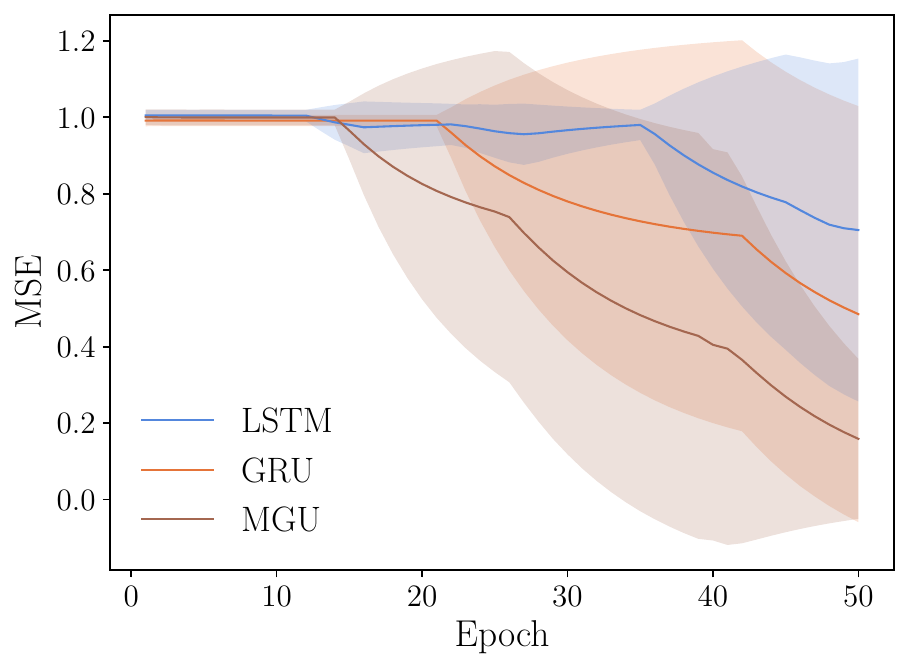}
        \includegraphics[width=.49\textwidth]{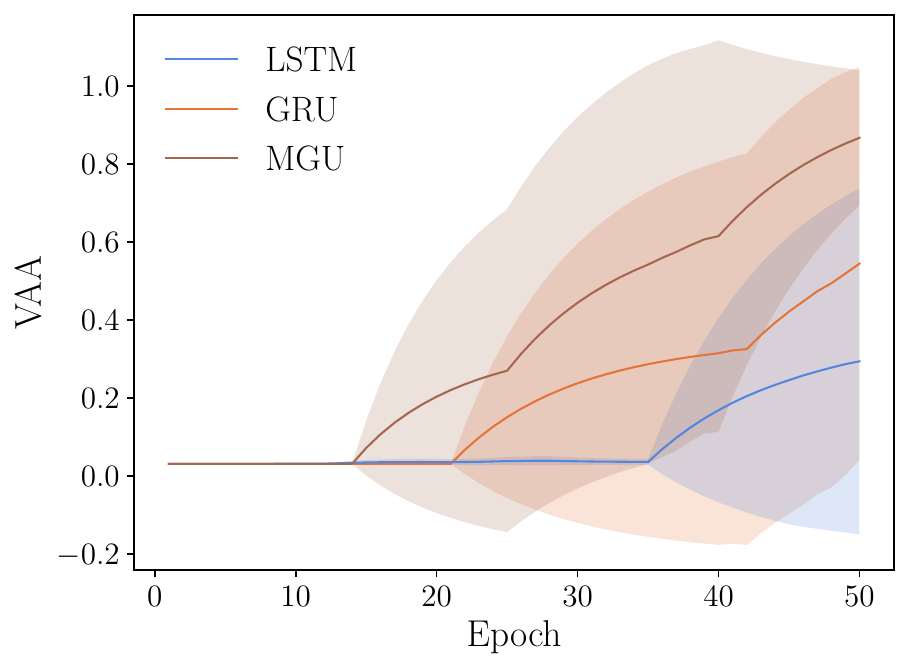}
        \caption{$S = 1, H = 256, \alpha = 1e-4, B = 32$}
    \end{subfigure}
    \begin{subfigure}{.49\textwidth}
        \centering
        \includegraphics[width=.49\textwidth]{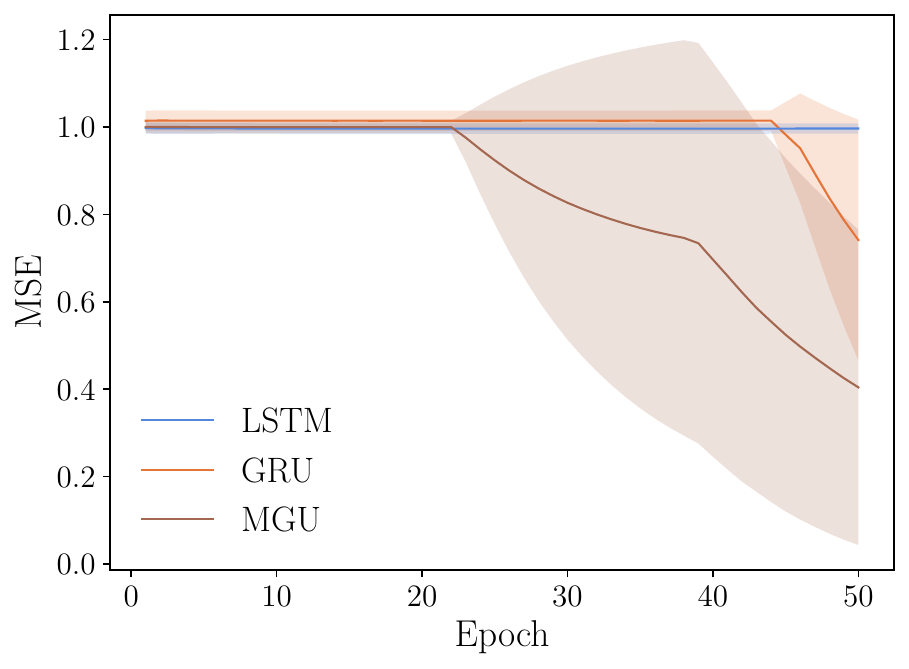}
        \includegraphics[width=.49\textwidth]{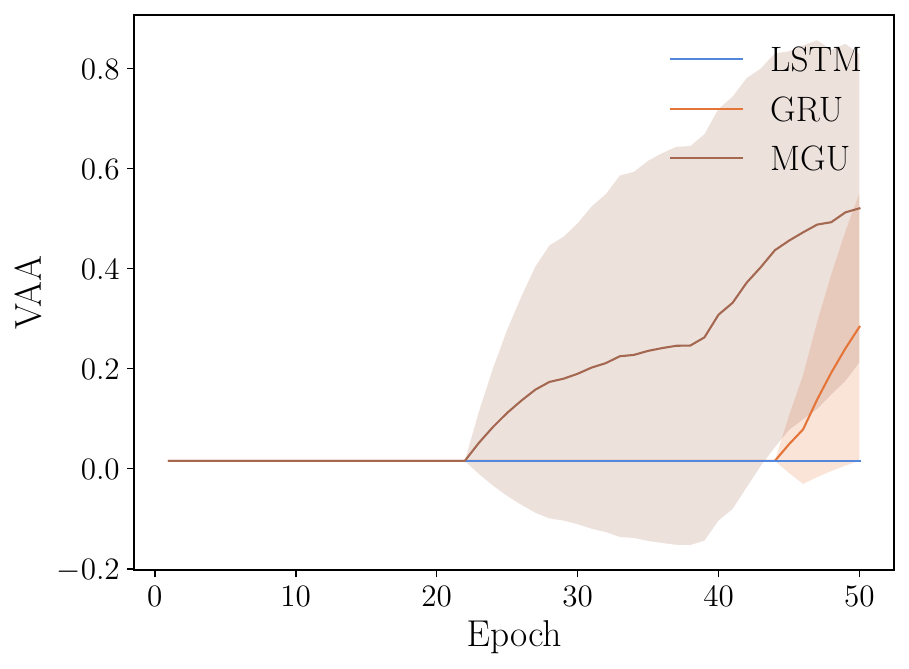}
        \caption{$S = 1, H = 256, \alpha = 1e-4, B = 64$}
    \end{subfigure}
    \caption{%
        Evolution of the validation loss (left) and of the VAA (right) of LSTM, GRU and MGU networks, for the copy first input benchmark.}
    \label{fig:copyfirstinput_generalisation_vaa_1}
\end{figure}

\begin{figure}[p]
    \begin{subfigure}{.49\textwidth}
        \centering
        \includegraphics[width=.49\textwidth]{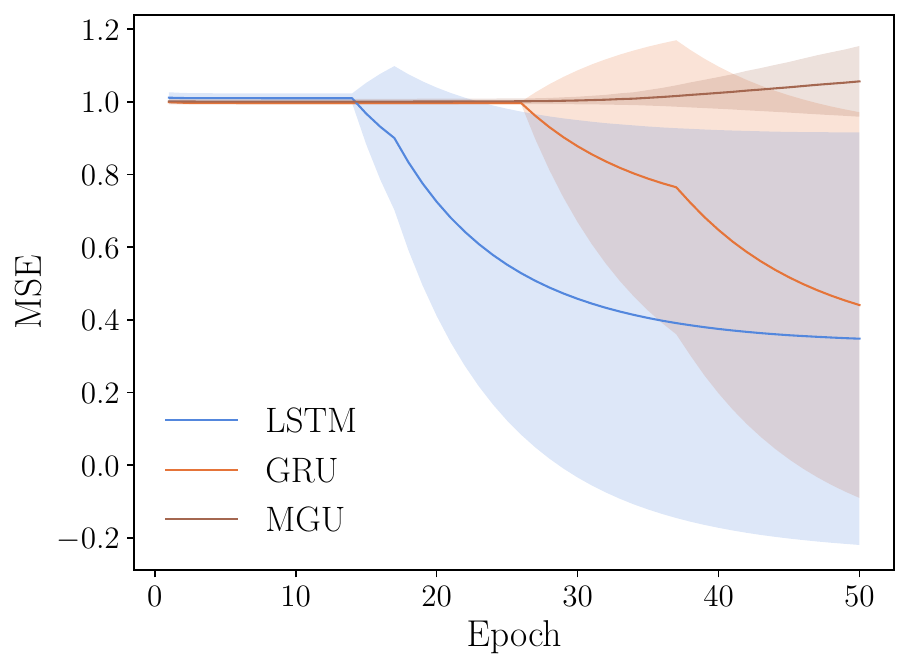}
        \includegraphics[width=.49\textwidth]{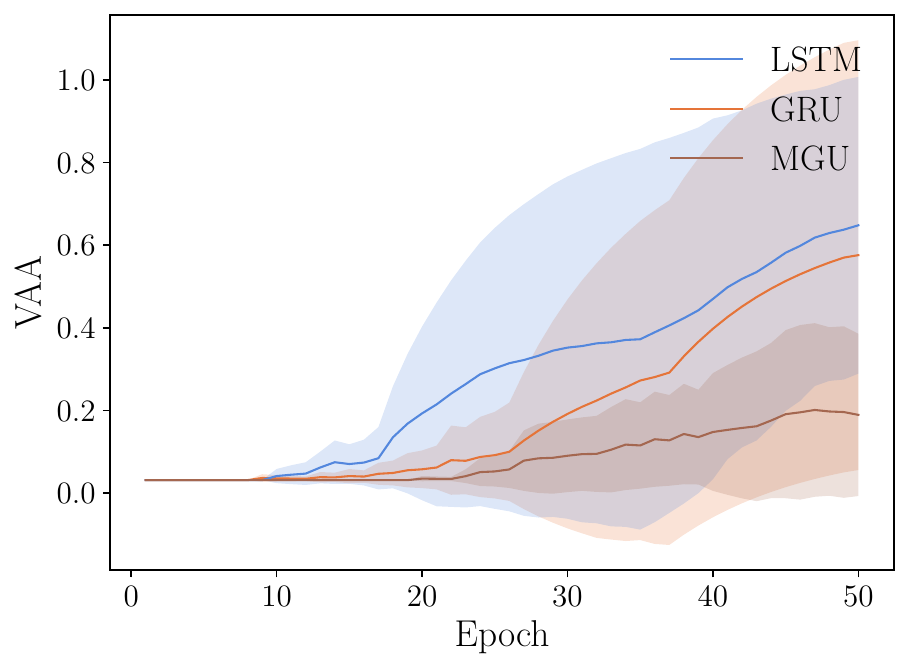}
        \caption{$S = 2, H = 64, \alpha = 1e-3, B = 32$}
    \end{subfigure}
    \begin{subfigure}{.49\textwidth}
        \centering
        \includegraphics[width=.49\textwidth]{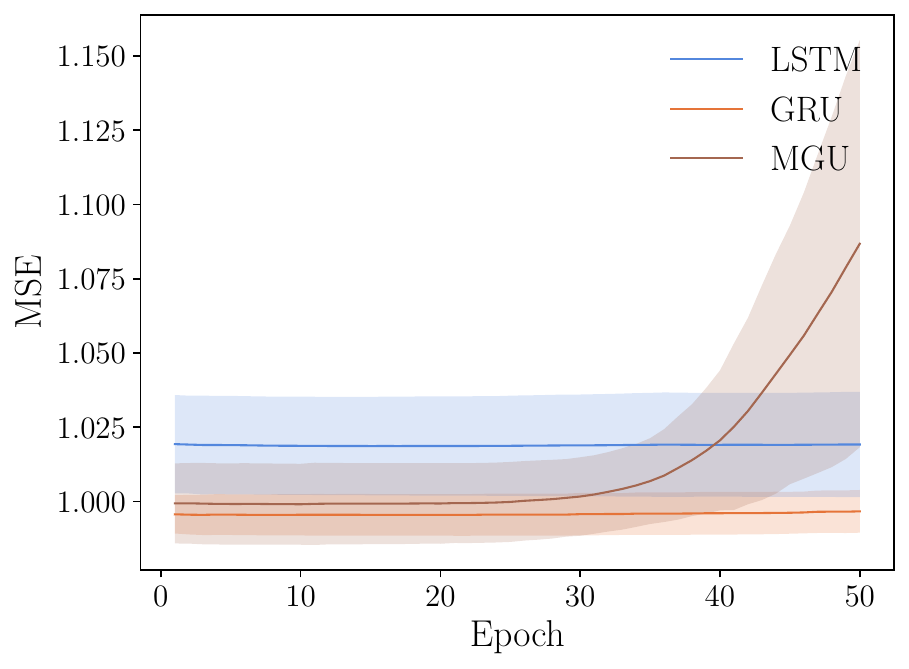}
        \includegraphics[width=.49\textwidth]{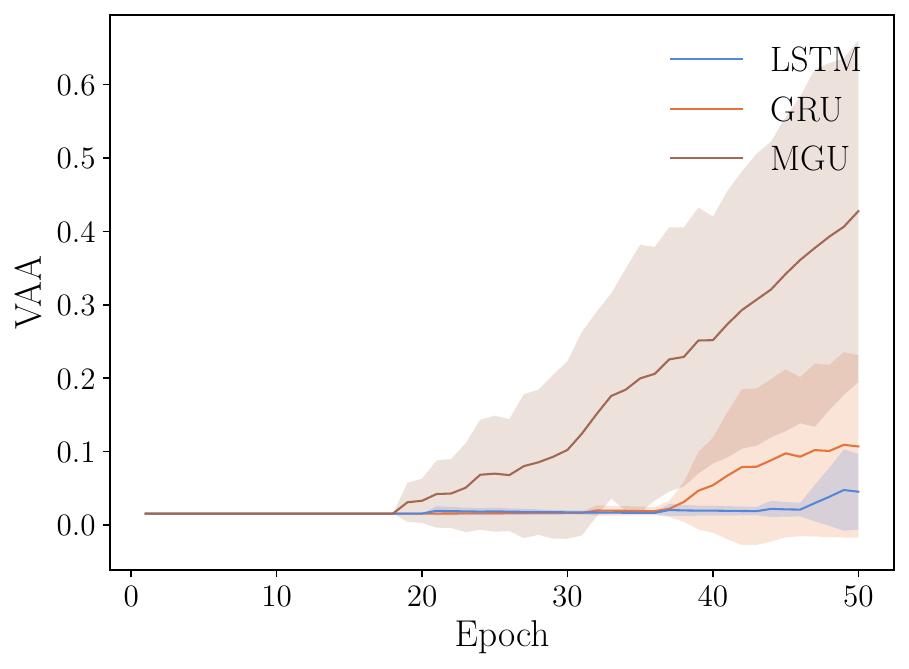}
        \caption{$S = 2, H = 64, \alpha = 1e-3, B = 64$}
    \end{subfigure}
    \begin{subfigure}{.49\textwidth}
        \centering
        \includegraphics[width=.49\textwidth]{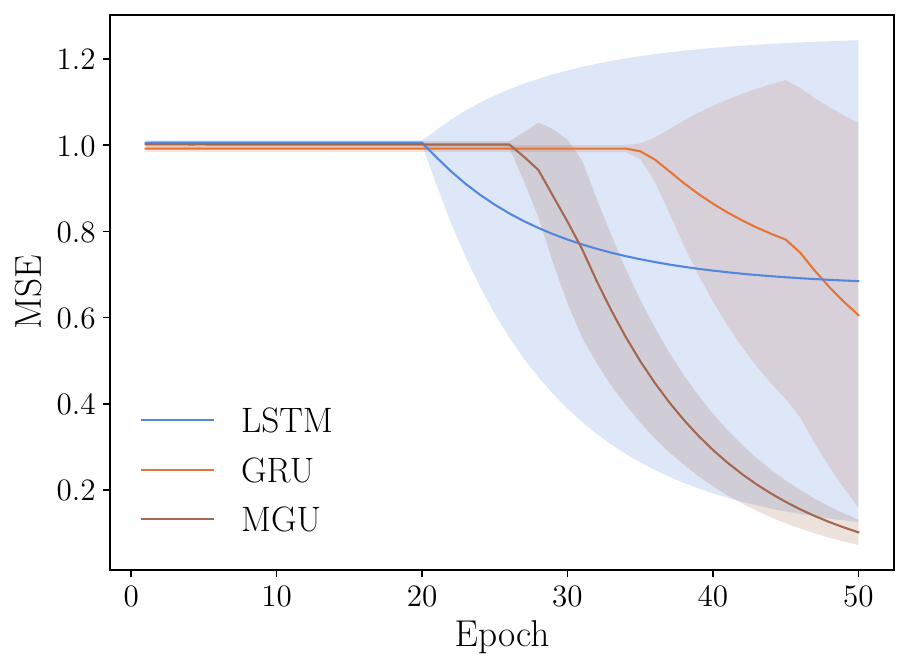}
        \includegraphics[width=.49\textwidth]{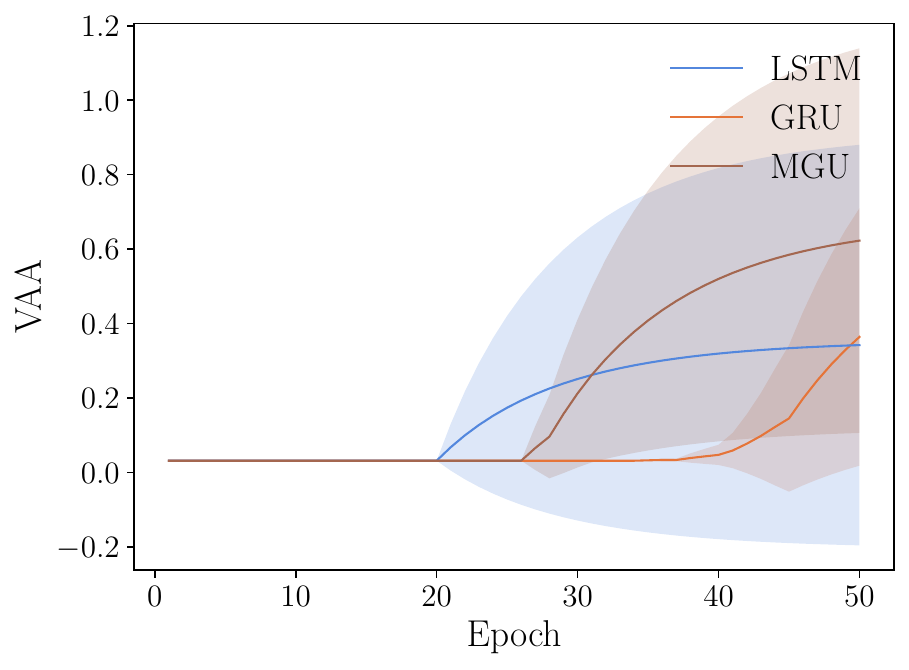}
        \caption{$S = 2, H = 64, \alpha = 1e-4, B = 32$}
    \end{subfigure}
    \begin{subfigure}{.49\textwidth}
        \centering
        \includegraphics[width=.49\textwidth]{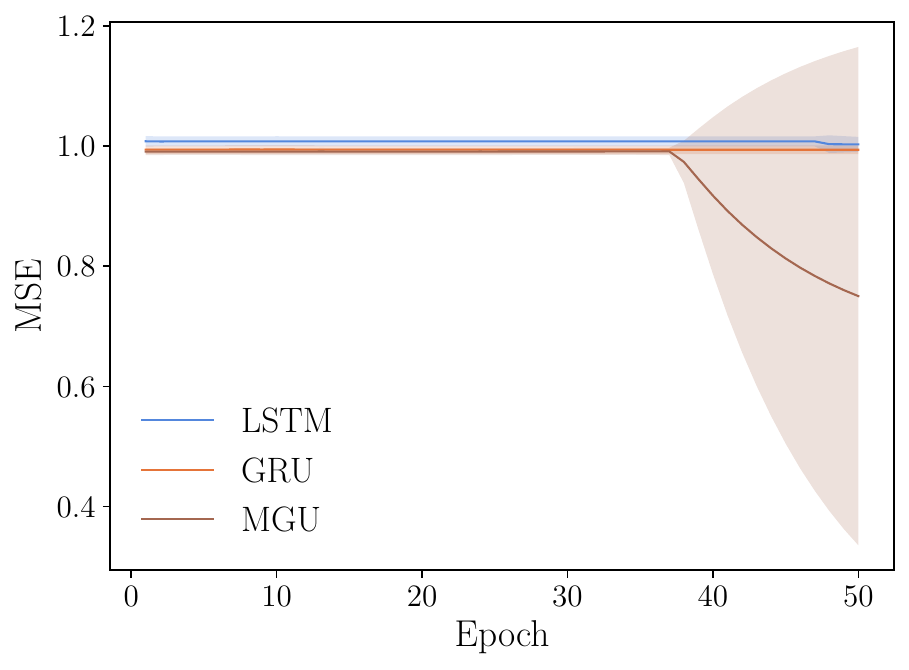}
        \includegraphics[width=.49\textwidth]{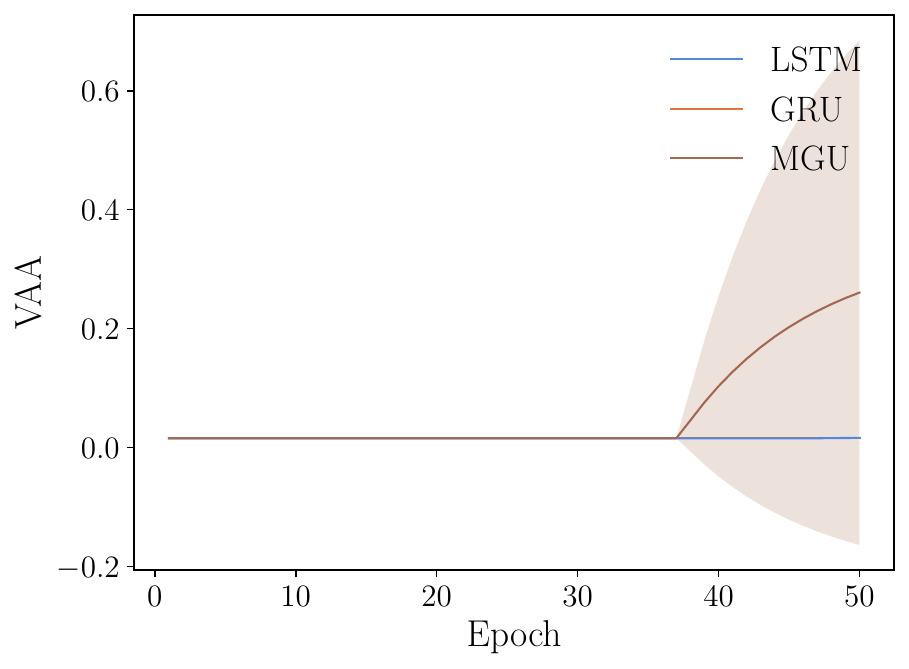}
        \caption{$S = 2, H = 64, \alpha = 1e-4, B = 64$}
    \end{subfigure}
    \begin{subfigure}{.49\textwidth}
        \centering
        \includegraphics[width=.49\textwidth]{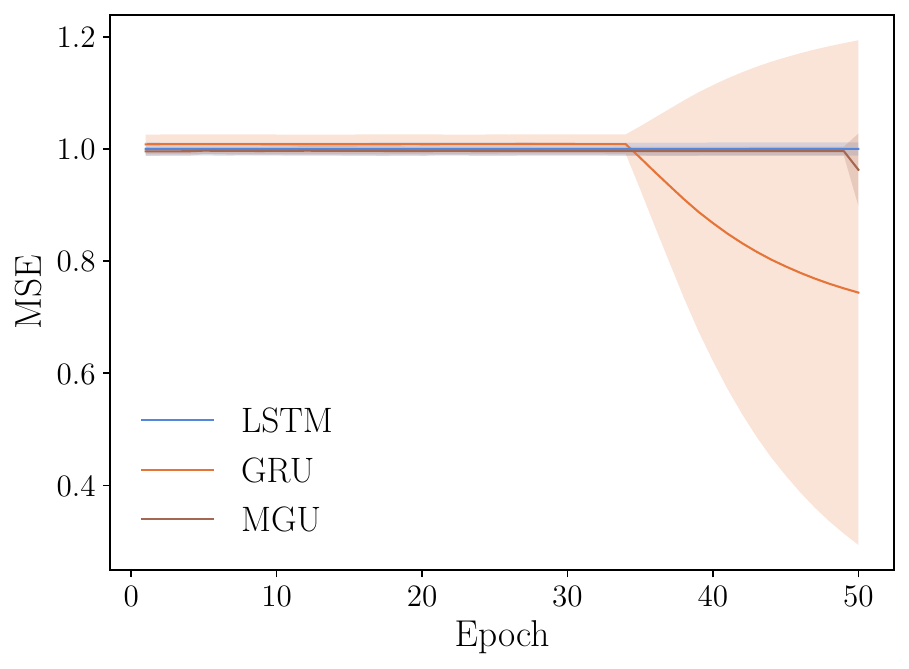}
        \includegraphics[width=.49\textwidth]{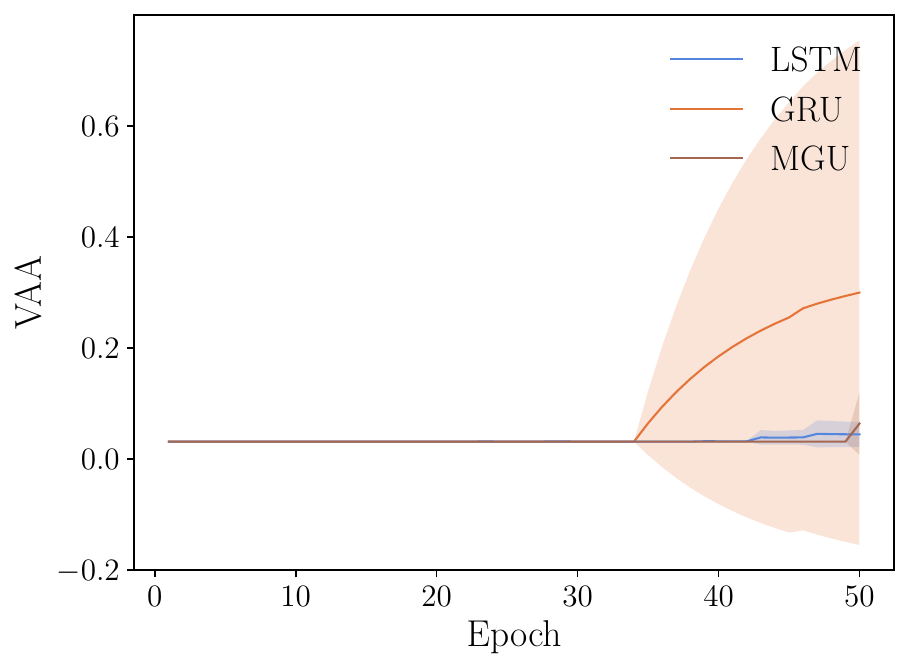}
        \caption{$S = 2, H = 256, \alpha = 1e-3, B = 32$}
    \end{subfigure}
    \begin{subfigure}{.49\textwidth}
        \centering
        \includegraphics[width=.49\textwidth]{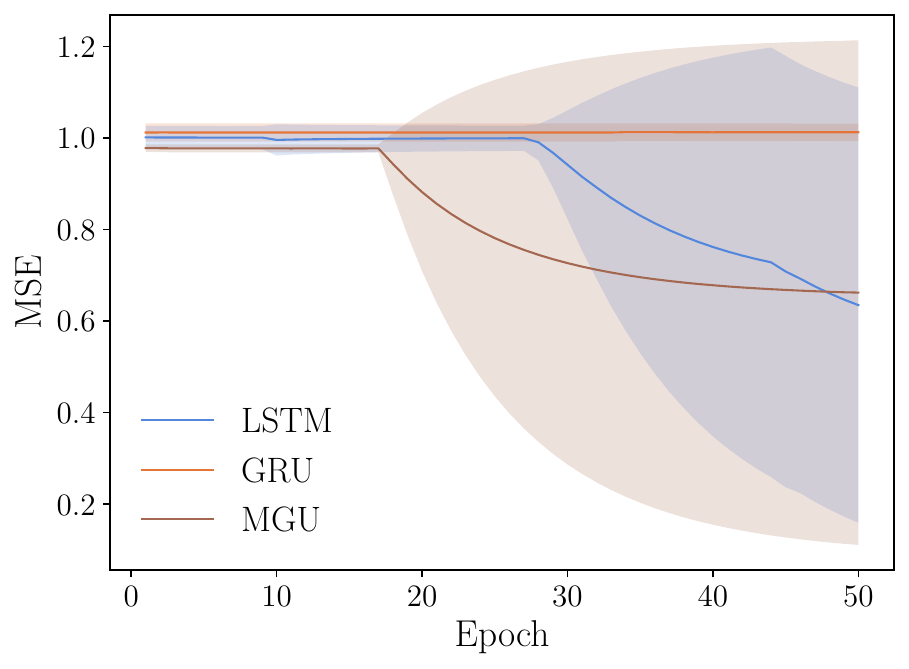}
        \includegraphics[width=.49\textwidth]{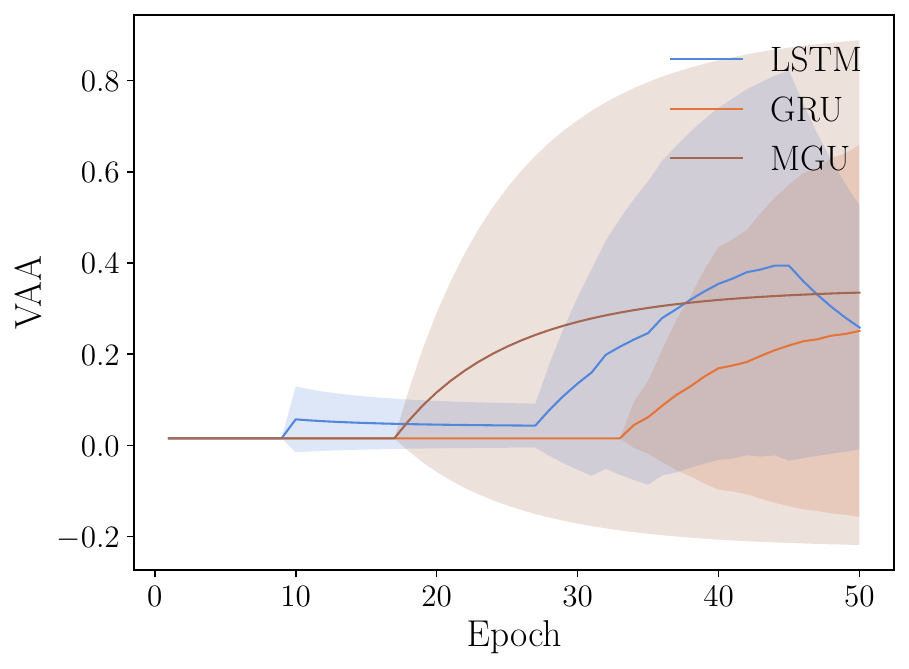}
        \caption{$S = 2, H = 256, \alpha = 1e-3, B = 64$}
    \end{subfigure}
    \begin{subfigure}{.49\textwidth}
        \centering
        \includegraphics[width=.49\textwidth]{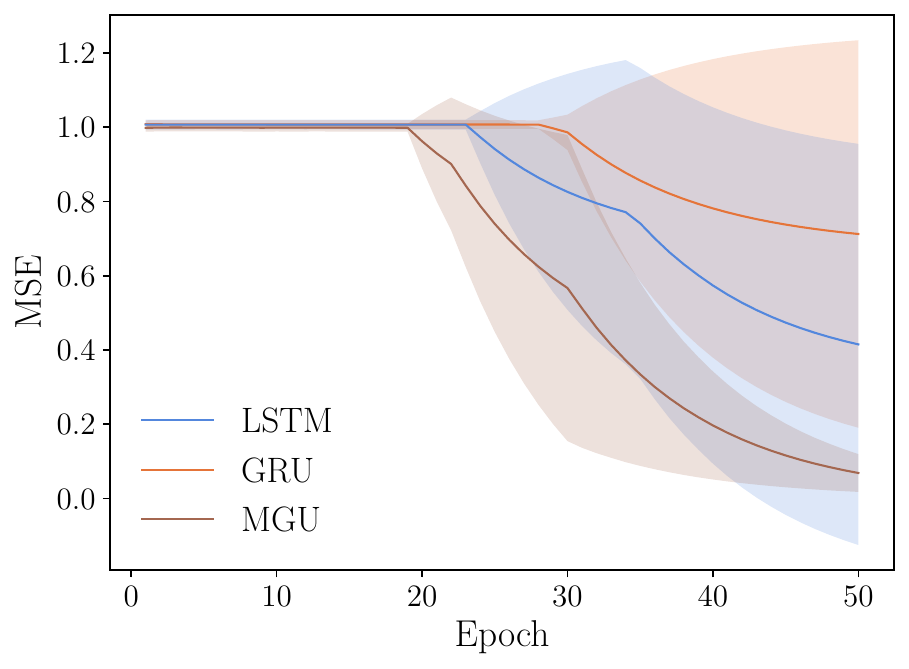}
        \includegraphics[width=.49\textwidth]{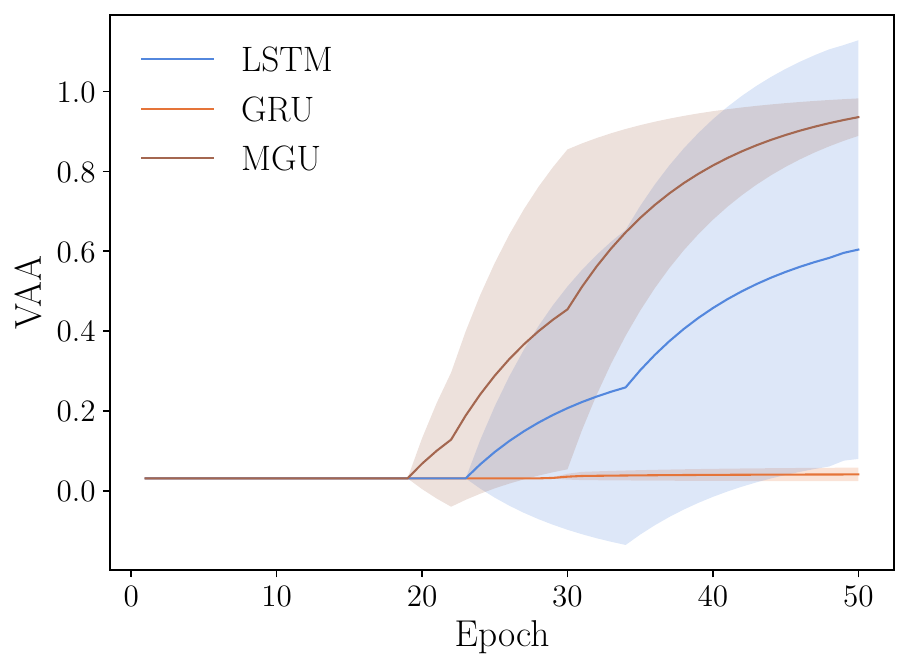}
        \caption{$S = 2, H = 256, \alpha = 1e-4, B = 32$}
    \end{subfigure}
    \begin{subfigure}{.49\textwidth}
        \centering
        \includegraphics[width=.49\textwidth]{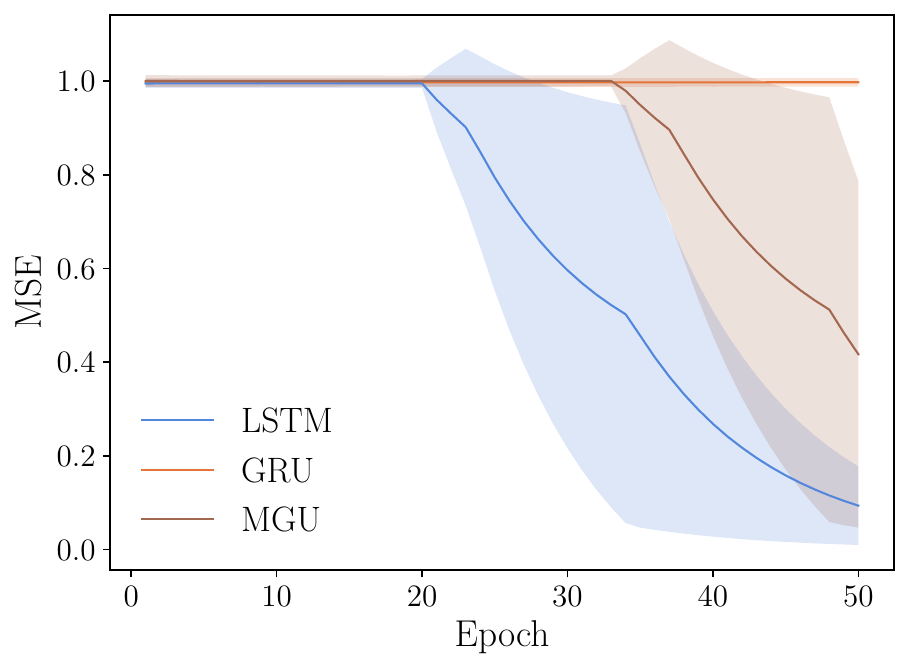}
        \includegraphics[width=.49\textwidth]{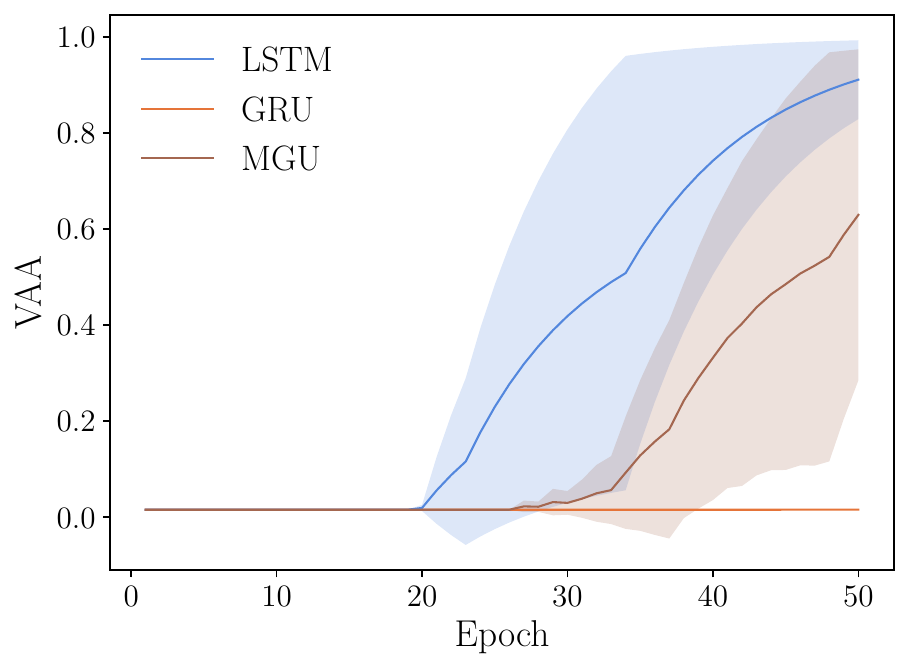}
        \caption{$S = 2, H = 256, \alpha = 1e-4, B = 64$}
    \end{subfigure}
    \caption{%
        Evolution of the validation loss (left) and of the VAA (right) of LSTM, GRU and MGU networks, for the copy first input benchmark.}
    \label{fig:copyfirstinput_generalisation_vaa_2}
\end{figure}

In \autoref{fig:copyfirstinput_generalisation_vaa_1} and \autoref{fig:copyfirstinput_generalisation_vaa_2}, we can see the evolution of the loss on the validation set and of the VAA for different hyperparameters. There is a clear correlation between learning and multistability, for all choices of hyperparameters. More precisely, it can be seen that learning loss decrease generally starts when the VAA starts increasing. Moreover, the loss is highly correlated with the VAA.

\subsection{Generalisation of the warmup procedure} \label{app:generalisation_warmup}

\begin{figure}[p]
    \centering
    \begin{subfigure}{.49\textwidth}
        \centering
        \includegraphics[width=.49\textwidth]{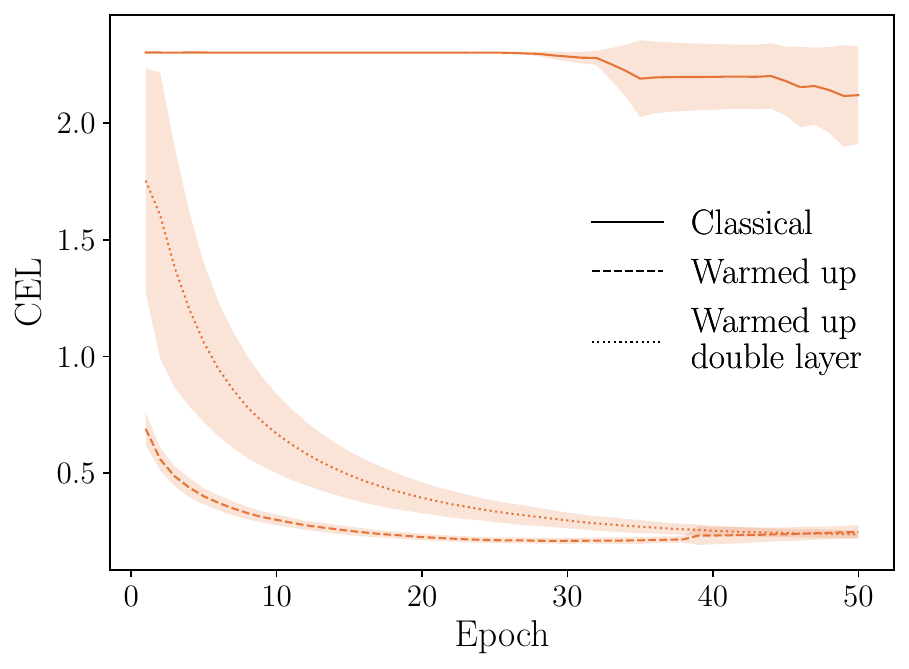}
        \includegraphics[width=.49\textwidth]{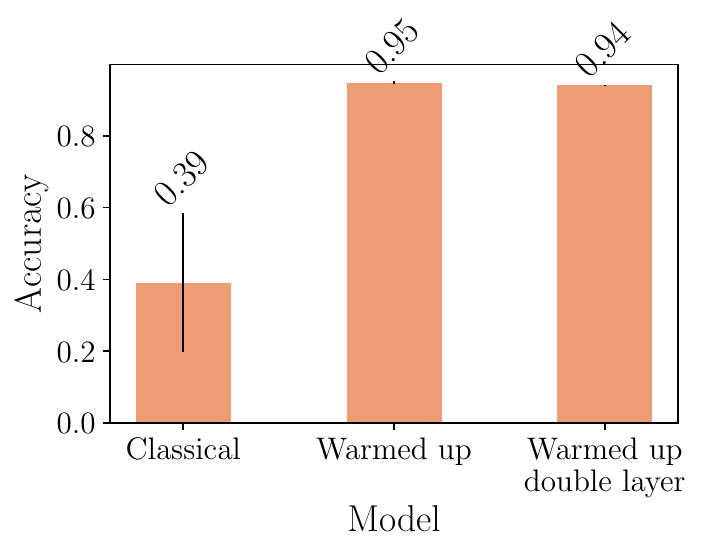}
        \caption{$S = 1, H = 64, \alpha = 1e-3, B = 32$}
    \end{subfigure}
    \begin{subfigure}{.49\textwidth}
        \centering
        \includegraphics[width=.49\textwidth]{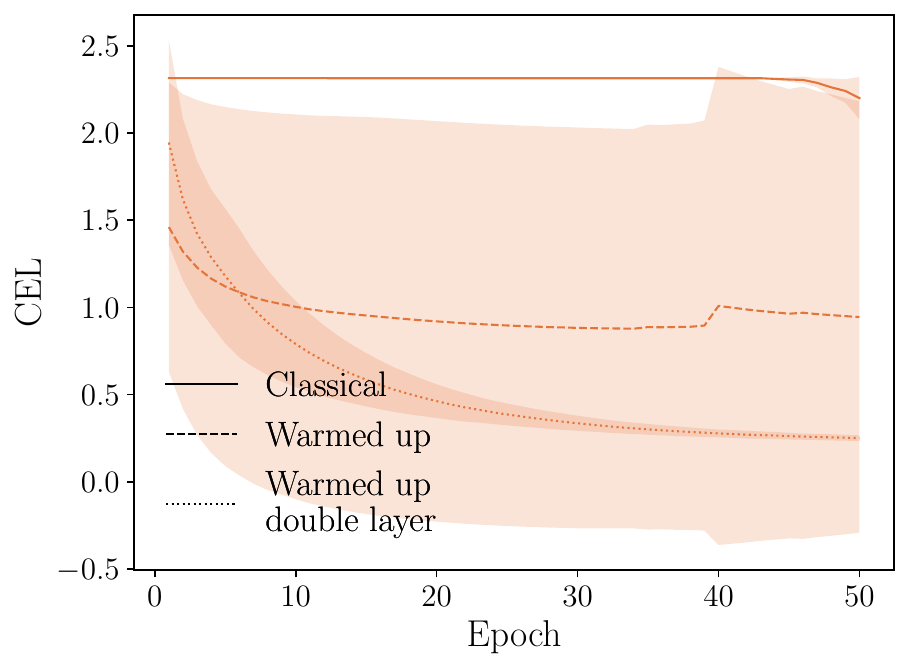}
        \includegraphics[width=.49\textwidth]{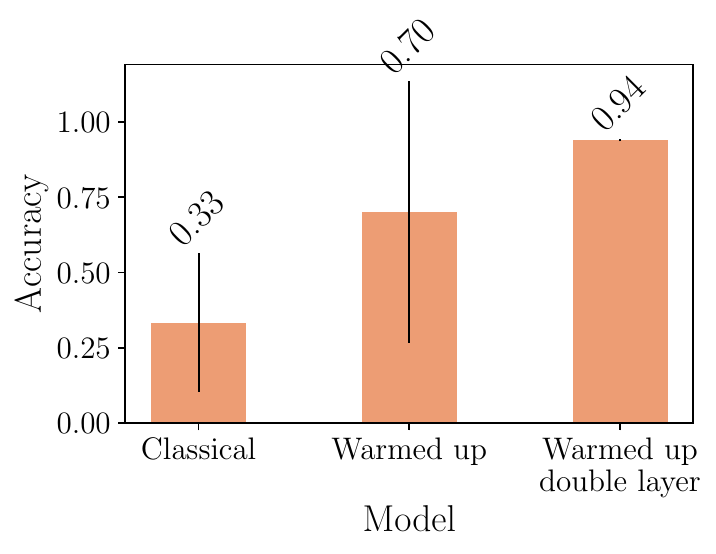}
        \caption{$S = 1, H = 64, \alpha = 1e-3, B = 64$}
    \end{subfigure}
    \begin{subfigure}{.49\textwidth}
        \centering
        \includegraphics[width=.49\textwidth]{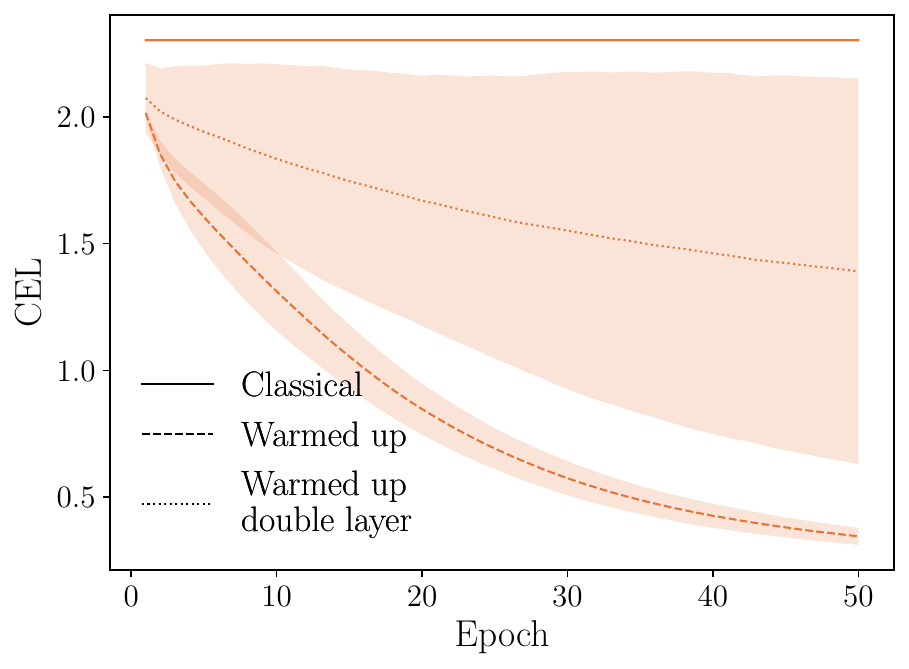}
        \includegraphics[width=.49\textwidth]{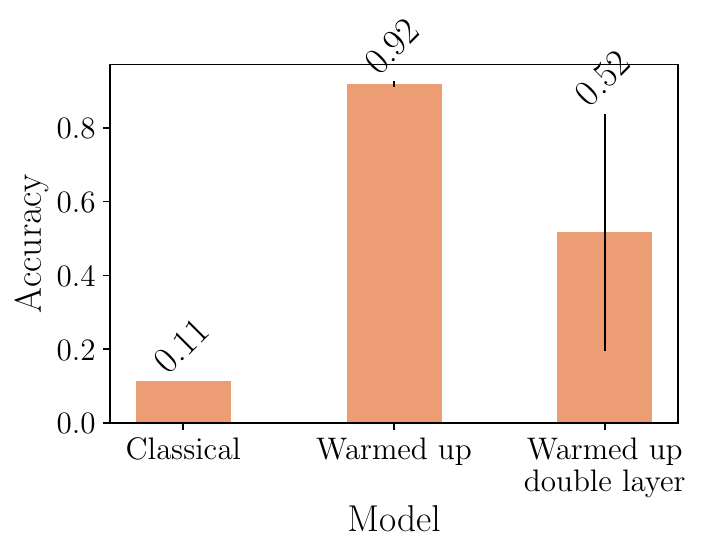}
        \caption{$S = 1, H = 64, \alpha = 1e-4, B = 32$}
    \end{subfigure}
    \begin{subfigure}{.49\textwidth}
        \centering
        \includegraphics[width=.49\textwidth]{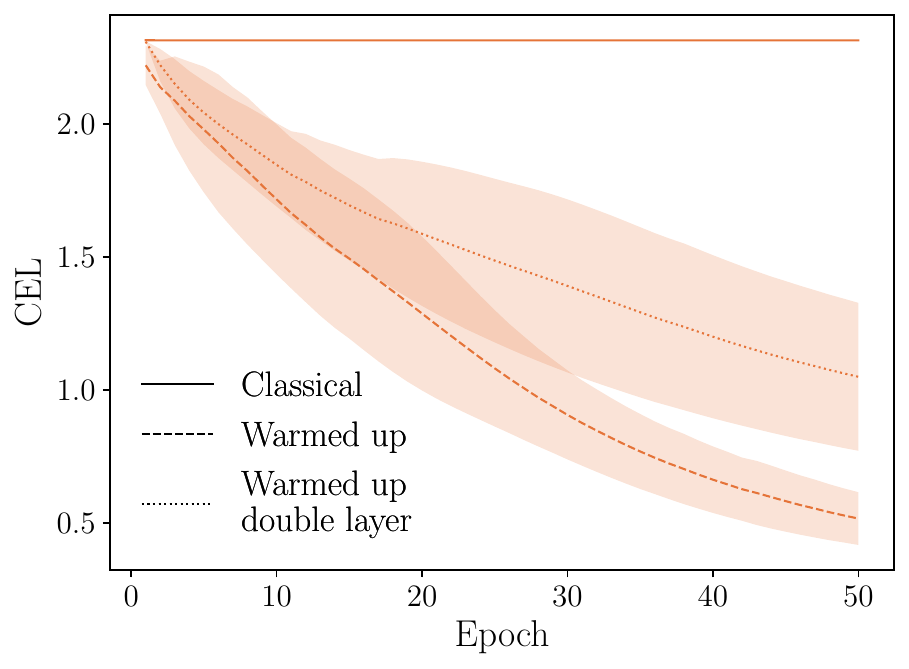}
        \includegraphics[width=.49\textwidth]{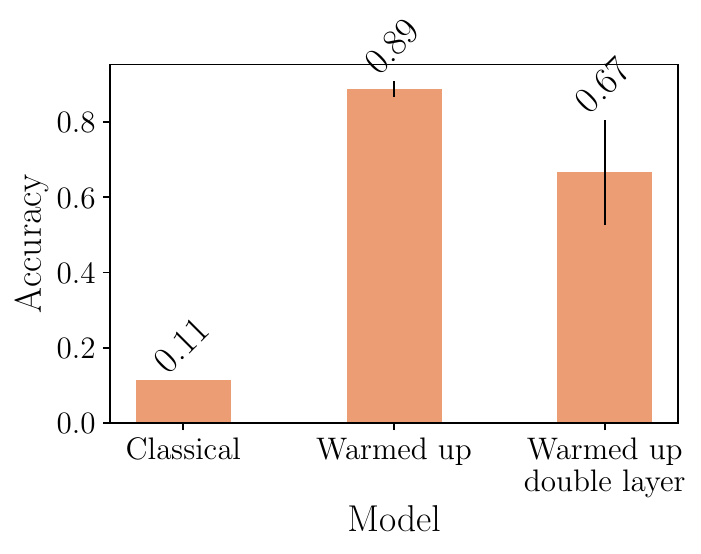}
        \caption{$S = 1, H = 64, \alpha = 1e-4, B = 64$}
    \end{subfigure}
    \begin{subfigure}{.49\textwidth}
        \centering
        \includegraphics[width=.49\textwidth]{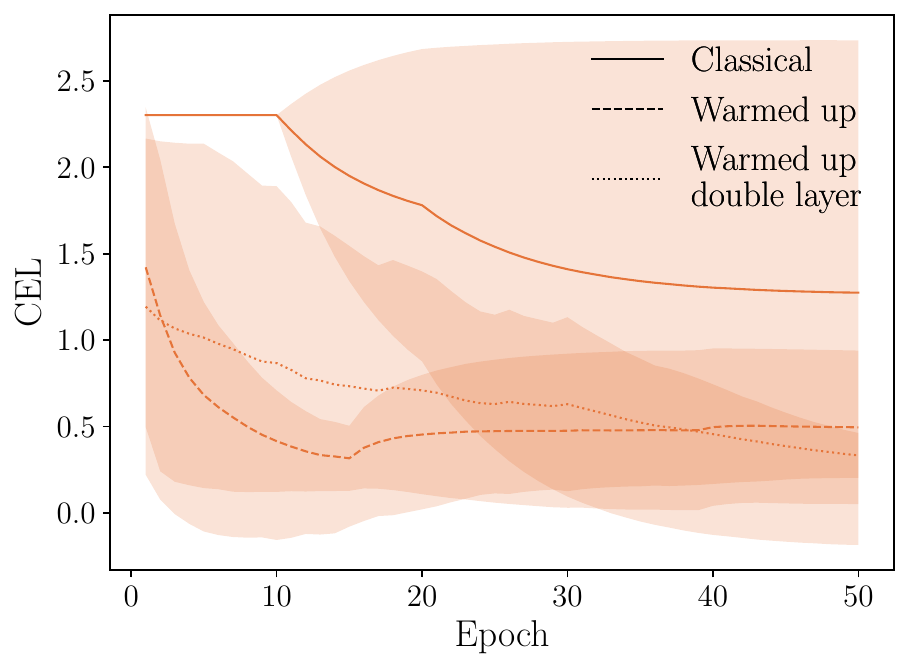}
        \includegraphics[width=.49\textwidth]{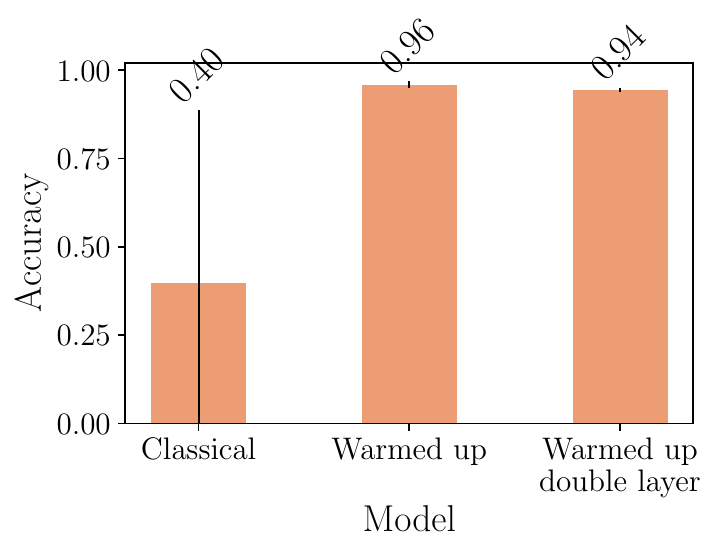}
        \caption{$S = 1, H = 256, \alpha = 1e-3, B = 32$}
    \end{subfigure}
    \begin{subfigure}{.49\textwidth}
        \centering
        \includegraphics[width=.49\textwidth]{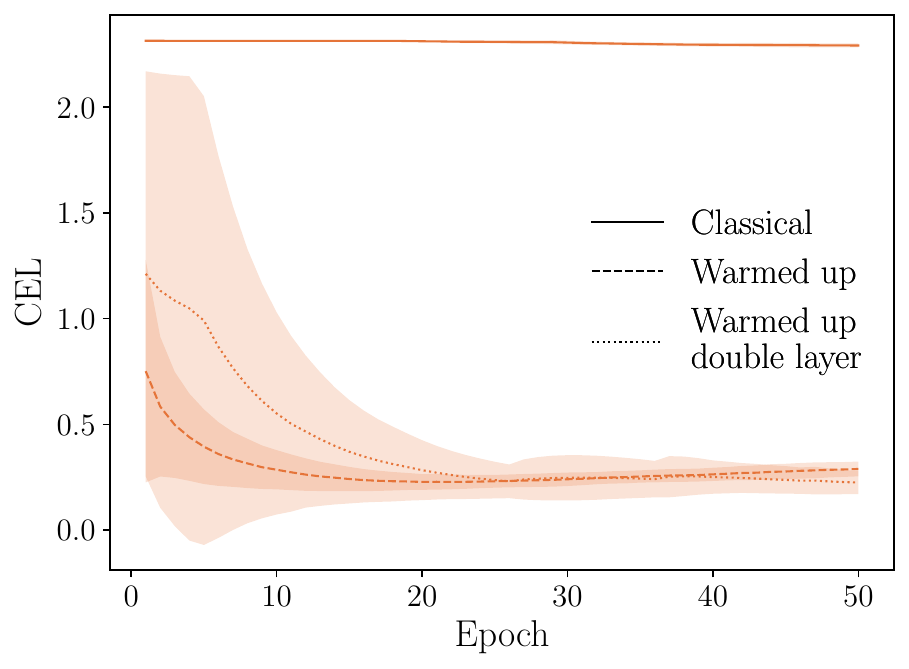}
        \includegraphics[width=.49\textwidth]{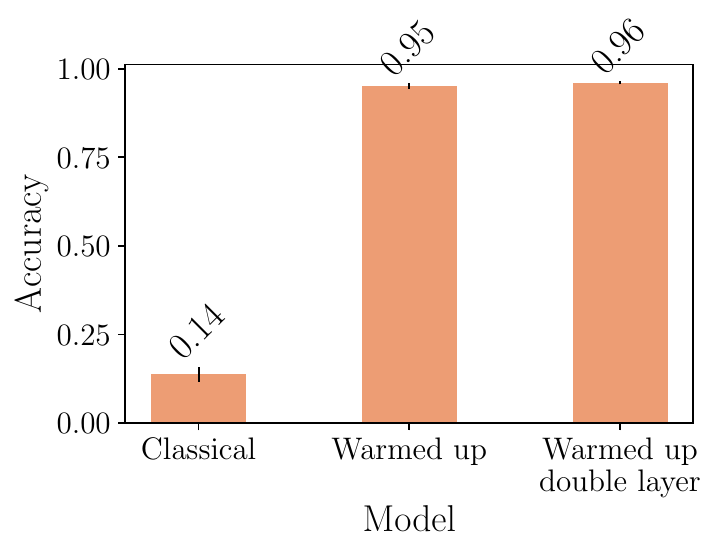}
        \caption{$S = 1, H = 256, \alpha = 1e-3, B = 64$}
    \end{subfigure}
    \begin{subfigure}{.49\textwidth}
        \centering
        \includegraphics[width=.49\textwidth]{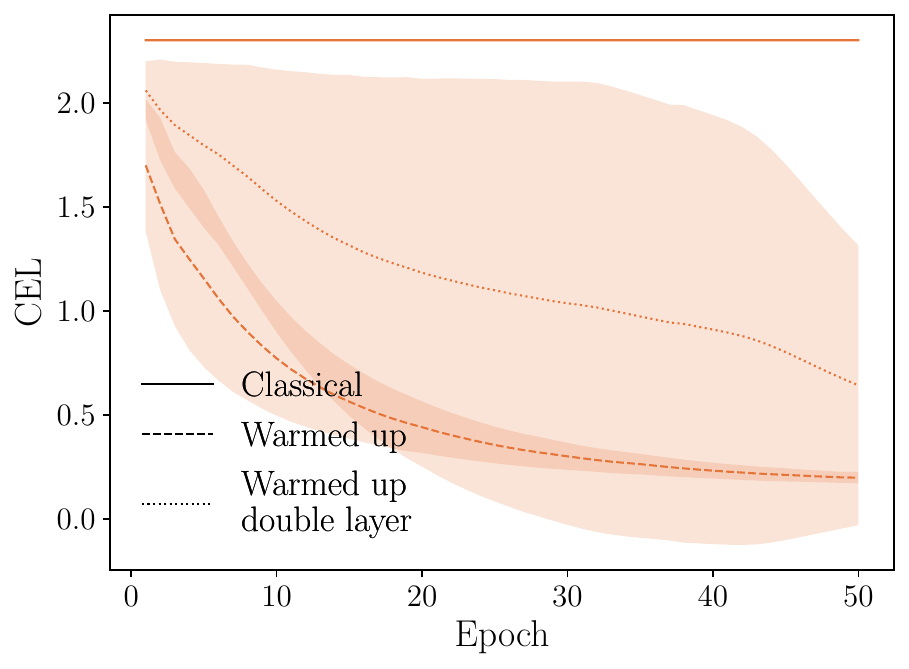}
        \includegraphics[width=.49\textwidth]{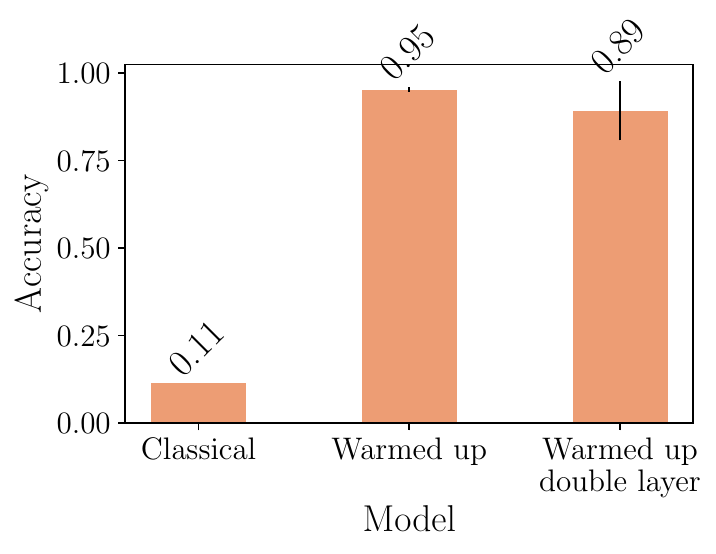}
        \caption{$S = 1, H = 256, \alpha = 1e-4, B = 32$}
    \end{subfigure}
    \begin{subfigure}{.49\textwidth}
        \centering
        \includegraphics[width=.49\textwidth]{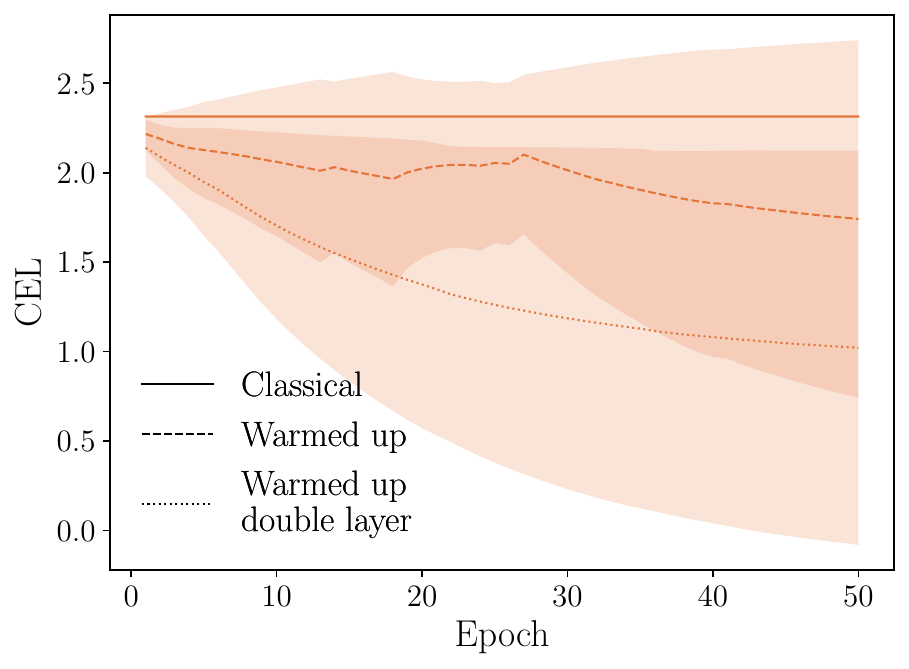}
        \includegraphics[width=.49\textwidth]{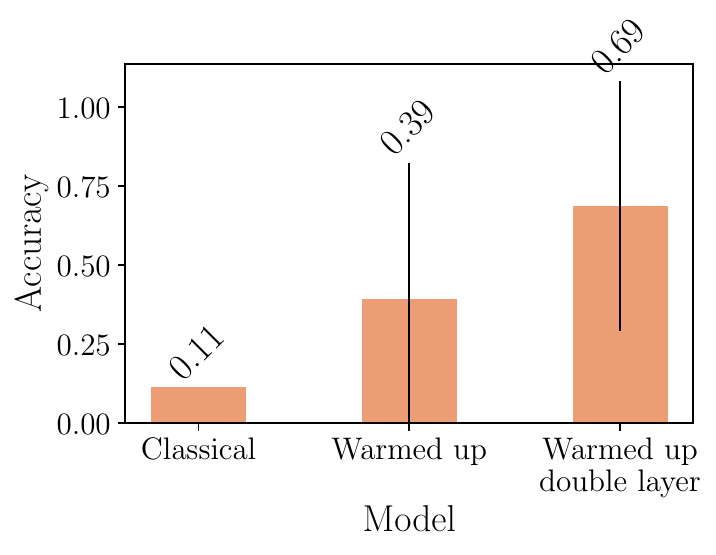}
        \caption{$S = 1, H = 256, \alpha = 1e-4, B = 64$}
    \end{subfigure}
    \caption{%
        Evolution of the validation loss (left) and test set accuracy after 50 epochs (right) of GRU networks, for the permuted line-sequential MNIST benchmark with $N = 472$.}
    \label{fig:copyfirstinput_generalisation_warmup_1}
\end{figure}

\begin{figure}[p]
    \vspace{6ex}
    \begin{subfigure}{.49\textwidth}
        \centering
        \includegraphics[width=.49\textwidth]{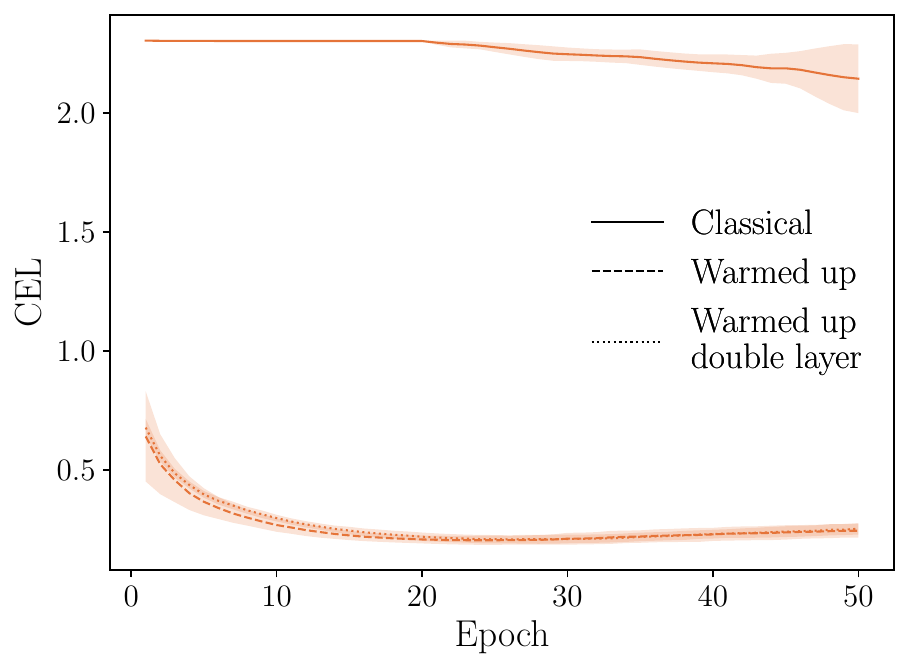}
        \includegraphics[width=.49\textwidth]{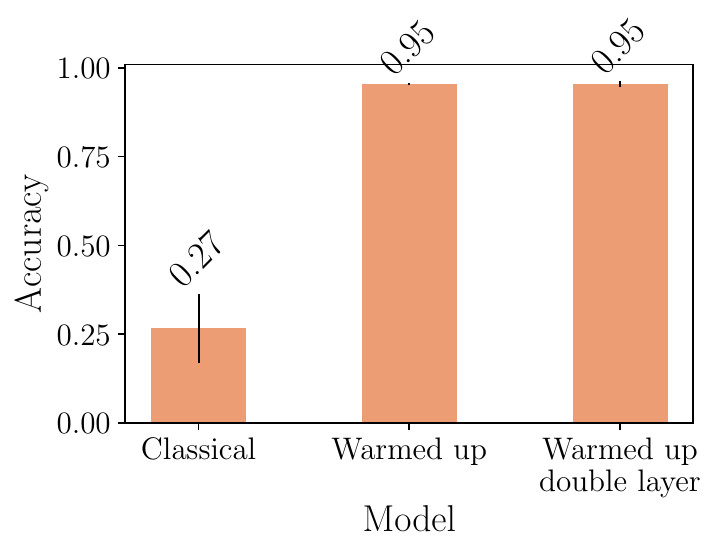}
        \caption{$S = 2, H = 64, \alpha = 1e-3, B = 32$}
    \end{subfigure}
    \begin{subfigure}{.49\textwidth}
        \centering
        \includegraphics[width=.49\textwidth]{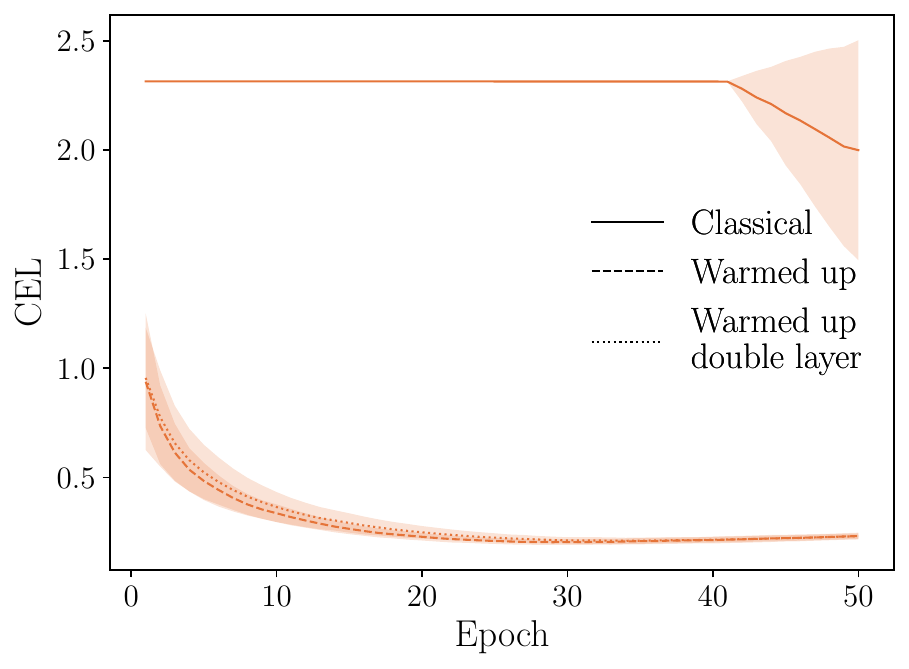}
        \includegraphics[width=.49\textwidth]{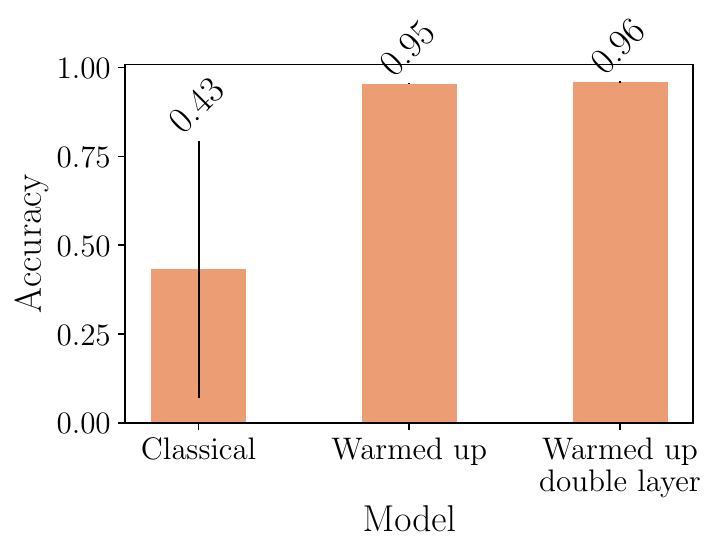}
        \caption{$S = 2, H = 64, \alpha = 1e-3, B = 64$}
    \end{subfigure}
    \begin{subfigure}{.49\textwidth}
        \centering
        \includegraphics[width=.49\textwidth]{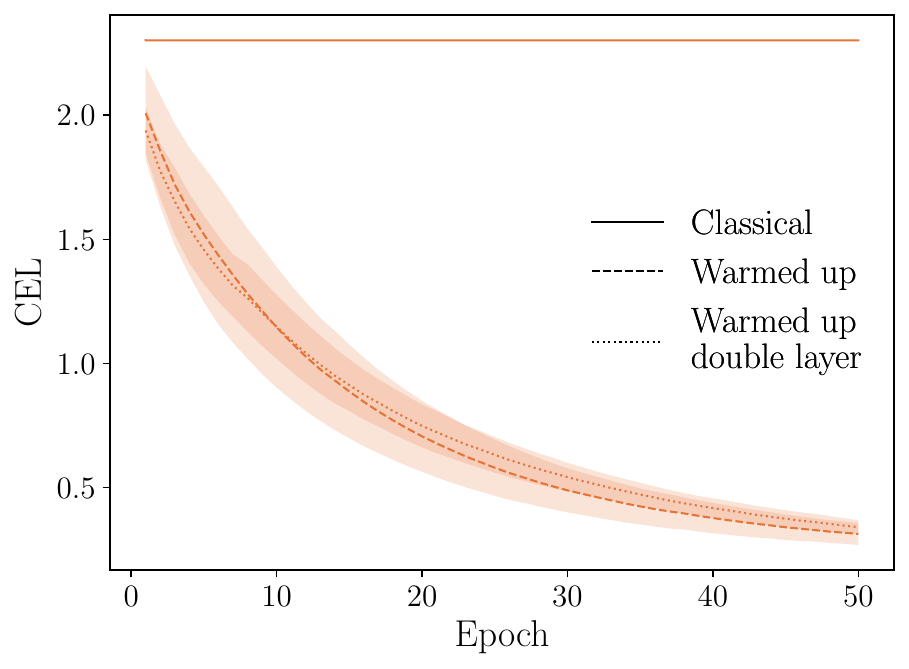}
        \includegraphics[width=.49\textwidth]{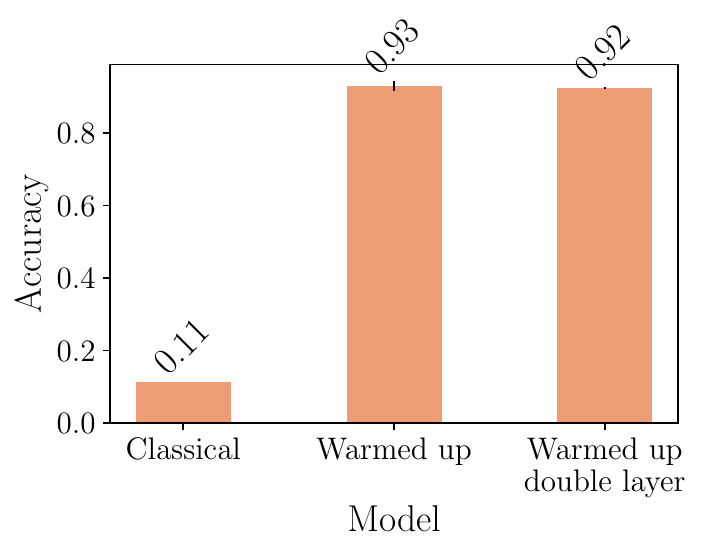}
        \caption{$S = 2, H = 64, \alpha = 1e-4, B = 32$}
    \end{subfigure}
    \begin{subfigure}{.49\textwidth}
        \centering
        \includegraphics[width=.49\textwidth]{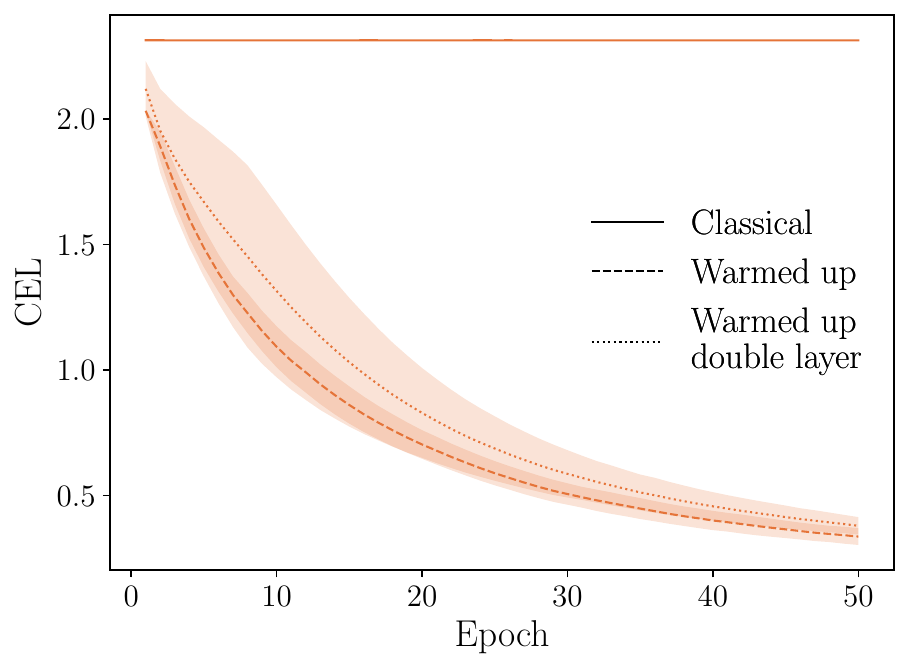}
        \includegraphics[width=.49\textwidth]{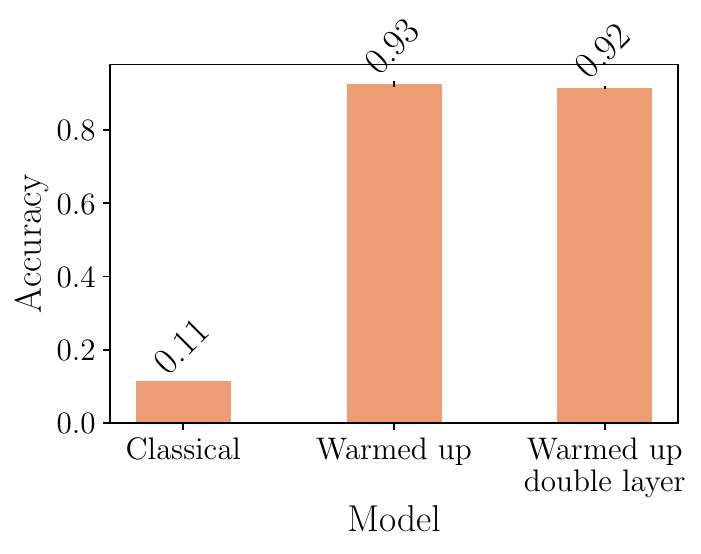}
        \caption{$S = 2, H = 64, \alpha = 1e-4, B = 64$}
    \end{subfigure}
    \begin{subfigure}{.49\textwidth}
        \centering
        \includegraphics[width=.49\textwidth]{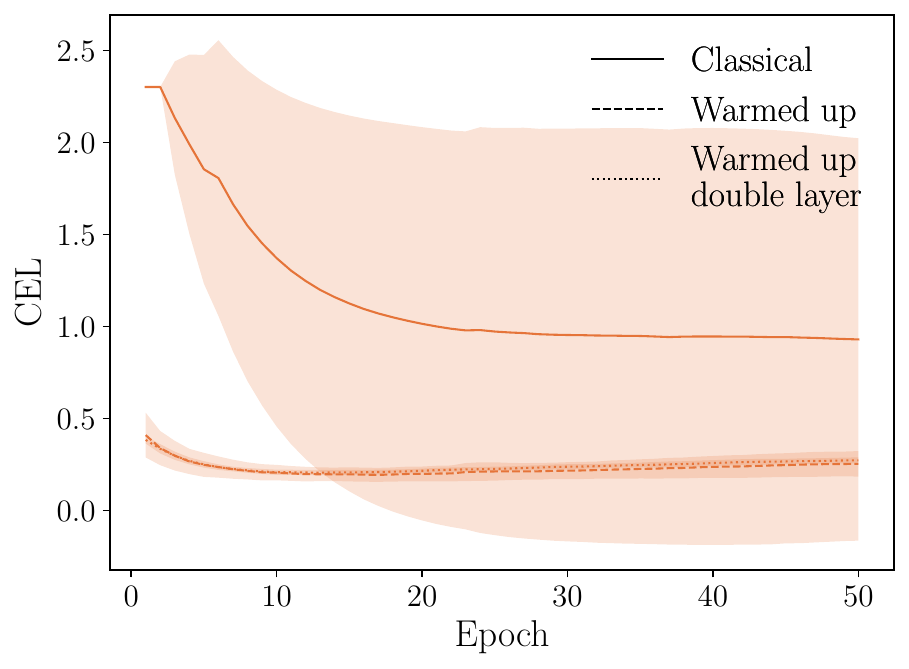}
        \includegraphics[width=.49\textwidth]{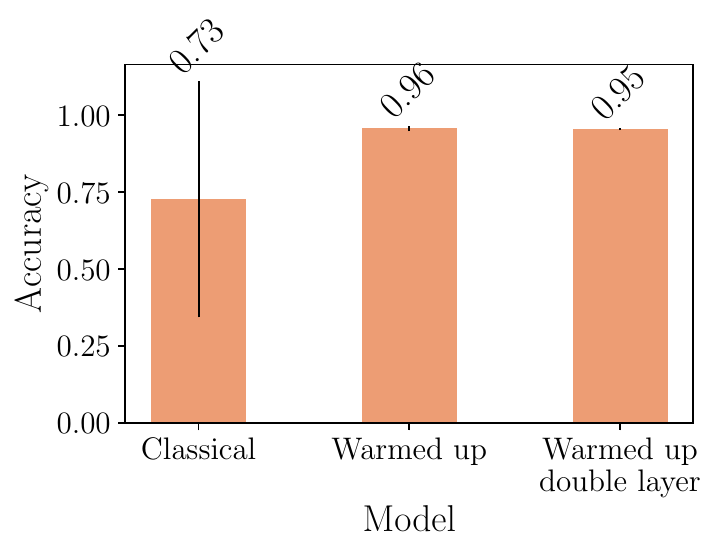}
        \caption{$S = 2, H = 256, \alpha = 1e-3, B = 32$}
    \end{subfigure}
    \begin{subfigure}{.49\textwidth}
        \centering
        \includegraphics[width=.49\textwidth]{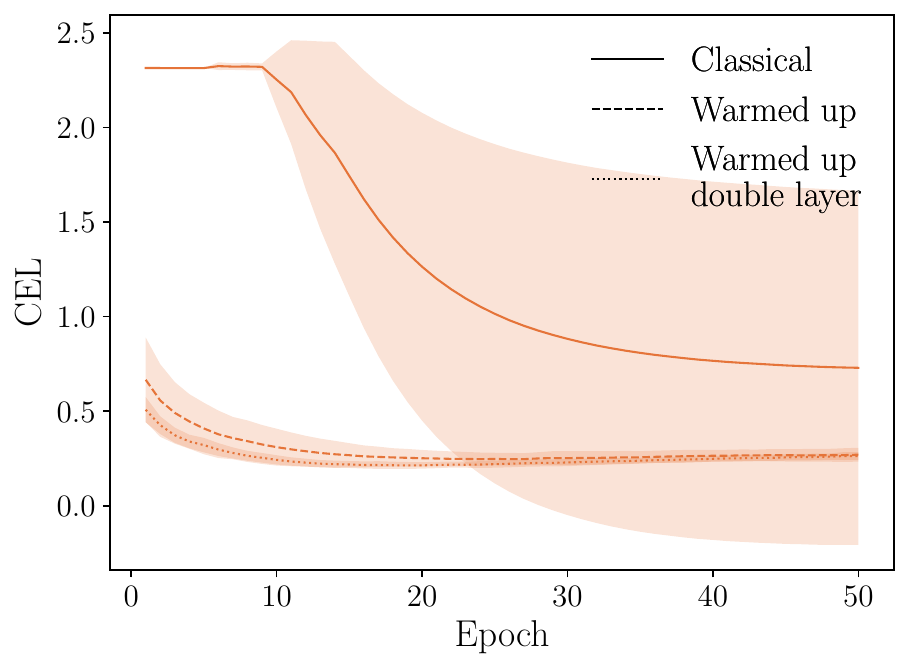}
        \includegraphics[width=.49\textwidth]{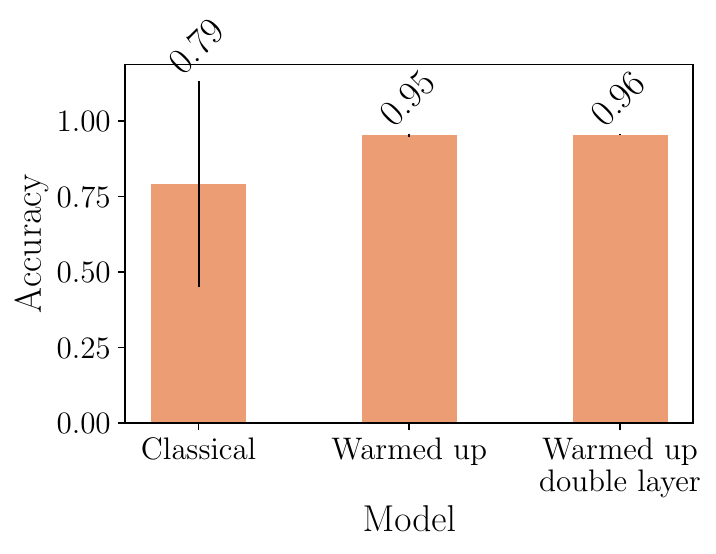}
        \caption{$S = 2, H = 256, \alpha = 1e-3, B = 64$}
    \end{subfigure}
    \begin{subfigure}{.49\textwidth}
        \centering
        \includegraphics[width=.49\textwidth]{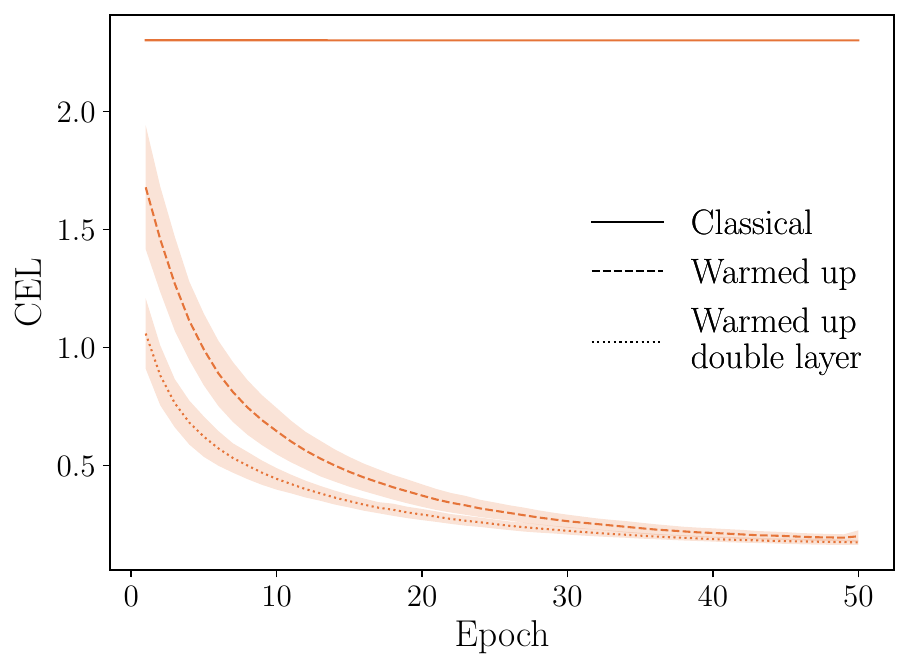}
        \includegraphics[width=.49\textwidth]{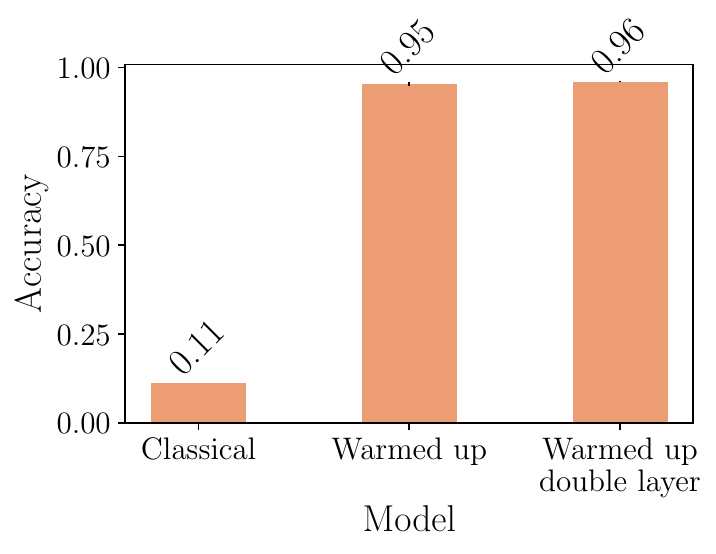}
        \caption{$S = 2, H = 256, \alpha = 1e-4, B = 32$}
    \end{subfigure}
    \begin{subfigure}{.49\textwidth}
        \centering
        \includegraphics[width=.49\textwidth]{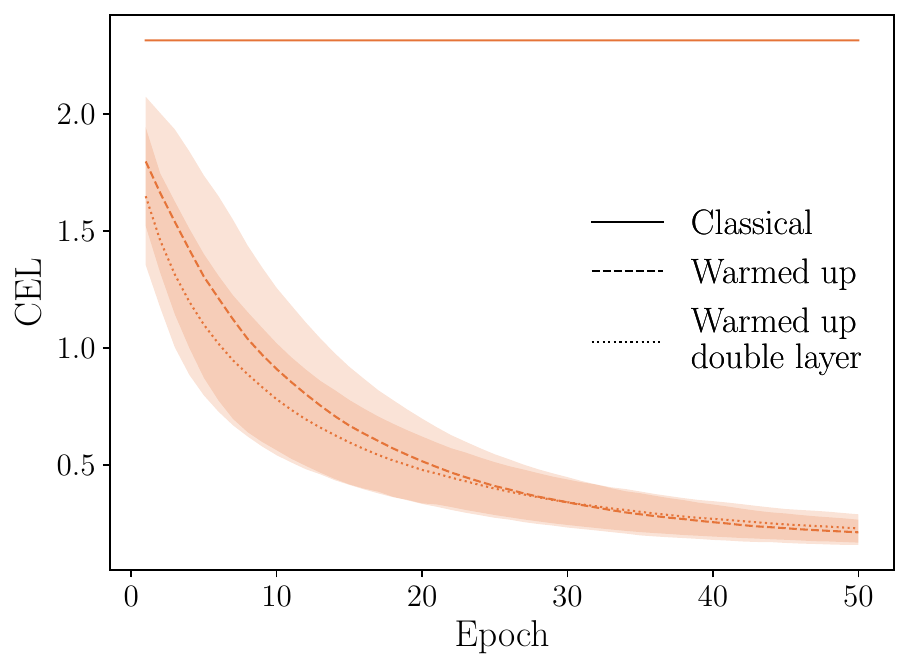}
        \includegraphics[width=.49\textwidth]{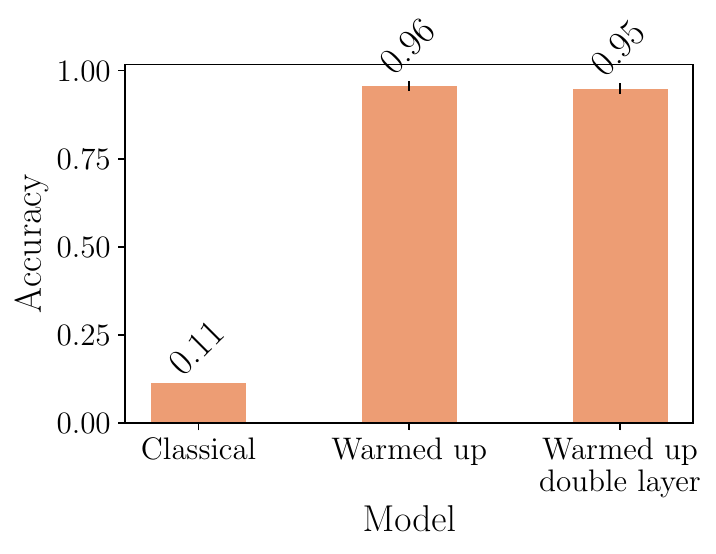}
        \caption{$S = 2, H = 256, \alpha = 1e-4, B = 64$}
    \end{subfigure}
    \caption{%
        Evolution of the validation loss (left) and test set accuracy after 50 epochs (right) of GRU networks, for the permuted line-sequential MNIST benchmark with $N = 472$.}
    \label{fig:copyfirstinput_generalisation_warmup_2}
    \vspace{6ex}
\end{figure}

In \autoref{fig:copyfirstinput_generalisation_warmup_1} and \autoref{fig:copyfirstinput_generalisation_warmup_2}, we can see the evolution of the loss on the validation set and the test set accuracy for different hyperparameters.
It can be seen from those figures that the warmup procedure and the double layer architecture with partial warmup both improve on the classically initialised GRU architecture.
Those improvements are consistent over all hyperparameters choices.
It can be noted that the warmup procedure is sometimes better than the double layer architecture in terms of speed of convergence, notably when using a single RNN layer and a small hidden size.

\subsection{Impact of the parameter \texorpdfstring{$k$}{k} in the warmup procedure} \label{app:impact_k}

\begin{figure}[p]
    \centering
    \begin{subfigure}{.32\textwidth}
        \centering
        \includegraphics[width=\linewidth]{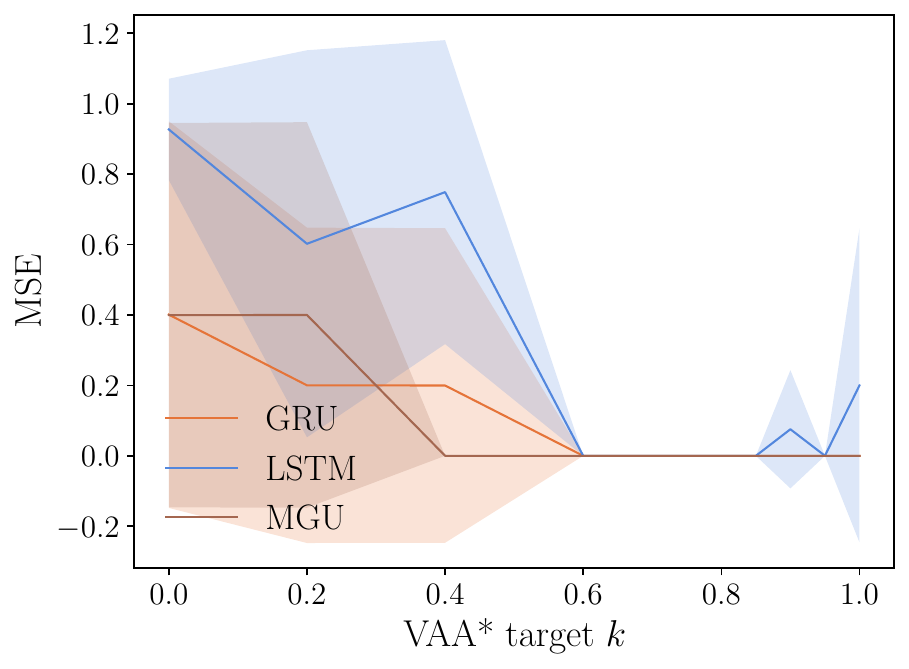}
        \caption{$T = 50$}
    \end{subfigure}
    \begin{subfigure}{.32\textwidth}
        \centering
        \includegraphics[width=\linewidth]{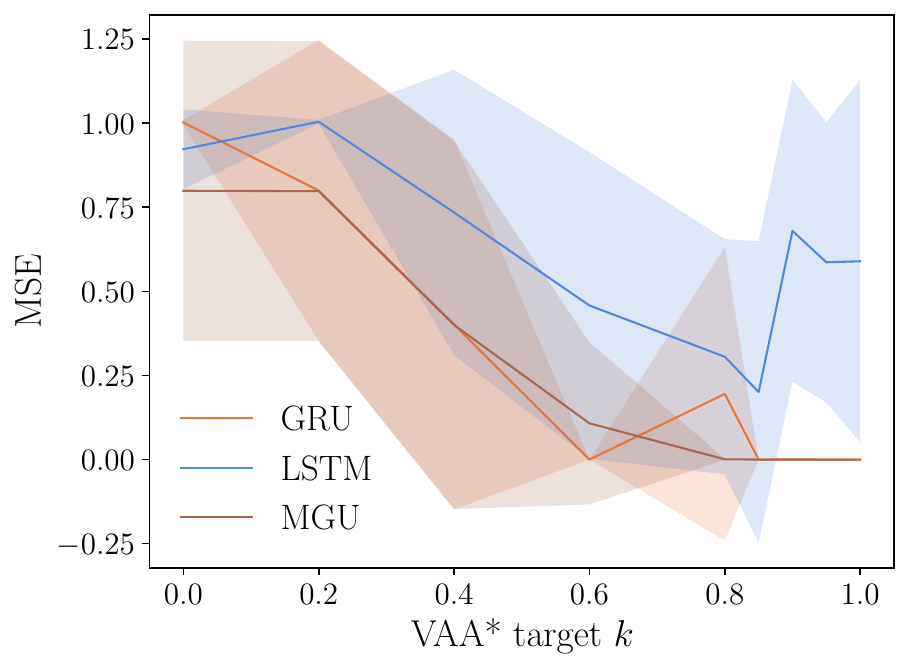}
        \caption{$T = 300$}
    \end{subfigure}
    \begin{subfigure}{.32\textwidth}
        \centering
        \includegraphics[width=\linewidth]{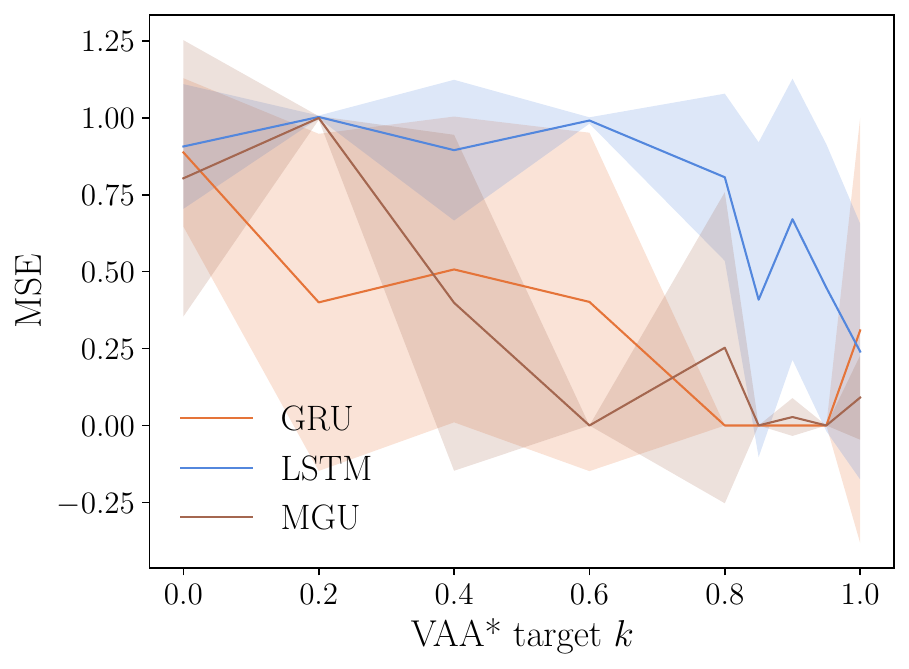}
        \caption{$T = 600$}
    \end{subfigure}
    \caption{Mean squared error (± standard deviation) of different architecture for different value of target \vaastar{} $k$ on the copy first input test set for different values of $T$.}
    \label{fig:impact_k}
\end{figure}

In \autoref{fig:impact_k}, we can see the impact of the target \vaastar{} $k$ used in the warmup procedure on the final test loss, for the copy first input benchmark for different sequence lengths $T$. It can be seen that for this benchmark with long time dependencies, the higher $k$, the lower the MSE.

\section{Hyperparameters optimization} \label{app:hyperparameters}

In this section, we report the best hyperparameters obtained for each cell version and the final test loss obtained for those hyperparameters, in \autoref{tab:hp_denoising} for the denoising benchmark with $N = 100$ and in \autoref{tab:hp_mnist} for the line-sequential MNIST benchmark with $N = 472$.
The hyperparameter selection procedure is described hereafter.
First, the dataset is split into the learning set and the test set.
Then, the learning set is split into three sets: the training set, the validation set and the selection set.
The network is then trained according to the standard procedure: the final weights are those that have obtained the lowest loss on the validation set, throughout the training on the training set.
Those weights are then evaluated on the selection set.
This procedure is repeated five times for each set of hyperparameters, with different splits of the learning set each time.
Note that those $5$ different splits are the same for all cell versions.
The set of hyperparameters having obtained the lowest loss on average on the selection set is selected.
Using those hyperparameters, the cells are then trained $5$ times on the learning set, using a standard training-validation split, and the average score obtained on the test set is reported.
The sets of hyperparameters that are considered are given by a grid search.

\begin{table}[h]
    \caption{Denoising benchmark: Test MSE after hyperparameter selection}
    \label{tab:hp_denoising}
    \centering
    \begin{tabular}{ccccccc}
    \toprule
     &  & $L$ & $H$ & $\alpha$ & $B$ & MSE \\
    \midrule
    \multirow[c]{3}{*}{LSTM} & Classical & 3 & 512 & \num{1e-03} & 32 & $0.9970 \pm 0.0089$ \\
    & Double & 3 & 256 & \num{5e-04} & 64 & \boldmath $0.0004 \pm 0.0001$ \\
    & Warmup & 3 & 256 & \num{1e-03} & 64 & $0.0010 \pm 0.0009$ \\
    \cline{1-7}
    \multirow[c]{3}{*}{GRU} & Classical & 1 & 512 & \num{1e-03} & 32 & $0.2656 \pm 0.4593$ \\
    & Double & 2 & 256 & \num{1e-03} & 32 & \boldmath $0.0002 \pm 0.0001$ \\
    & Warmup & 1 & 512 & \num{5e-04} & 64 & \boldmath $0.0002 \pm 0.0001$ \\
    \cline{1-7}
    \multirow[c]{3}{*}{MGU} & Classical & 3 & 512 & \num{1e-03} & 32 & $0.3356 \pm 0.5695$ \\
    & Double & 2 & 256 & \num{5e-04} & 32 & \boldmath $0.0003 \pm 0.0000$ \\
    & Warmup & 1 & 256 & \num{1e-03} & 32 & \boldmath $0.0003 \pm 0.0002$ \\
    \cline{1-7}
    \multirow[c]{3}{*}{Chrono} & Classical & 1 & 512 & \num{1e-03} & 32 & \boldmath $0.0003 \pm 0.0002$ \\
    & Double & 1 & 512 & \num{1e-03} & 32 & $0.0004 \pm 0.0003$ \\
    & Warmup & 1 & 512 & \num{1e-03} & 32 & $0.0004 \pm 0.0002$ \\
    \cline{1-7}
    BRC & Classical & 3 & 256 & \num{1e-03} & 32 & \boldmath $0.0006 \pm 0.0001$ \\
    NBRC & Classical & 1 & 512 & \num{1e-03} & 32 & \boldmath $0.0001 \pm 0.0000$ \\
    \bottomrule
    \end{tabular}
\end{table}

\begin{table}[h]
    \caption{Line Sequential MNIST: Test accuracy after hyperparameter selection}
    \label{tab:hp_mnist}
    \centering
    \begin{tabular}{ccccccc}
    \toprule
     &  & $L$ & $H$ & $\alpha$ & $B$ & Accuracy \\
    \midrule
    \multirow[c]{3}{*}{LSTM} & Classical & 2 & 512 & \num{1e-03} & 64 & $0.6693 \pm 0.4807$ \\
    & Double & 2 & 256 & \num{5e-04} & 32 & \boldmath $0.9519 \pm 0.0058$ \\
    & Warmup & 3 & 256 & \num{5e-04} & 32 & $0.9475 \pm 0.0008$ \\
    \cline{1-7}
    \multirow[c]{3}{*}{GRU} & Classical & 3 & 512 & \num{5e-04} & 32 & \boldmath $0.9578 \pm 0.0087$ \\
    & Double & 2 & 512 & \num{1e-04} & 32 & $0.9549 \pm 0.0011$ \\
    & Warmup & 1 & 512 & \num{1e-04} & 32 & $0.9555 \pm 0.0053$ \\
    \cline{1-7}
    \multirow[c]{3}{*}{MGU} & Classical & 2 & 512 & \num{5e-04} & 64 & \boldmath $0.9576 \pm 0.0085$ \\
    & Double & 2 & 512 & \num{5e-04} & 64 & $0.9562 \pm 0.0045$ \\
    & Warmup & 2 & 256 & \num{5e-04} & 32 & $0.9485 \pm 0.0073$ \\
    \cline{1-7}
    \multirow[c]{3}{*}{Chrono} & Classical & 1 & 256 & \num{1e-03} & 32 & $0.9545 \pm 0.0020$ \\
    & Double & 1 & 256 & \num{1e-03} & 32 & \boldmath $0.9575 \pm 0.0029$ \\
    & Warmup & 1 & 256 & \num{1e-03} & 32 & $0.9562 \pm 0.0017$ \\
    \cline{1-7}
    BRC & Classical & 2 & 512 & \num{5e-04} & 32 & \boldmath $0.9589 \pm 0.0064$ \\
    NBRC & Classical & 2 & 512 & \num{5e-04} & 64 & \boldmath $0.9600 \pm 0.0006$ \\
    \bottomrule
    \end{tabular}
\end{table}

\section{Performance of the double-layer architecture without partial warmup} \label{app:dlu_control}

\begin{figure}[ht]
    \centering
    \begin{subfigure}{.49\textwidth}
        \centering
        \includegraphics[width=\textwidth]{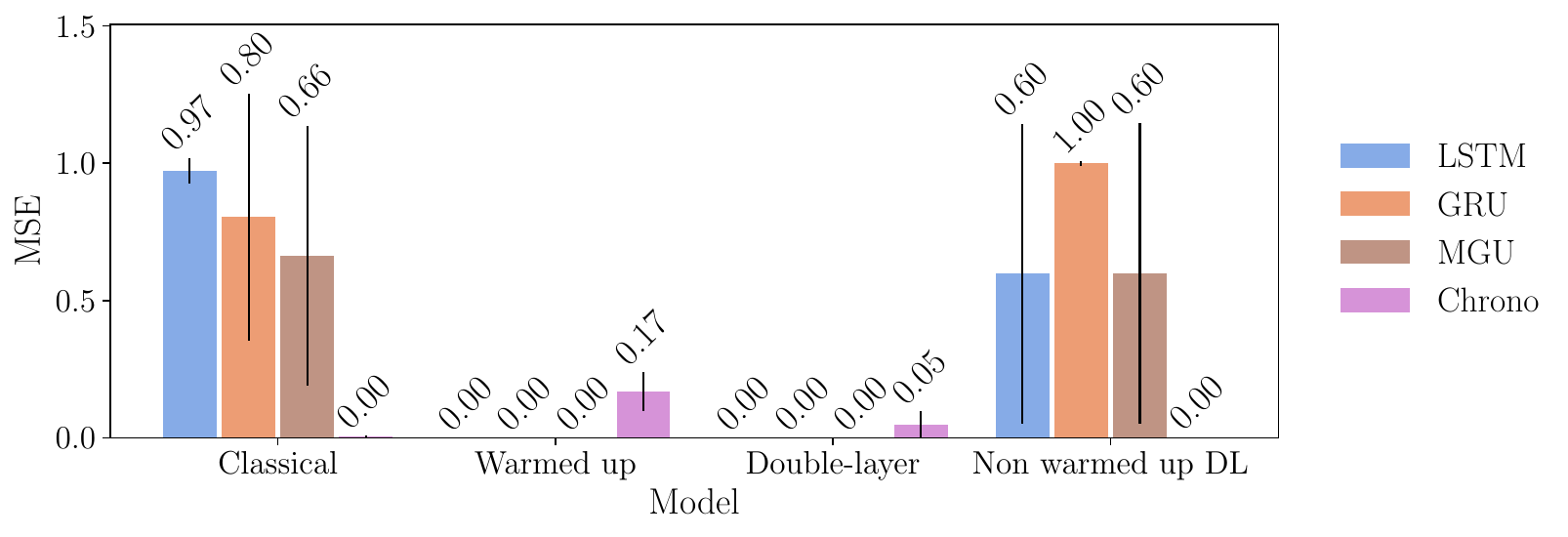}
        \caption{$T = 50$}
        \label{fig:bar_copy_50_double}
    \end{subfigure}
    \begin{subfigure}{.49\textwidth}
        \centering
        \includegraphics[width=\textwidth]{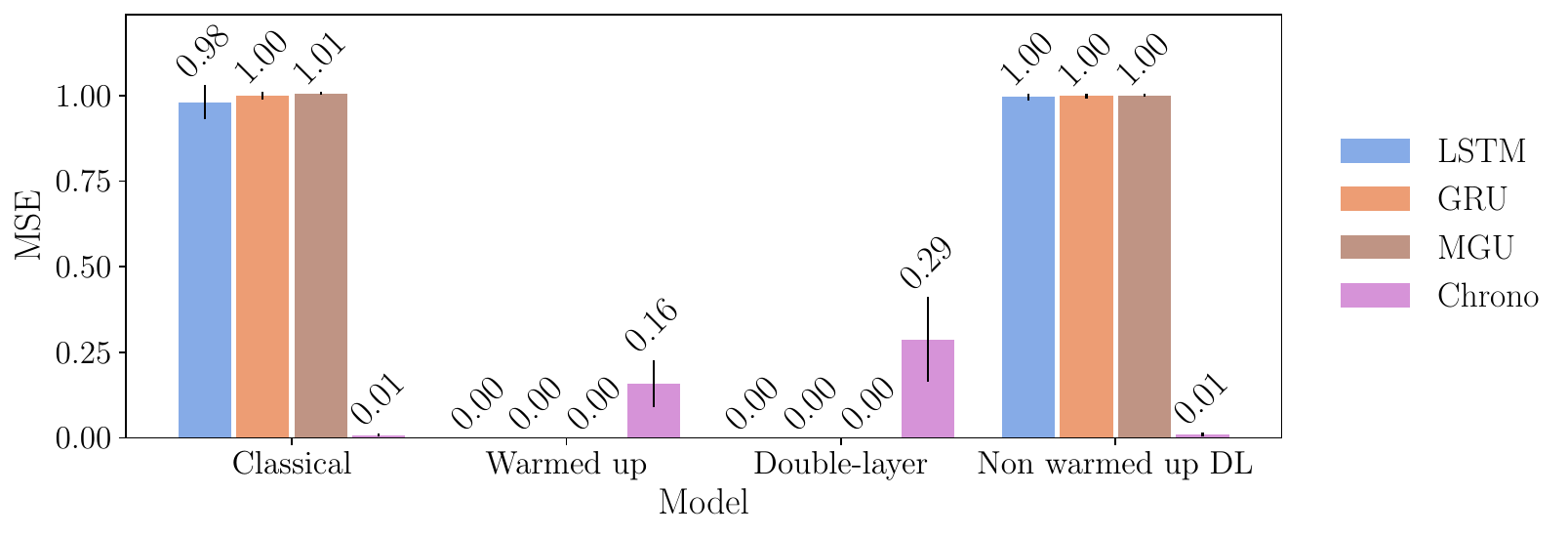}
        \caption{$T = 300$}
        \label{fig:bar_copy_300_double}
    \end{subfigure}
    \begin{subfigure}{.49\textwidth}
        \centering
        \includegraphics[width=\textwidth]{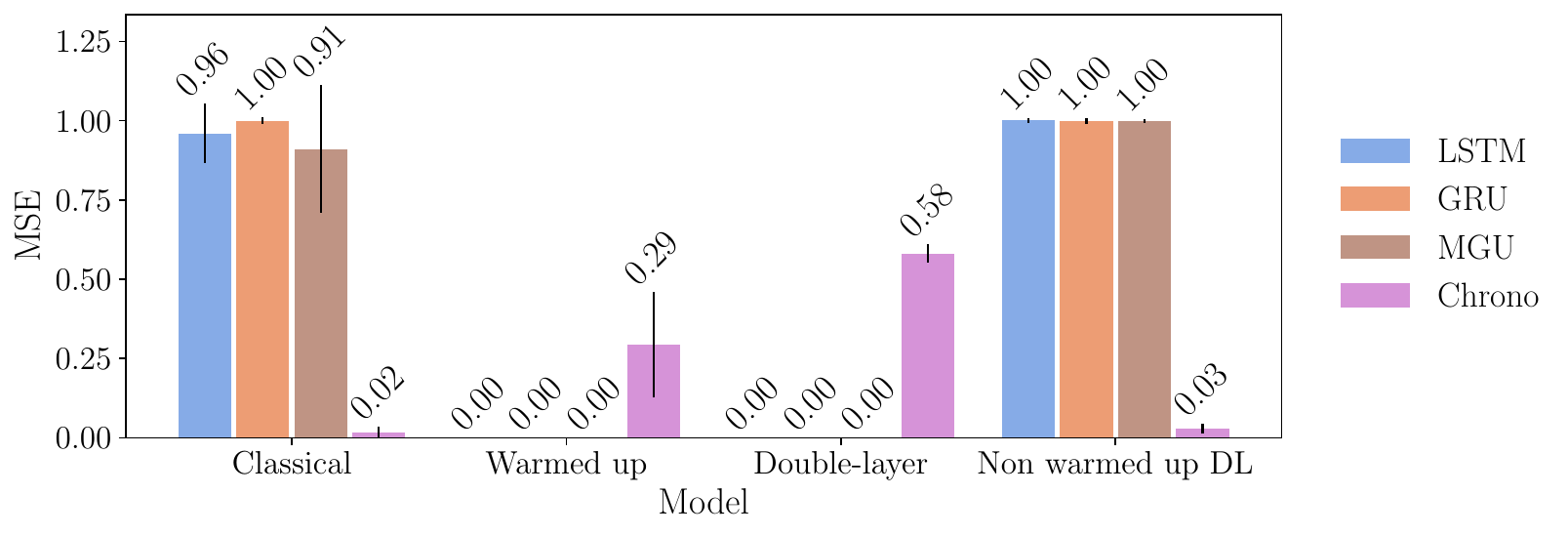}
        \caption{$T = 600$}
        \label{fig:bar_copy_600_double}
    \end{subfigure}
    \caption{%
        Test MSE loss for the copy first input benchmark with different sequence lengths $T$.
        Mean and standard deviation are reported after 50 epochs.}
    \label{tab:double_bench1}
\end{figure}

\begin{figure}[ht]
    \centering
    \begin{subfigure}{.49\textwidth}
        \centering
        \includegraphics[width=\textwidth]{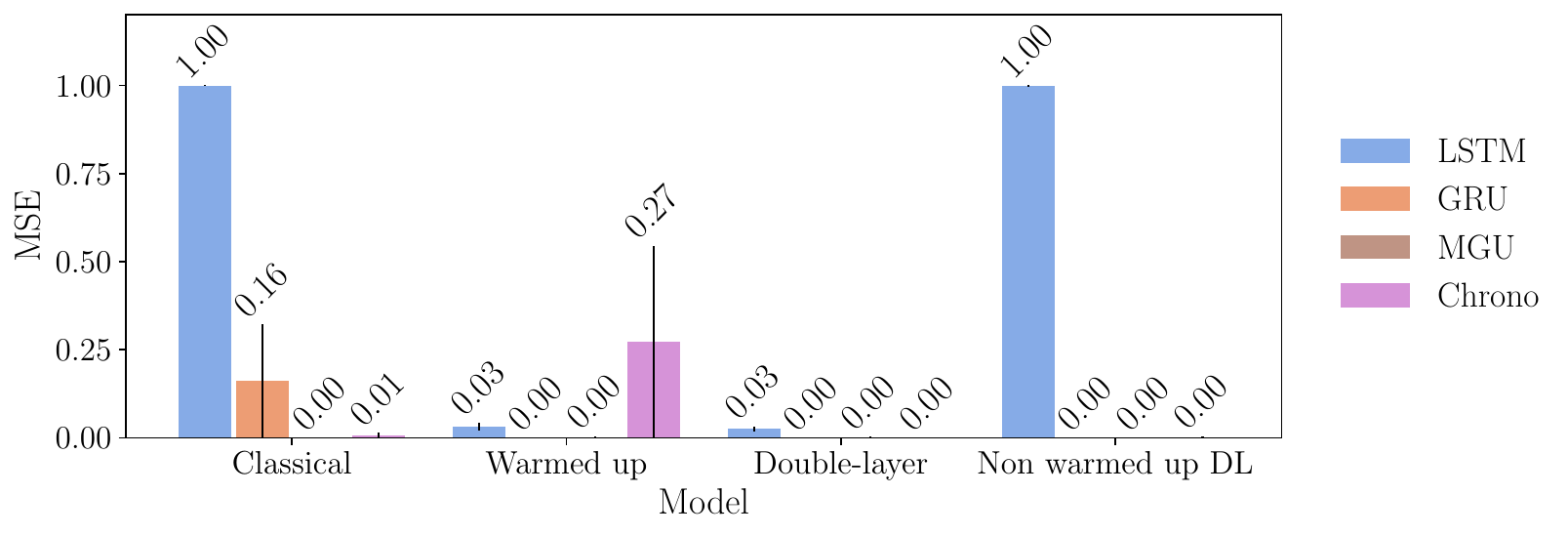}
        \caption{$N = 5$}
        \label{fig:bar_denoising_5_double}
    \end{subfigure}
    \begin{subfigure}{.49\textwidth}
        \centering
        \includegraphics[width=\textwidth]{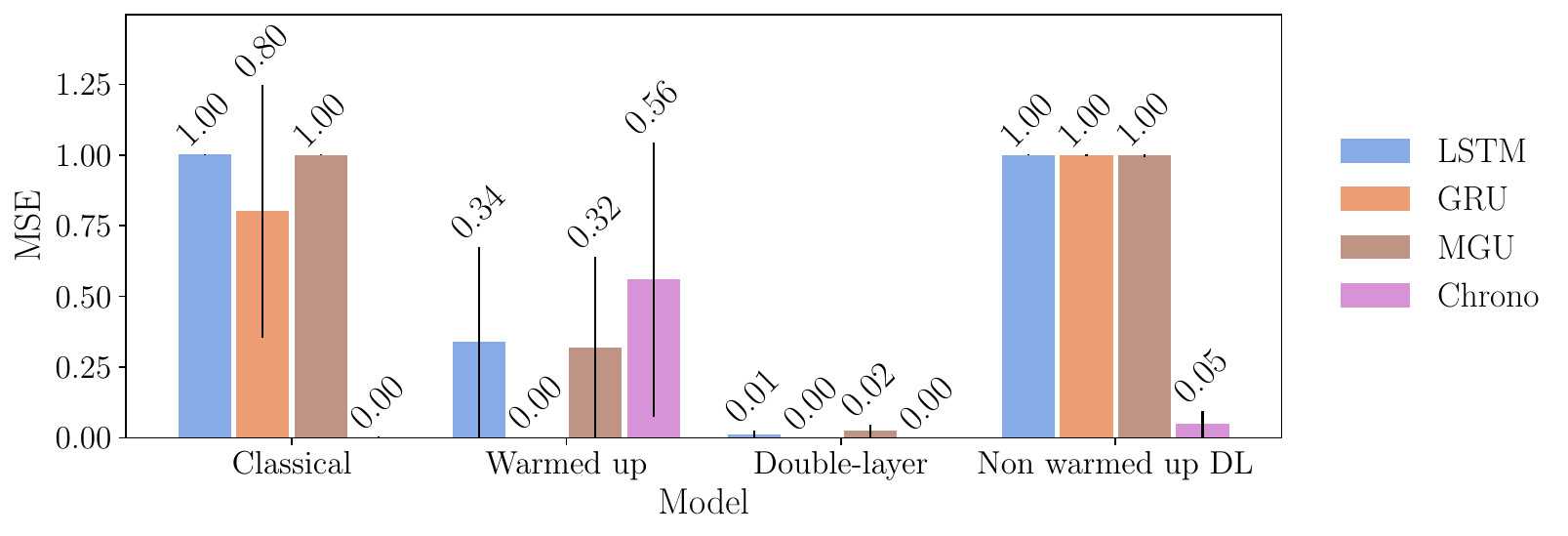}
        \caption{$N = 100$}
        \label{fig:bar_denoising_100_double}
    \end{subfigure}
    \caption{%
        Test MSE loss for the denoising benchmark with different forgetting periods $N$ and $T = 200$.
        Mean and standard deviation are reported after 50 epochs.}
    \label{tab:double_bench2}
\end{figure}

In this section, we show the performance of all cells on the copy first input and denoising benchmarks including the double-layer architecture without partial warmup.
As can be seen from \autoref{tab:double_bench1} and \autoref{tab:double_bench2} the double-layer architecture without partial warmup generally performs worse than the classic architecture.
This ablation study confirms that the partial warmup is the most important factor for the double-layer architecture performance.

\section{RNNs with auxiliary losses} \label{app:aux_losses}

\citet{trinh_learning_2018} introduces a new method to pretrain and train RNNs on very long sequences using auxiliary losses to teach the networks to correctly encode input sequences.
To do this, input sequences are split into small sequences.
After each small sequence, the final hidden state is given to a decoder RNN, which must then either reconstruct the sequence or predict the next timesteps.
This method has been tested on several image classification datasets where images are fed pixel by pixel, and it has achieved good results.
We were wondering if this method also promotes multistability when applied to benchmarks with long-term dependencies. Therefore, we tested it on the copy, denoising and permuted line-sequential MNIST benchmarks.
We adapt this implementation\footnote{https://github.com/younggyoseo/rnn-auxiliary-loss} to run our tests.
For each experiment we ran $20$ epochs of pretraining and $50$ epochs of training.
The \vaa{} is computed at the end of each epoch.
As for the auxiliary loss, we have used the reconstruction loss, \ie{} the MSE between the real sequence and the reconstructed sequence. During training, the auxiliary loss is also taken into account to make the gradient step.

\begin{figure}[ht]
    \centering
    \begin{subfigure}{.49\textwidth}
        \centering
        \includegraphics[width=.49\textwidth]{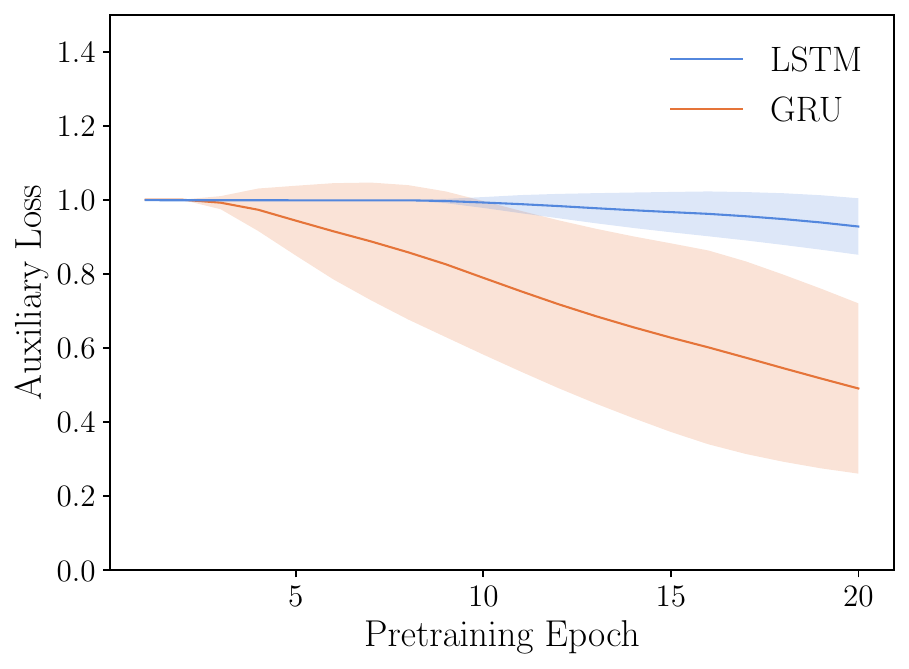}
        \includegraphics[width=.49\textwidth]{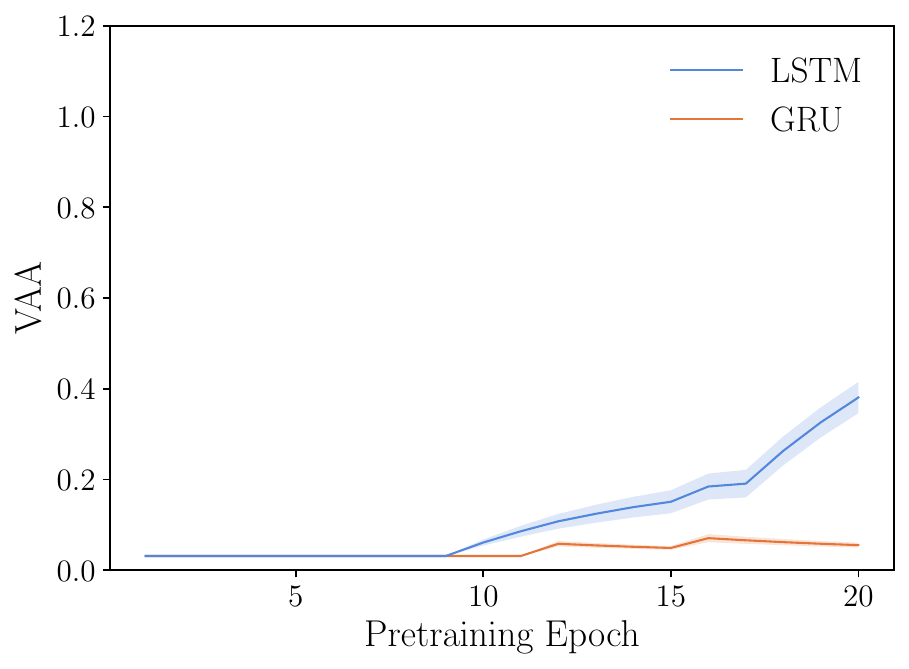}
        \caption{$T = 50$ and the reconstruction loss is evaluated on \num{30} timesteps.}
    \end{subfigure}
    \begin{subfigure}{.49\textwidth}
        \centering
        \includegraphics[width=.49\textwidth]{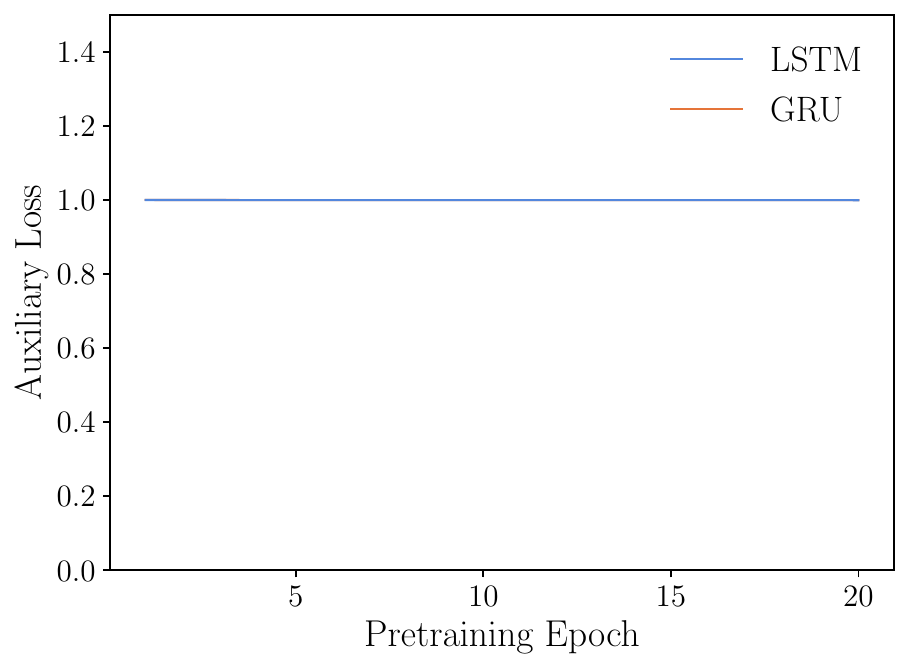}
        \includegraphics[width=.49\textwidth]{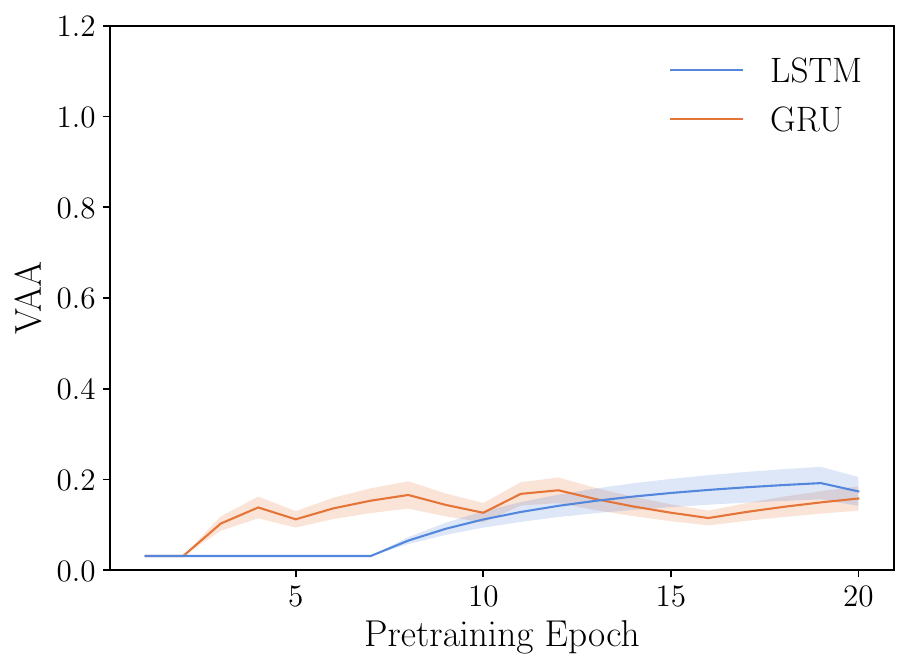}
        \caption{$T = 300$ and the reconstruction loss is evaluated on \num{200} timesteps.}
    \end{subfigure}
    \begin{subfigure}{.49\textwidth}
        \centering
        \includegraphics[width=.49\textwidth]{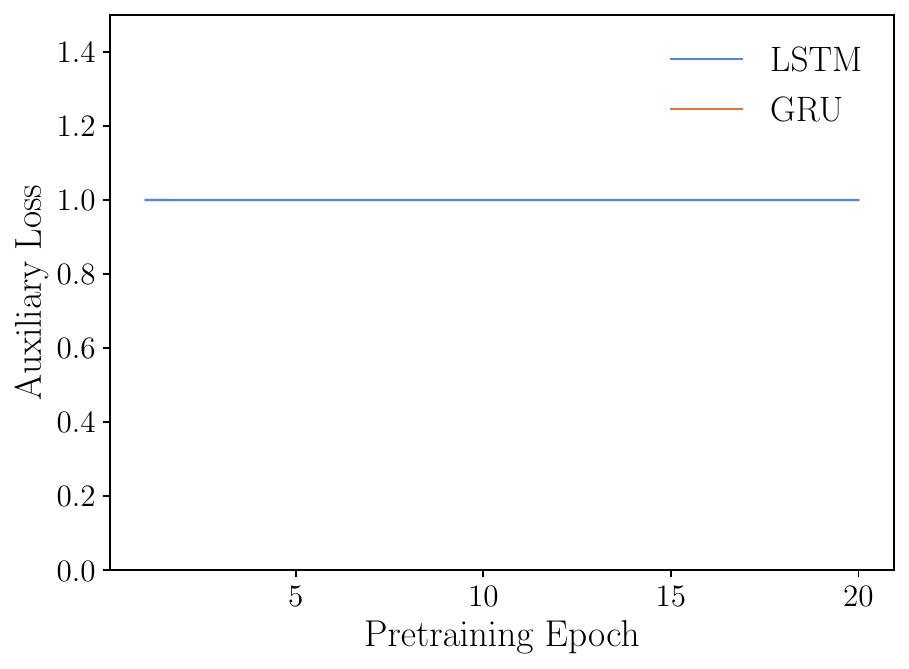}
        \includegraphics[width=.49\textwidth]{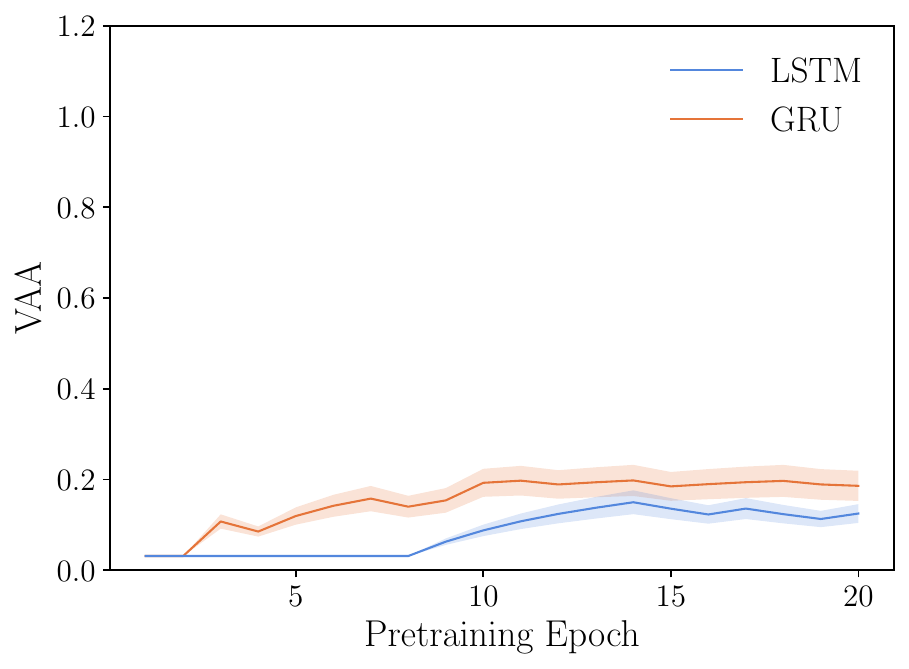}
        \caption{$T = 600$ and the reconstruction loss is evaluated on \num{400} timesteps.}
    \end{subfigure}
    \caption{Pretraining on the copy first input benchmark for different sequence lengths.}
    \label{fig:rw_pt_copy}
\end{figure}

\begin{figure}[ht]
    \centering
    \begin{subfigure}{\textwidth}
        \centering
        \includegraphics[width=.24\textwidth]{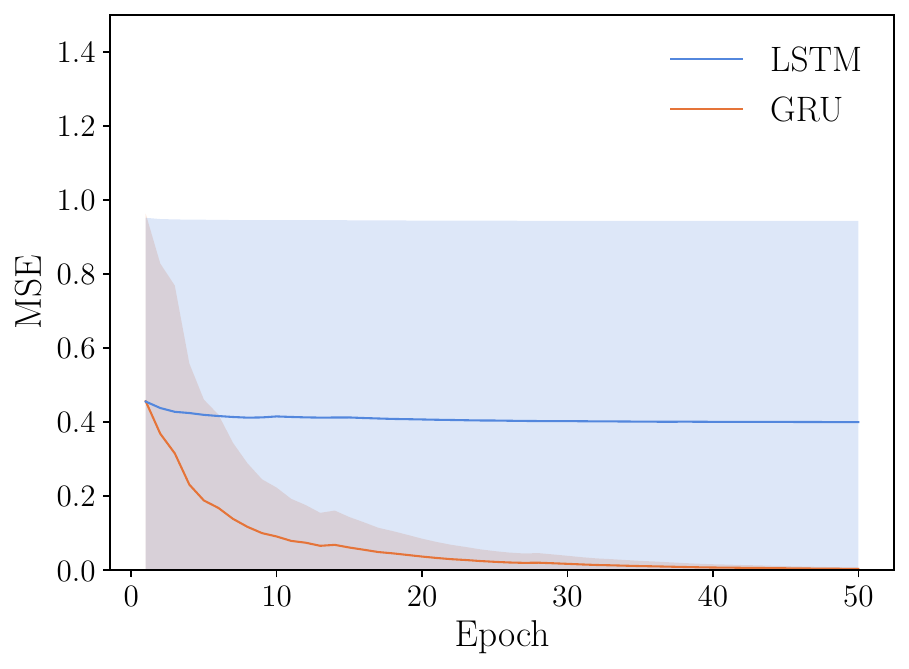}
        \includegraphics[width=.24\textwidth]{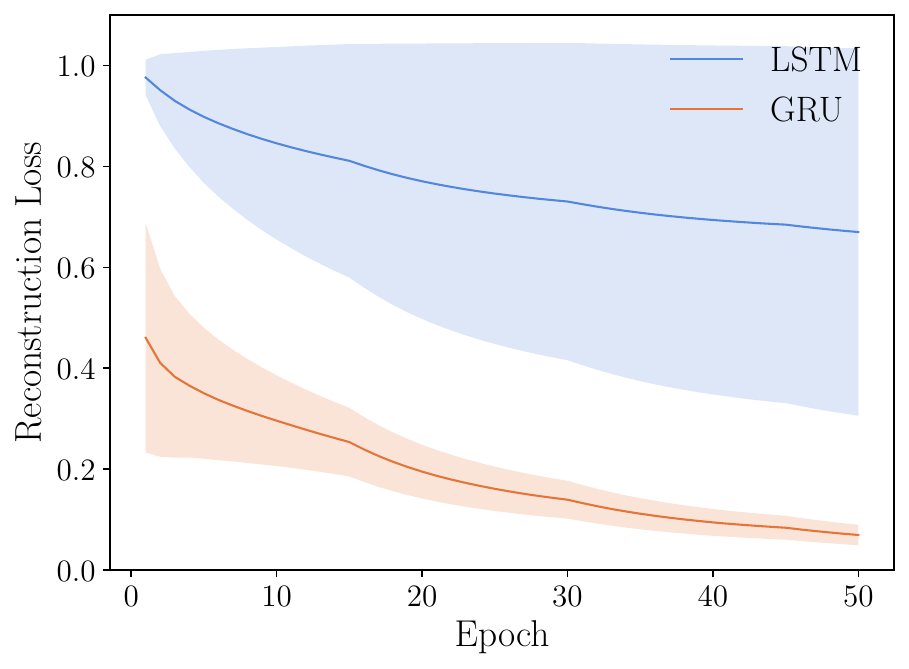}
        \includegraphics[width=.24\textwidth]{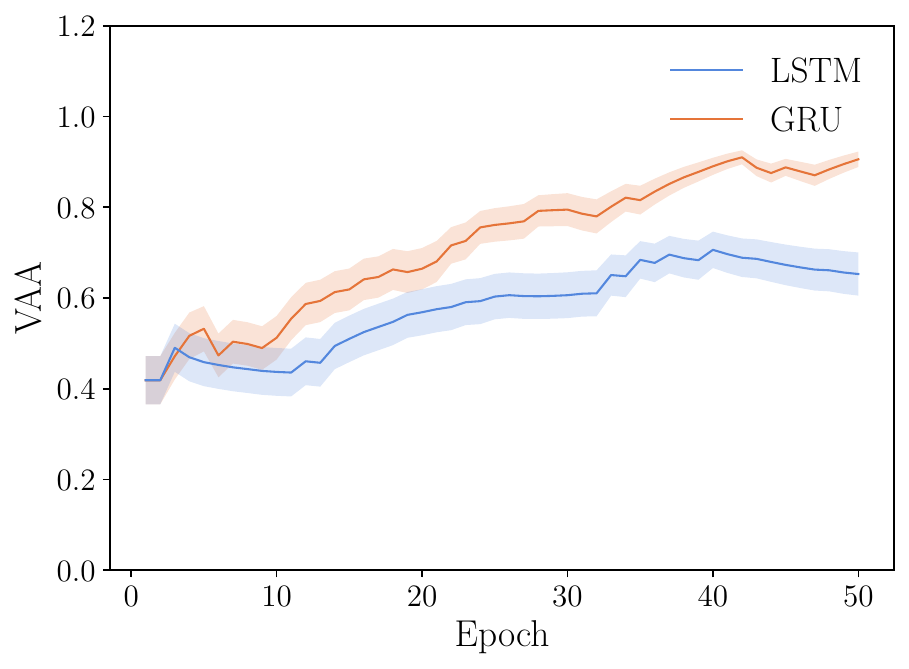}
        \includegraphics[width=.22\textwidth]{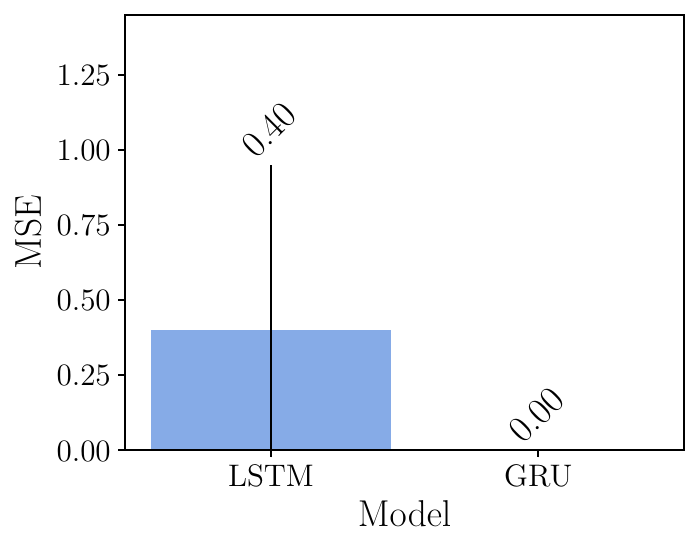}
        \caption{$T = 50$ and the reconstruction loss is evaluated on \num{30} timesteps.}
    \end{subfigure}
    \begin{subfigure}{\textwidth}
        \centering
        \includegraphics[width=.24\textwidth]{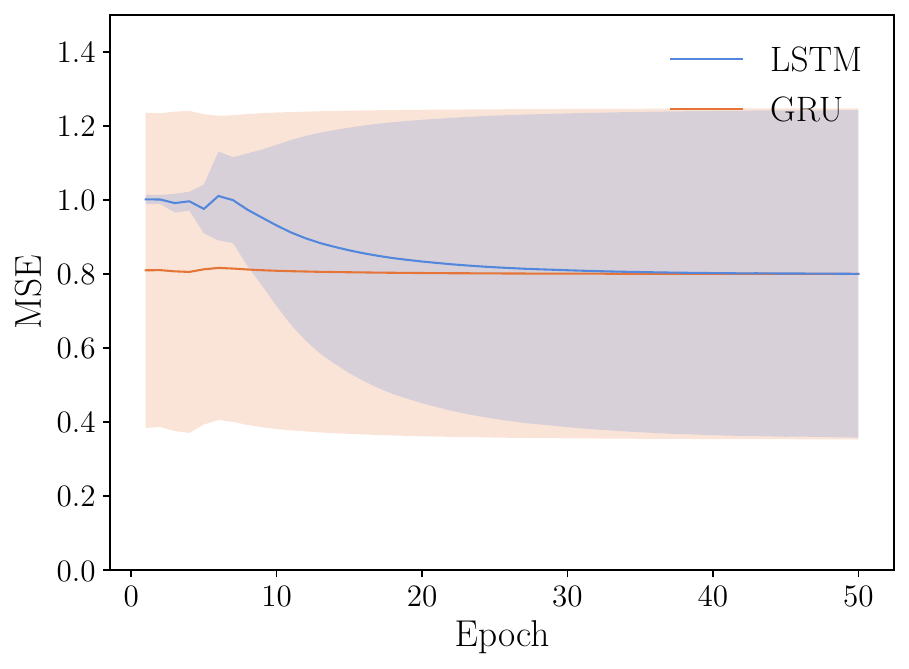}
        \includegraphics[width=.24\textwidth]{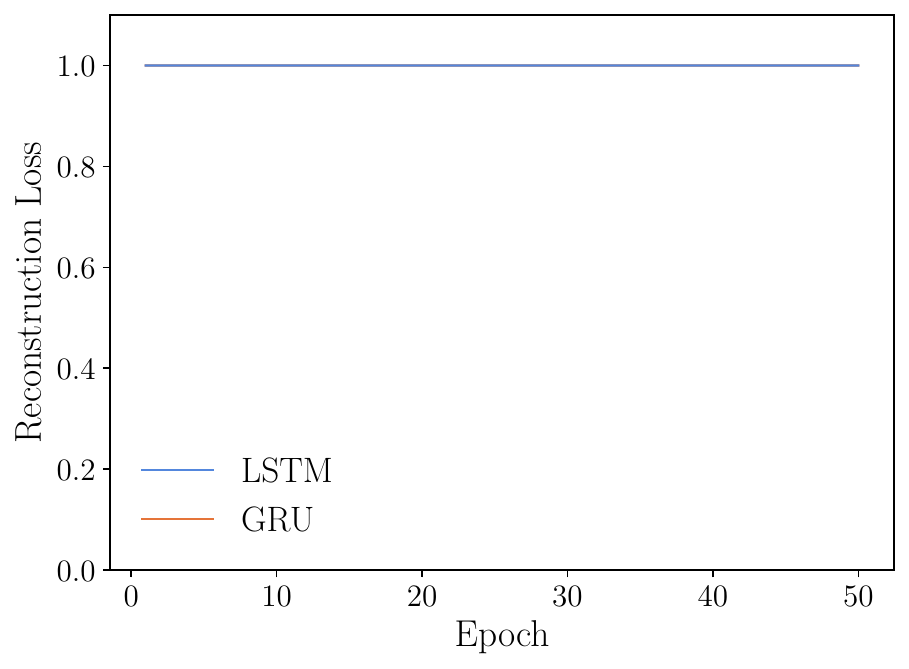}
        \includegraphics[width=.24\textwidth]{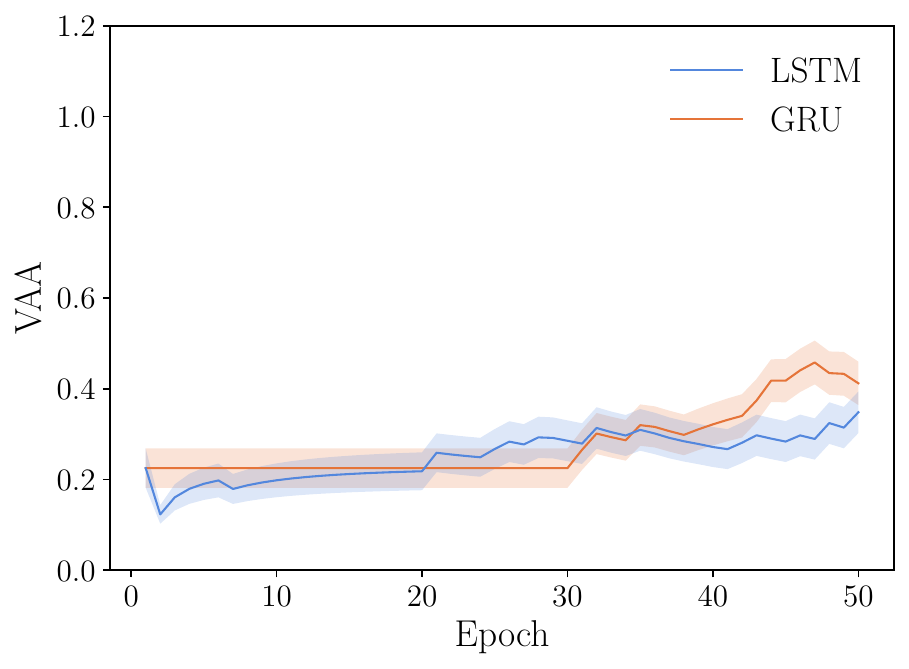}
        \includegraphics[width=.22\textwidth]{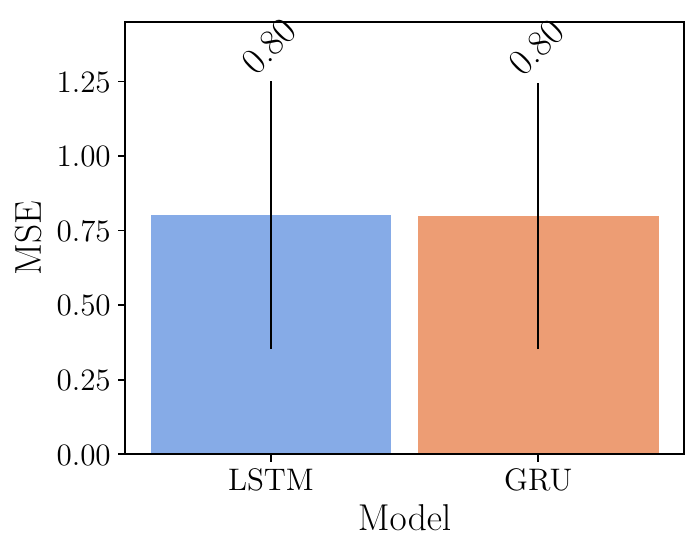}
        \caption{$T = 300$ and the reconstruction loss is evaluated on \num{200} timesteps.}
    \end{subfigure}
    \begin{subfigure}{\textwidth}
        \centering
        \includegraphics[width=.24\textwidth]{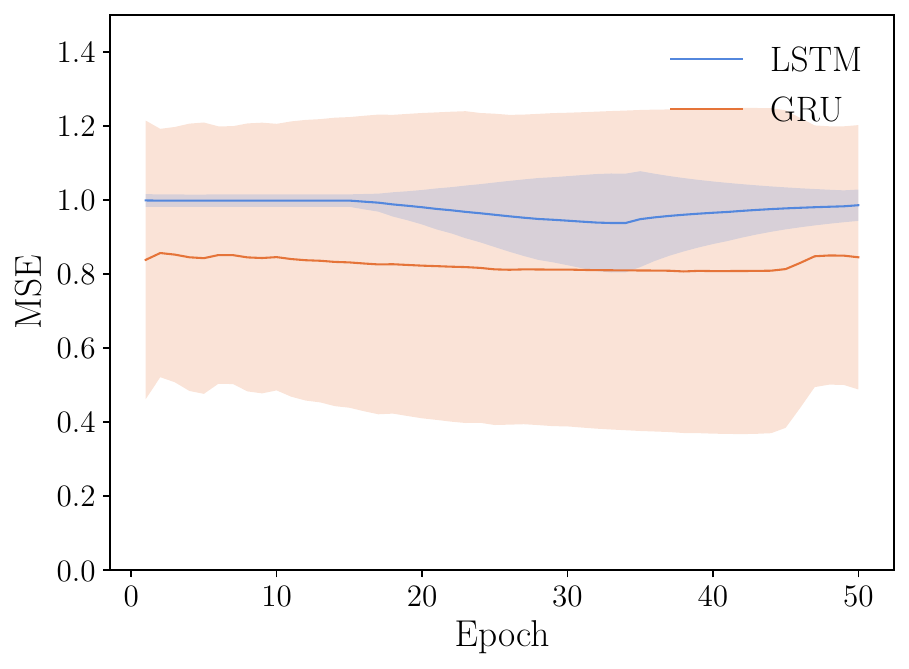}
        \includegraphics[width=.24\textwidth]{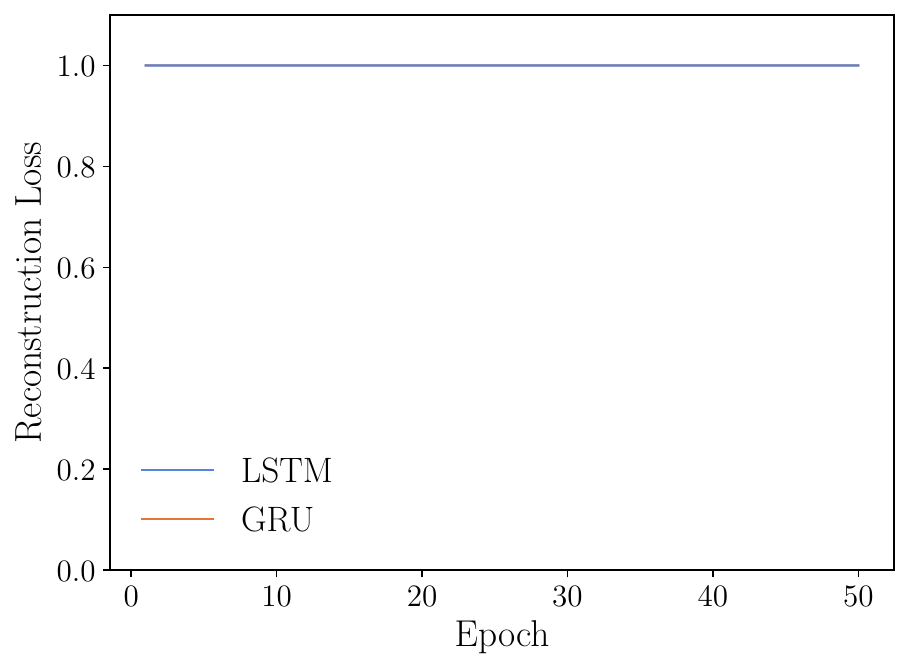}
        \includegraphics[width=.24\textwidth]{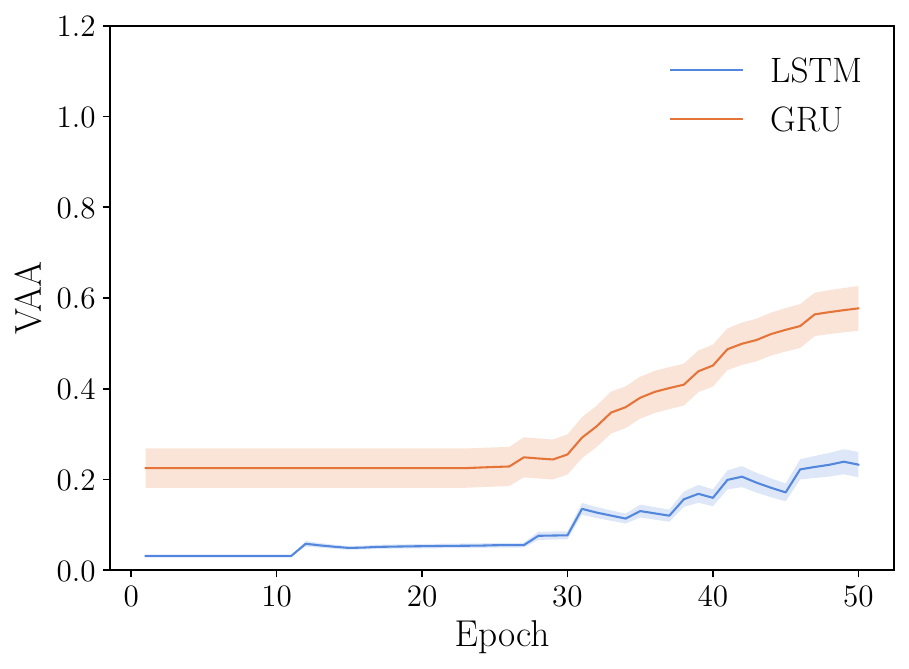}
        \includegraphics[width=.22\textwidth]{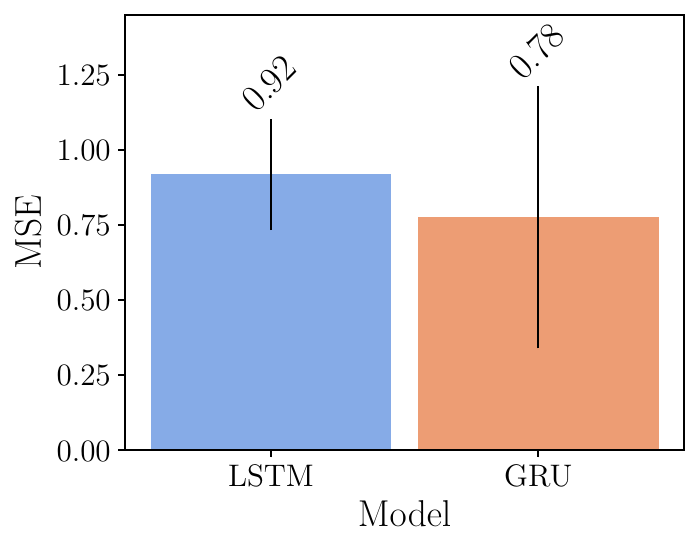}
        \caption{$T = 600$ and the reconstruction loss is evaluated on \num{400} timesteps.}
    \end{subfigure}
    \caption{Training on the copy first input benchmark for different sequence lengths.}
    \label{fig:rw_copy}
\end{figure}

\begin{figure}[ht]
    \vspace{20ex}
    \centering
    \begin{subfigure}{.49\textwidth}
        \centering
        \includegraphics[width=.49\textwidth]{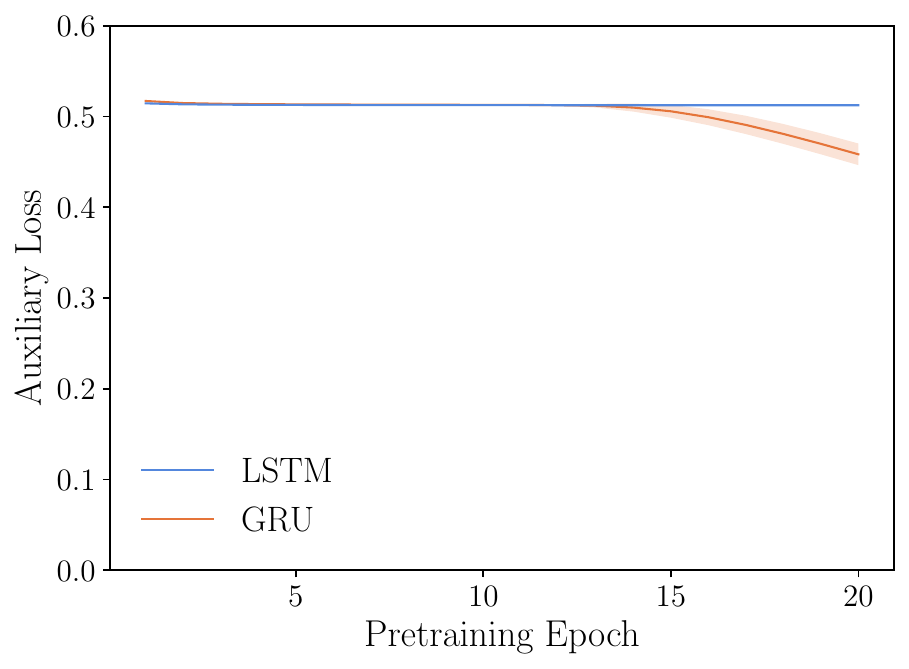}
        \includegraphics[width=.49\textwidth]{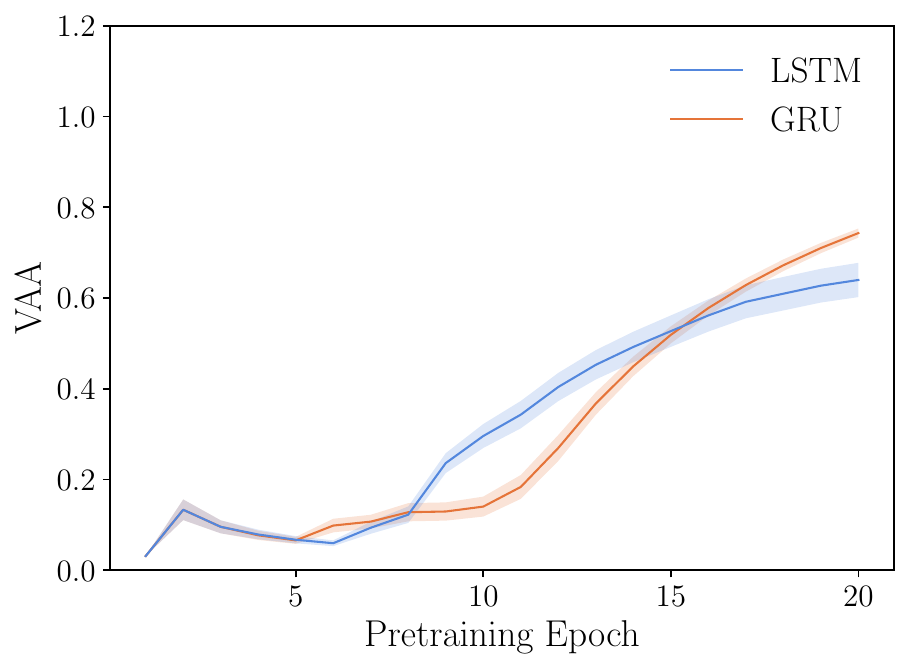}
        \caption{$N = 5$}
    \end{subfigure}
    \begin{subfigure}{.49\textwidth}
        \centering
        \includegraphics[width=.49\textwidth]{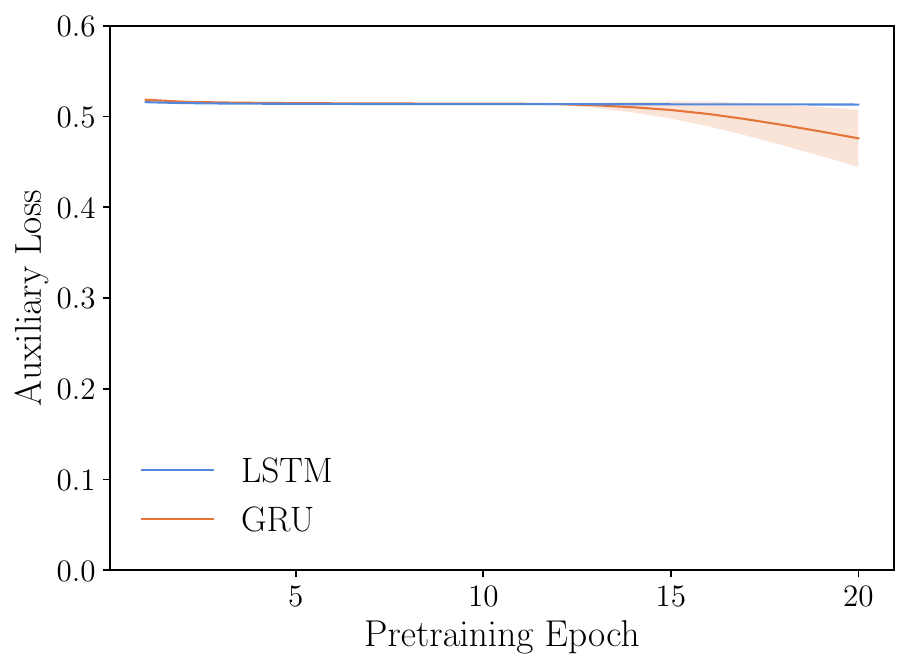}
        \includegraphics[width=.49\textwidth]{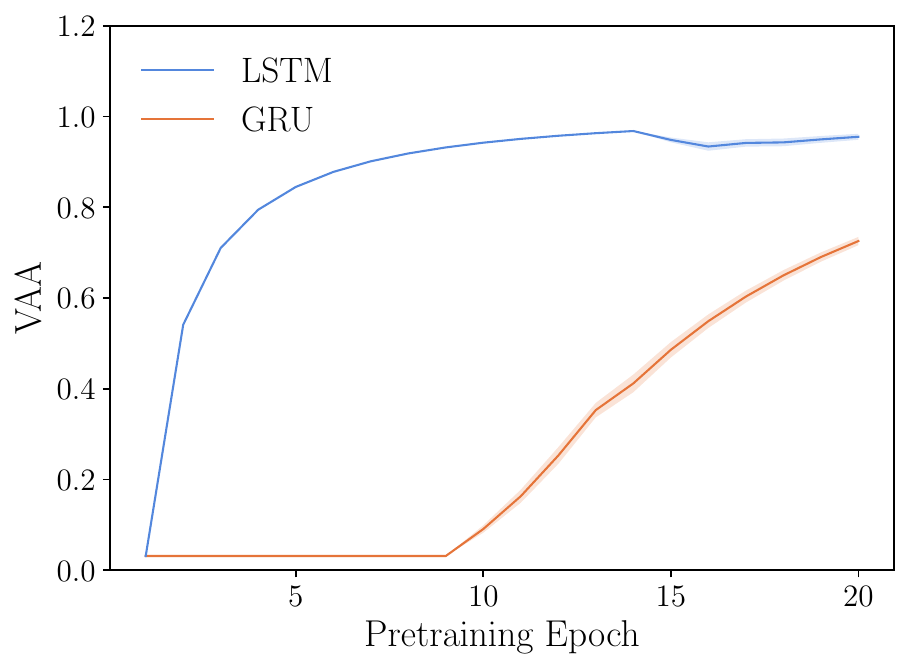}
        \caption{$N = 100$}
    \end{subfigure}
    \caption{Pretraining on the denoising benchmark for different forgetting periods and $T = 200$. The reconstruction loss is evaluated on \num{150} timesteps.}
    \label{fig:rw_pt_denoising}
\end{figure}

\begin{figure}[ht]
    \centering
    \begin{subfigure}{\textwidth}
        \centering
        \includegraphics[width=.24\textwidth]{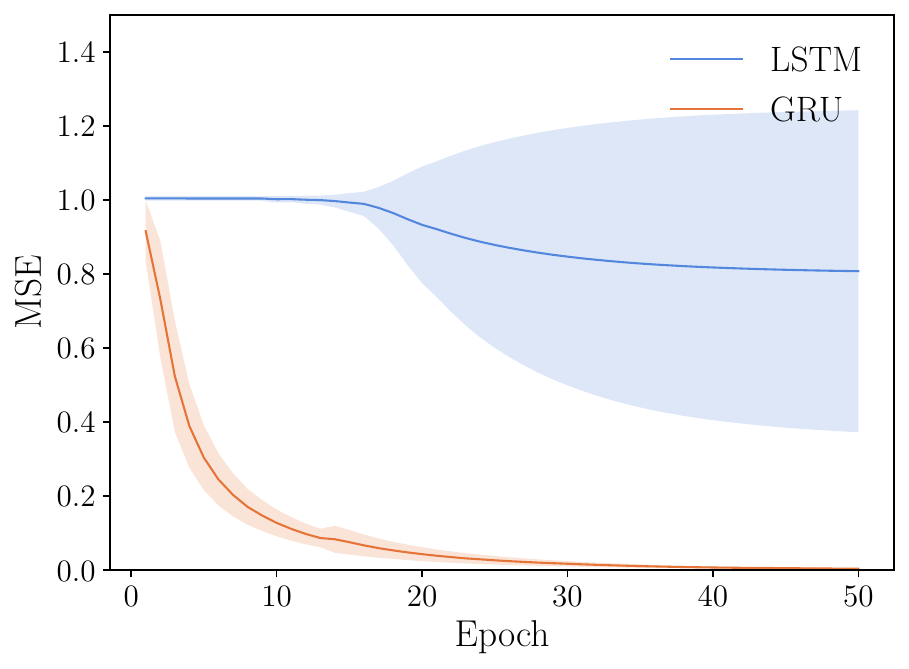}
        \includegraphics[width=.24\textwidth]{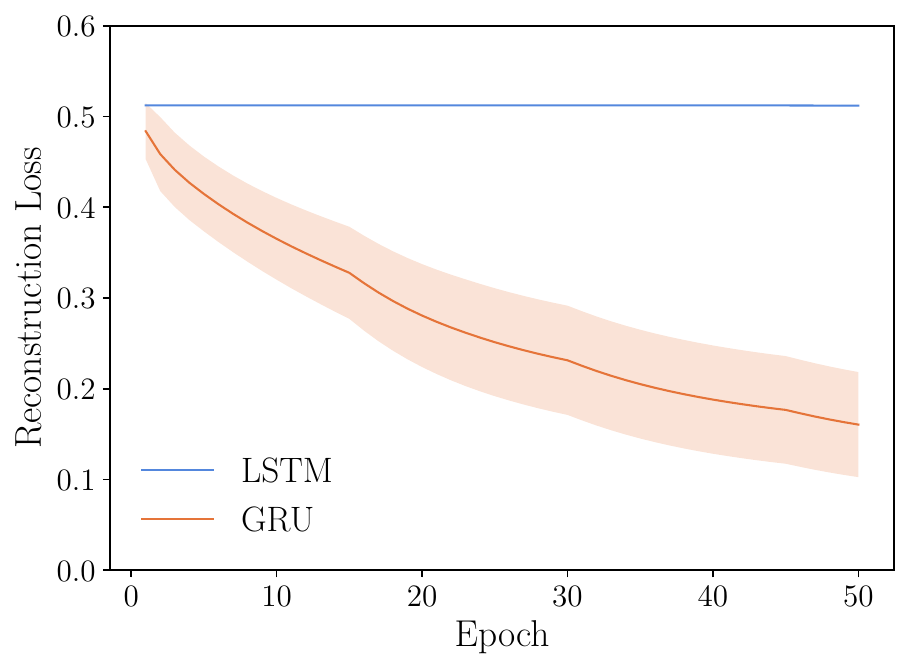}
        \includegraphics[width=.24\textwidth]{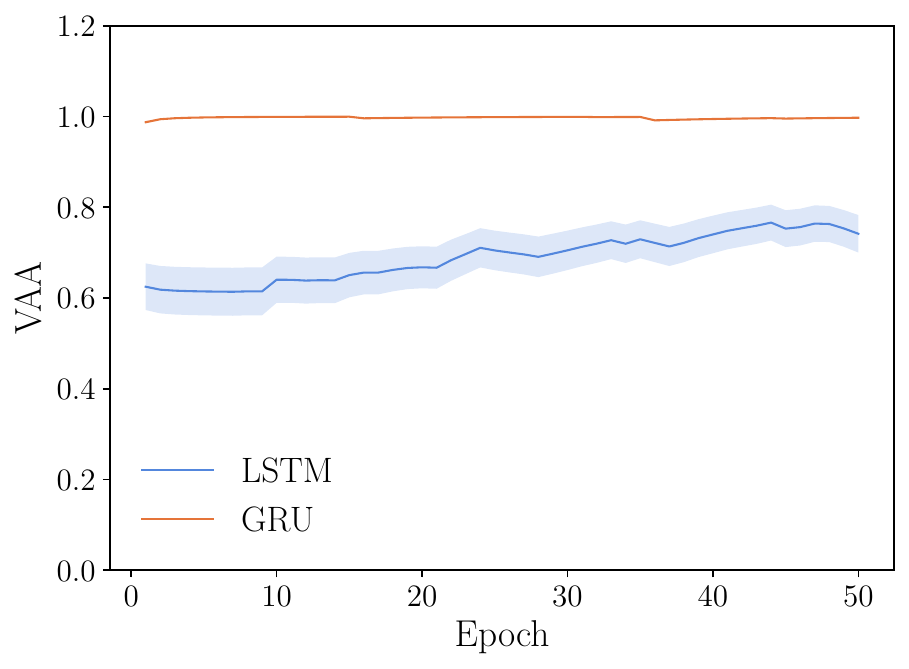}
        \includegraphics[width=.22\textwidth]{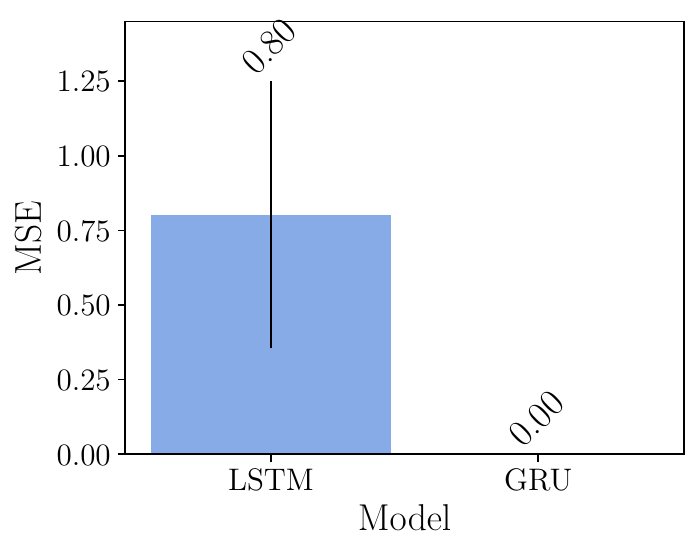}
        \caption{$N = 5$}
    \end{subfigure}
    \begin{subfigure}{\textwidth}
        \centering
        \includegraphics[width=.24\textwidth]{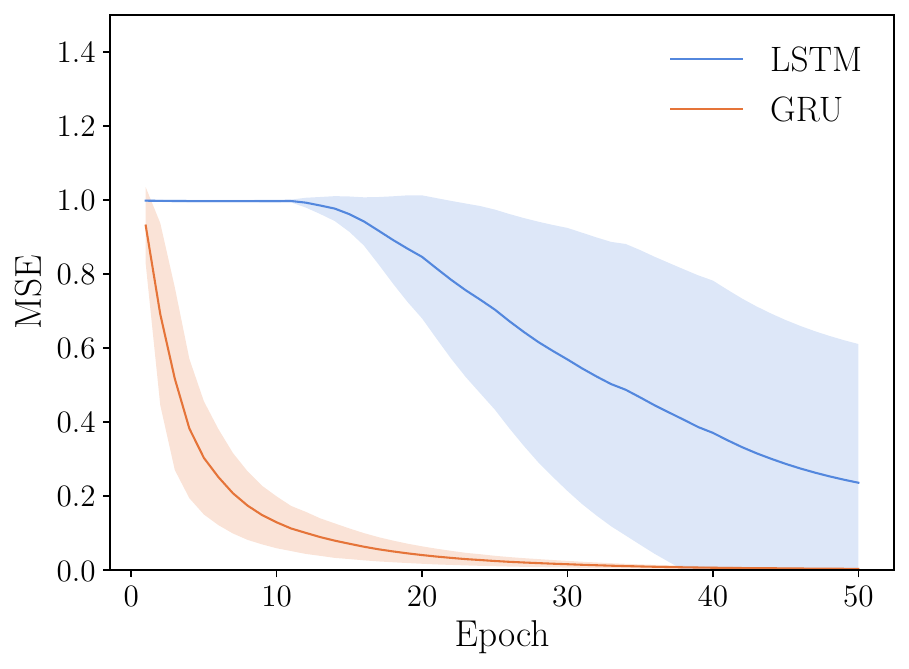}
        \includegraphics[width=.24\textwidth]{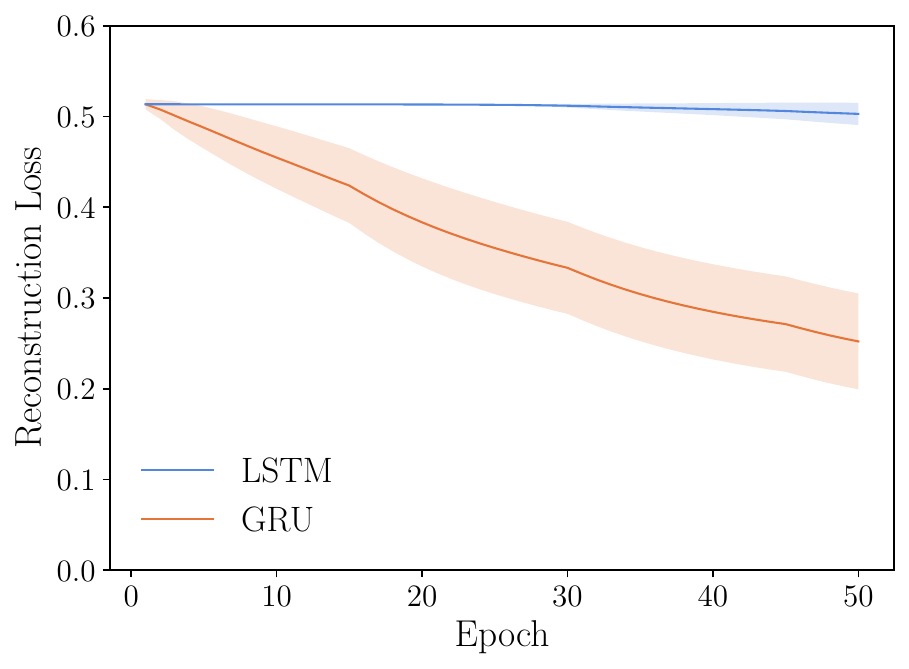}
        \includegraphics[width=.24\textwidth]{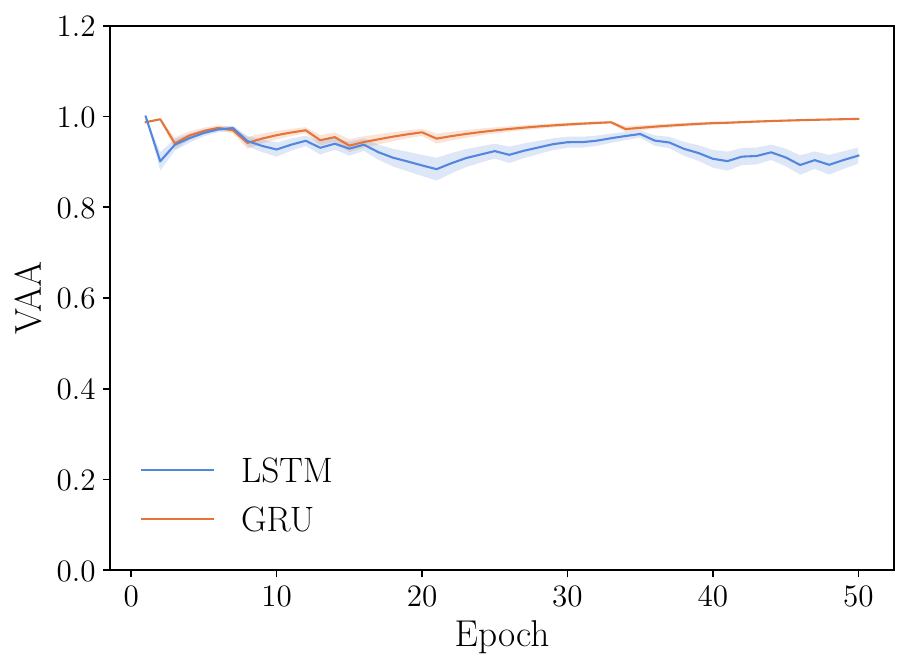}
        \includegraphics[width=.22\textwidth]{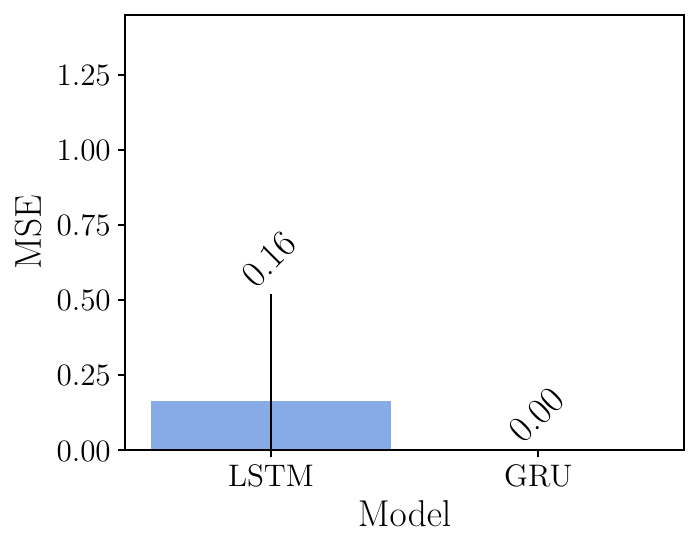}
        \caption{$N = 100$}
    \end{subfigure}
    \caption{Training on the denoising benchmark for different forgetting periods and $T = 200$. The reconstruction loss is evaluated on \num{150} timesteps.}
    \label{fig:rw_denoising}
    \vspace{20ex}
\end{figure}

\begin{figure}[ht]
    \vspace{20ex}
    \centering
    \begin{subfigure}{.49\textwidth}
        \centering
        \includegraphics[width=.49\textwidth]{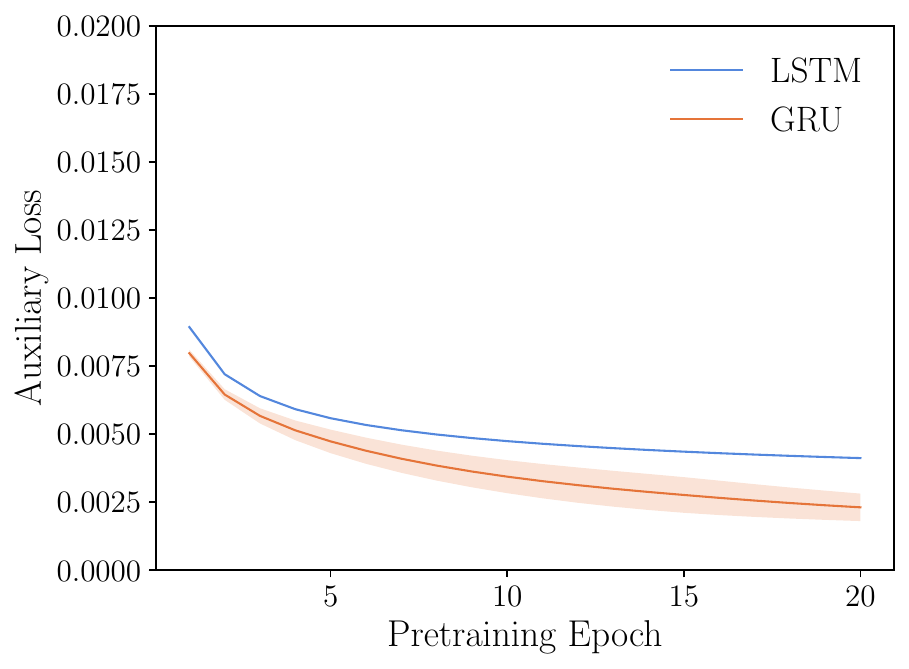}
        \includegraphics[width=.49\textwidth]{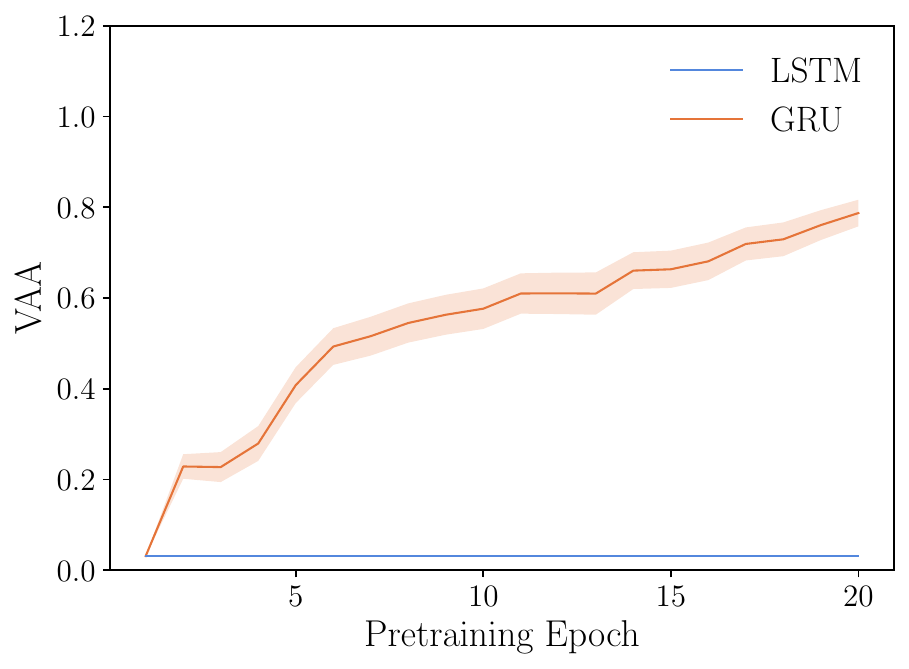}
        \caption{$N = 72$ and the reconstruction loss is evaluated on \num{50} timesteps.}
    \end{subfigure}
    \begin{subfigure}{.49\textwidth}
        \centering
        \includegraphics[width=.49\textwidth]{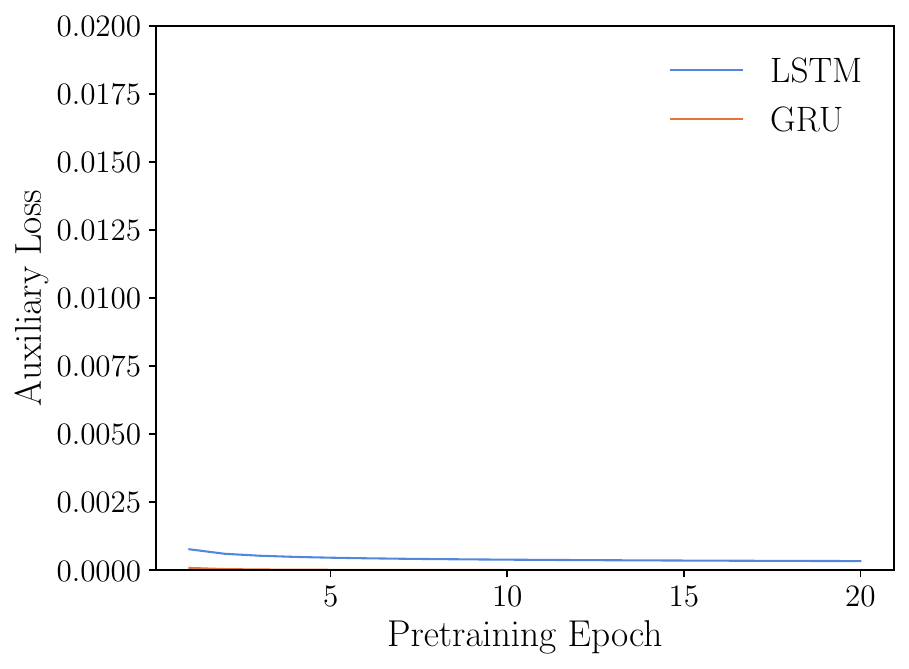}
        \includegraphics[width=.49\textwidth]{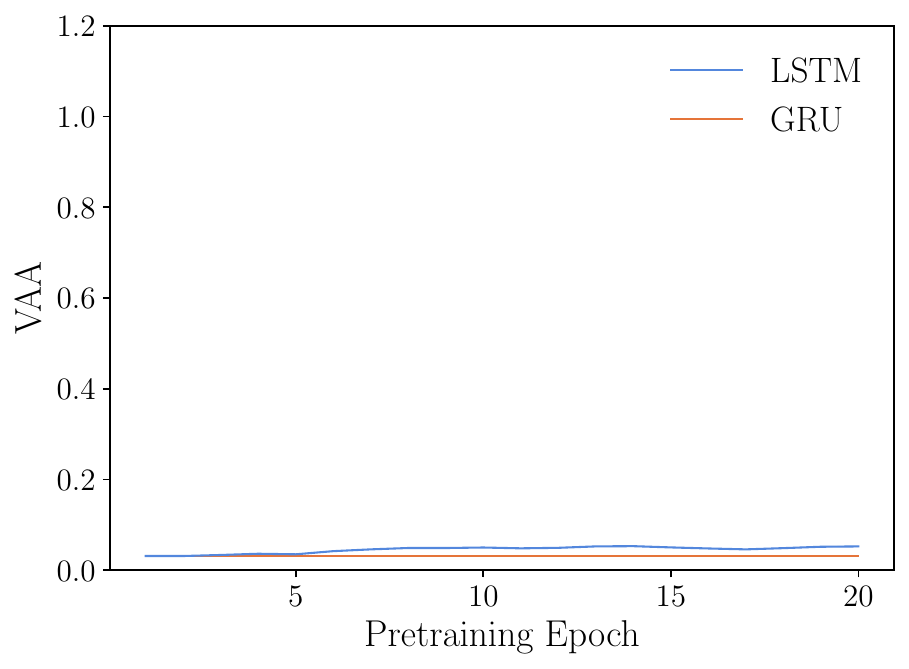}
        \caption{$N = 472$ and the reconstruction loss is evaluated on \num{350} timesteps.}
    \end{subfigure}
    \caption{Pretraining on the permuted-line sequential MNIST benchmark for different forgetting periods.}
    \label{fig:rw_pt_row_mnist}
\end{figure}

\begin{figure}[ht]
    \centering
    \begin{subfigure}{\textwidth}
        \centering
        \includegraphics[width=.24\textwidth]{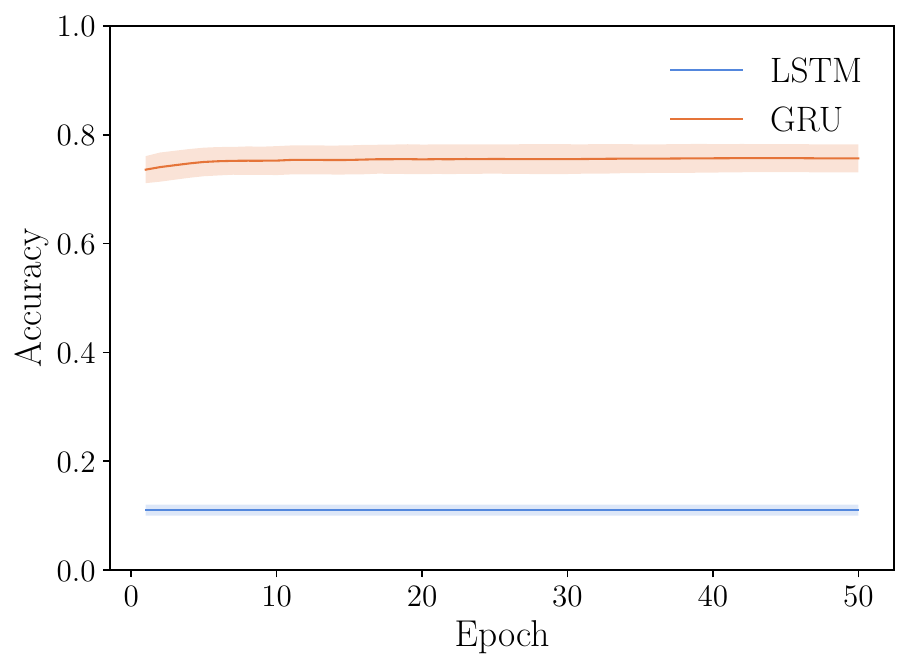}
        \includegraphics[width=.24\textwidth]{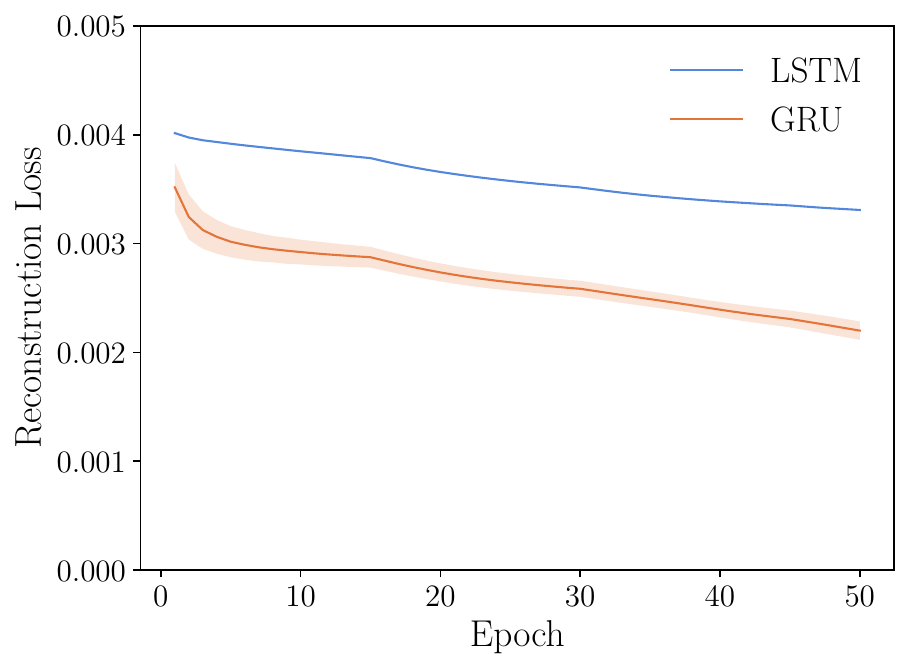}
        \includegraphics[width=.24\textwidth]{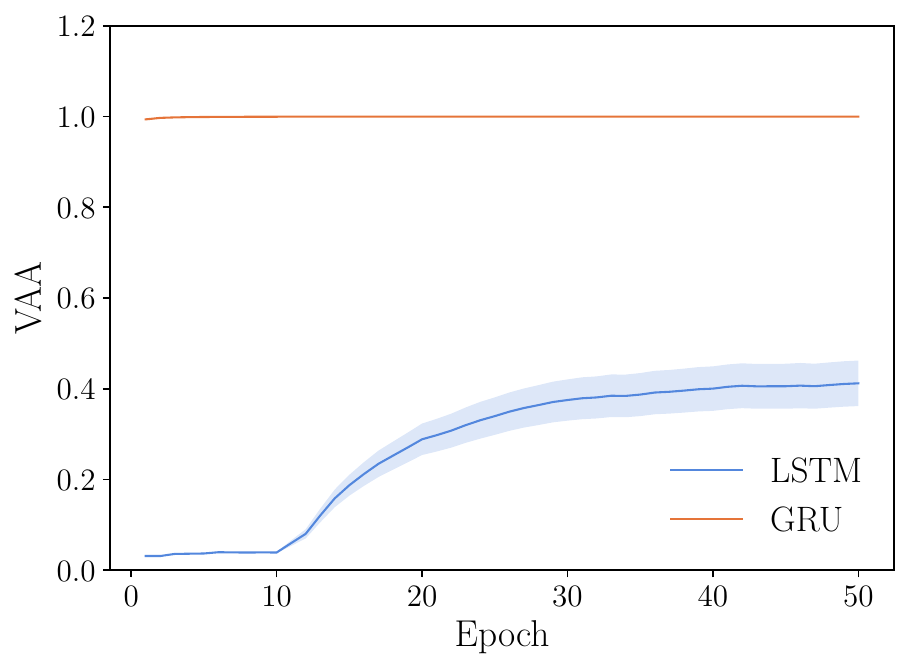}
        \includegraphics[width=.22\textwidth]{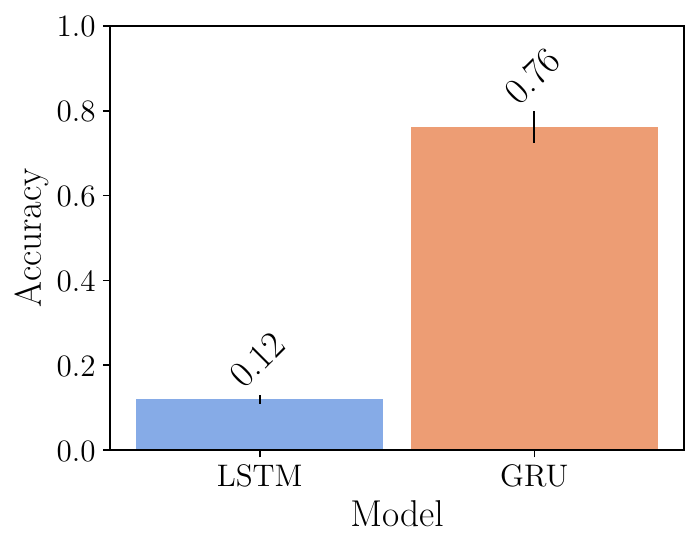}
        \caption{$N = 72$ and the reconstruction loss is evaluated on \num{50} timesteps.}
    \end{subfigure}
    \begin{subfigure}{\textwidth}
        \centering
        \includegraphics[width=.24\textwidth]{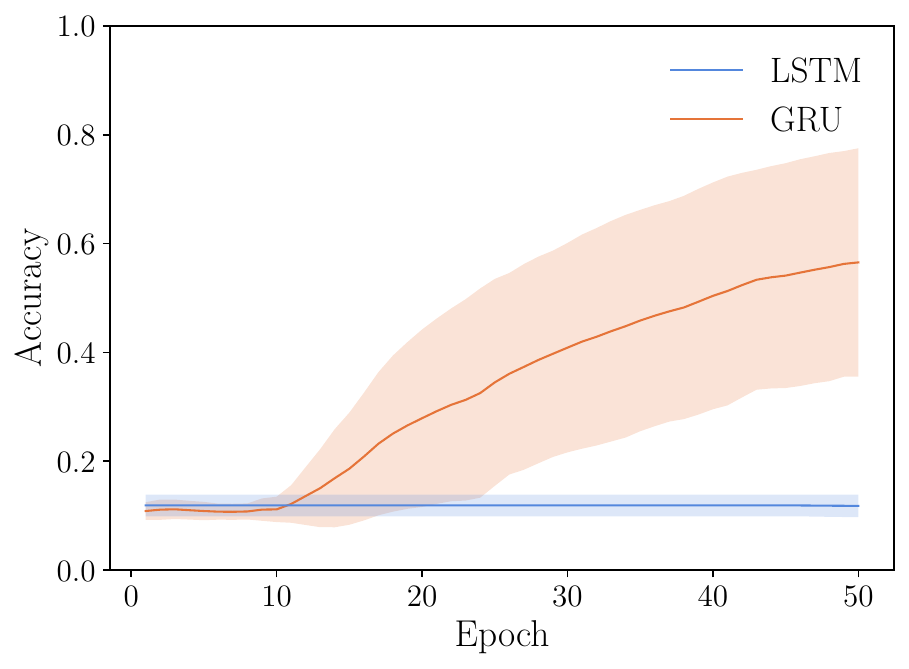}
        \includegraphics[width=.24\textwidth]{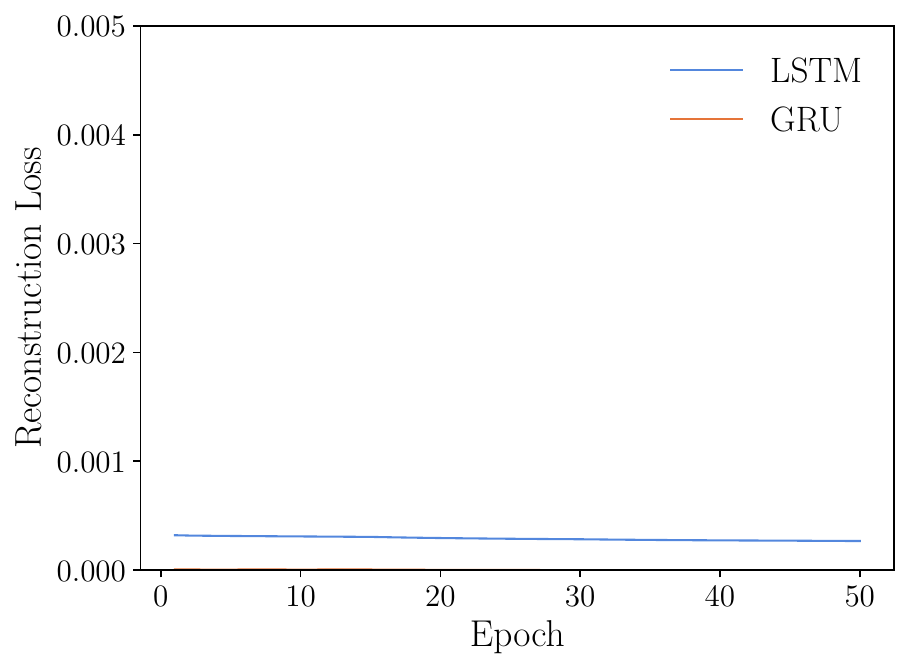}
        \includegraphics[width=.24\textwidth]{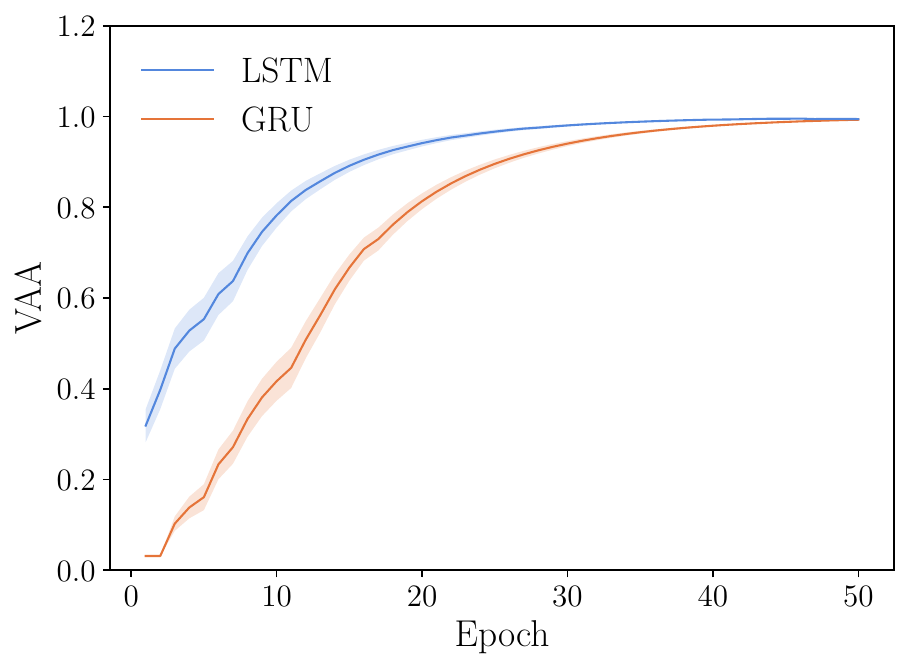}
        \includegraphics[width=.22\textwidth]{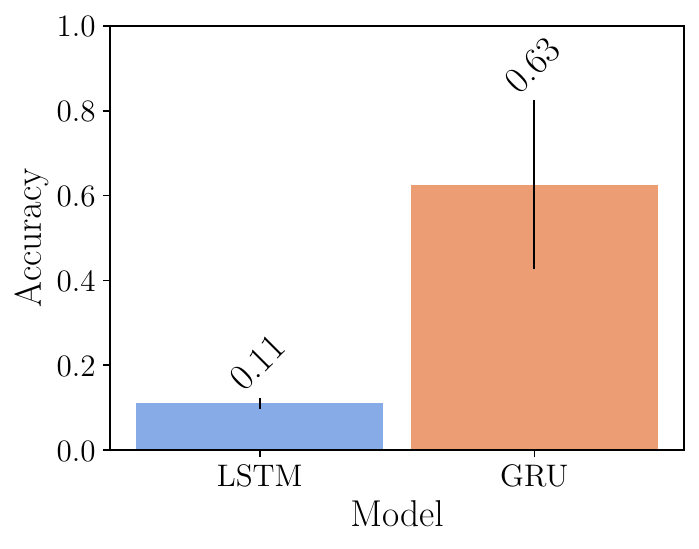}
        \caption{$N = 472$ and the reconstruction loss is evaluated on \num{350} timesteps.}
    \end{subfigure}
    \caption{Training on the permuted-line sequential MNIST benchmark for different forgetting periods.}
    \label{fig:rw_row_mnist}
    \vspace{20ex}
\end{figure}

\autoref{fig:rw_pt_copy}, \autoref{fig:rw_pt_denoising} and \autoref{fig:rw_pt_row_mnist} show the evolutions of the reconstruction loss and the \vaa{} on the three benchmarks during the pretraining, while \autoref{fig:rw_copy}, \autoref{fig:rw_denoising} and \autoref{fig:rw_row_mnist} show the evolution of the validation MSE (or accuracy on validation set for the permuted-line sequential MNIST), the reconstruction loss and the \vaa{} during training as well as the testing MSE/accuracy.

Globally, adding this auxiliary loss promotes multistability: the \vaa{} increases during the pretraining, but not in all cases. The RNNs have more difficult to decrease the reconstruction loss when the sequences are longer, \ie{} when there is more noise. However, this method increases the performances, especially when the sequences are not too long. For instance, GRU achieves very good results on the denoising benchmark, no matter the forgetting period. Concerning the permuted sequential-line MNIST benchmark, the models manage to obtain very low reconstruction losses. This is due to the padding of black pixels, which can be easily reconstructed. However, GRU managed to learn on this benchmark, which indicates that the auxiliary loss has been useful.

To conclude this section, adding an auxiliary loss to teach the models to reconstruct and thus to correctly encode the input sequences has led to good results. Indeed, that method has also been shown to promote multistability. However, its big drawback is its duration: for each experiment, we have run 20 pretraining epochs, which can take a lot of time especially for big datasets like MNIST. In comparison, in the experiments of \autoref{sec:warmup_experiments}, each warmup consisted of \num{100} gradient steps, which is much more lightweight. Also, this approach leads to equivalent or worse results than the warmup, depending on the benchmarks and the sequence lengths. Finally, concerning the benchmarks, it is important to make a distinction: in \citep{trinh_learning_2018}, this method has been tested on benchmarks with long sequences but where each timestep contain information. It is quite different from the benchmarks we have used here where large portions of the sequences consist of noise.

\end{document}